\documentclass{bmvc2k}
\usepackage{amsmath}
\usepackage{amssymb}
\usepackage{amsfonts}
\newcommand{\STAB}[1]{\begin{tabular}{@{}c@{}}#1\end{tabular}}

\newcommand{\new}[1]{{\color{black} {#1}}}

\definecolor{napiergreen}{rgb}{0.16, 0.5, 0.0}

\title{Learning from Silence and Noise for Visual Sound Source Localization}

\addauthor{Xavier Juanola}{xavier.juanola@upf.edu}{1}
\addauthor{Giovana Morais}{giovana.morais@nyu.edu}{2}
\addauthor{Magdalena Fuentes}{mfuentes@nyu.edu}{2}
\addauthor{Gloria Haro}{gloria.haro@upf.edu}{1}

\addinstitution{
 Intelligent Multimodal Vision Analysis\\
 Universitat Pompeu Fabra\\
 Barcelona, Spain
}
\addinstitution{
 MARL-IDM\\
 New York University,\\
 New York, USA
}

\runninghead{Juanola et al.}{SSL-SaN: Learning from Silence and Noise for VSSL}

\def\eg{\emph{e.g}\bmvaOneDot}

\def\ie{\emph{i.e}\bmvaOneDot}

\def\real{\mathbb{R}}

\begin{document}

\maketitle

\begin{abstract}
Visual sound source localization is a fundamental perception task that aims to detect the location of sounding sources in a video given its audio. Despite recent progress, we identify two shortcomings in current methods: 1) most approaches perform poorly in cases with low audio-visual semantic correspondence such as silence, noise, and offscreen sounds, i.e. in the presence of negative audio; and 2) most prior evaluations are limited to positive cases, where both datasets and metrics convey scenarios with a single visible sound source in the scene. To address this, we introduce three key contributions. First, we propose a new training strategy that incorporates silence and noise, which improves performance in positive cases, while being more robust against negative sounds. Our resulting self-supervised model, SSL-SaN, achieves state-of-the-art performance compared to other self-supervised models, both in sound localization and cross-modal retrieval. Second, we propose a new metric that quantifies the trade-off between alignment and separability of auditory and visual features across positive and negative audio-visual pairs. Third, we present IS3+, an extended and improved version of the IS3 synthetic dataset with negative audio. 
 Our data, metrics and code are available on the \href{https://xavijuanola.github.io/SSL-SaN/}{Project page}.
\end{abstract}

\section{Introduction}
\label{sec:intro}

Humans have an incredible capacity for multimodal integration, particularly in processing audio-visual information from the environment. 
\new{While sound localization is primarily handled 
by the auditory system, visual data aids in disambiguation and accuracy
\cite{risoud2018sound}}. 
Inspired by human perceptual abilities, multimodal approaches have demonstrated that combining auditory and visual information not only improves traditional perception tasks, such as object detection and activity recognition \cite{eliav2024audio, cheng2024integrating, salas2024multimodal}, but also enables entirely new capabilities. 
For example, audio-visual scene synthesis allows systems to reconstruct missing modalities \cite{jamaludin2019you, montesinos2023speech, sung2023sound}, predict soundscapes for silent videos \cite{du2023conditional}, or even anticipate visual scenes based on sound  \cite{liang2023av, sung2024sound2vision, shi2021survey}. Other examples that benefited from multimodality are audio-visual source separation~\cite{afouras2020self, gao2021visualvoice,montesinos2022vovit}, comprehensive scene understanding \cite{lei2023recent, fichna2021effect} and visual sound source localization.

Visual sound source localization (VSSL) aims to localize sounding sources within a video.
Pioneering approaches used networks pretrained in ImageNet \cite{deng2009imagenet} as backbones and finetuned them for audio-visual correspondence ~\cite{LVS, EZ-VSL, SLAVC}, resulting in weakly-supervised models \cite{SLAVC}. Recently, new methods, such as \cite{SSLTIE, senocak2024aligning}, shifted from weakly-supervised learning to self-supervised learning, training models from scratch and using different data augmentation techniques, \eg geometrical transformations, to avoid overfitting and improve robustness. However, these augmentations are mostly done in positive cases (\ie when there is a sounding source that is \textit{visible} in the image), and thus these models still struggle when presented with negative cases \cite{LVS, juanola2024critical}. 

Moreover, current datasets used to benchmark VSSL methods, such as VGGSound \cite{chen2020vggsound}, VGG Sound Sources (VGG-SS) \cite{LVS}, and Flickr \cite{arandjelovic2017look}, feature mostly sounding objects that are in the foreground and centered in the image frame \cite{RCGrad}. As a result, there is no strong incentive for models to use the audio content to localize the sound source in the image. Instead, models tend to identify the object regardless of sound \cite{oya2020we}. Recent efforts have been made to mitigate this by introducing synthetic evaluation sets with multiple sources \cite{senocak2024aligning}, but these still suffer from inaccurate image-audio pairings, and model evaluation is mostly done on positive cases.  

To address these shortcomings, we introduce three main contributions: (1) A novel training strategy that incorporates negative audio samples (silence and noise) during training, with two additional loss terms that ensure the network learns to ignore these non-informative sounds; 
(2) The IS3+ benchmark, an improved version of IS3 \cite{senocak2024aligning}, in which we substitute incorrect image-audio pairs that were present in the original test set with correct audio from Adobe Sound Effects \cite{adobe_sfx} and a clean subset of IS3 audio samples. In addition, we introduce the evaluation of negative audio cases, namely in the presence of noise, silence, and offscreen sounds; and (3) A new metric that quantifies the discriminative power (or separability) of auditory and visual features, which correlates with performance in both sound localization and cross-modal retrieval tasks. Through a comprehensive evaluation of sound localization models in both positive and negative audio samples, we show that our model SSL-SaN (Sound Source Localization with Silence and Noise), trained with our new strategy, yields state-of-the-art results.  
Code and data are available at \href{https://xavijuanola.github.io/SSL-SaN/}{https://xavijuanola.github.io/SSL-SaN/}. 


\section{Related Work}
\label{sec:rel_work}

Pioneering works \cite{fisher2000learning, hershey1999audio, kidron2005pixels} learn to capture correspondences between audio and visual features using classical machine learning methods, such as canonical correlation analysis \cite{hardoon2004canonical}. More recent methods adopted deep neural networks for representation learning by leveraging the synchronization between audio and video as a signal for self-supervised learning \cite{owens2018audio, korbar2018cooperative}. Currently, the most widely used approach for sound source localization is cross-modal attention \cite{senocak2018learning, senocak2019learning, tian2018audio} with contrastive loss \cite{chopra2005learning, oord2018representation, senocak2018learning, owens2018audio, LVS}. These approaches seek to localize objects by aligning audio and visual representation spaces. Some methods use additional semantic labels to pretrain audio and vision with classification loss \cite{senocak2022less} or  refine audio-visual feature alignment \cite{qian2020multiple}. \new{Knowledge distillation from pretrained object detection and sound
classification models was used in \cite{yaghoubi2023acoustic}.} LVS \cite{LVS} used a contrastive loss with hard negative mining to learn the audio-visual co-occurrence map discriminatively. 
EZ-VSL \cite{EZ-VSL} introduced a multiple instance contrastive learning framework that focuses only on the most aligned regions when matching the audio to the video by combining the attention-based localization output with a pretrained visual feature activation map. \new{ACL \cite{park2024can} leverages the CLIPSeg model \cite{luddecke2022image}, which is an image segmentation model trained in a supervised way.}

Most works in the literature were trained and evaluated with datasets that in majority feature a single-sounding object present in the scene at a given time \cite{arandjelovic2017look, arandjelovic2018objects, LVS, SSLTIE, EZ-VSL, SLAVC, owens2018audio, ramaswamy2020see, senocak2018learning, senocak2024aligning, FNAC, song2024enhancing, RCGrad}. This setting is rare in real-life scenarios, where there are multiple objects sounding at the same time (\ie a mixture of sounds), silent objects, sounds produced by objects that are not visually present in the scene or occluded by other objects (offscreen sounds), noise, etc. There has been an increasing interest in working with mixtures of sounds \cite{hu2022mix, kim2024learning, mahmud2024t, mo2023audio, qian2020multiple}, but very few worked with silent objects \cite{hu2020discriminative, liu2022visual, SLAVC} or offscreen sounds \cite{liu2022visual}. DSOL \cite{hu2020discriminative} and IEr \cite{liu2022visual} proposed a method to suppress localization of silent objects by creating an Audio-Instance-Identifier module, which identifies the sounds present in the audio, and 
filters out possible offscreen sounds and silent objects. These methods, however, rely on the number of sound sources, which is not available in most large-scale datasets, or ``in-the-wild'' data. Given the complexity of the problem and the limitations of the datasets available for training, many models in this domain are prone to overfitting. As a result, early stopping is commonly adopted as a practical regularization technique to prevent overfitting and improve generalization performance on unseen data. SLAVC \cite{SLAVC}, on the other hand, proposes a framework that solves overfitting and the need for early stopping by adding extreme visual dropout and momentum encoders. Another key  strategy to overcome data limitations is data augmentation 
\cite{chen2020simple, chen2021exploring, grill2020bootstrap, he2020momentum}. Following this, SSL-TIE \cite{SSLTIE} presents a neural network composed of an image and an audio encoder, trained with contrastive learning and geometrical consistency, ensuring that the audio-visual similarity maps undergo the same geometrical transformation as the input images. SSL-Align \cite{senocak2024aligning} proposes a novel method that utilizes semantic alignment with multi-views and semantically similar samples. The authors present both a fully self-supervised and a weakly-supervised  variation --with supervisedly pretrained audio and image encoders-- of their model. 

Except for a few works \cite{SLAVC, juanola2024critical, hamilton2024separating}, most evaluations are predominantly done on positive cases (\ie visible \textit{and} audible sources). This not only results in incomplete assessments of the model capabilities, but also reinforces biases in models, as they are implicitly optimized and judged under conditions that favor co-occurring signals. To address this, we combine image and audio augmentations with additional negative audio samples (silence and noise) {and additional loss terms} during training, resulting in a more robust and accurate self-supervised model. This fully self-supervised model outperforms previous self-supervised VSSL models, as we demonstrate in a comprehensive evaluation including both positive and negative cases. {In a concurrent work \cite{li2025audio}, and in a different context --supervised audio-visual segmentation-- the inclusion of silence and noise during training has also been shown to be beneficial.}


\section{Method}
\label{sec:method}

Most of the  VSSL models 
 (\eg \cite{arandjelovic2018objects,RCGrad, LVS, EZ-VSL, FNAC, SSLTIE, senocak2024aligning, senocak2022learning, park2023marginnce, senocak2022less, hu2020discriminative}) use a two-stream network 
with an audio encoder that extracts 
auditory features, $\mathbf{a}_i \in \real^c$, from the $i$-th audio segment and a visual encoder that extracts visual features, $\mathbf{v}_j \in \real^{c \times h \times w}$, from the $j$-th image (usually the central frame of the $j$-th audio segment). Then, an audio-visual similarity map is computed by cosine similarity:
\begin{equation}
\label{eq:sim_map}
    S(\mathbf{a}_i, \mathbf{v}_j) = \frac{\mathbf{a}_i \cdot \mathbf{v}_j }{\|\mathbf{a}_i\| \cdot \|\mathbf{v}_j \|} \in [-1, 1]^{h \times w},
\end{equation}
and  a global audio-visual correspondence value  is computed from $S$ by a certain pooling operation \cite{arandjelovic2018objects, LVS, park2023marginnce, EZ-VSL, SSLTIE, hu2020discriminative} and eventually semantic  projection heads \cite{senocak2024aligning}. The network is trained in a self-supervised way by contrastive learning; forming positive and negative audio-visual pairs just by taking $i=j$ and $i\neq j$, respectively. 
Typically, \eg \cite{LVS, EZ-VSL, FNAC, SLAVC, SSLTIE, hu2020discriminative, senocak2022less, park2023marginnce, senocak2022learning}, the audio and visual encoders are ResNet18 networks. The audio encoder is trained from scratch and the visual encoder is initialized with ImageNet pretrained weights, resulting in weakly-supervised models \cite{SLAVC}. Recent works, \cite{SSLTIE, senocak2024aligning}, have shown that state-of-the-art results in VSSL can be achieved by training both encoders from scratch (\ie in a fully self-supervised way). Both of them use (different) data augmentation techniques. Additionally, \cite{SSLTIE} uses an equivariance loss term and  \cite{senocak2024aligning} uses a semantic projection head on top of the encoders. As baseline model we use SSL-TIE \cite{SSLTIE}, since it is a state-of-the-art self-supervised model \cite{juanola2024critical} whose code is open-source. \new{However, our proposed training strategy can be applied to any localization model trained in a contrastive way.}

\subsection{Learning from Silence and Noise}
\label{sec:sil_noi}

Our aim is to design a sound localization model that is robust to negative sounds such as silence and noise. To that end, we introduce two modifications during training. First, we pair each image in the batch  with these two types of negative audio samples, thus creating two new negative audio-visual pairs. We define \textit{silence} as an empty audio and \textit{noise} as an audio with random values following a Gaussian distribution with zero mean and standard deviation $\sigma$=1. Second, we add two new loss terms forcing an empty similarity map for these two negative audio-visual pairs.  
More concretely, for every $j$-th image in the batch we define: 
$$\mathcal{L}_{S} =   \left\lVert S(\mathbf{a}^S, \mathbf{v}_j)\right\rVert _2^2,$$
the square $L_2$ norm of the audio-visual similarity map between the visual features from the $j$-th image, $\mathbf{v}_j$, and the silence feature, $\mathbf{a}^S$, and analogously, for the same visual features and the audio features extracted from a realization of the noise, $\mathbf{a}^{N}_j$, that is:
$$\mathcal{L}_{N} = \left\lVert S(\mathbf{a}^{N}_j, \mathbf{v}_j)\right\rVert _2^2.$$ 
Intuitively, these terms penalize the model localizing any sound in the presence of silence and noise, enforcing a predictable behavior in the presence of negative audio. 
Moreover, as shown in the experimental section, learning from silence and noise also improves the results with a  positive audio.

\subsection{New evaluation set: IS3+}
\label{sec:eval_set}

The current VSSL benchmarks, such as VGG-SS \cite{LVS}, contain videos gathered from YouTube, where typically a single object dominates the scene, both in the auditory and visual modalities. Consequently, the task of sound localization can be solved by using objectness cues in the image, without the need for the audio characteristics 
\cite{oya2020we}. This fact motivated the proposal, in \cite{senocak2024aligning}, of a new VSSL benchmark: \textit{Interactive-Synthetic Sound Source (IS3)}. 
IS3 spans 118 object categories and includes 3,240 images that have been synthetically generated by diffusion models \cite{rombach2022high}, simulating scenes with multiple objects in diverse sizes. Each image in IS3 is paired with two audio samples from VGG-SS, corresponding to the two most visible objects present in the image, resulting in 6,480 audio-visual instances. Although IS3 improves upon the existing benchmarks in terms of visual data, it still inherits the weaknesses of VGG-SS in the audio modality. VGG-SS videos are in-the-wild videos from YouTube and in a significant number of videos the audio signal does not contain the sound of the object in the scene. Typical examples of wrong audio samples are music instead of the original audio, an offscreen sound that masks the sound of interest, or a mixture of different types of sounds (see Supplementary material for more details). Thus, we propose an improved version of IS3, named IS3+, where each IS3 image is paired with two \textit{clean} audio samples corresponding to the two main objects in the image. 

To do so, we use a combination of audio samples from Adobe Sound
Effects (Adobe SFX)~\cite{adobe_sfx},
and clean samples from IS3. 
We manually selected samples from Adobe SFX through careful review and
semantic matching of the audio content with IS3 categories.
When we did not have enough Adobe SFX audio samples for a given class, we selected clean audio samples from IS3.
The criteria for audio selection, in both Adobe SFX and IS3, were
1) to have audio that matches the necessary category, and 
2) to avoid as much background noise as possible.
We also simplified the dataset categories in which the image did not
match the class. For example, the ``playing harpsichord" category only had images of
harps, therefore we replaced it with ``harp''. 
Similarly, we simplified categories such as ``cat growling", 
``cat caterwauling" by ``cat" as the images did not reflect these actions (the full mapping is in the Supplementary material). Finally, we create IS3+ by taking the simplified IS3 annotations and replacing the audio with a randomly sampled item within the correct category from the clean audio pool. We normalize and downsample audio samples to 16kHz.

\vspace{-0.4cm}
\section{Experiments and Results}
\label{sec:experiments}

\subsection{Experimental Setup}
\label{sec:setup}

\textbf{Datasets.} Our method is trained using the \textit{VGGSound-144K}~\cite{senocak2018learning}, a fixed subset of 144K videos randomly selected from VGGSound \cite{chen2020vggsound}.  
We test models' performance on \textit{VGG Sound Sources} (\textit{VGG-SS}) \cite{LVS} (a subset of VGG-Sound with annotated bounding boxes for sound sources), IS3 \cite{senocak2024aligning},  IS3+ (presented in Section \ref{sec:eval_set}) \new{and AVS-Bench S4 test set \cite{zhou2024avss}. We evaluate the VGG-SS test set using the bounding boxes from the annotations, while we use the segmentation masks present in IS3, IS3$^+$ and AVS-Bench S4.} 

\noindent \textbf{Evaluation Metrics.} The current standard metric to evaluate sound source localization performance is the consensus Intersection over Union (cIoU) proposed by \cite{senocak2018learning}.
To evaluate the cIoU, a threshold is necessary to binarize the audio-visual similarity map \eqref{eq:sim_map} and convert it to a localization mask. 
Recently, \cite{senocak2024aligning} proposed the \textit{Adaptive} cIoU, which sets the top $B$ pixels to $1$, where $B$ is the area of the ground truth bounding box or mask. This metric assumes that the object is always visible and that the object's size is known, assumptions that do not hold in most cases. On the other hand, \cite{juanola2024critical} proposed a \textit{Universal threshold} that does not assume a visible object in the scene, and is designed to be discriminative of positive vs.\;negative audio cases. This threshold is set to be the $3${rd} quartile of the maximum audio-visual similarity in negative cases, so that it filters false positives without any  assumption on the object size. 
We report cIoU with both the Universal threshold (Uth) and the adaptive one (Adap.). We also report the Area Under the Curve (AUC), which measures the integral of the success ratio (proportion of samples with cIoU$ > \tau $) as a function of the threshold $\tau$ varying from 0 to 1.

Besides these metrics, we report the percentage of Image Area ({pIA}) \cite{juanola2024critical}, which is designed to assess models' performance with negative audio by measuring the proportion of the image area that has been activated in the model's localization mask in the presence of noise, silence and offscreen sounds separately. We also report \textit{AUC$_N$} which is analogous to AUC for the case of negative audio (in this case, the success ratio is defined as proportion of samples with pIA $\leq \tau$). 
 Finally, we evaluate overall model performance across both positive and negative cases with \textit{F$_{LOC}$ and F$_{AUC}$} \cite{juanola2024critical}. They both compute the harmonic mean between a positive and a negative metric (F$_{LOC}$ uses the cIoU and pIA while F$_{AUC}$ uses AUC and AUC$_N$).

\noindent \textbf{Implementation}. Following \cite{SSLTIE}, we resize input images into a resolution of $224 \times 224$, and we represent audio samples by log-Mel Spectrograms, extracted from 3 seconds of audio at a sample rate of 16 kHz. The log-Mel Spectrograms are computed using 512 as the size of the FFT windows, 239 for the hop length between STFT windows, and 257 mel filterbanks. We use ResNet18 \cite{he2016deep} for the audio and image encoders. We train all models from scratch.
The models are trained with a batch size of 32, the Adam optimizer \cite{kingma2014adam} with a learning rate of $1e^{-4}$ and ReduceOnPlateau as the learning rate scheduler.

\subsection{Comparison to prior work}
\label{sec:results}

We compare our model to different prior works:  RCGrad \cite{RCGrad},  LVS \cite{LVS},  EZ-VSL \cite{EZ-VSL},  FNAC \cite{FNAC},  SLAVC \cite{SLAVC}, \new{ACL \cite{park2024can},} SSL-Align \cite{senocak2024aligning}, and  SSL-TIE \cite{SSLTIE}. \new{SSL-TIE and a version of SSL-Align  are, as ours, the only fully sef-supervised models. The rest are weakly-supervised, since they leverage image encoders/decoders pretrained in a supervised way on annotated datasets such as ImageNet \cite{deng2009imagenet} or PhraseCut \cite{wu2020phrasecut} (in case of ACL).
We exclude  object guided localization (OGL) postprocessing \cite{EZ-VSL}, as it is not meaningful for negative sounds.}

\begin{figure*}[htb] 
\centerline{\includegraphics[width=\textwidth]{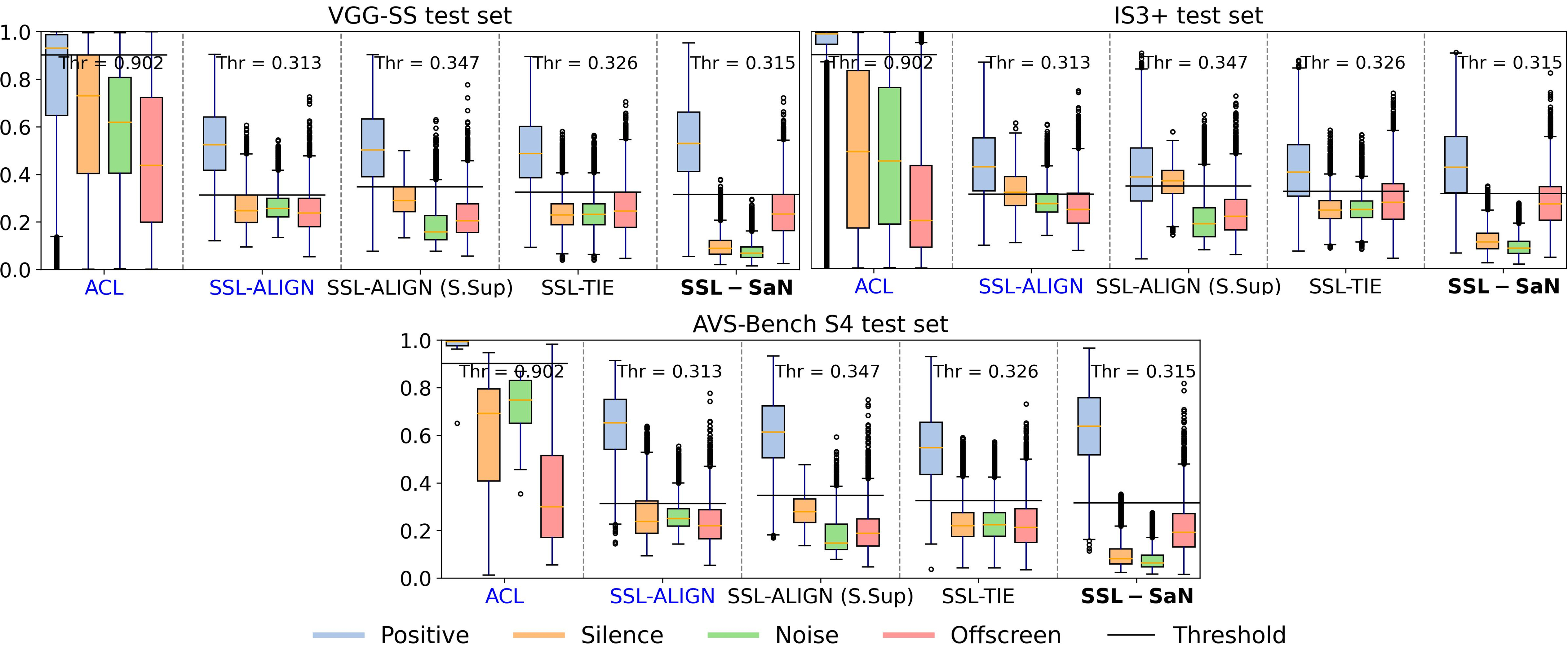}}
\caption{Distribution of the maximum values of the audio-visual similarity maps across different models and  datasets, for both positive and negative audio samples. The Universal threshold for each model is indicated. Weakly-supervised models are indicated in blue.}
\label{fig:boxplots}
\end{figure*}

\noindent \textbf{Modality Alignment and Separability.}  
A good VSSL model should align auditory and visual features corresponding to a positive audio-visual pair and separate them in case of a negative  pair, giving rise to a high value of audio-visual similarity in the first case and a low similarity in the second. We adopt the visualization of Figure \ref{fig:boxplots} from \cite{juanola2024critical},  showing the distribution of the maximum values of the similarity maps of the positive and the different negative cases. As can be seen,  SSL-SaN is the model that better learns to separate the positives from the negatives in general, considering the three datasets (the rest of models are shown in the Supplementary material). 
We propose a new metric, denoted as \textit{separability}: 
 \begin{equation} \label{eq:separability}
 \mathcal{S}ep = Q_1^+ - Q_3^-
  \end{equation}
  where $Q_1^+$ denotes the 1st quartile of the maximum audio-visual similarities of the positive pairs and $Q_3^-$ the 3rd quartile of the maximum audio-visual similarities of the negative pairs. $\mathcal{S}ep$  is a real number and the larger the value, the better the VSSL model is at discriminating positive audio-visual pairs from negative ones. A negative value indicates that the interquartile ranges  of audio-visual similarity values for positive pairs ($Q_3^+ - Q_1^+$) and negative pairs ($Q_3^- - Q_1^-$) overlap. This last scenario is not desirable.

    \begin{table}[ht]
    \centering
    \resizebox{0.7\columnwidth}{!}{
    \begin{tabular}{c|lccccc}
    \hline
    \textbf{Test set} & \textbf{Model} & \textbf{Self Supervised} & \textbf{Magnitude $\downarrow$} & \textbf{Alignment $\uparrow$} & \textbf{Separability$_{\pm}$ $\uparrow$} \\
    \hline

    \multirow{8}{*}{\STAB{\rotatebox[origin=c]{90}{\textbf{VGG-SS}}}}

    &LVS {\scriptsize [CVPR, 2021]}
    &\ding{56}
    &\textit{0.92} 
    &0.19
    &-0.1124 \\ 

    &EZ-VSL {\scriptsize [ECCV, 2022]}
    &\ding{56}
    &1.14 
    &0.20
    &-0.0339 \\ 

    &FNAC {\scriptsize [CVPR, 2023]} 
    &\ding{56}
    &1.25  
    &0.03
    &-0.0094 \\ 

    &SLAVC {\scriptsize [NEURIPS, 2022]} 
    &\ding{56}
    &1.14  
    &0.22
    &-0.0427 \\ 

    &SSL-Align {\scriptsize [ICCV, 2023]} 
    &\ding{56}
    &0.94  
    &\textit{0.41}
    &\textit{0.1034} \\ 
    
    &ACL {\scriptsize [WACV, 2025]}
    &\ding{56}
    &1.31 
    &0.14
    &-0.2551 \\ 

    &\cellcolor[gray]{0.9}SSL-TIE {\scriptsize [ACMMM, 2022]}
    &\cellcolor[gray]{0.9}\ding{51}
    &\cellcolor[gray]{0.9}0.96 
    &\cellcolor[gray]{0.9}0.37
    &\cellcolor[gray]{0.9}\underline{0.0606} \\ 

    &\cellcolor[gray]{0.9}SSL-Align {\scriptsize [ICCV, 2023]}
    &\cellcolor[gray]{0.9}\ding{51}
    &\cellcolor[gray]{0.9}\underline{0.95} 
    &\cellcolor[gray]{0.9}\underline{0.39}
    &\cellcolor[gray]{0.9}0.0428 \\ 

    &\cellcolor[gray]{0.9}\textbf{Ours $\rightarrow$ SSL-SaN}
    &\cellcolor[gray]{0.9}\ding{51}
    &\cellcolor[gray]{0.9}\textbf{0.93} 
    &\cellcolor[gray]{0.9}\textbf{0.40}
    &\cellcolor[gray]{0.9}\textbf{0.0971} \\ 

    \hline 

    \multirow{8}{*}{\STAB{\rotatebox[origin=c]{90}{\textbf{IS3}}}}

    &LVS {\scriptsize [CVPR, 2021]}
    &\ding{56}
    &\textit{0.95} 
    &0.11
    &-0.1113 \\ 

    &EZ-VSL {\scriptsize [ECCV, 2022]}
    &\ding{56}
    &1.14 
    &0.17
    &-0.0423 \\ 

    &FNAC {\scriptsize [CVPR, 2023]} 
    &\ding{56}
    &1.24  
    &0.02
    &-0.0241 \\ 

    &SLAVC {\scriptsize [NEURIPS, 2022]} 
    &\ding{56}
    &1.17  
    &0.15
    &-0.0733 \\ 

    &SSL-Align {\scriptsize [ICCV, 2023]} 
    &\ding{56}
    &0.98  
    &\textit{0.32}
    &-0.0608 \\ 
    
    &ACL {\scriptsize [WACV, 2025]}
    &\ding{56}
    &1.29  
    &0.17
    &\textit{0.1112} \\ 

    &\cellcolor[gray]{0.9}SSL-TIE {\scriptsize [ACMMM, 2022]}
    &\cellcolor[gray]{0.9}\ding{51}
    &\cellcolor[gray]{0.9}\underline{0.98}  
    &\cellcolor[gray]{0.9}\underline{0.28}
    &\cellcolor[gray]{0.9}\underline{-0.0530} \\ 

    &\cellcolor[gray]{0.9}SSL-Align {\scriptsize [ICCV, 2023]} 
    &\cellcolor[gray]{0.9}\ding{51} 
    &\cellcolor[gray]{0.9}0.99   
    &\cellcolor[gray]{0.9}0.26  
    &\cellcolor[gray]{0.9}-0.1287 \\ 

    &\cellcolor[gray]{0.9}\textbf{Ours $\rightarrow$ SSL-SaN} 
    &\cellcolor[gray]{0.9}\ding{51}  
    &\cellcolor[gray]{0.9}\textbf{0.97}  
    &\cellcolor[gray]{0.9}\textbf{0.31}  
    &\cellcolor[gray]{0.9}\textbf{-0.0327} \\ 

    \hline 

    \multirow{8}{*}{\STAB{\rotatebox[origin=c]{90}{\textbf{IS3+}}}}

    &LVS {\scriptsize [CVPR, 2021]}  
    &\ding{56}  
    &\textit{0.96} 
    &0.10  
    &-0.1148 \\ 

    &EZ-VSL {\scriptsize [ECCV, 2022]}  
    &\ding{56}  
    &1.14 
    &0.17  
    &-0.0423 \\ 

    &FNAC {\scriptsize [CVPR, 2023]}  
    &\ding{56}  
    &1.24  
    &0.01  
    &-0.0231 \\ 

    &SLAVC {\scriptsize [NEURIPS, 2022]}  
    &\ding{56}  
    &1.16 
    &0.15  
    &-0.0692 \\ 

    &SSL-Align {\scriptsize [ICCV, 2023]}  
    &\ding{56}  
    &0.98 
    &\textit{0.33}  
    &-0.0480 \\ 
    
    &ACL {\scriptsize [WACV, 2025]}
    &\ding{56}
    &1.29 
    &0.17
    &\textit{0.0717} \\ 

    &\cellcolor[gray]{0.9}SSL-TIE {\scriptsize [ACMMM, 2022]}  
    &\cellcolor[gray]{0.9}\ding{51}  
    &\cellcolor[gray]{0.9}\underline{0.99}  
    &\cellcolor[gray]{0.9}\underline{0.28}  
    &\cellcolor[gray]{0.9}\underline{-0.0604} \\ 

    &\cellcolor[gray]{0.9}SSL-Align {\scriptsize [ICCV, 2023]} 
    &\cellcolor[gray]{0.9}\ding{51} 
    &\cellcolor[gray]{0.9}\underline{0.99} 
    &\cellcolor[gray]{0.9}0.26  
    &\cellcolor[gray]{0.9}-0.1279 \\ 

    &\cellcolor[gray]{0.9}\textbf{Ours $\rightarrow$ SSL-SaN} 
    &\cellcolor[gray]{0.9}\ding{51} 
    &\cellcolor[gray]{0.9}\textbf{0.97} 
    &\cellcolor[gray]{0.9}\textbf{0.31} 
    &\cellcolor[gray]{0.9}\textbf{-0.0327} \\ 
    
    \hline 

    \multirow{8}{*}{\STAB{\rotatebox[origin=c]{90}{\textbf{AVS-Bench S4}}}}

    &LVS {\scriptsize [CVPR, 2021]}  
    &\ding{56}  
    &0.88 
    &0.28  
    &-0.1085 \\ 

    &EZ-VSL {\scriptsize [ECCV, 2022]}  
    &\ding{56}  
    &1.12 
    &0.18  
    &-0.0180 \\ 

    &FNAC {\scriptsize [CVPR, 2023]}  
    &\ding{56}  
    &1.23 
    &0.01  
    &-0.0017 \\ 

    &SLAVC {\scriptsize [NEURIPS, 2022]}  
    &\ding{56}  
    &1.15 
    &0.14  
    &-0.0327 \\ 

    &SSL-Align {\scriptsize [ICCV, 2023]}  
    &\ding{56}  
    &\textit{0.86} 
    &\textit{0.49}  
    &0.2161 \\ 
    
    &ACL {\scriptsize [WACV, 2025]}
    &\ding{56}
    &1.28 
    &0.18
    &0.0575 \\ 

    &\cellcolor[gray]{0.9}SSL-TIE {\scriptsize [ACMMM, 2022]}  
    &\cellcolor[gray]{0.9}\ding{51}  
    &\cellcolor[gray]{0.9}0.93 
    &\cellcolor[gray]{0.9}0.41  
    &\cellcolor[gray]{0.9}0.1442 \\ 

    &\cellcolor[gray]{0.9}SSL-Align {\scriptsize [ICCV, 2023]} 
    &\cellcolor[gray]{0.9}\ding{51} 
    &\cellcolor[gray]{0.9}\underline{0.88} 
    &\cellcolor[gray]{0.9}\textbf{0.48} 
    &\cellcolor[gray]{0.9}\underline{0.1729} \\ 

    &\cellcolor[gray]{0.9}\textbf{Ours $\rightarrow$ SSL-SaN} 
    &\cellcolor[gray]{0.9}\ding{51} 
    &\cellcolor[gray]{0.9}\textbf{\textit{0.86}} 
    &\cellcolor[gray]{0.9}\underline{0.47} 
    &\cellcolor[gray]{0.9}\textbf{\textit{0.2479}} \\ 
    
    \hline
    \end{tabular}}
    \vspace{0.2cm}
    \caption{Results of the cross-modal alignment and separability analysis. Best results in \textit{italics}. Best results of self-supervised  models in  \textbf{bold} and the second-best  \underline{underlined}.}
    \label{tab:modality_gap}
    \end{table}

Table \ref{tab:modality_gap} reports the separability metric in VGG-SS, \new{IS3}, IS3+ \new{and AVS-Bench S4} test sets. 
To better understand how models align both modalities in a positive pair, we also report the metrics of magnitude and alignment \cite{zhang2023diagnosing, goel2022cyclip}. \textit{Alignment} measures how close the audio and image embeddings are for positive pairs using cosine similarity, while \textit{magnitude} quantifies the $L_2$ distance between both embeddings. As shown in Table \ref{tab:modality_gap}, the LVS model achieves the best scores in both magnitude and alignment \new{in VGG-SS, IS3 and IS3$^+$, and second best in S4}, which at first glance could suggest an excellent reduction of the modality gap. However, this result seems to indicate that the audio and image embeddings are collapsing into the same region of the embedding space,  rather than aligning them in a semantically meaningful way. This collapse reduces their discriminative power and is reflected in LVS's poor performance on the separability  metric. A reversed pattern is observed with FNAC, which achieves a relatively good separability score, but its poor alignment and high magnitude suggest that the model struggles to structure the modality representations meaningfully. 
In contrast, SSL-SaN model achieves a better balance,  obtaining the
best results in magnitude and alignment among the  self-supervised models \new{in VGG-SS, IS3 and IS3$^+$, and best magnitude and second best alignment in S4}.  Most importantly, it achieves the best score, among the self-supervised models, in the separability metric \new{in all test sets}, which highlights a strong distinction between positive and negative samples. These results demonstrate the strength of our fully self-supervised training approach, which encourages the learning of semantically rich and well-structured audio-visual representations.

    \begin{table*}[ht]
    \centering
    \resizebox{\textwidth}{!}{ 
\begin{tabular}{@{}c|lcccccccccccccccc@{}}\hline
    & & & \multicolumn{4}{c}{{\bf Positive audio input}} & & \multicolumn{6}{c}{{\bf Negative audio input}} & & \multicolumn{2}{c}{{\bf Global metric}} &\\
    & & & \multicolumn{2}{c}{{\bf cIoU}} & \multicolumn{2}{c}{{\bf AUC}} & & \multicolumn{2}{c}{{\bf Silence}} & \multicolumn{2}{c}{{\bf Noise}} & \multicolumn{2}{c}{{\bf Offscreen sound}} & &  & \\
    \cline{4-7} \cline{9-10} \cline{11-12} \cline{13-14} \cline{16-17}
    {\bf Test set} & {\bf Model} & {\bf Self Sup.} & {\bf Uth $\uparrow$} & {\bf Adap. $\uparrow$} & {\bf Uth $\uparrow$} & {\bf Adap. $\uparrow$} & & {\bf pIA $\downarrow$} & {\bf AUC$_{\text{N}}$ $\uparrow$} & {\bf pIA $\downarrow$} & {\bf AUC$_{\text{N}}$ $\uparrow$} & {\bf pIA $\downarrow$} & {\bf AUC$_{\text{N}}$ $\uparrow$} & & {\bf F$_{\text{LOC}}$ $\uparrow$} & {\bf F$_{\text{AUC}}$ $\uparrow$}\\
    \hline

    \multirow{9}{*}{\STAB{\rotatebox[origin=c]{90}{\textbf{VGG-SS}}}}

    & RCGrad {\scriptsize [INTERSPEECH, 2022]}
        & \ding{56} 
        &  11.71 
        &  37.04 
        &  12.08 
        &  37.26 
        & 
        &  4.42 
        &  95.80 
        &  4.42 
        &  95.80 
        &  4.41 
        &  95.82 
        & 
        &  20.86 
        &  21.45 \\ 

    & LVS {\scriptsize [CVPR, 2021]}
        & \ding{56} 
        & 3.90 
        & 39.43 
        & 5.91 
        & 41.27 
        & 
        & 1.59 
        & 98.30 
        & 0.05 
        & 99.93 
        & 0.70 
        & 99.23 
        & 
        & 7.51 
        & 11.15 \\
\hline
    \end{tabular}%
    }
    \vspace{0.1cm}
    \caption{Localization  results on the VGG-SS, IS3$^{+}$ and S4 extended test sets. The best value across all models is shown in \textit{italics}. Within the self-supervised subset, the best value is in \textbf{bold} and the second-best is \underline{underlined}.}
    \label{tab:all_metrics}
    \end{table*}

\noindent \textbf{Localization results.} As reported  in Table \ref{tab:all_metrics}, our model outperforms all prior self-supervised works in the three test sets in terms of cIoU-Uth, metrics of silence and noise, and more importantly, the global metrics. It is second best in offscreen metrics. For  cIoU-Adap. it is the best in S4 and second best in the rest of datasets. s
Thanks to the addition of loss terms specifically addressing silence and noise during training, our model completely filters out silence and noise. 
Interestingly, the weakly-supervised version of SSL-Align achieves by far the best results on positive sounds, but not on the negative ones (across all datasets). In fact, it performs worse on negative sounds compared to its fully self-supervised version. We hypothesize that the weakly-supervised version, which uses a pretrained visual encoder from  ImageNet, is more prone to leveraging objectness cues in the image \cite{oya2020we}. This appears to be  beneficial for positive sound but detrimental for negative ones. 
\new{ACL achieves the best results in positive, offscreen and global metrics in IS3, IS3+ and S4. It  leverages CLIPSeg \cite{luddecke2022image}, an   image segmentation model based on CLIP with a conditioned transformer decoder trained for object segmentation in a supervised way, with both positive and negative segmentation queries. Thus, it computes powerful and robust features for localization (highly related to segmentation) but less competitive for cross-modal retrieval, as shown in the next results.}

\noindent \textbf{Cross-modal retrieval.} As in \cite{SSLTIE, senocak2024aligning}, we evaluate VSSL models using a cross-modal retrieval task, to better assess  how they capture the semantic correspondence between the audio and visual modalities. For this evaluation, we use Precision and Accuracy  at top-$K$ retrieved results, P@$K$ and A@$K$, respectively. Precision refers to the proportion of the top-$K$ retrieved items that belong to the same category as the query, while Accuracy considers a retrieval successful if at least one item among the top-$K$ matches the query's category. Table~\ref{tab:retrieval_paper_main} reports  results for VGG-SS \new{and AVS-Bench S4}  (IS3 and IS3+  ones are in the Supp. mat.). 
Our model outperforms the self-supervised SSL-Align model \new{at all $K$ values in VGG-SS, IS3 and IS3$^+$ for both Precision and Accuracy in Image to Audio. Our model outperforms the self-supervised SSL-Align at high $K$ for Precision in both Image to Audio and Audio to Image. On the other hand, the self-supervised version of SSL-Align outperforms our model in Audio to Image  for all test sets except at high $K$ values of Precision for S4 test set.}
\new{Interestingly, our model improves by a large margin its baseline method, SSL-TIE, across all metrics and datasets, showing the benefits of including silence and noise in the learning stage.}

    \begin{table}[ht]
    \centering
    \resizebox{\columnwidth}{!}{
    \begin{tabular}{c|lcccccccccccccccc}
    \hline
    & & & & \multicolumn{6}{c}{I $\rightarrow$ A} & & \multicolumn{6}{c}{A $\rightarrow$ I} \\
    \cline{5-10} \cline{12-17}
    \textbf{Test set} & \textbf{Model} & \textbf{Self Supervised} & & \textbf{P@1 $\uparrow$} & \textbf{P@5 $\uparrow$} & \textbf{P@10 $\uparrow$} & \textbf{A@1 $\uparrow$} & \textbf{A@5 $\uparrow$} & \textbf{A@10 $\uparrow$} && \textbf{P@1 $\uparrow$} & \textbf{P@5 $\uparrow$} & \textbf{P@10 $\uparrow$} & \textbf{A@1 $\uparrow$} & \textbf{A@5 $\uparrow$} & \textbf{A@10 $\uparrow$}  \\
    \hline

    \multirow{8}{*}{\STAB{\rotatebox[origin=c]{90}{\textbf{VGG-SS}}}}
    
    & LVS {\scriptsize [CVPR, 2021]} & 
    \ding{56} & 
    & 
    2.96 & 
    2.84 & 
    2.83 & 
    2.96 & 
    10.37 & 
    16.47 & 
    &
    4.02 & 
    3.92 & 
    3.54& 
    4.02 & 
    13.71 & 
    20.22 \\ 

    & EZ-VSL {\scriptsize [ECCV, 2022]} & 
    \ding{56} & 
    & 
    2.11 & 
    2.27 & 
    2.16 & 
    2.11 & 
    8.21 & 
    13.06 & 
    &
    3.94 & 
    3.51 & 
    3.24& 
    3.94 & 
    14.19 & 
    22.55 \\ 

    & FNAC {\scriptsize [CVPR, 2023]} & 
    \ding{56} & 
    & 
    2.03  & 
    1.83 & 
    2.06 & 
    2.03 & 
    6.65 & 
    14.26 & 
    &
    2.08 & 
    1.66 & 
    1.84& 
    2.08 & 
    7.51 & 
    14.92 \\ 

    & SLAVC {\scriptsize [NEURIPS, 2022]} & 
    \ding{56} & 
    & 
    3.54  & 
    3.14 & 
    2.93 & 
    3.54 & 
    10.80 & 
    16.02 & 
    &
    4.07 & 
    3.51 & 
    3.32& 
    4.07 & 
    14.72 & 
    24.91 \\ 

    & SSL-Align {\scriptsize [ICCV, 2023]} & 
    \ding{56} & 
    & 
    \textit{27.95}  & 
    \textit{25.38} & 
    \textit{23.13} & 
    \textit{27.95} & 
    \textit{49.00} & 
    58.64 & 
    &
    \textit{32.04} & 
    \textit{29.15} & 
    \textit{26.26} & 
    \textit{32.04} & 
    \textit{56.93} & 
    \textit{67.08} \\ 
    
    & ACL {\scriptsize [WACV, 2025]}   & 
    \ding{56} & 
     & 
    10.07 & 
    9.36 & 
    8.92 & 
    10.07 & 
    29.60 & 
    43.36 & 
    &
    13.35 & 
    12.92 & 
    12.44& 
    13.35 & 
    33.52 & 
    43.87 \\ 

    & \cellcolor[gray]{0.9}SSL-TIE {\scriptsize [ACMMM, 2022]}     & 
    \cellcolor[gray]{0.9}\ding{51} & 
    \cellcolor[gray]{0.9} &
    \cellcolor[gray]{0.9}16.25 & 
    \cellcolor[gray]{0.9}14.87 & 
    \cellcolor[gray]{0.9}13.83 & 
    \cellcolor[gray]{0.9}16.25 & 
    \cellcolor[gray]{0.9}35.81 & 
    \cellcolor[gray]{0.9}47.51 & 
    \cellcolor[gray]{0.9}&
    \cellcolor[gray]{0.9}15.12 & 
    \cellcolor[gray]{0.9}14.82 & 
    \cellcolor[gray]{0.9}13.81& 
    \cellcolor[gray]{0.9}15.12 & 
    \cellcolor[gray]{0.9}35.91 & 
    \cellcolor[gray]{0.9}47.41 \\ 

    & \cellcolor[gray]{0.9}SSL-Align {\scriptsize [ICCV, 2023]}   & 
    \cellcolor[gray]{0.9}\ding{51} & 
    \cellcolor[gray]{0.9} & 
    \cellcolor[gray]{0.9}\underline{22.65} & 
    \cellcolor[gray]{0.9}\underline{20.38} & 
    \cellcolor[gray]{0.9}\underline{19.04} & 
    \cellcolor[gray]{0.9}\underline{22.65} & 
    \cellcolor[gray]{0.9}\underline{44.53} & 
    \cellcolor[gray]{0.9}\underline{56.50} & 
    \cellcolor[gray]{0.9}&
    \cellcolor[gray]{0.9}\textbf{26.67} & 
    \cellcolor[gray]{0.9}\textbf{23.82} & 
    \cellcolor[gray]{0.9}\textbf{21.40}& 
    \cellcolor[gray]{0.9}\textbf{26.67} & 
    \cellcolor[gray]{0.9}\textbf{51.26} & 
    \cellcolor[gray]{0.9}\textbf{61.85} \\ 

    & \cellcolor[gray]{0.9}\textbf{Ours $\rightarrow$ SSL-SaN}    & 
    \cellcolor[gray]{0.9}\ding{51} & 
    \cellcolor[gray]{0.9} & 
    \cellcolor[gray]{0.9}\textbf{24.61}  & 
    \cellcolor[gray]{0.9}\textbf{22.94} & 
    \cellcolor[gray]{0.9}\textbf{20.88} & 
    \cellcolor[gray]{0.9}\textbf{24.61} & 
    \cellcolor[gray]{0.9}\textbf{48.57} & 
    \cellcolor[gray]{0.9}\textbf{\textit{58.97}} & 
    \cellcolor[gray]{0.9}&
    \cellcolor[gray]{0.9}\underline{24.49} & 
    \cellcolor[gray]{0.9}\underline{22.97} & 
    \cellcolor[gray]{0.9}\underline{21.25}& 
    \cellcolor[gray]{0.9}\underline{24.49} & 
    \cellcolor[gray]{0.9}\underline{46.74} & 
    \cellcolor[gray]{0.9}\underline{57.41} \\ 
    
    \hline 
    
    \multirow{8}{*}{\STAB{\rotatebox[origin=c]{90}{\textbf{AVS-Bench S4}}}}
    
    & LVS {\scriptsize [CVPR, 2021]} & 
    \ding{56} & 
    & 
    20.81 & 
    23.89 & 
    25.15 & 
    20.81 & 
    56.49 & 
    70.81 & 
    &
    33.38 & 
    30.81 & 
    30.26& 
    33.38 & 
    54.59 & 
    66.22 \\ 

    & EZ-VSL {\scriptsize [ECCV, 2022]} & 
    \ding{56} & 
    & 
    2.97 & 
    3.70 & 
    3.81 & 
    2.97 & 
    13.65 & 
    23.92 & 
    &
    5.14 & 
    5.14 & 
    4.86& 
    5.14 & 
    5.41 & 
    9.19 \\ 

    & FNAC {\scriptsize [CVPR, 2023]} & 
    \ding{56} & 
    & 
    5.41  & 
    4.92 & 
    4.86 & 
    5.41 & 
    11.89 & 
    20.95 & 
    &
    4.59 & 
    4.62 & 
    5.12& 
    4.59 & 
    5.14 & 
    9.86 \\ 

    & SLAVC {\scriptsize [NEURIPS, 2022]} & 
    \ding{56} & 
    & 
    5.95  & 
    5.22 & 
    5.01 & 
    5.95 & 
    15.68 & 
    24.32 & 
    &
    6.76 & 
    6.62 & 
    6.47& 
    6.76 & 
    7.16 & 
    12.70 \\ 

    & SSL-Align {\scriptsize [ICCV, 2023]} & 
    \ding{56} & 
    & 
    \textit{85.27}  & 
    \textit{84.08} & 
    \textit{82.38} & 
    \textit{85.27} & 
    \textit{91.76} & 
    93.24 & 
    &
    \textit{87.16} & 
    \textit{86.00} & 
    \textit{84.70} & 
    \textit{87.16} & 
    \textit{94.86} & 
    \textit{96.76} \\ 
    
    & ACL {\scriptsize [WACV, 2025]}   & 
    \ding{56} & 
     & 
    75.14 & 
    73.95 & 
    71.68 & 
    75.14 & 
    91.08 & 
    94.05 & 
    &
    72.57 & 
    72.57 & 
    72.22& 
    72.57 & 
    72.57 & 
    80.14 \\ 

    & \cellcolor[gray]{0.9}SSL-TIE {\scriptsize [ACMMM, 2022]}     & 
    \cellcolor[gray]{0.9}\ding{51} & 
    \cellcolor[gray]{0.9} &
    \cellcolor[gray]{0.9}73.38 & 
    \cellcolor[gray]{0.9}75.27 & 
    \cellcolor[gray]{0.9}74.92 & 
    \cellcolor[gray]{0.9}73.38 & 
    \cellcolor[gray]{0.9}85.27 & 
    \cellcolor[gray]{0.9}90.00 & 
    \cellcolor[gray]{0.9}&
    \cellcolor[gray]{0.9}66.76 & 
    \cellcolor[gray]{0.9}66.76 & 
    \cellcolor[gray]{0.9}66.69& 
    \cellcolor[gray]{0.9}66.76 & 
    \cellcolor[gray]{0.9}66.76 & 
    \cellcolor[gray]{0.9}77.03 \\ 

    & \cellcolor[gray]{0.9}SSL-Align {\scriptsize [ICCV, 2023]}   & 
    \cellcolor[gray]{0.9}\ding{51} & 
    \cellcolor[gray]{0.9} & 
    \cellcolor[gray]{0.9}\textbf{80.00} & 
    \cellcolor[gray]{0.9}\underline{78.54} & 
    \cellcolor[gray]{0.9}\underline{77.43} & 
    \cellcolor[gray]{0.9}\textbf{80.00} & 
    \cellcolor[gray]{0.9}\textbf{\textit{91.76}} & 
    \cellcolor[gray]{0.9}\textbf{\textit{94.59}} & 
    \cellcolor[gray]{0.9}&
    \cellcolor[gray]{0.9}\textbf{81.89} & 
    \cellcolor[gray]{0.9}\underline{80.65} & 
    \cellcolor[gray]{0.9}\underline{79.28}& 
    \cellcolor[gray]{0.9}\textbf{81.89} & 
    \cellcolor[gray]{0.9}\textbf{91.08} & 
    \cellcolor[gray]{0.9}\textbf{94.73} \\ 

    & \cellcolor[gray]{0.9}\textbf{Ours $\rightarrow$ SSL-SaN}    & 
    \cellcolor[gray]{0.9}\ding{51} & 
    \cellcolor[gray]{0.9} & 
    \cellcolor[gray]{0.9}\underline{79.86}  & 
    \cellcolor[gray]{0.9}\textbf{80.81} & 
    \cellcolor[gray]{0.9}\textbf{81.50} & 
    \cellcolor[gray]{0.9}\underline{79.86} & 
    \cellcolor[gray]{0.9}\underline{89.59} & 
    \cellcolor[gray]{0.9}\underline{92.43} & 
    \cellcolor[gray]{0.9}&
    \cellcolor[gray]{0.9}\underline{81.35} & 
    \cellcolor[gray]{0.9}\textbf{81.38} & 
    \cellcolor[gray]{0.9}\textbf{81.82}& 
    \cellcolor[gray]{0.9}\underline{81.35} & 
    \cellcolor[gray]{0.9}\underline{81.49} & 
    \cellcolor[gray]{0.9}\underline{87.16} \\ 
    
    \hline
    \end{tabular}}
    \vspace{0.1cm}
    \caption{Results of the cross-modal retrieval. 
    Best results in \textit{italics}. 
    Best results of self-supervised  models in  \textbf{bold} and the second-best  \underline{underlined}.}
    \label{tab:retrieval_paper_main}
    \end{table}

\begin{figure}[!ht]
\label{fig:inferences}
\begin{center}
\begin{tabular}{cccccc}
& \scriptsize \hspace{-0.35cm} \raisebox{-0.17cm}{\includegraphics[height=0.5cm]{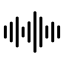}} Dinosaur & \scriptsize \hspace{-0.5cm} \raisebox{-0.17cm}{\includegraphics[height=0.5cm]{Figures/equalizer.png}} Firetruck & \scriptsize \hspace{-0.5cm} \raisebox{-0.17cm}{\includegraphics[height=0.5cm]{Figures/equalizer.png}} Silence & \scriptsize \hspace{-0.6cm} \raisebox{-0.17cm}{\includegraphics[height=0.5cm]{Figures/equalizer.png}} Noise & \hspace{-0.52cm} \raisebox{-0.17cm}{\includegraphics[height=0.5cm]{Figures/equalizer.png}} \scriptsize Offscreen \\
\vspace{0.05cm}
\rotatebox[origin=c]{90}{\scriptsize SSL-TIE } & 
\hspace{-0.45cm} \includegraphics[align=c, width=1.45cm]{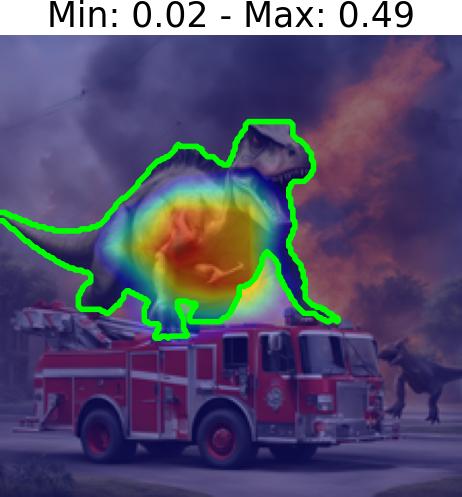}&
\hspace{-0.45cm} \includegraphics[align=c, width=1.45cm]{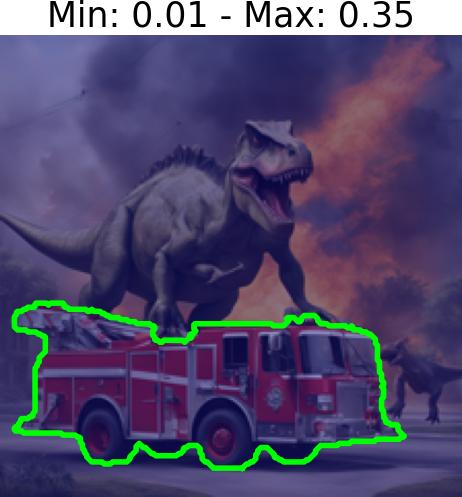}&
\hspace{-0.45cm} \includegraphics[align=c, width=1.45cm]{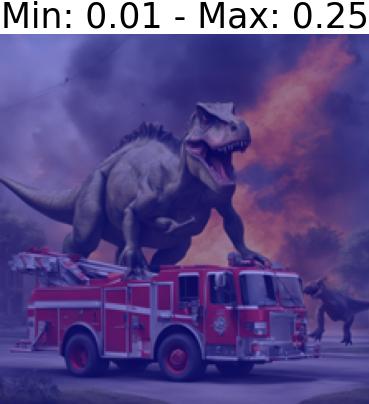}&
\hspace{-0.45cm} \includegraphics[align=c, width=1.45cm]{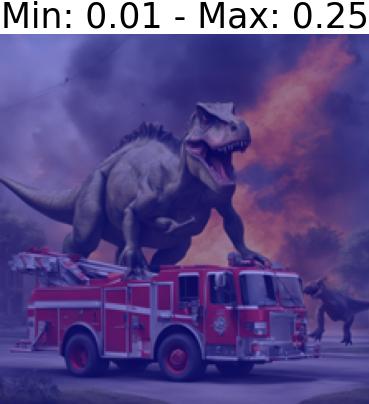}&
\hspace{-0.45cm} \includegraphics[align=c, width=1.45cm]{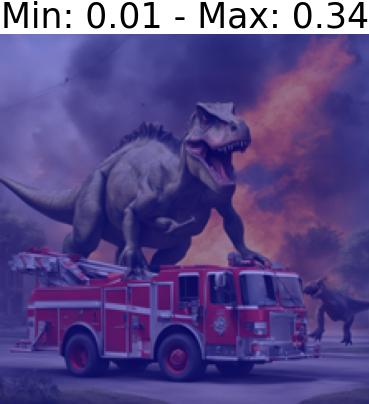}\\
\vspace{0.05cm}
\rotatebox[origin=c]{90}{\scriptsize SSL-Align (S.S.) } & 
\hspace{-0.45cm} \includegraphics[align=c, width=1.45cm]{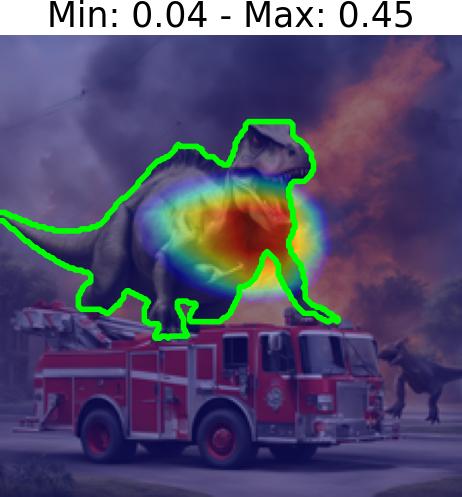}&
\hspace{-0.45cm} \includegraphics[align=c, width=1.45cm]{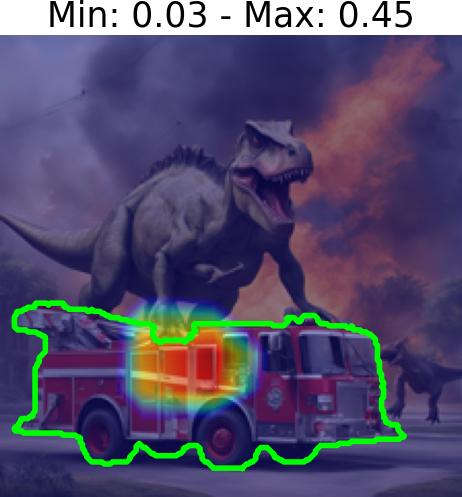}&
\hspace{-0.45cm} \includegraphics[align=c, width=1.45cm]{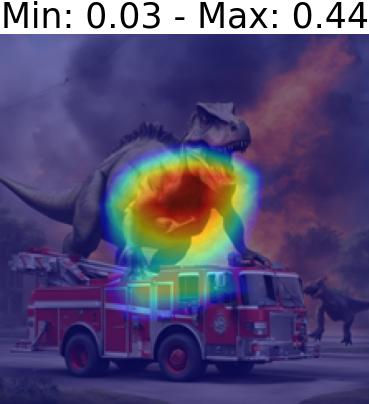}&
\hspace{-0.45cm} \includegraphics[align=c, width=1.45cm]{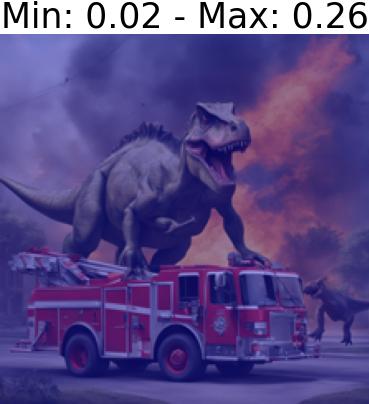}&
\hspace{-0.45cm} \includegraphics[align=c, width=1.45cm]{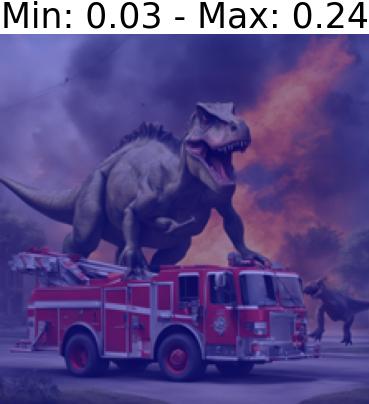}\\ \vspace{0.05cm}
\rotatebox[origin=c]{90}{\scriptsize \textbf{SSL-SaN} } & 
\hspace{-0.45cm} \includegraphics[align=c, width=1.45cm]{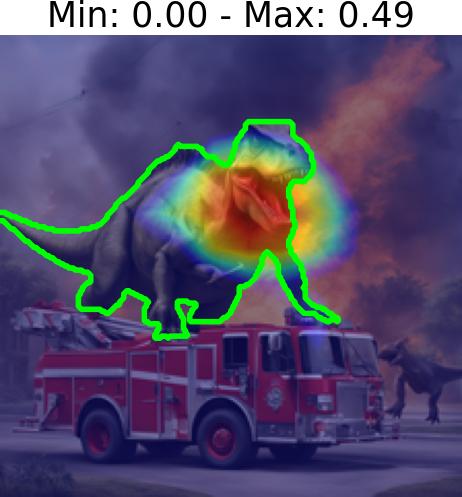}&
\hspace{-0.45cm} \includegraphics[align=c, width=1.45cm]{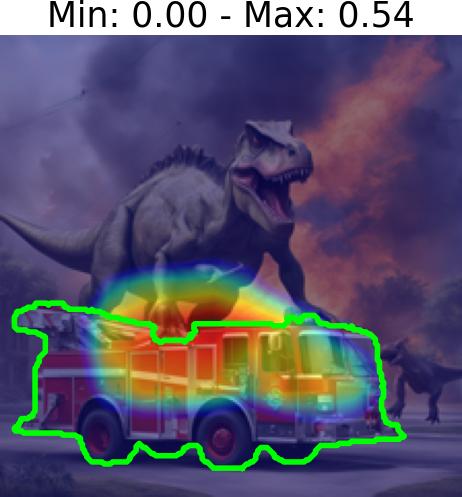}&
\hspace{-0.45cm} \includegraphics[align=c, width=1.45cm]{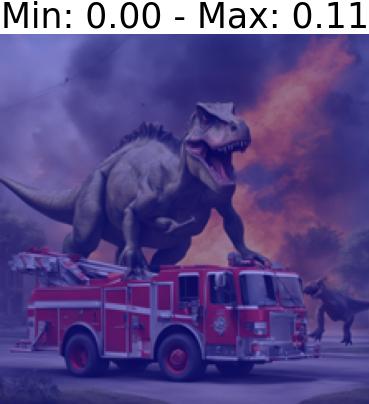}&
\hspace{-0.45cm} \includegraphics[align=c, width=1.45cm]{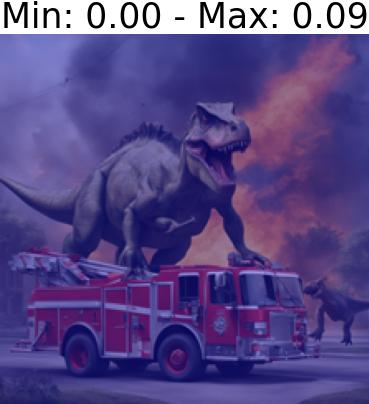}&
\hspace{-0.45cm} \includegraphics[align=c, width=1.45cm]{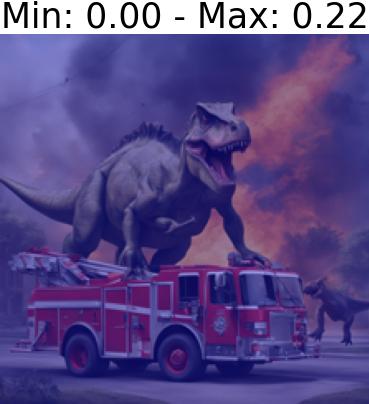} 
\end{tabular}
\end{center}
\vspace{-0.1cm}
\caption{Localization results of  different models in both \textit{positive} and \textit{negative} audio samples in the IS3+ test set. The audio-visual similarities below the universal threshold of each model are clipped, normalized to the interval $[0, 1]$, and overlaid with the original image. The boundaries of the ground-truth mask are shown in green. The minimum and maximum values of the audio-visual similarities are reported on top of each image.}
\label{fig:inferences_main}
\end{figure}

\noindent \textbf{Qualitative results.} We show in Fig.  \ref{fig:inferences_main} the localization maps of the self-supervised models for an example of IS3+. The similarity maps are filtered using the univeral threshold  \cite{juanola2024critical}. Unlike previous models, which fail to correctly localize both sound sources (dinosaur and firetruck) while filtering out silence, noise, and offscreen sounds, our approach successfully achieves both. More qualitative results (including failure cases and cross-modal retrieval results) are in the Supplementary material. 

\subsection{Ablation}
\label{sec:ablations}

First, we study the contribution of the new negative pairs with \textit{silence} and \textit{noise} as well as the loss terms $\mathcal{L}_S$ and $\mathcal{L}_N$.
Table \ref{tab:ablation_sn_lsln} shows this study  on the \new{AVS-Bench S4} test set. The best metrics, except for the case of offscreen sounds \new{and separability}, are obtained when considering silence, noise and $\mathcal{L}_S + \mathcal{L}_N$ during training. The best result for offscreen is achieved when considering only silence and $\mathcal{L}_S$, but this comes at the cost of  worsening the results on positive sounds, silence and noise. \new{The best result for separability is in the model trained with noise, but not with $\mathcal{L}_N$. The second best result for separability is obtained with the model using silence, noise and  $\mathcal{L}_S + \mathcal{L}_N$. The results of this ablation on all test sets is present in the Supplementary material.} 
Finally, Table \ref{tab:ablation_lambda} in the Supplementary material shows an ablation study  of the weight $\lambda_{SN}$ that  multiplies the new loss term $(\mathcal{L}_S + \mathcal{L}_N)$. Best results are achieved for $\lambda_{SN}=1$.

    \begin{table}[ht]
    \begin{center}
    \resizebox{0.7\columnwidth}{!}{
    \begin{tabular}{@{}c|lcccccclcccc@{}}\hline
    & &  & & & & & \multicolumn{3}{c}{{\bf Negative audio input}} &\\
    \cline{8-10}
    {\bf Test set} & {\bf S} & {\bf N} & {\bf $\mathcal{L}_S$} & {\bf $\mathcal{L}_N$} & & {\bf cIoU$_{Uth}$ $\uparrow$} & {\bf pIA$_{\text{S}}$ $\downarrow$} & {\bf pIA$_{\text{N}}$ $\downarrow$}& {\bf pIA$_{\text{O}}$ $\downarrow$}& {\bf F$_{\text{LOC}}$ $\uparrow$}& {\bf Sep $\uparrow$} \\
    \hline

    \multirow{7}{*}{\STAB{\rotatebox[origin=c]{90}{\textbf{AVS-Bench S4}}}}
        &  
        &  
        &  
        &  
        &
        & \small 31.57
        & \small 2.53
        & \small 2.49
        & \small 1.00
        & \small 47.76 
        & \small 0.2384 \\

        & \ding{51} 
        & 
        & 
        & 
        &
        & \small 31.76
        & \small 2.39
        & \small 1.85
        & \small \underline{0.82}
        & \small 48.01 
        & \small 0.2350 \\

        & \ding{51} 
        &  
        & \ding{51} 
        &  
        &
        & \small \underline{32.49}
        & \small 2.54
        & \small 2.58
        & \small \textbf{0.81}
        & \small \underline{48.80} 
        & \small 0.2458 \\

        &  
        & \ding{51} 
        &  
        &  
        &
        & \small 31.97
        & \small 2.26
        & \small 2.59
        & \small 0.88
        & \small 48.22 
        & \small \textbf{0.2558} \\

        &  
        & \ding{51} 
        & 
        & \ding{51} 
        &
        & \small 32.07
        & \small 2.21
        & \small 2.26
        & \small 1.07
        & \small 48.34 
        & \small 0.2375 \\

        & \ding{51} 
        & \ding{51} 
        &  
        &  
        &
        & \small 32.05
        & \small \underline{1.75}
        & \small \underline{1.01}
        & \small 0.84
        & \small 48.40 
        & \small 0.2393 \\

        & \ding{51} 
        & \ding{51} 
        & \ding{51} 
        & \ding{51} 
        &
        & \small \textbf{32.76}
        & \small \textbf{0.05}
        & \small \textbf{0.00}
        & \small 0.95
        & \small \textbf{49.31} 
        & \small \underline{0.2479} \\
    
    \hline
    \end{tabular}%
    }
    \vspace{-0.1cm}
    \end{center}
    \caption{Ablation table on the use of the silence (S) and  noise (N) samples during training, and the new loss terms ($\mathcal{L}_S$ and $\mathcal{L}_N$). Best results in \textbf{bold}, second best \underline{underlined}.}
    \label{tab:ablation_sn_lsln}
    \end{table}

\section{Conclusion}
\label{sec:conclusion}
We present SSL-SaN, a simple yet effective approach that leverages silence and noise audio samples to enhance cross-modal retrieval, sound localization and  robustness to negative audio samples. Additionally, we present IS3+, an improved version of IS3 that corrects mismatched image-audio pairs, improving  evaluation reliability. 
Our model is comprehensively evaluated on four benchmark datasets using positive, negative, and global metrics. We further propose a new metric to quantify cross-modal alignment and feature separability, which also predicts sound localization and retrieval performance by capturing how well positive audio-visual pairs are distinguished from negative ones. Our approach outperforms recent state-of-the-art self-supervised models on most metrics and datasets.

\clearpage
\newpage
\section*{Acknowledgements}
\label{sec:acknowledgements}
The authors acknowledge support by Maria de Maeztu project ref. CEX2021-001195-M/AEI /10.13039/501100011033. X. J. has received financial support through FPI scholarship PRE2022-101321. G. H. has received financial support through Fulbright Program and Ministerio de Universidades (Spain) funding for mobility stays of professors and researchers in foreign higher education and research centers. M. F. has received support from the National Science Foundation under Grant Number 2152119. This work was supported in part through the NYU IT High Performance Computing resources, services, and staff expertise.

\bibliography{bibliography}

\begin{thebibliography}{69}
\providecommand{\natexlab}[1]{#1}
\providecommand{\url}[1]{\texttt{#1}}
\expandafter\ifx\csname urlstyle\endcsname\relax
  \providecommand{\doi}[1]{doi: #1}\else
  \providecommand{\doi}{doi: \begingroup \urlstyle{rm}\Url}\fi

\bibitem[{Adobe}(2023)]{adobe_sfx}
{Adobe}.
\newblock Adobe audition sound effects, 2023.
\newblock URL \url{https://www.adobe.com/products/audition/offers/adobeauditiondlcsfx.html}.
\newblock Accessed: [25-Jan-2025].

\bibitem[Afouras et~al.(2020)Afouras, Owens, Chung, and Zisserman]{afouras2020self}
Triantafyllos Afouras, Andrew Owens, Joon~Son Chung, and Andrew Zisserman.
\newblock Self-supervised learning of audio-visual objects from video.
\newblock In \emph{European Conference on Computer Vision}, pages 208--224. Springer, 2020.

\bibitem[Arandjelovic and Zisserman(2017)]{arandjelovic2017look}
Relja Arandjelovic and Andrew Zisserman.
\newblock Look, listen and learn.
\newblock In \emph{IEEE International Conference on Computer Vision}, pages 609--617, 2017.

\bibitem[Arandjelovic and Zisserman(2018)]{arandjelovic2018objects}
Relja Arandjelovic and Andrew Zisserman.
\newblock Objects that sound.
\newblock In \emph{European Conference on Computer Vision}, pages 435--451, 2018.

\bibitem[Chen et~al.(2020{\natexlab{a}})Chen, Xie, Vedaldi, and Zisserman]{chen2020vggsound}
Honglie Chen, Weidi Xie, Andrea Vedaldi, and Andrew Zisserman.
\newblock Vggsound: A large-scale audio-visual dataset.
\newblock In \emph{IEEE International Conference on Acoustics, Speech and Signal Processing}, pages 721--725, 2020{\natexlab{a}}.

\bibitem[Chen et~al.(2021)Chen, Xie, Afouras, Nagrani, Vedaldi, and Zisserman]{LVS}
Honglie Chen, Weidi Xie, Triantafyllos Afouras, Arsha Nagrani, Andrea Vedaldi, and Andrew Zisserman.
\newblock Localizing visual sounds the hard way.
\newblock In \emph{IEEE/CVF conference on computer vision and pattern recognition}, pages 16867--16876, 2021.

\bibitem[Chen et~al.(2020{\natexlab{b}})Chen, Kornblith, Norouzi, and Hinton]{chen2020simple}
Ting Chen, Simon Kornblith, Mohammad Norouzi, and Geoffrey Hinton.
\newblock A simple framework for contrastive learning of visual representations.
\newblock In \emph{International Conference on Machine Learning}, pages 1597--1607. PMLR, 2020{\natexlab{b}}.

\bibitem[Chen and He(2021)]{chen2021exploring}
Xinlei Chen and Kaiming He.
\newblock Exploring simple siamese representation learning.
\newblock In \emph{IEEE/CVF conference on computer vision and pattern recognition}, pages 15750--15758, 2021.

\bibitem[Cheng et~al.(2024)Cheng, Wang, Zheng, Chen, Huang, Zhang, Chen, and Li]{cheng2024integrating}
Luyao Cheng, Hui Wang, Siqi Zheng, Yafeng Chen, Rongjie Huang, Qinglin Zhang, Qian Chen, and Xihao Li.
\newblock Integrating audio, visual, and semantic information for enhanced multimodal speaker diarization.
\newblock \emph{arXiv preprint arXiv:2408.12102}, 2024.

\bibitem[Chopra et~al.(2005)Chopra, Hadsell, and LeCun]{chopra2005learning}
Sumit Chopra, Raia Hadsell, and Yann LeCun.
\newblock Learning a similarity metric discriminatively, with application to face verification.
\newblock In \emph{IEEE/CVF Conference on Computer Vision and Pattern Recognition}, volume~1, pages 539--546, 2005.

\bibitem[Deng et~al.(2009)Deng, Dong, Socher, Li, Li, and Fei-Fei]{deng2009imagenet}
Jia Deng, Wei Dong, Richard Socher, Li-Jia Li, Kai Li, and Li~Fei-Fei.
\newblock Imagenet: A large-scale hierarchical image database.
\newblock In \emph{IEEE/CVF Conference on Computer Vision and Pattern Recognition}, pages 248--255, 2009.

\bibitem[Du et~al.(2023)Du, Chen, Salamon, Russell, and Owens]{du2023conditional}
Yuexi Du, Ziyang Chen, Justin Salamon, Bryan Russell, and Andrew Owens.
\newblock Conditional generation of audio from video via foley analogies.
\newblock In \emph{IEEE/CVF Conference on Computer Vision and Pattern Recognition}, pages 2426--2436, 2023.

\bibitem[Eliav and Gannot(2024)]{eliav2024audio}
Amit Eliav and Sharon Gannot.
\newblock Audio-visual approach for multimodal concurrent speaker detection.
\newblock \emph{arXiv preprint arXiv:2407.01774}, 2024.

\bibitem[Fichna et~al.(2021)Fichna, Biberger, Seeber, and Ewert]{fichna2021effect}
Stefan Fichna, Thomas Biberger, Bernhard~U Seeber, and Stephan~D Ewert.
\newblock Effect of acoustic scene complexity and visual scene representation on auditory perception in virtual audio-visual environments.
\newblock In \emph{2021 Immersive and 3D Audio: from Architecture to Automotive (I3DA)}, pages 1--9. IEEE, 2021.

\bibitem[Fisher~III et~al.(2000)Fisher~III, Darrell, Freeman, and Viola]{fisher2000learning}
John~W Fisher~III, Trevor Darrell, William Freeman, and Paul Viola.
\newblock Learning joint statistical models for audio-visual fusion and segregation.
\newblock \emph{Advances in neural information processing systems}, 13, 2000.

\bibitem[Gao and Grauman(2021)]{gao2021visualvoice}
Ruohan Gao and Kristen Grauman.
\newblock Visualvoice: Audio-visual speech separation with cross-modal consistency.
\newblock In \emph{IEEE/CVF Conference on Computer Vision and Pattern Recognition}, pages 15490--15500. IEEE, 2021.

\bibitem[Goel et~al.(2022)Goel, Bansal, Bhatia, Rossi, Vinay, and Grover]{goel2022cyclip}
Shashank Goel, Hritik Bansal, Sumit Bhatia, Ryan Rossi, Vishwa Vinay, and Aditya Grover.
\newblock Cyclip: Cyclic contrastive language-image pretraining.
\newblock \emph{Advances in Neural Information Processing Systems}, 35:\penalty0 6704--6719, 2022.

\bibitem[Grill et~al.(2020)Grill, Strub, Altch{\'e}, Tallec, Richemond, Buchatskaya, Doersch, Avila~Pires, Guo, Gheshlaghi~Azar, et~al.]{grill2020bootstrap}
Jean-Bastien Grill, Florian Strub, Florent Altch{\'e}, Corentin Tallec, Pierre Richemond, Elena Buchatskaya, Carl Doersch, Bernardo Avila~Pires, Zhaohan Guo, Mohammad Gheshlaghi~Azar, et~al.
\newblock Bootstrap your own latent-a new approach to self-supervised learning.
\newblock \emph{Advances in neural information processing systems}, 33:\penalty0 21271--21284, 2020.

\bibitem[Hamilton et~al.(2024)Hamilton, Zisserman, Hershey, and Freeman]{hamilton2024separating}
Mark Hamilton, Andrew Zisserman, John~R Hershey, and William~T Freeman.
\newblock Separating the" chirp" from the" chat": Self-supervised visual grounding of sound and language.
\newblock In \emph{IEEE/CVF Conference on Computer Vision and Pattern Recognition}, pages 13117--13127, 2024.

\bibitem[Hardoon et~al.(2004)Hardoon, Szedmak, and Shawe-Taylor]{hardoon2004canonical}
David~R Hardoon, Sandor Szedmak, and John Shawe-Taylor.
\newblock Canonical correlation analysis: An overview with application to learning methods.
\newblock \emph{Neural computation}, 16\penalty0 (12):\penalty0 2639--2664, 2004.

\bibitem[He et~al.(2016)He, Zhang, Ren, and Sun]{he2016deep}
Kaiming He, Xiangyu Zhang, Shaoqing Ren, and Jian Sun.
\newblock Deep residual learning for image recognition.
\newblock In \emph{IEEE/CVF conference on computer vision and pattern recognition}, pages 770--778, 2016.

\bibitem[He et~al.(2020)He, Fan, Wu, Xie, and Girshick]{he2020momentum}
Kaiming He, Haoqi Fan, Yuxin Wu, Saining Xie, and Ross Girshick.
\newblock Momentum contrast for unsupervised visual representation learning.
\newblock In \emph{IEEE/CVF conference on computer vision and pattern recognition}, pages 9729--9738, 2020.

\bibitem[Hershey and Movellan(1999)]{hershey1999audio}
John Hershey and Javier Movellan.
\newblock Audio vision: Using audio-visual synchrony to locate sounds.
\newblock \emph{Advances in neural information processing systems}, 12, 1999.

\bibitem[Hu et~al.(2020)Hu, Qian, Jiang, Tan, Wen, Ding, Lin, and Dou]{hu2020discriminative}
Di~Hu, Rui Qian, Minyue Jiang, Xiao Tan, Shilei Wen, Errui Ding, Weiyao Lin, and Dejing Dou.
\newblock Discriminative sounding objects localization via self-supervised audiovisual matching.
\newblock \emph{Advances in Neural Information Processing Systems}, 33:\penalty0 10077--10087, 2020.

\bibitem[Hu et~al.(2022)Hu, Chen, and Owens]{hu2022mix}
Xixi Hu, Ziyang Chen, and Andrew Owens.
\newblock Mix and localize: Localizing sound sources in mixtures.
\newblock In \emph{IEEE/CVF Conference on Computer Vision and Pattern Recognition}, pages 10483--10492, 2022.

\bibitem[Jamaludin et~al.(2019)Jamaludin, Chung, and Zisserman]{jamaludin2019you}
Amir Jamaludin, Joon~Son Chung, and Andrew Zisserman.
\newblock You said that?: Synthesising talking faces from audio.
\newblock \emph{International Journal of Computer Vision}, 127:\penalty0 1767--1779, 2019.

\bibitem[Juanola et~al.(2025)Juanola, Haro, and Fuentes]{juanola2024critical}
Xavier Juanola, Gloria Haro, and Magdalena Fuentes.
\newblock A critical assessment of visual sound source localization models including negative audio.
\newblock In \emph{IEEE International Conference on Acoustics, Speech and Signal Processing}, pages 1--5, 2025.

\bibitem[Kidron et~al.(2005)Kidron, Schechner, and Elad]{kidron2005pixels}
Einat Kidron, Yoav~Y Schechner, and Michael Elad.
\newblock Pixels that sound.
\newblock In \emph{IEEE/CVF Conference on Computer Vision and Pattern Recognition}, volume~1, pages 88--95, 2005.

\bibitem[Kim et~al.(2024)Kim, Um, Lee, and Kim]{kim2024learning}
Dongjin Kim, Sung~Jin Um, Sangmin Lee, and Jung~Uk Kim.
\newblock Learning to visually localize sound sources from mixtures without prior source knowledge.
\newblock In \emph{IEEE/CVF Conference on Computer Vision and Pattern Recognition}, pages 26467--26476, 2024.

\bibitem[Kingma(2014)]{kingma2014adam}
Diederik~P Kingma.
\newblock Adam: A method for stochastic optimization.
\newblock \emph{arXiv preprint arXiv:1412.6980}, 2014.

\bibitem[Korbar et~al.(2018)Korbar, Tran, and Torresani]{korbar2018cooperative}
Bruno Korbar, Du~Tran, and Lorenzo Torresani.
\newblock Cooperative learning of audio and video models from self-supervised synchronization.
\newblock \emph{Advances in Neural Information Processing Systems}, 31, 2018.

\bibitem[Lei et~al.(2023)Lei, Wang, Chen, Wang, Wang, and Yang]{lei2023recent}
Yinjie Lei, Zixuan Wang, Feng Chen, Guoqing Wang, Peng Wang, and Yang Yang.
\newblock Recent advances in multi-modal 3d scene understanding: A comprehensive survey and evaluation.
\newblock \emph{arXiv preprint arXiv:2310.15676}, 2023.

\bibitem[Li et~al.(2025)Li, Zhao, Huang, Guo, and Tian]{li2025audio}
Jia Li, Wenjie Zhao, Ziru Huang, Yunhui Guo, and Yapeng Tian.
\newblock Do audio-visual segmentation models truly segment sounding objects?
\newblock \emph{arXiv preprint arXiv:2502.00358}, 2025.

\bibitem[Liang et~al.(2023)Liang, Huang, Tian, Kumar, and Xu]{liang2023av}
Susan Liang, Chao Huang, Yapeng Tian, Anurag Kumar, and Chenliang Xu.
\newblock Av-nerf: Learning neural fields for real-world audio-visual scene synthesis.
\newblock \emph{Advances in Neural Information Processing Systems}, 36:\penalty0 37472--37490, 2023.

\bibitem[Liu et~al.(2022{\natexlab{a}})Liu, Ju, Xie, and Zhang]{SSLTIE}
Jinxiang Liu, Chen Ju, Weidi Xie, and Ya~Zhang.
\newblock Exploiting transformation invariance and equivariance for self-supervised sound localisation.
\newblock In \emph{Proceedings of the 30th ACM International Conference on Multimedia}, pages 3742--3753, 2022{\natexlab{a}}.

\bibitem[Liu et~al.(2022{\natexlab{b}})Liu, Qian, Zhou, Hu, Lin, Liu, Zhou, and Zhou]{liu2022visual}
Xian Liu, Rui Qian, Hang Zhou, Di~Hu, Weiyao Lin, Ziwei Liu, Bolei Zhou, and Xiaowei Zhou.
\newblock Visual sound localization in the wild by cross-modal interference erasing.
\newblock In \emph{Proceedings of the AAAI Conference on Artificial Intelligence}, volume~36, pages 1801--1809, 2022{\natexlab{b}}.

\bibitem[L{\"u}ddecke and Ecker(2022)]{luddecke2022image}
Timo L{\"u}ddecke and Alexander Ecker.
\newblock Image segmentation using text and image prompts.
\newblock In \emph{IEEE/CVF conference on computer vision and pattern recognition}, pages 7086--7096, 2022.

\bibitem[Mahmud et~al.(2024)Mahmud, Tian, and Marculescu]{mahmud2024t}
Tanvir Mahmud, Yapeng Tian, and Diana Marculescu.
\newblock T-vsl: Text-guided visual sound source localization in mixtures.
\newblock In \emph{IEEE/CVF Conference on Computer Vision and Pattern Recognition}, pages 26742--26751, 2024.

\bibitem[Mo and Morgado(2022{\natexlab{a}})]{EZ-VSL}
Shentong Mo and Pedro Morgado.
\newblock Localizing visual sounds the easy way.
\newblock In \emph{European Conference on Computer Vision}, pages 218--234. Springer, 2022{\natexlab{a}}.

\bibitem[Mo and Morgado(2022{\natexlab{b}})]{SLAVC}
Shentong Mo and Pedro Morgado.
\newblock A closer look at weakly-supervised audio-visual source localization.
\newblock \emph{Advances in Neural Information Processing Systems}, 35:\penalty0 37524--37536, 2022{\natexlab{b}}.

\bibitem[Mo and Tian(2023)]{mo2023audio}
Shentong Mo and Yapeng Tian.
\newblock Audio-visual grouping network for sound localization from mixtures.
\newblock In \emph{IEEE/CVF Conference on Computer Vision and Pattern Recognition}, pages 10565--10574, 2023.

\bibitem[Montesinos et~al.(2022)Montesinos, Kadandale, and Haro]{montesinos2022vovit}
Juan~F Montesinos, Venkatesh~S Kadandale, and Gloria Haro.
\newblock Vovit: Low latency graph-based audio-visual voice separation transformer.
\newblock In \emph{European Conference on Computer Vision}, pages 310--326. Springer, 2022.

\bibitem[Montesinos et~al.(2023)Montesinos, Michelsanti, Haro, Tan, and Jensen]{montesinos2023speech}
Juan~F Montesinos, Daniel Michelsanti, Gloria Haro, Zheng-Hua Tan, and Jesper Jensen.
\newblock Speech inpainting: Context-based speech synthesis guided by video.
\newblock In \emph{Interspeech}, pages 4459--4463, 2023.

\bibitem[Oord et~al.(2018)Oord, Li, and Vinyals]{oord2018representation}
Aaron van~den Oord, Yazhe Li, and Oriol Vinyals.
\newblock Representation learning with contrastive predictive coding.
\newblock \emph{arXiv preprint arXiv:1807.03748}, 2018.

\bibitem[Owens and Efros(2018)]{owens2018audio}
Andrew Owens and Alexei~A Efros.
\newblock Audio-visual scene analysis with self-supervised multisensory features.
\newblock In \emph{European Conference on Computer Vision}, pages 631--648, 2018.

\bibitem[Oya et~al.(2020)Oya, Iwase, Natsume, Itazuri, Yamaguchi, and Morishima]{oya2020we}
Takashi Oya, Shohei Iwase, Ryota Natsume, Takahiro Itazuri, Shugo Yamaguchi, and Shigeo Morishima.
\newblock Do we need sound for sound source localization?
\newblock In \emph{Asian Conference on Computer Vision}, 2020.

\bibitem[Park et~al.(2023)Park, Senocak, and Chung]{park2023marginnce}
Sooyoung Park, Arda Senocak, and Joon~Son Chung.
\newblock Marginnce: Robust sound localization with a negative margin.
\newblock In \emph{IEEE International Conference on Acoustics, Speech and Signal Processing}, pages 1--5, 2023.

\bibitem[Park et~al.(2024)Park, Senocak, and Chung]{park2024can}
Sooyoung Park, Arda Senocak, and Joon~Son Chung.
\newblock Can clip help sound source localization?
\newblock In \emph{IEEE/CVF Winter Conference on Applications of Computer Vision}, pages 5711--5720, 2024.

\bibitem[Qian et~al.(2020)Qian, Hu, Dinkel, Wu, Xu, and Lin]{qian2020multiple}
Rui Qian, Di~Hu, Heinrich Dinkel, Mengyue Wu, Ning Xu, and Weiyao Lin.
\newblock Multiple sound sources localization from coarse to fine.
\newblock In \emph{European Conference on Computer Vision}, pages 292--308. Springer, 2020.

\bibitem[Ramaswamy and Das(2020)]{ramaswamy2020see}
Janani Ramaswamy and Sukhendu Das.
\newblock See the sound, hear the pixels.
\newblock In \emph{IEEE/CVF winter conference on applications of computer vision}, pages 2970--2979, 2020.

\bibitem[Risoud et~al.(2018)Risoud, Hanson, Gauvrit, Renard, Lemesre, Bonne, and Vincent]{risoud2018sound}
Michael Risoud, J-N Hanson, Fanny Gauvrit, Christian Renard, P-E Lemesre, N-X Bonne, and Christophe Vincent.
\newblock Sound source localization.
\newblock \emph{European annals of otorhinolaryngology, head and neck diseases}, 135\penalty0 (4):\penalty0 259--264, 2018.

\bibitem[Rombach et~al.(2022)Rombach, Blattmann, Lorenz, Esser, and Ommer]{rombach2022high}
Robin Rombach, Andreas Blattmann, Dominik Lorenz, Patrick Esser, and Bj{\"o}rn Ommer.
\newblock High-resolution image synthesis with latent diffusion models.
\newblock In \emph{IEEE/CVF Conference on Computer Vision and Pattern Recognition}, pages 10684--10695, 2022.

\bibitem[Salas-C{\'a}ceres et~al.(2024)Salas-C{\'a}ceres, Lorenzo-Navarro, Freire-Obreg{\'o}n, and Castrill{\'o}n-Santana]{salas2024multimodal}
Jos{\'e} Salas-C{\'a}ceres, Javier Lorenzo-Navarro, David Freire-Obreg{\'o}n, and Modesto Castrill{\'o}n-Santana.
\newblock Multimodal emotion recognition based on a fusion of audiovisual information with temporal dynamics.
\newblock \emph{Multimedia Tools and Applications}, pages 1--17, 2024.

\bibitem[Senocak et~al.(2018)Senocak, Oh, Kim, Yang, and Kweon]{senocak2018learning}
Arda Senocak, Tae-Hyun Oh, Junsik Kim, Ming-Hsuan Yang, and In~So Kweon.
\newblock Learning to localize sound source in visual scenes.
\newblock In \emph{IEEE/CVF Conference on Computer Vision and Pattern Recognition}, pages 4358--4366, 2018.

\bibitem[Senocak et~al.(2019)Senocak, Oh, Kim, Yang, and Kweon]{senocak2019learning}
Arda Senocak, Tae-Hyun Oh, Junsik Kim, Ming-Hsuan Yang, and In~So Kweon.
\newblock Learning to localize sound sources in visual scenes: Analysis and applications.
\newblock \emph{IEEE transactions on pattern analysis and machine intelligence}, 43\penalty0 (5):\penalty0 1605--1619, 2019.

\bibitem[Senocak et~al.(2022{\natexlab{a}})Senocak, Ryu, Kim, and Kweon]{senocak2022learning}
Arda Senocak, Hyeonggon Ryu, Junsik Kim, and In~So Kweon.
\newblock Learning sound localization better from semantically similar samples.
\newblock In \emph{IEEE International Conference on Acoustics, Speech and Signal Processing}, pages 4863--4867, 2022{\natexlab{a}}.

\bibitem[Senocak et~al.(2022{\natexlab{b}})Senocak, Ryu, Kim, and Kweon]{senocak2022less}
Arda Senocak, Hyeonggon Ryu, Junsik Kim, and In~So Kweon.
\newblock Less can be more: Sound source localization with a classification model.
\newblock In \emph{IEEE/CVF conference on computer vision and pattern recognition}, pages 3308--3317, 2022{\natexlab{b}}.

\bibitem[Senocak et~al.(2024)Senocak, Ryu, Kim, Oh, Pfister, and Chung]{senocak2024aligning}
Arda Senocak, Hyeonggon Ryu, Junsik Kim, Tae-Hyun Oh, Hanspeter Pfister, and Joon~Son Chung.
\newblock Aligning sight and sound: Advanced sound source localization through audio-visual alignment.
\newblock \emph{arXiv preprint arXiv:2407.13676}, 2024.

\bibitem[Shi(2021)]{shi2021survey}
Zhaofeng Shi.
\newblock A survey on audio synthesis and audio-visual multimodal processing.
\newblock \emph{arXiv preprint arXiv:2108.00443}, 2021.

\bibitem[Song et~al.(2024)Song, Zhang, Wang, Fan, and Zhang]{song2024enhancing}
Zengjie Song, Jiangshe Zhang, Yuxi Wang, Junsong Fan, and Zhaoxiang Zhang.
\newblock Enhancing sound source localization via false negative elimination.
\newblock \emph{IEEE Transactions on Pattern Analysis and Machine Intelligence}, 2024.

\bibitem[Sun et~al.(2023)Sun, Zhang, Wang, Liu, Zhong, Feng, Guo, Zhang, and Barnes]{FNAC}
Weixuan Sun, Jiayi Zhang, Jianyuan Wang, Zheyuan Liu, Yiran Zhong, Tianpeng Feng, Yandong Guo, Yanhao Zhang, and Nick Barnes.
\newblock Learning audio-visual source localization via false negative aware contrastive learning.
\newblock In \emph{IEEE/CVF Conference on Computer Vision and Pattern Recognition}, pages 6420--6429, 2023.

\bibitem[Sung-Bin et~al.(2023)Sung-Bin, Senocak, Ha, Owens, and Oh]{sung2023sound}
Kim Sung-Bin, Arda Senocak, Hyunwoo Ha, Andrew Owens, and Tae-Hyun Oh.
\newblock Sound to visual scene generation by audio-to-visual latent alignment.
\newblock In \emph{IEEE/CVF Conference on Computer Vision and Pattern Recognition}, pages 6430--6440, 2023.

\bibitem[Sung-Bin et~al.(2024)Sung-Bin, Senocak, Ha, and Oh]{sung2024sound2vision}
Kim Sung-Bin, Arda Senocak, Hyunwoo Ha, and Tae-Hyun Oh.
\newblock Sound2vision: Generating diverse visuals from audio through cross-modal latent alignment.
\newblock \emph{arXiv preprint arXiv:2412.06209}, 2024.

\bibitem[Tian et~al.(2018)Tian, Shi, Li, Duan, and Xu]{tian2018audio}
Yapeng Tian, Jing Shi, Bochen Li, Zhiyao Duan, and Chenliang Xu.
\newblock Audio-visual event localization in unconstrained videos.
\newblock In \emph{European Conference on Computer Vision}, pages 247--263, 2018.

\bibitem[Wu et~al.(2020)Wu, Lin, Cohen, Bui, and Maji]{wu2020phrasecut}
Chenyun Wu, Zhe Lin, Scott Cohen, Trung Bui, and Subhransu Maji.
\newblock Phrasecut: Language-based image segmentation in the wild.
\newblock In \emph{IEEE/CVF Conference on Computer Vision and Pattern Recognition}, pages 10216--10225, 2020.

\bibitem[Wu et~al.(2022)Wu, Fuentes, Seetharaman, and Bello]{RCGrad}
Ho-Hsiang Wu, Magdalena Fuentes, Prem Seetharaman, and Juan~Pablo Bello.
\newblock How to listen? rethinking visual sound localization.
\newblock \emph{Interspeech}, 2022.

\bibitem[Yaghoubi et~al.(2023)Yaghoubi, Kelm, Gerkmann, and Frintrop]{yaghoubi2023acoustic}
Ehsan Yaghoubi, Andre~Peter Kelm, Timo Gerkmann, and Simone Frintrop.
\newblock Acoustic and visual knowledge distillation for contrastive audio-visual localization.
\newblock In \emph{Proceedings of the 25th International Conference on Multimodal Interaction}, pages 15--23, 2023.

\bibitem[Zhang et~al.(2023)Zhang, HaoChen, Huang, Wang, Zou, and Yeung]{zhang2023diagnosing}
Yuhui Zhang, Jeff~Z HaoChen, Shih-Cheng Huang, Kuan-Chieh Wang, James Zou, and Serena Yeung.
\newblock Diagnosing and rectifying vision models using language.
\newblock \emph{arXiv preprint arXiv:2302.04269}, 2023.

\bibitem[Zhou et~al.(2024)Zhou, Shen, Wang, Zhang, Sun, Zhang, Birchfield, Guo, Kong, Wang, and Zhong]{zhou2024avss}
Jinxing Zhou, Xuyang Shen, Jianyuan Wang, Jiayi Zhang, Weixuan Sun, Jing Zhang, Stan Birchfield, Dan Guo, Lingpeng Kong, Meng Wang, and Yiran Zhong.
\newblock Audio-visual segmentation with semantics.
\newblock \emph{International Journal of Computer Vision}, pages 1--21, 2024.

\end{thebibliography}

\clearpage
\appendix

\renewcommand{\thefigure}{A.\arabic{figure}}
\renewcommand{\thetable}{A.\arabic{table}}
\renewcommand{\theequation}{A.\arabic{equation}}
\setcounter{figure}{0}
\setcounter{table}{0}
\setcounter{equation}{0}

\section{Supplementary material}

\subsection{Architecture}
\label{sup:architecture}
\new{

Our model, as well as all of the models presented and compared to ours (except for ACL \cite{park2024can} that uses transformers) use ResNet18 as the backbone for both the audio and image encoders. ResNet18 is composed of an initial convolutional layer followed by four residual stages, each made up of two \textit{BasicBlocks}. Each BasicBlock introduces residual connections that help mitigate the vanishing gradient problem in deep neural networks, enabling more effective training of deeper models.

\subsubsection{Input Branches.} In our implementation, we include two separate input branches to handle different modalities:
\begin{itemize}
    \item \texttt{conv1\_a}: used for audio inputs (1 channel)
    \item \texttt{conv1}: used for RGB images (3 channels)
\end{itemize}
Each of these branches consists of a $7{\times}7$ convolutional layer with stride 2 and padding 3, followed by batch normalization and ReLU activation. After this initial stage, all inputs share the same residual backbone.

\subsubsection{ResNet18 Structure.} 
The full ResNet18 architecture consists of:
\begin{itemize}
    \item Initial convolution + batch norm + ReLU
    \item $3{\times}3$ max pooling with stride 2
    \item \textbf{Layer 1:} Two BasicBlocks with 64 filters
    \item \textbf{Layer 2:} Two BasicBlocks with 128 filters (first block includes downsampling)
    \item \textbf{Layer 3:} Two BasicBlocks with 256 filters (first block includes downsampling)
    \item \textbf{Layer 4:} Two BasicBlocks with 512 filters (first block includes downsampling)
    \item Adaptive average pooling
    \item Fully connected classification layer (removed in our model)
\end{itemize}

\subsubsection{BasicBlock.} The \textit{BasicBlock} is the fundamental building unit of ResNet18. Each BasicBlock contains:
\begin{itemize}
    \item A $3{\times}3$ convolution, followed by batch normalization and ReLU
    \item A second $3{\times}3$ convolution, followed by batch normalization
    \item A residual (skip) connection that adds the input of the block to its output
\end{itemize}

If the input and output dimensions differ (e.g., due to a change in the number of channels or stride), a parallel \texttt{downsample} path is introduced using a $1{\times}1$ convolution and batch normalization to match dimensions before the addition.

The output of the BasicBlock is computed as:
\[
\text{Output} = \text{ReLU}(\text{BN}_2(\text{Conv}_2(\text{ReLU}(\text{BN}_1(\text{Conv}_1(x)))) + \text{Residual}(x))
\]
This structure allows gradients to propagate more easily during training and facilitates the learning of identity mappings when needed.

\subsubsection{Adaptation.} In our model variants (e.g., SSL-SaN), the final fully connected layer is removed, and the ResNet18 backbone serves as a feature extractor. 

}
\subsection{Computational Efficiency and Resource Requirements}
\label{sup:computational_efficiency}

\subsubsection{Training Cost}

We report training time statistics for our proposed method (SSL-SaN) and compare them with SSL-TIE (backbone of our model).

\begin{table}[h]
\centering
\resizebox{\textwidth}{!}{%
\begin{tabular}{l|ccccc|c}
\hline
\textbf{Method} & \textbf{Training Epoch} & \textbf{Validation Epoch} & \textbf{Retrieval Phase} & \textbf{Epochs}  & \textbf{Total (Epoch)}  & \textbf{Total} \\
\hline
SSL-TIE  & 970.94 s. (16.18 min)  & 19.46 s. (0.32 min) & 995.88 s. (16.60 min)  & 100 & 1986.28 s. (33.10 min) & 55.17 h (\textbf{2.30 day})\\
\textbf{Ours $\rightarrow$ SSL-SaN}  & 2093.79 s. (34.90 min) & 22.14 s. (0.37 min) & 1029.84 s. (17.16 min) & 120 & 3145.77 s. (52.43 min) & 104.86 h (\textbf{4.37 day})\\
\hline
\end{tabular}%
}
\vspace{0.01cm}
\caption{Training cost comparison including number of training epochs}
\label{tab:training_cost}
\end{table}
The increase in training time seen in Table \ref{tab:training_cost} is due to the two additional audio samples (silence and noise) that the model processes for each image-audio pair, as well as the extra loss computations involved.

\subsubsection{Inference Time}

The average inference time per sample is reported in the Table  \ref{tab:inference_time}.

\begin{table}[h]
\centering
\begin{tabular}{l|c}
\hline
\textbf{Method} & \textbf{Inference Time (seconds)} \\
\hline
LVS     & 0.42 \\
EZ-VSL     & 0.27 \\
FNAC       & 0.31 \\
SLAVC      & 0.32 \\
SSL-TIE    & 0.83 \\
SSL-Align  & 0.45 \\
ACL     & 0.47 \\
\textbf{Ours $\rightarrow$ SSL-SaN}    & 0.82 \\
\hline
\end{tabular}
\vspace{0.2cm}
\caption{Average inference time per sample}
\label{tab:inference_time}
\end{table}

\subsubsection{Resource Requirements}

The number of parameters (in millions) for image and audio encoders, and the full models are listed in the Table \ref{tab:resource_requirements}. Notice how ACL, which is based on transformer encoders and decoder, has significantly much more parameters than the rest of models, that are based on ResNet18 encoders.

\begin{table}[h]
\centering
\begin{tabular}{l|c|c|c}
\hline
\textbf{Method} & \textbf{Image Encoder} & \textbf{Audio Encoder} & \textbf{Full Model} \\
\hline
LVS     & 11.18 M & 11.17 M & 23.95 M \\
EZ-VSL     & 11.18 M & 11.17 M & 22.87 M \\
FNAC       & 11.18 M & 11.17 M & 22.87 M \\
SLAVC      & 11.18 M & 11.17 M & 46.79 M \\
SSL-TIE    & 11.18 M & 11.17 M & 23.95 M \\
SSL-Align  & 11.18 M & 11.17 M & 23.95 M \\
ACL        & 85.05 M & 89.79 M & 248.34 M \\
\textbf{Ours $\rightarrow$ SSL-SaN}    & 11.18 M & 11.17 M & 23.95 M \\
\hline
\end{tabular}
\vspace{0.2cm}
\caption{Resource requirements (number of parameters in millions)}
\label{tab:resource_requirements}
\end{table}
\subsection{Data augmentations}
\label{sec:data_augmentations}

Following \cite{SSLTIE}, we apply augmentations to both audio and image and an additional loss term, during the training process, namely: \\
\textbf{Spectrogram masking}. The Mel Spectrograms are randomly replaced with zeros along the two axes with random widths (time and frequency masking). \\
\textbf{Appearence transformations $(\mathcal{T}_{\text{app}})$}. The appearance of images is changed with some random transformations: color jittering, gaussian blur, and grayscale. \\
\textbf{Geometrical transformations $(\mathcal{T}_{\text{geo}})$}. Images are transformed with geometrical transformations such as  crop, resize, rotation, and horizontal flip. The visual features extracted from the geometrically-transformed $j$-th image are denoted as $\mathbf{v}^{\mathcal{T}_{geo}}_j$. 

Ideally, the similarity map of the geometrically-transformed image with its corresponding $j$-th audio, should match the geometrically-transformed similarity map of the non-transformed image with the same audio. 
The loss term that expresses this geometrical equivariance is:
\begin{equation}
\label{sup_eq:L_geo}
    \mathcal{L}_{\text{geo}} = \left\lVert S \left( \mathbf{a}_j, \mathbf{v}^{\mathcal{T}_{geo}}_j \right) - \mathcal{T}_{geo} \left( S \left( \mathbf{a}_j, \mathbf{v}_j \right) \right) \right\rVert _2^2
\end{equation}

\noindent\textbf{Similar Audio}. To identify similar audio samples to the target audio, the auadio encoder of the VSSL model is used. 
At each training epoch, the similarity between the embeddings of every audio sample and all other audio samples is measured using a suitable similarity metric (e.g., cosine similarity). For each audio waveform $W_{f_{i}}$, the audio with the highest similarity score, denoted as $W_{f_{i}}^{sim}$, is selected as the most similar audio.

\noindent\textbf{Similar Audio Mixing (SAM)}. After identifying the most similar audio, we apply Similar Audio Mixing (SAM). SAM combines the original audio waveform $W_{f_{i}}$ with its most similar audio waveform $W_{f_{i}}^{sim}$ using a mixing coefficient $\alpha$. This mixing is formulated as follows:

\begin{equation}
    \label{sup_eq:W_SAM}
    W^{\text{SAM}}_{f_i} = \left( 1 -\alpha \right) W_{f_i} + \alpha W^{\text{sim}}_{f_i},
\end{equation}
where the the mixing coefficient $\alpha$ starts at 0, preserving the original audio characteristics. As training progresses through epochs, $\alpha$ gradually increases after each epoch, reaching a maximum value of 0.5 at epoch 50; thus increasingly emphasizing the contribution of the similar audio. After the first epoch, the original audio is completely replaced by the mixed audio $W_{f_{i}}^{SAM}$ for subsequent training, thereby progressively leveraging similar audio characteristics to enhance model learning.
\subsection{IS3+}
\label{sup:is3+}
The IS3+ dataset is publicly available in  project page:  \href{https://xavijuanola.github.io/SSL-SaN/}{https://xavijuanola.github.io/SSL-SaN/}.

\subsubsection{IS3+ data curation}
\label{ap:wrong_class_examples}

\begin{figure}[ht]
    \centering

    \begin{minipage}{0.3\linewidth}
        \centering
        \includegraphics[width=\linewidth]{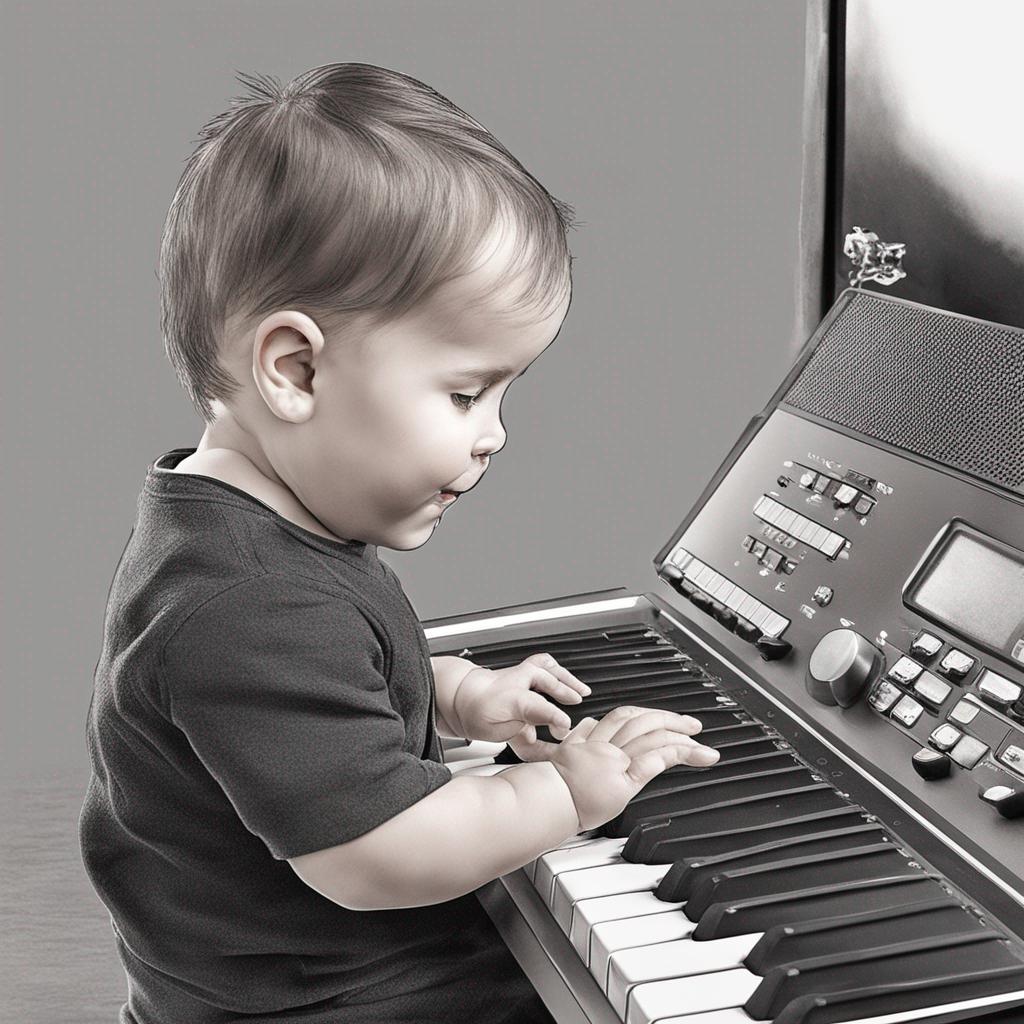}
        \caption*{Image of ``baby'' and ``computer keyboard''. ``computer keyboard" was replaced by "electronic organ''}
    \end{minipage}
    \hfill
    \begin{minipage}{0.3\linewidth}
        \centering
        \includegraphics[width=\linewidth]{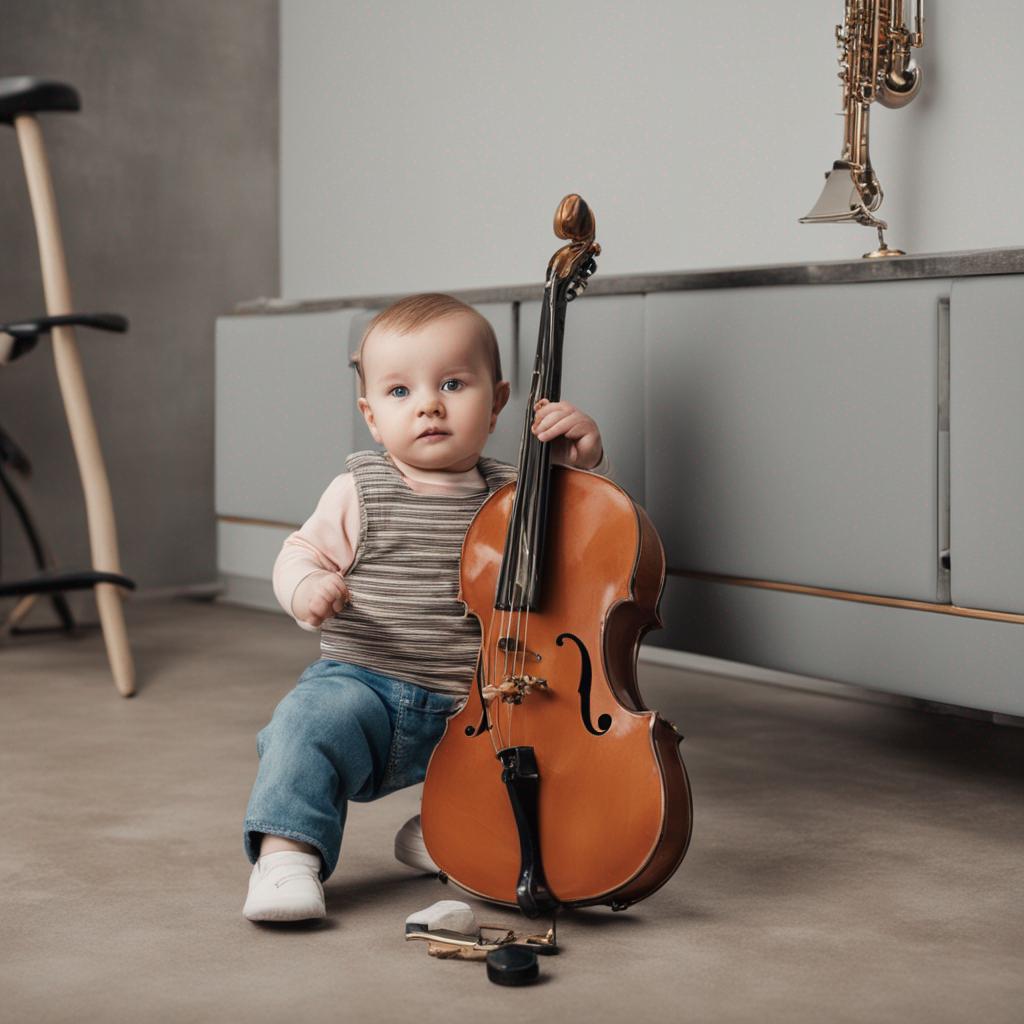}
        \caption*{Image of ``baby'' and ``saxophone''. ``saxophone'' was replaced by ``cello''.}
    \end{minipage}
    \hfill
    \begin{minipage}{0.3\linewidth}
        \centering
        \includegraphics[width=\linewidth]{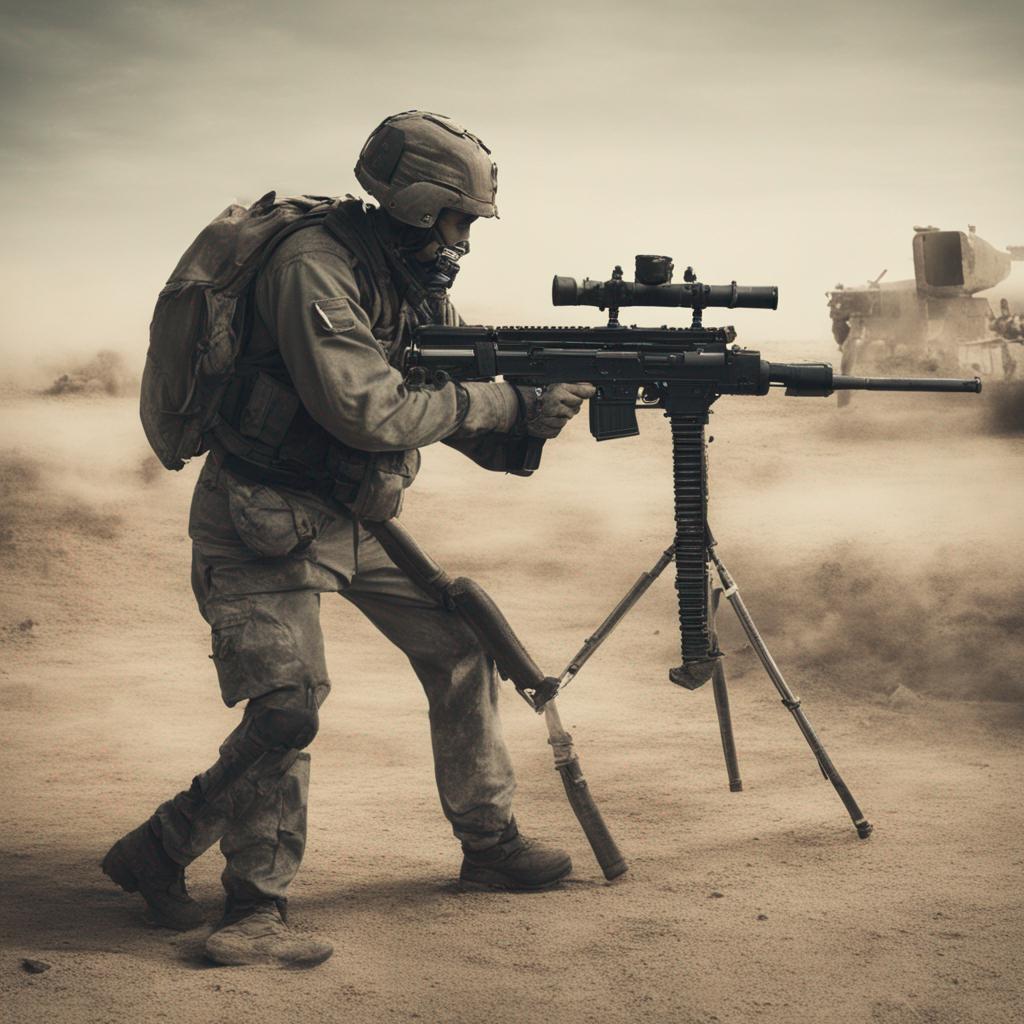}
        \caption*{Image of ``machine gun'' and ``people''. ``people" was replaced by ``male voice''}
    \end{minipage}

    \vskip 1em

    \begin{minipage}{0.45\linewidth}
        \centering
        \includegraphics[width=\linewidth]{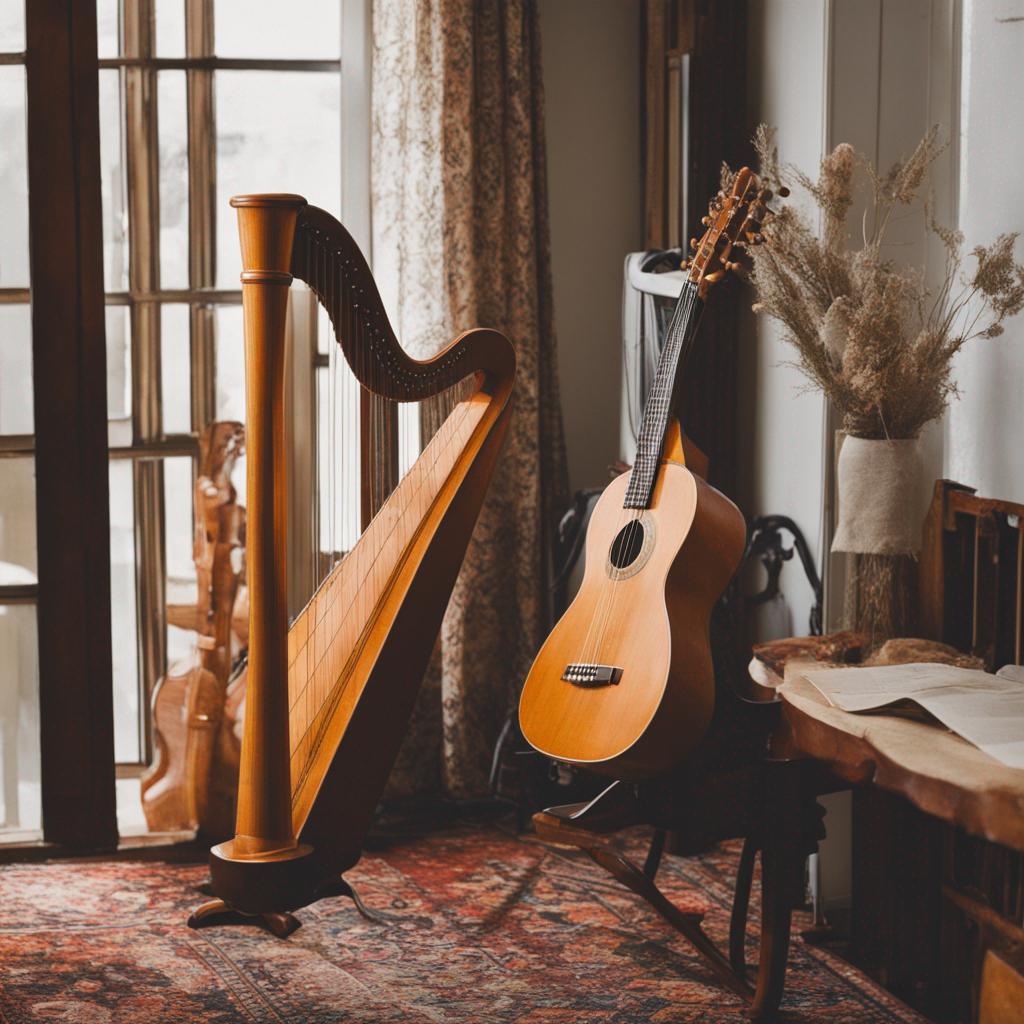}
        \caption*{Image of ``harp'' and ``ukelele''. ``ukelele'' was replaced by ``acoustic guitar''.}
    \end{minipage}
    \hfill
    \begin{minipage}{0.45\linewidth}
        \centering
        \includegraphics[width=\linewidth]{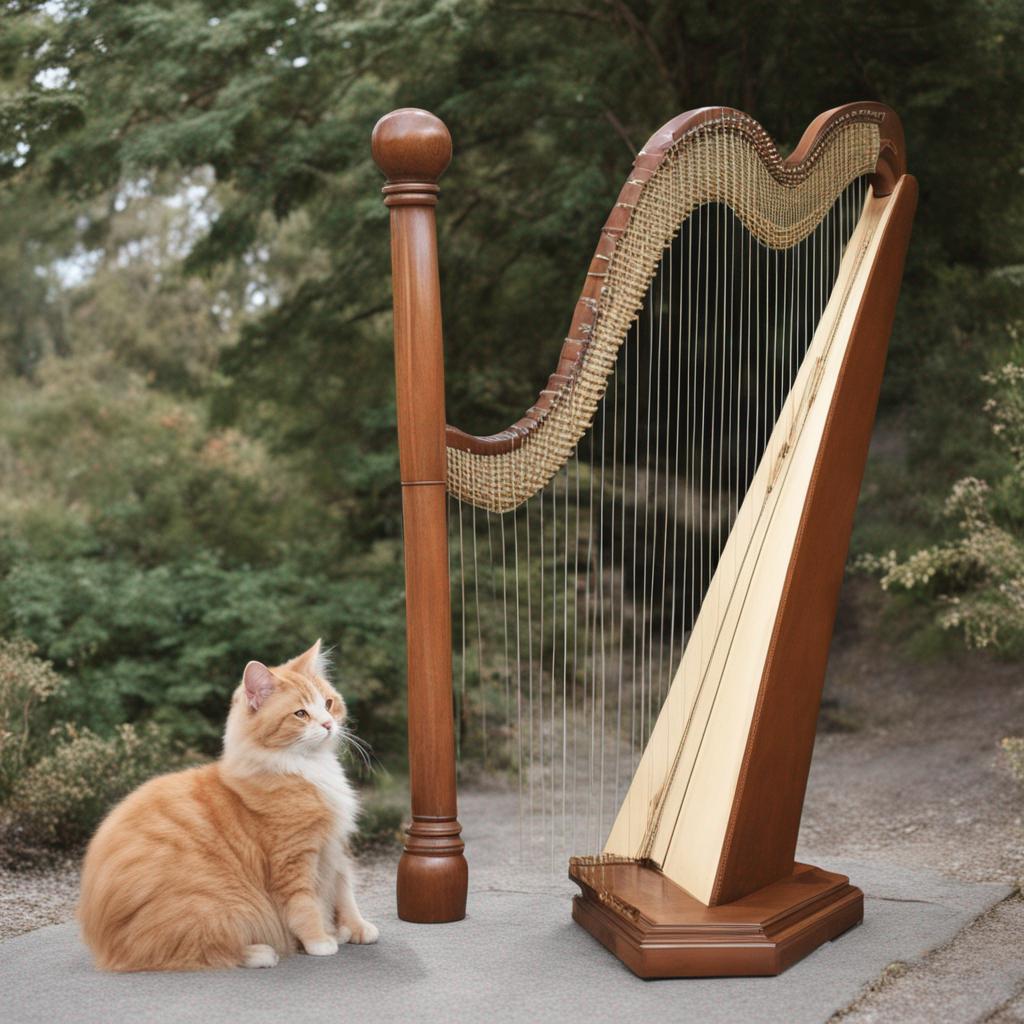}
        \caption*{Image of ``harpsichord'' and "cat". ``harpsichord'' was replaced by ``harp''.}
    \end{minipage}

    \vspace{0.4cm}
    \caption{Examples of mismatches between original class labels and image content. Labels were corrected through manual curation.}
    \label{fig:wrong_classes}
\end{figure}

Figure \ref{fig:wrong_classes} shows examples where the class labels in IS3 were incorrect, along with the new labels assigned to the corresponding images. These corrections were made through manual curation.

\subsubsection{IS3 category simplification}
\label{ap:is3_category_simplification}

Table \ref{tab:IS3+_class_mapping} shows the full class mapping from old labels to the ones assigned. There were some specific labels impossible to distinguish with a single image, so we merged them into a single one. One example could be: \textit{``chicken clucking''} and \textit{``chicken crowing''} are simplified as \textit{``chicken''}.

\begin{table}[ht]
\begin{center}
\scriptsize
\begin{tabular}{|p{0.3\textwidth} |p{0.3\textwidth} |p{0.3\textwidth}|}
\hline
playing accordion $\rightarrow$ \textbf{accordion} & car engine starting $\rightarrow$ \textbf{vehicle} & fox barking $\rightarrow$ \textbf{fox} \\
\hline
playing acoustic guitar $\rightarrow$ \textbf{acoustic guitar} & car passing by $\rightarrow$ \textbf{vehicle} & playing french horn $\rightarrow$ \textbf{french horn} \\
\hline
airplane $\rightarrow$ \textbf{airplane} & driving buses $\rightarrow$ \textbf{vehicle} & gibbon howling $\rightarrow$ \textbf{gibbon} \\
\hline
airplane, airplane flyby $\rightarrow$ \textbf{airplane} & opening or closing car electric windows $\rightarrow$ \textbf{vehicle} & goat bleating $\rightarrow$ \textbf{goat} \\
\hline
airplane flyby $\rightarrow$ \textbf{airplane} & race car, auto racing $\rightarrow$ \textbf{vehicle} & playing electric guitar $\rightarrow$ \textbf{guitar} \\
\hline
alarm clock ringing $\rightarrow$ \textbf{alarm clock} & cat caterwauling $\rightarrow$ \textbf{cat} & playing steel guitar, slide guitar $\rightarrow$ \textbf{guitar} \\
\hline
alligators, crocodiles hissing $\rightarrow$ \textbf{alligator} & cat growling $\rightarrow$ \textbf{cat} & cap gun shooting $\rightarrow$ \textbf{gun} \\
\hline
baby laughter $\rightarrow$ \textbf{baby} & cat hissing $\rightarrow$ \textbf{cat} & machine gun shooting $\rightarrow$ \textbf{gun} \\
\hline
baby crying $\rightarrow$ \textbf{baby} & cat meowing $\rightarrow$ \textbf{cat} & hair dryer drying $\rightarrow$ \textbf{hair dryer} \\
\hline
playing banjo $\rightarrow$ \textbf{banjo} & cat purring $\rightarrow$ \textbf{cat} & playing harpsichord $\rightarrow$ \textbf{harp} \\
\hline
playing bassoon $\rightarrow$ \textbf{bassoon} & playing cello $\rightarrow$ \textbf{cello} & playing harp $\rightarrow$ \textbf{harp} \\
\hline
barn swallow calling $\rightarrow$ \textbf{bird} & chainsawing trees $\rightarrow$ \textbf{chainsaw} & hedge trimmer running $\rightarrow$ \textbf{hedge trimmer} \\
\hline
bird chirping, tweeting $\rightarrow$ \textbf{bird} & cheetah chirrup $\rightarrow$ \textbf{cheetah} & helicopter $\rightarrow$ \textbf{helicopter} \\
\hline
bird wings flapping $\rightarrow$ \textbf{bird} & chicken clucking $\rightarrow$ \textbf{chicken} & horse clip-clop $\rightarrow$ \textbf{horse} \\
\hline
black capped chickadee calling $\rightarrow$ \textbf{bird} & chicken crowing $\rightarrow$ \textbf{chicken} & ice cream truck, ice cream van $\rightarrow$ \textbf{ice cream truck} \\
\hline
canary calling $\rightarrow$ \textbf{bird} & child singing $\rightarrow$ \textbf{child} & lathe spinning $\rightarrow$ \textbf{lathe/engine} \\
\hline
wood thrush calling $\rightarrow$ \textbf{bird} & child speech, kid speaking $\rightarrow$ \textbf{child} & lawn mowing $\rightarrow$ \textbf{lawn mower} \\
\hline
blowtorch igniting $\rightarrow$ \textbf{blowtorch} & chimpanzee pant-hooting $\rightarrow$ \textbf{chimpanzee} & lions growling $\rightarrow$ \textbf{lion} \\
\hline
typing on computer keyboard $\rightarrow$ \textbf{computer keyboard} & chinchilla barking $\rightarrow$ \textbf{chinchilla} & lions roaring $\rightarrow$ \textbf{lion} \\
\hline
playing cornet $\rightarrow$ \textbf{cornet} & chipmunk chirping $\rightarrow$ \textbf{chipmunk} & male speech, man speaking $\rightarrow$ \textbf{male voice} \\
\hline
bull bellowing $\rightarrow$ \textbf{cow} & church bell ringing $\rightarrow$ \textbf{church bell} & playing mandolin $\rightarrow$ \textbf{mandolin} \\
\hline
cattle mooing $\rightarrow$ \textbf{cow} & cricket chirping $\rightarrow$ \textbf{cricket} & missile launch $\rightarrow$ \textbf{missile} \\
\hline
cow lowing $\rightarrow$ \textbf{cow} & playing cymbal $\rightarrow$ \textbf{cymbal} & motorboat, speedboat acceleration $\rightarrow$ \textbf{motorboat} \\
\hline
dinosaurs bellowing $\rightarrow$ \textbf{dinosaur} & dog barking $\rightarrow$ \textbf{dog} & driving motorcycle $\rightarrow$ \textbf{motorcycle} \\
\hline
dog baying $\rightarrow$ \textbf{dog} & dog bow-wow $\rightarrow$ \textbf{dog} & mouse squeaking $\rightarrow$ \textbf{mouse} \\
\hline
dog growling $\rightarrow$ \textbf{dog} & dog howling $\rightarrow$ \textbf{dog} & playing oboe $\rightarrow$ \textbf{oboe} \\
\hline
dog whimpering $\rightarrow$ \textbf{dog} & donkey, ass braying $\rightarrow$ \textbf{donkey} & ocean burbling $\rightarrow$ \textbf{ocean} \\
\hline
playing drum kit $\rightarrow$ \textbf{drums} & eagle screaming $\rightarrow$ \textbf{eagle} & orchestra $\rightarrow$ \textbf{orchestra} \\
\hline
electric grinder grinding $\rightarrow$ \textbf{electric grinder} & playing electronic organ $\rightarrow$ \textbf{electronic organ} & owl hooting $\rightarrow$ \textbf{owl} \\
\hline
elephant trumpeting $\rightarrow$ \textbf{elephant} & eletric blender running $\rightarrow$ \textbf{eletric blender} & parrot talking $\rightarrow$ \textbf{parrot} \\
\hline
elk bugling $\rightarrow$ \textbf{elk} & female singing $\rightarrow$ \textbf{female voice} & penguins braying $\rightarrow$ \textbf{penguin} \\
\hline
female speech, woman speaking $\rightarrow$ \textbf{female voice} & fireworks banging $\rightarrow$ \textbf{fireworks} & people crowd $\rightarrow$ \textbf{people crowd} \\
\hline
people eating crisps $\rightarrow$ \textbf{people eating crisps} & people marching $\rightarrow$ \textbf{people marching} & playing piano $\rightarrow$ \textbf{piano} \\
\hline
pigeon, dove cooing $\rightarrow$ \textbf{pigeon} & popping popcorn $\rightarrow$ \textbf{popcorn} & playing saxophone $\rightarrow$ \textbf{saxophone} \\
\hline
sea lion barking $\rightarrow$ \textbf{sea lion} & sheep bleating $\rightarrow$ \textbf{sheep} & playing shofar $\rightarrow$ \textbf{shofar} \\
\hline
fire truck siren $\rightarrow$ \textbf{siren} & police car (siren) $\rightarrow$ \textbf{siren} & skateboarding $\rightarrow$ \textbf{skateboarding} \\
\hline
slot machine $\rightarrow$ \textbf{slot machine} & snake hissing $\rightarrow$ \textbf{snake} & snake rattling $\rightarrow$ \textbf{snake} \\
\hline
driving snowmobile $\rightarrow$ \textbf{snowmobile} & splashing water $\rightarrow$ \textbf{stream} & squishing water $\rightarrow$ \textbf{stream} \\
\hline
subway, metro, underground $\rightarrow$ \textbf{subway} & tap dancing $\rightarrow$ \textbf{tap dance} & telephone bell ringing $\rightarrow$ \textbf{telephone bell} \\
\hline
playing timbales $\rightarrow$ \textbf{timbales} & tractor digging $\rightarrow$ \textbf{vehicle} & train horning $\rightarrow$ \textbf{train} \\
\hline
train wheels squealing $\rightarrow$ \textbf{train} & train whistling $\rightarrow$ \textbf{train} & playing trumpet $\rightarrow$ \textbf{trumpet} \\
\hline
turkey gobbling $\rightarrow$ \textbf{turkey} & playing ukulele $\rightarrow$ \textbf{ukulele} & vacuum cleaner cleaning floors $\rightarrow$ \textbf{vacuum cleaner} \\
\hline
waterfall burbling $\rightarrow$ \textbf{waterfall} & whale calling $\rightarrow$ \textbf{whale} & wind chime $\rightarrow$ \textbf{wind chime} \\
\hline
woodpecker pecking tree $\rightarrow$ \textbf{woodpecker} & & \\
\hline

\end{tabular}
\vspace{0.3cm}
\caption{Full mapping from original IS3 class labels to simplified IS3+ labels.}
\label{tab:IS3+_class_mapping}
\end{center}
\end{table}
\subsection{Modality Alignment and Separability}
\label{sup:modality_alignment}

\begin{figure*}[htb] 
\centerline{\includegraphics[width=\textwidth]{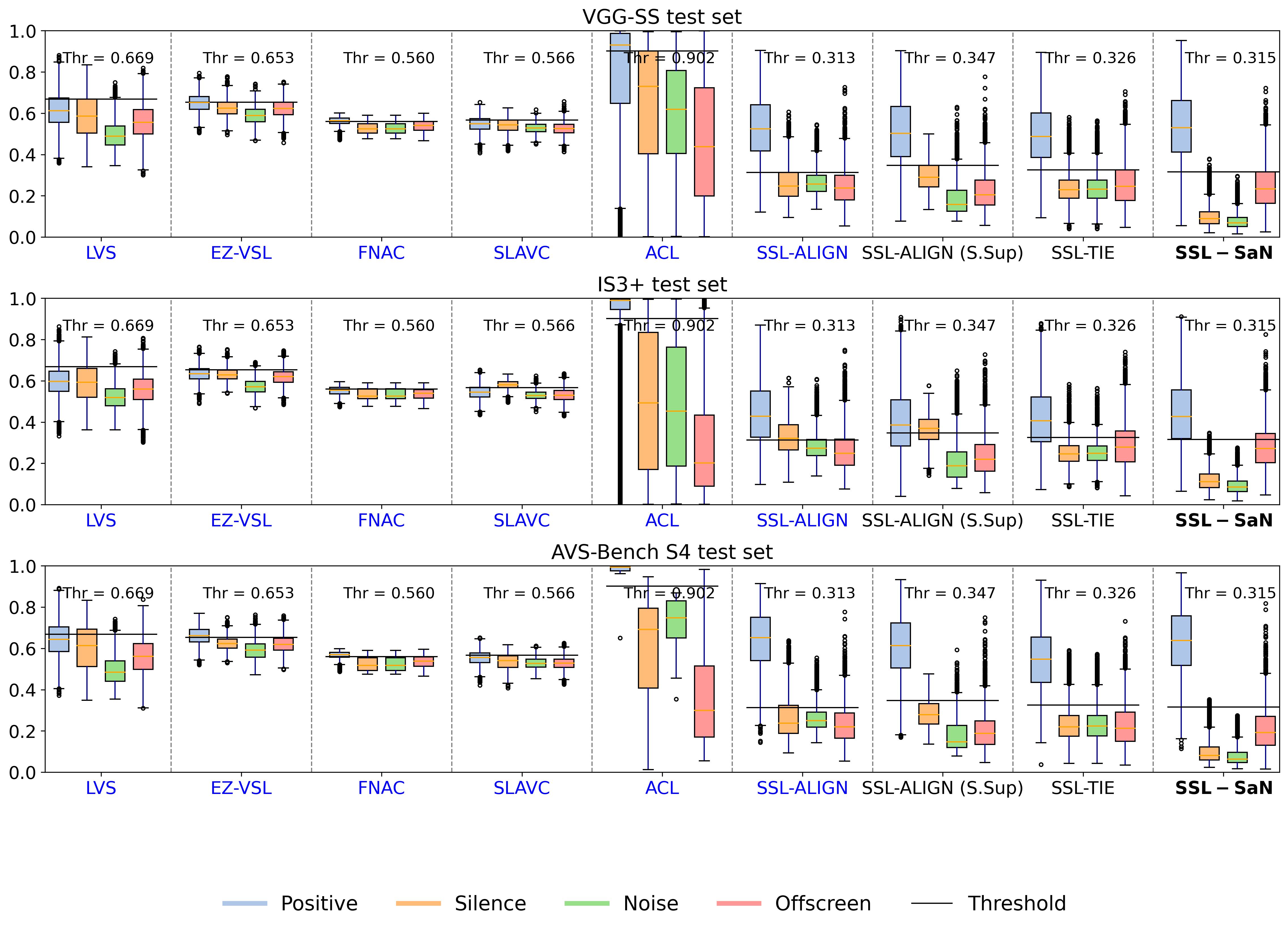}}
\caption{Distribution of the maximum audio-visual similarity values across different models and datasets for both positive and negative audio inputs. The Universal threshold is indicated in the top-right corner of each model panel. Models in blue are trained in a weakly-supervised manner while black are trained in a self-supervised manner.}
\label{sup_fig:boxplots}
\end{figure*}

Figure \ref{sup_fig:boxplots} shows the distribution of the maximum audio-visual similarity values for positive and negative audio-visual pairs across all models evaluated in the paper. We observe that in VGG-SS, the only models capable of clearly distinguishing between positives (blue) and negatives (silence, noise, and offscreen) are SSL-TIE, SSL-Align (both the fully self-supervised and the weakly-supervised versions), and SSL-SaN. 
On the other hand, LVS, EZ-VSL, FNAC and SLAVC 
provide overlapping similarity values for positive and negative audio-visual pairs, both in VGG-SS and IS3+, making difficult to establish a proper universal threshold that allows to distinguish both cases.  
In IS3+, both versions of SSL-Align are not good at filtering silence samples, but are better than
 SSL-TIE and SSL-SaN at distinuishing offscreen sounds from positive ones. \new{Although ACL does not provide a good separability from positive and negative pairs in VGG-SS, it does a much better job in the other two datasets. Actually, in these two datasets, it is the best model at separating positive pairs from negative offscreen pairs. This is due to the supervised training of CLIPSeg \cite{luddecke2022image} (the segmentation network used by ACL), which has been trained for object segmentation with both (annotated) positive and negative queries. On the other hand, it can be seen that our model is better than ACL at separating negative pairs with silence and noise. This suggests that ACL could benefit from our general strategy of introducing silence and noise in the contrastive learning.}

Among all the models, SSL-SaN shows a better balance in separating positive sounds from the three types of negative ones \new{across all datasets}.

\subsection{Cross-modal Retrieval}
\label{sup:cross_modal_retrieval}

In this section, we present the results of the cross-modal retrieval task on the same datasets as in Table \ref{tab:retrieval_paper_main} from the paper, but now including results from the IS3 dataset (omitted in the paper due to space constraints).

For IS3, we observe a similar trend as in IS3+, where our model achieves the best performance among the self-supervised models in the Image to Audio retrieval and SSL-Align is the best in Audio to Image.

    \begin{table}[ht]
    \centering
    \resizebox{\columnwidth}{!}{
\begin{tabular}{c|lcccccccccccccccc}
    \hline
    & & & & \multicolumn{6}{c}{I $\rightarrow$ A} & & \multicolumn{6}{c}{A $\rightarrow$ I} \\
    \cline{5-10} \cline{12-17}
    \textbf{Test set} & \textbf{Model} & \textbf{Self Supervised} & & \textbf{P@1 $\uparrow$} & \textbf{P@5 $\uparrow$} & \textbf{P@10 $\uparrow$} & \textbf{A@1 $\uparrow$} & \textbf{A@5 $\uparrow$} & \textbf{A@10 $\uparrow$} && \textbf{P@1 $\uparrow$} & \textbf{P@5 $\uparrow$} & \textbf{P@10 $\uparrow$} & \textbf{A@1 $\uparrow$} & \textbf{A@5 $\uparrow$} & \textbf{A@10 $\uparrow$}  \\
    \hline

    \multirow{8}{*}{\STAB{\rotatebox[origin=c]{90}{\textbf{VGG-SS}}}}
    
    & LVS {\scriptsize [CVPR, 2021]} & 
    \ding{56} & 
    & 
    2.96 & 
    2.84 & 
    2.83 & 
    2.96 & 
    10.37 & 
    16.47 & 
    &
    4.02 & 
    3.92 & 
    3.54& 
    4.02 & 
    13.71 & 
    20.22 \\ 

    & EZ-VSL {\scriptsize [ECCV, 2022]} & 
    \ding{56} & 
    & 
    2.11 & 
    2.27 & 
    2.16 & 
    2.11 & 
    8.21 & 
    13.06 & 
    &
    3.94 & 
    3.51 & 
    3.24& 
    3.94 & 
    14.19 & 
    22.55 \\
\hline
    \end{tabular}}
    \vspace{0.1cm}
    \caption{Results of the cross-modal retrieval analysis for same class in VGG-SS, IS3, IS3$^+$ and AVS-Bench S4.}
    \label{tab:prec_acc_vggss_class}
    \end{table}

\begin{figure}[ht]
\centering
\begin{tabular}{m{1.8cm} | m{1.8cm} m{1.8cm} m{1.8cm} m{1.8cm}}
\multicolumn{1}{c|}{\textbf{Query Image}} &
\multicolumn{4}{c}{\textbf{Retrieved Audios}} \\
& \centering \textbf{Top 1} & \centering \textbf{Top 2} & \centering \textbf{Top 3} & \centering \textbf{Top 4} \\
\end{tabular}

\vspace{0.2em}

\begin{tabular}{m{1.8cm} | m{1.8cm} m{1.8cm} m{1.8cm} m{1.8cm}}
\includegraphics[width=1.8cm]{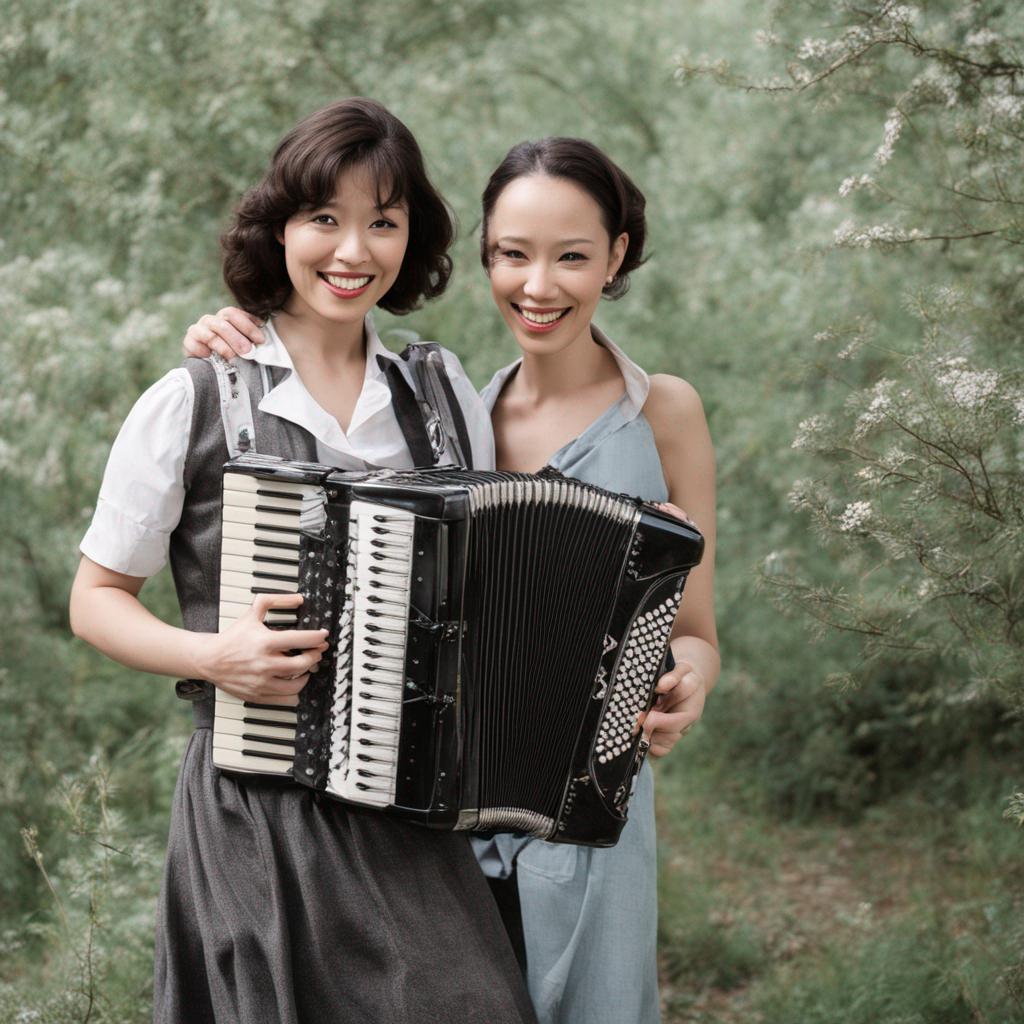} &
\includegraphics[width=1.8cm]{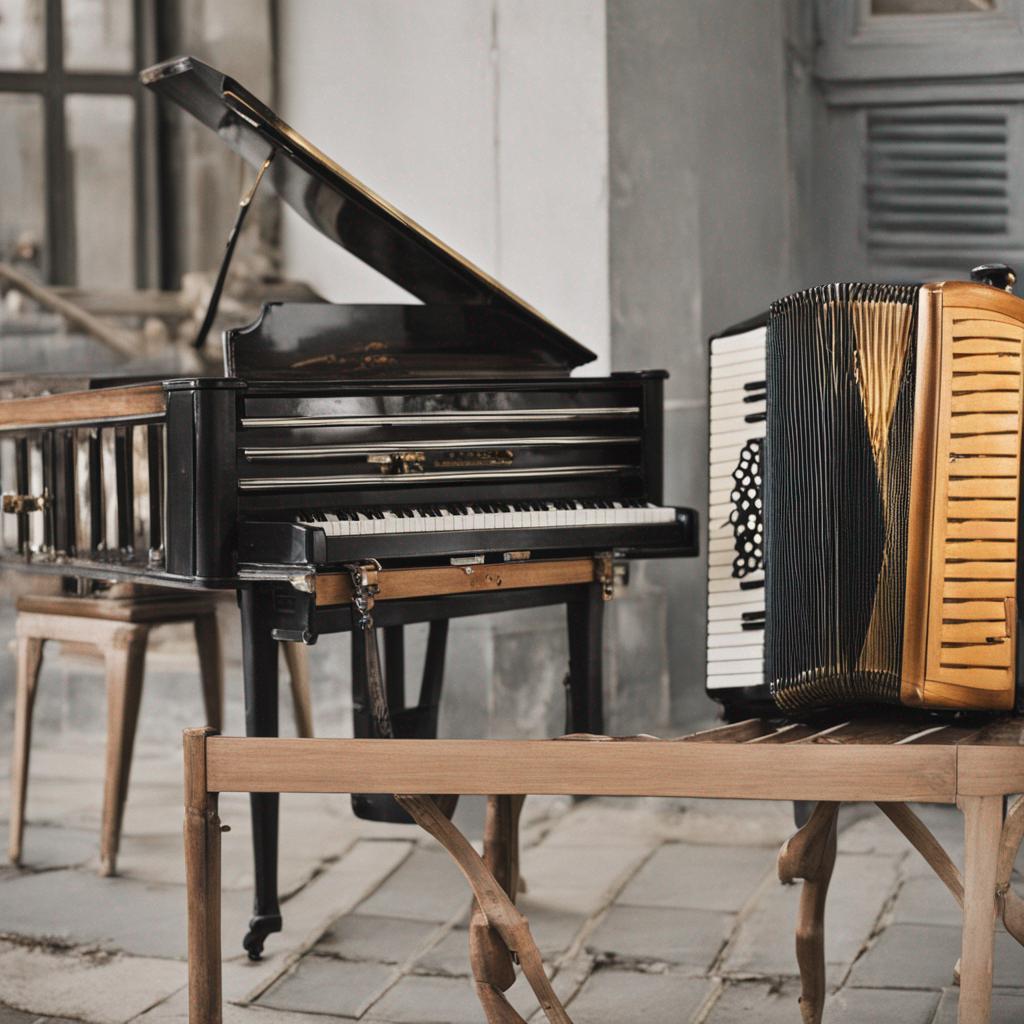} &
\includegraphics[width=1.8cm]{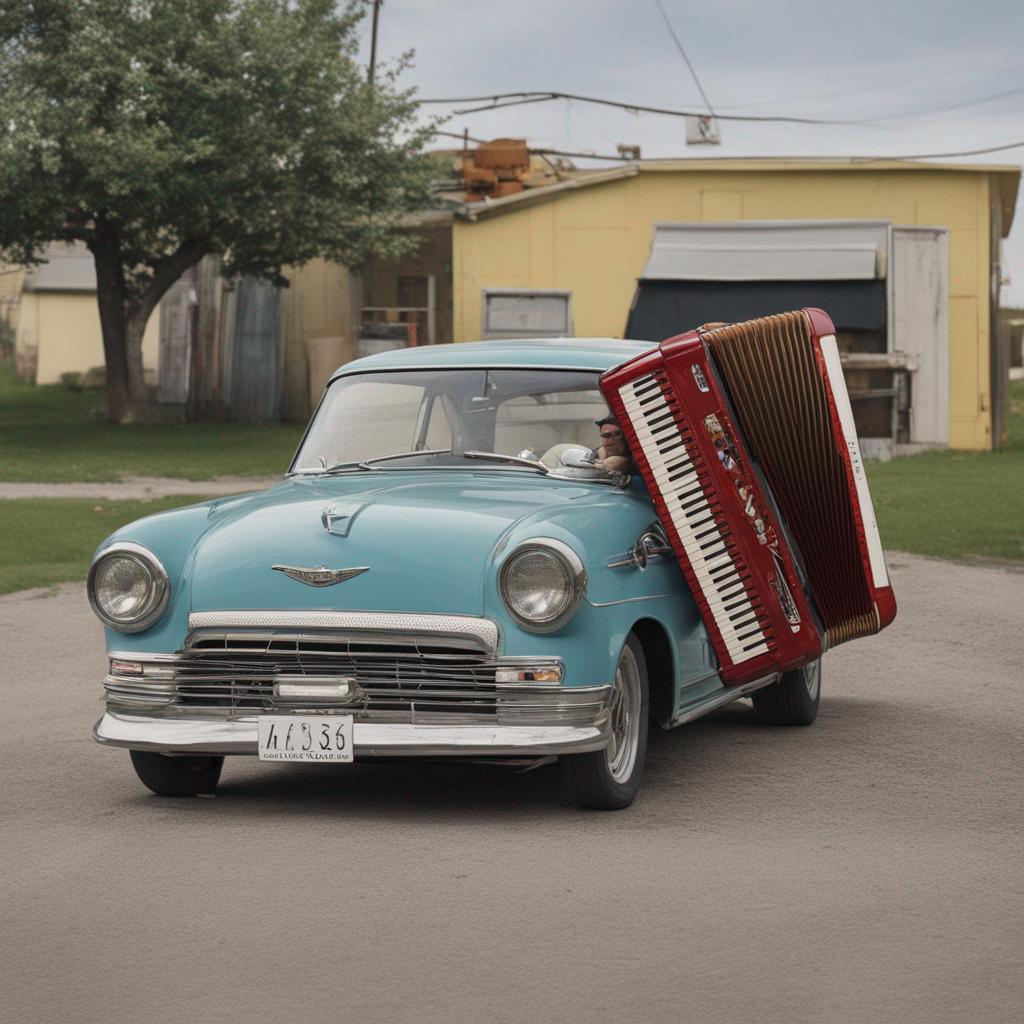} &
\includegraphics[width=1.8cm]{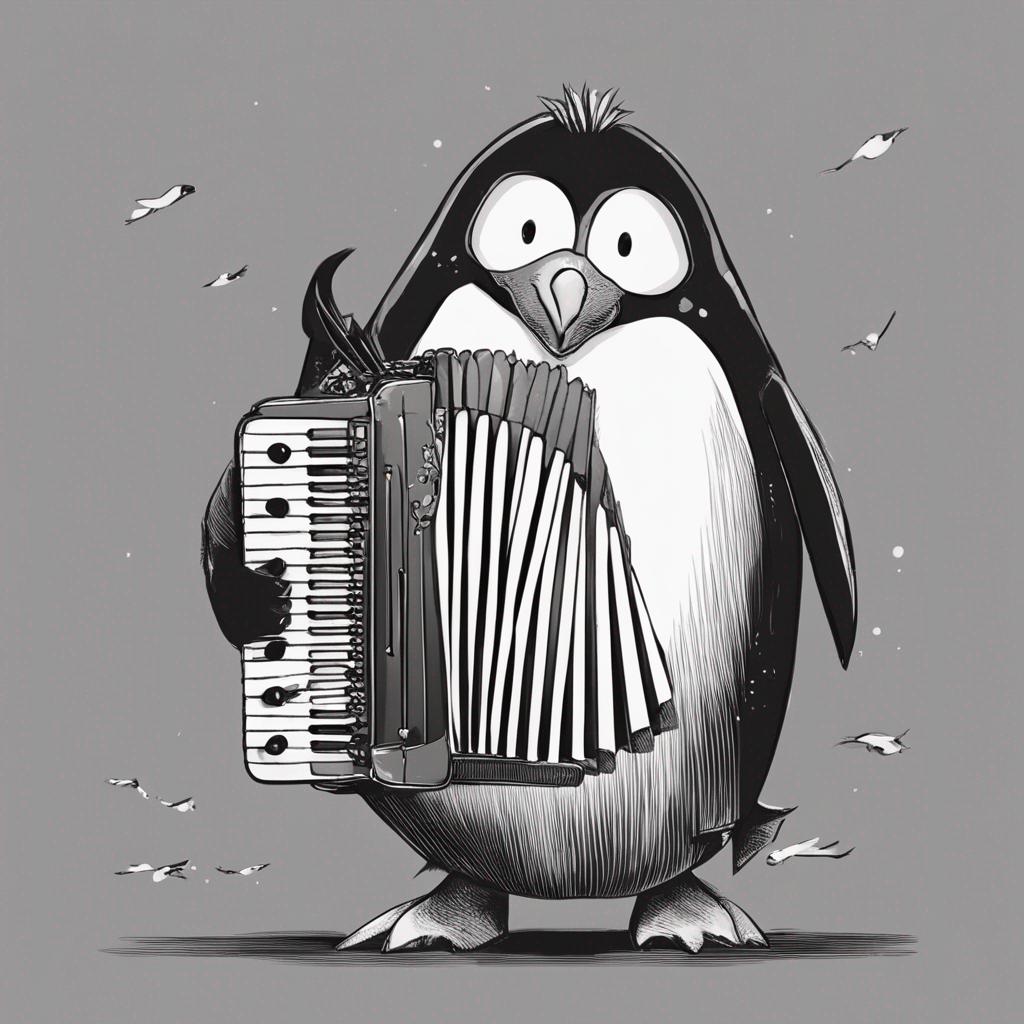} &
\includegraphics[width=1.8cm]{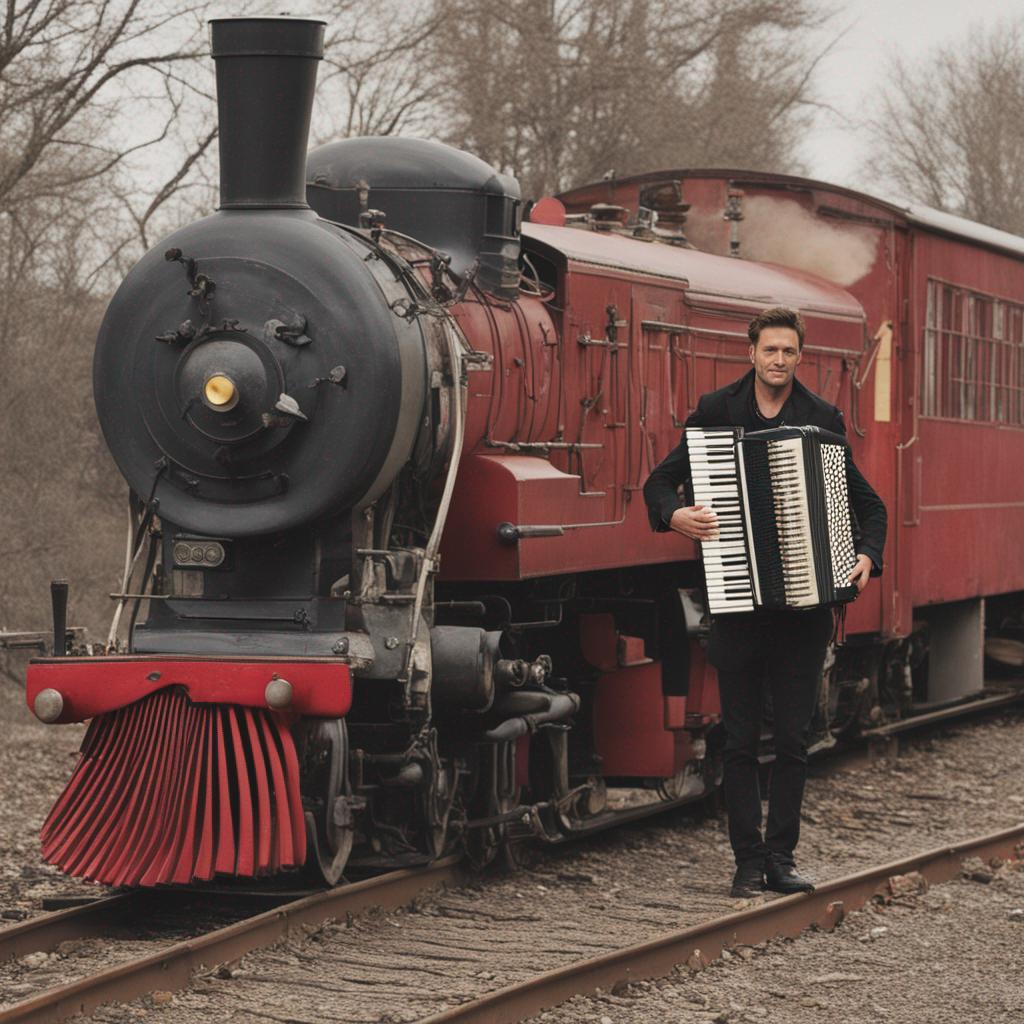} \\
& \scriptsize \raisebox{-0.17cm}{\includegraphics[height=0.5cm]{Figures/equalizer.png}} Accordion 
& \scriptsize \raisebox{-0.17cm}{\includegraphics[height=0.5cm]{Figures/equalizer.png}} Accordion 
& \scriptsize \raisebox{-0.17cm}{\includegraphics[height=0.5cm]{Figures/equalizer.png}} Accordion 
& \scriptsize \raisebox{-0.17cm}{\includegraphics[height=0.5cm]{Figures/equalizer.png}} Accordion \\

\includegraphics[width=1.8cm]{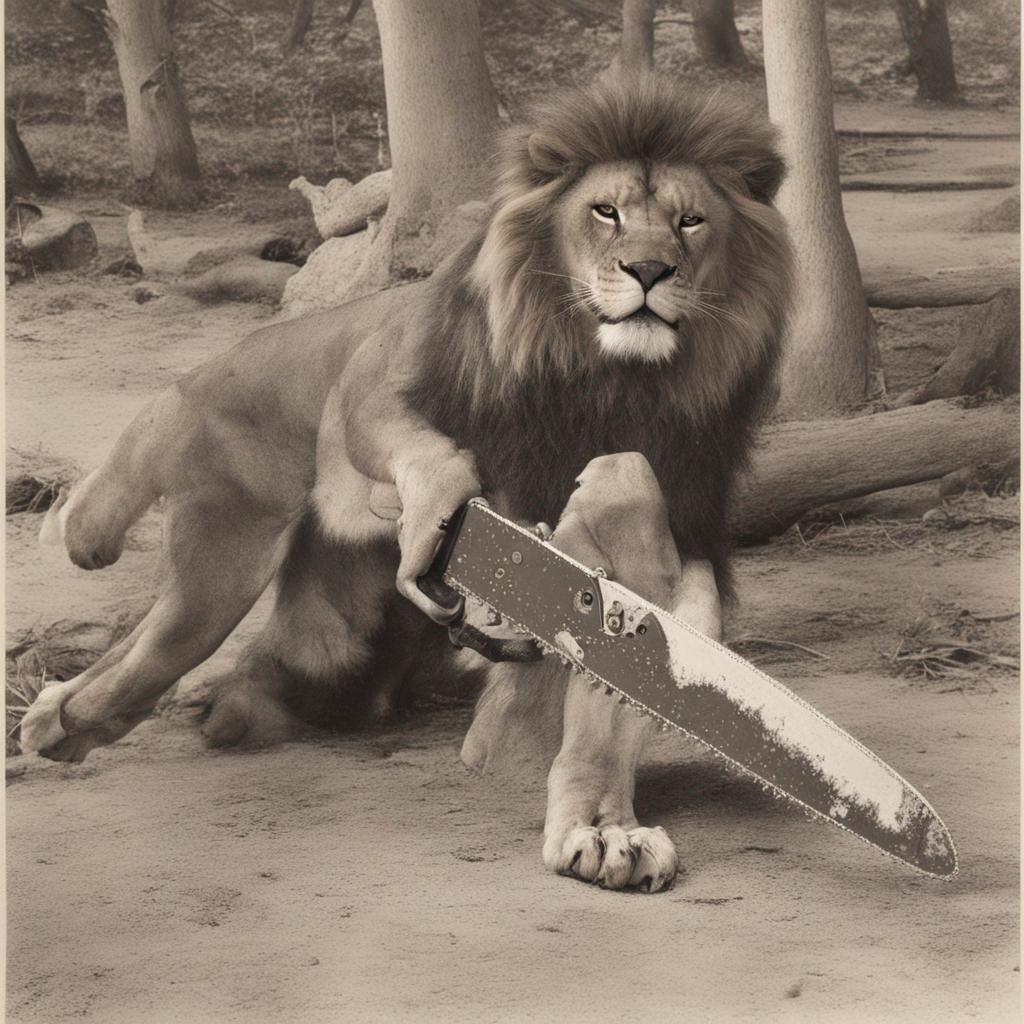} &
\includegraphics[width=1.8cm]{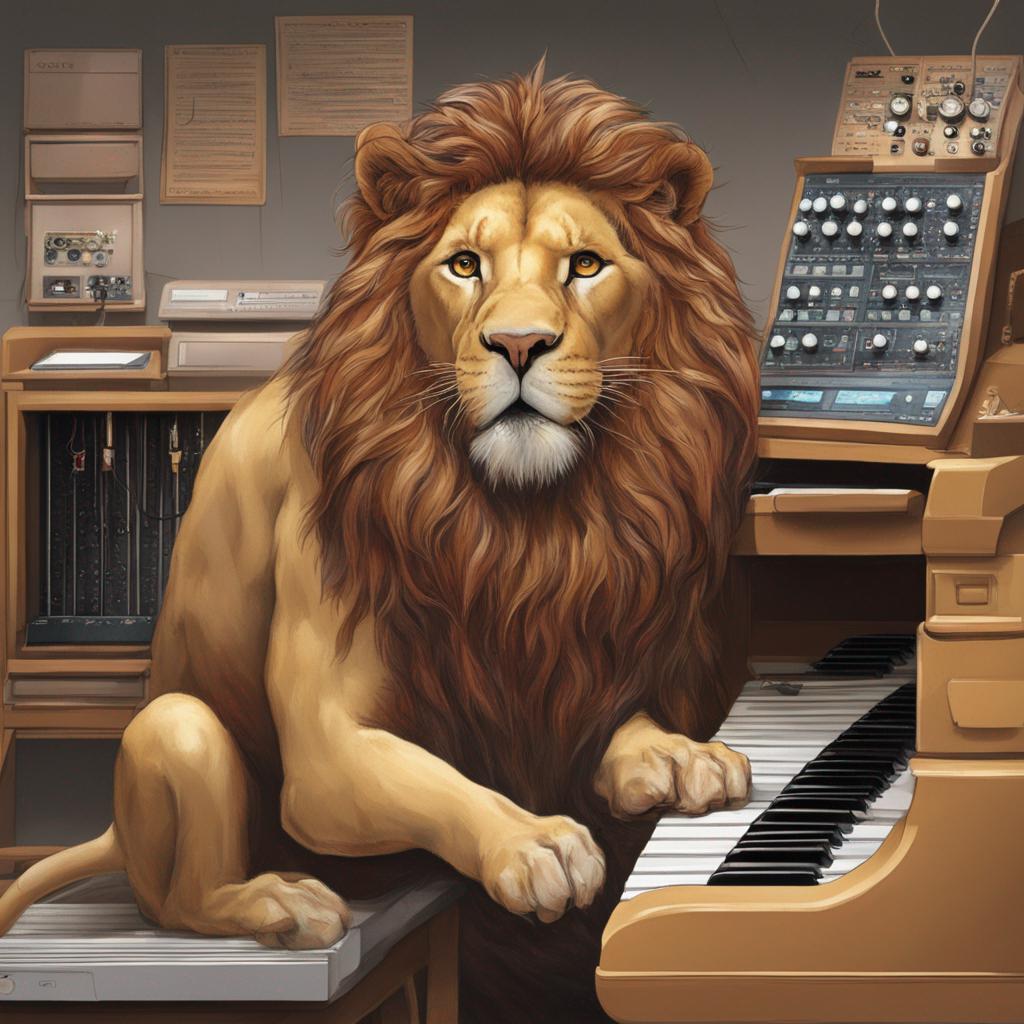} &
\includegraphics[width=1.8cm]{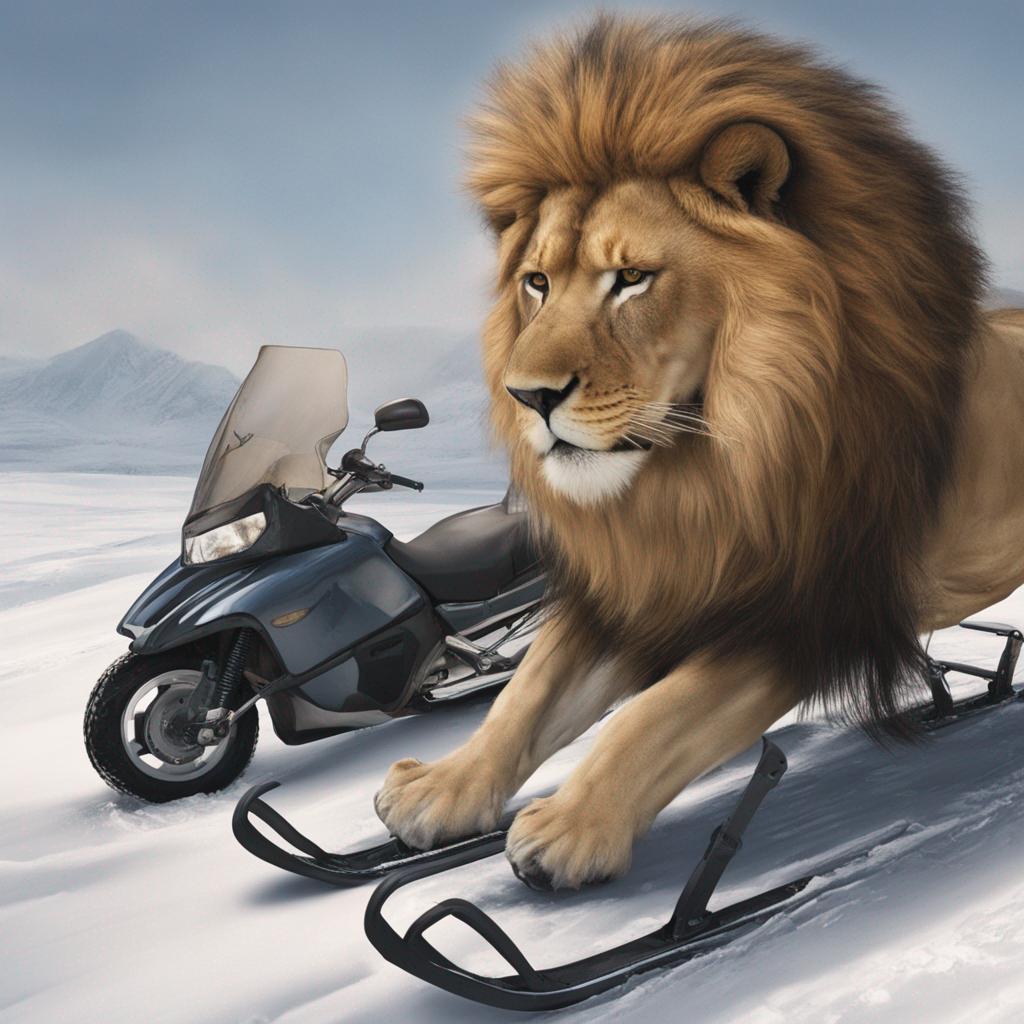} &
\includegraphics[width=1.8cm]{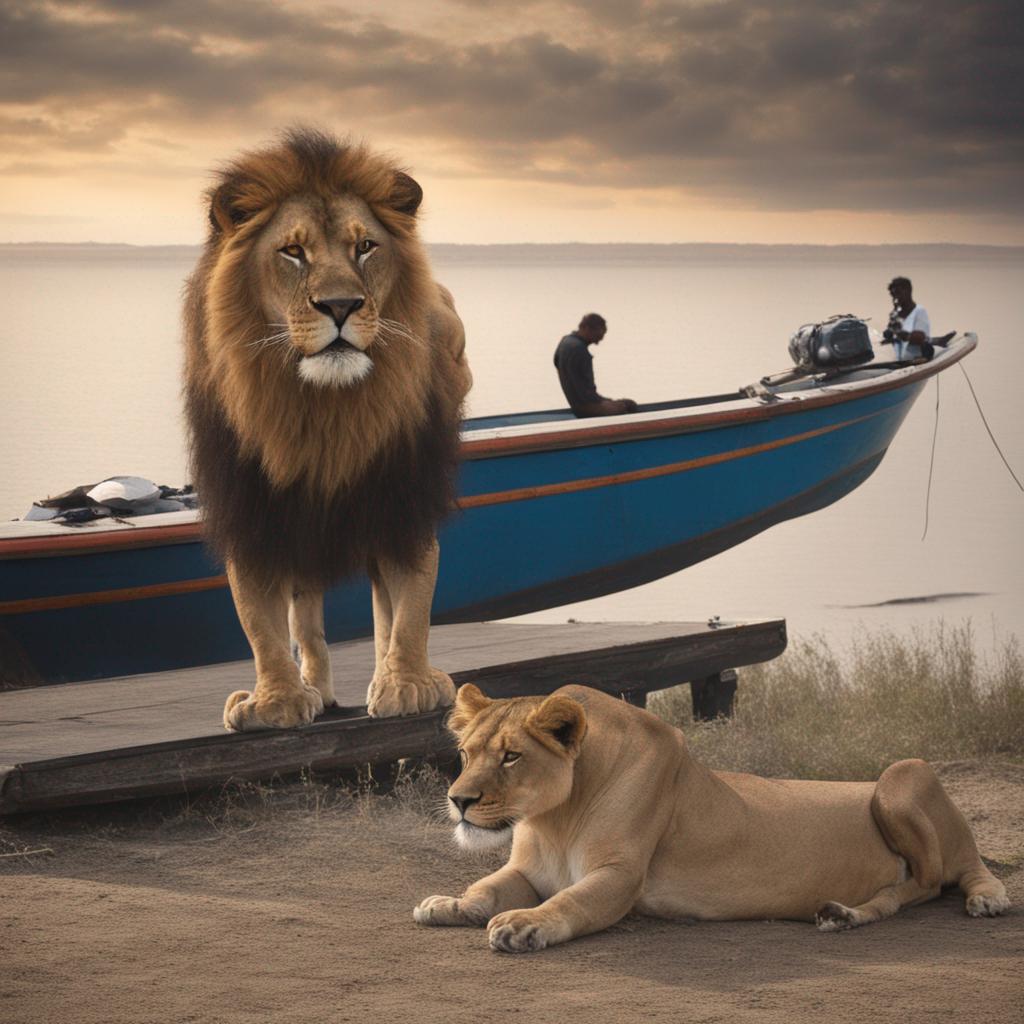} &
\includegraphics[width=1.8cm]{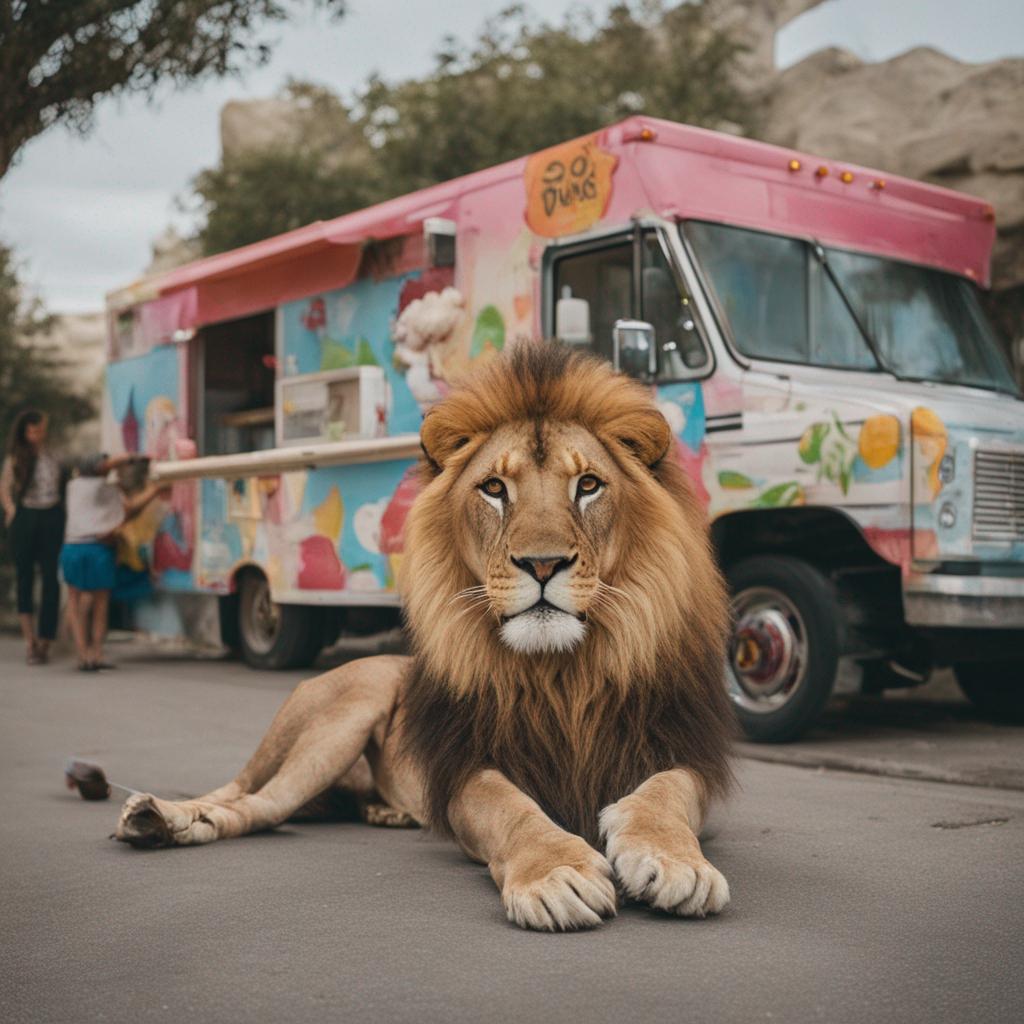} \\
& \scriptsize  \hspace{0.25cm}  \raisebox{-0.17cm}{\includegraphics[height=0.5cm]{Figures/equalizer.png}} Lion 
& \scriptsize  \hspace{0.25cm}  \raisebox{-0.17cm}{\includegraphics[height=0.5cm]{Figures/equalizer.png}} Lion 
& \scriptsize  \hspace{0.25cm}  \raisebox{-0.17cm}{\includegraphics[height=0.5cm]{Figures/equalizer.png}} Lion 
& \scriptsize  \hspace{0.25cm}  \raisebox{-0.17cm}{\includegraphics[height=0.5cm]{Figures/equalizer.png}} Lion \\

\includegraphics[width=1.8cm]{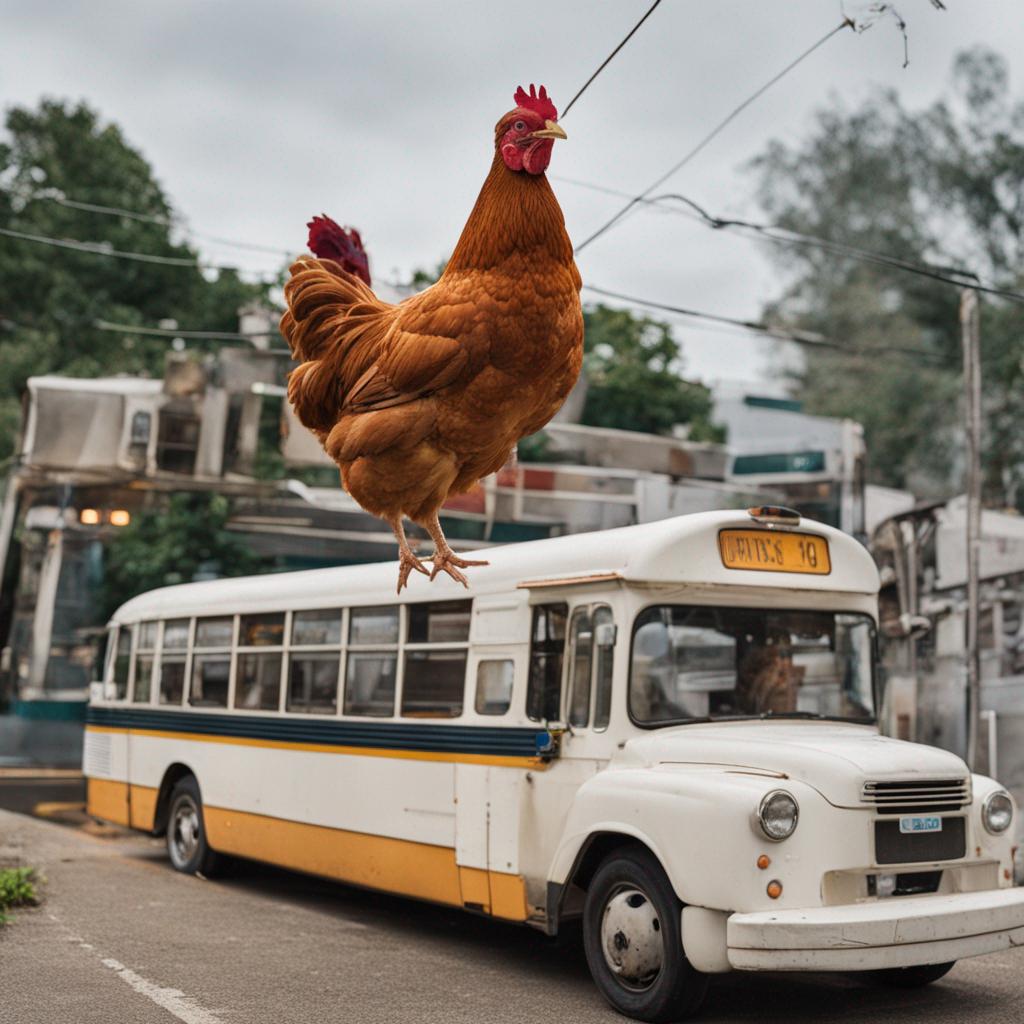} &
\includegraphics[width=1.8cm]{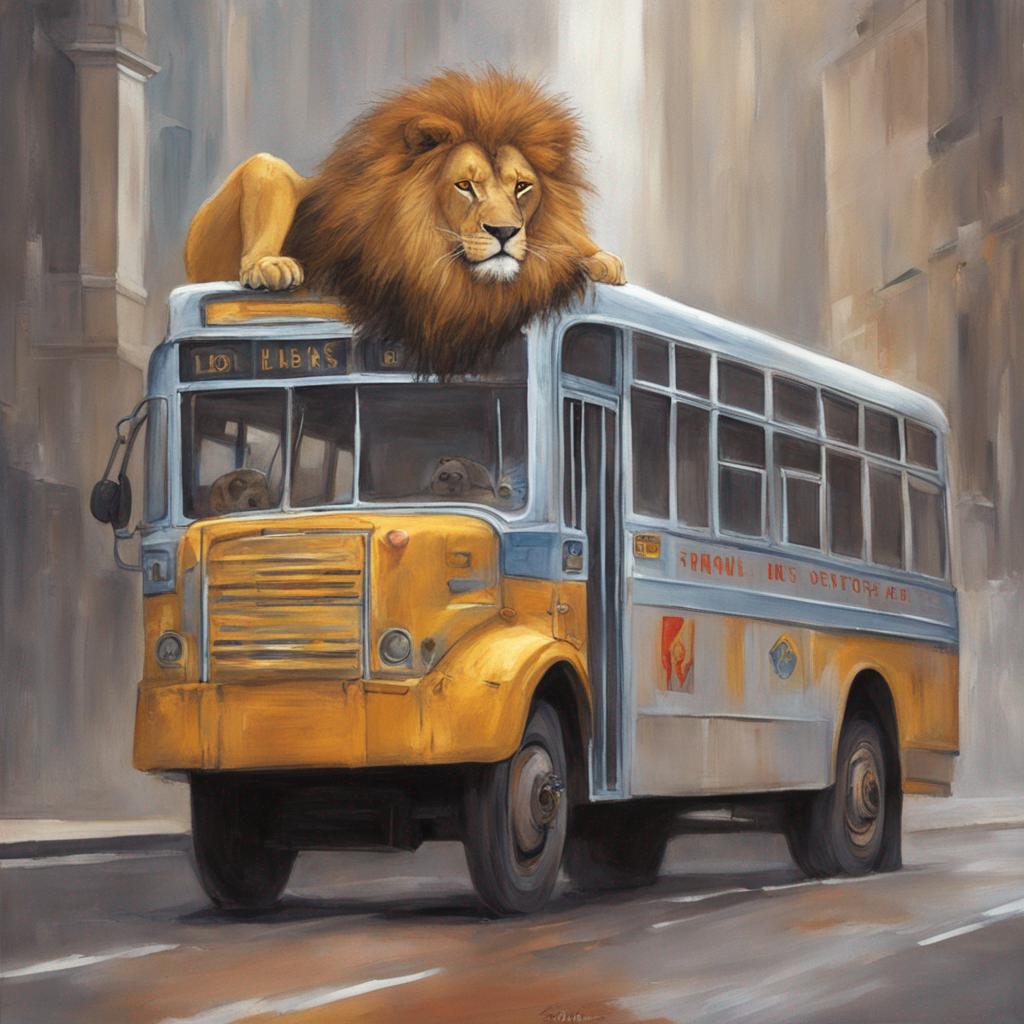} &
\includegraphics[width=1.8cm]{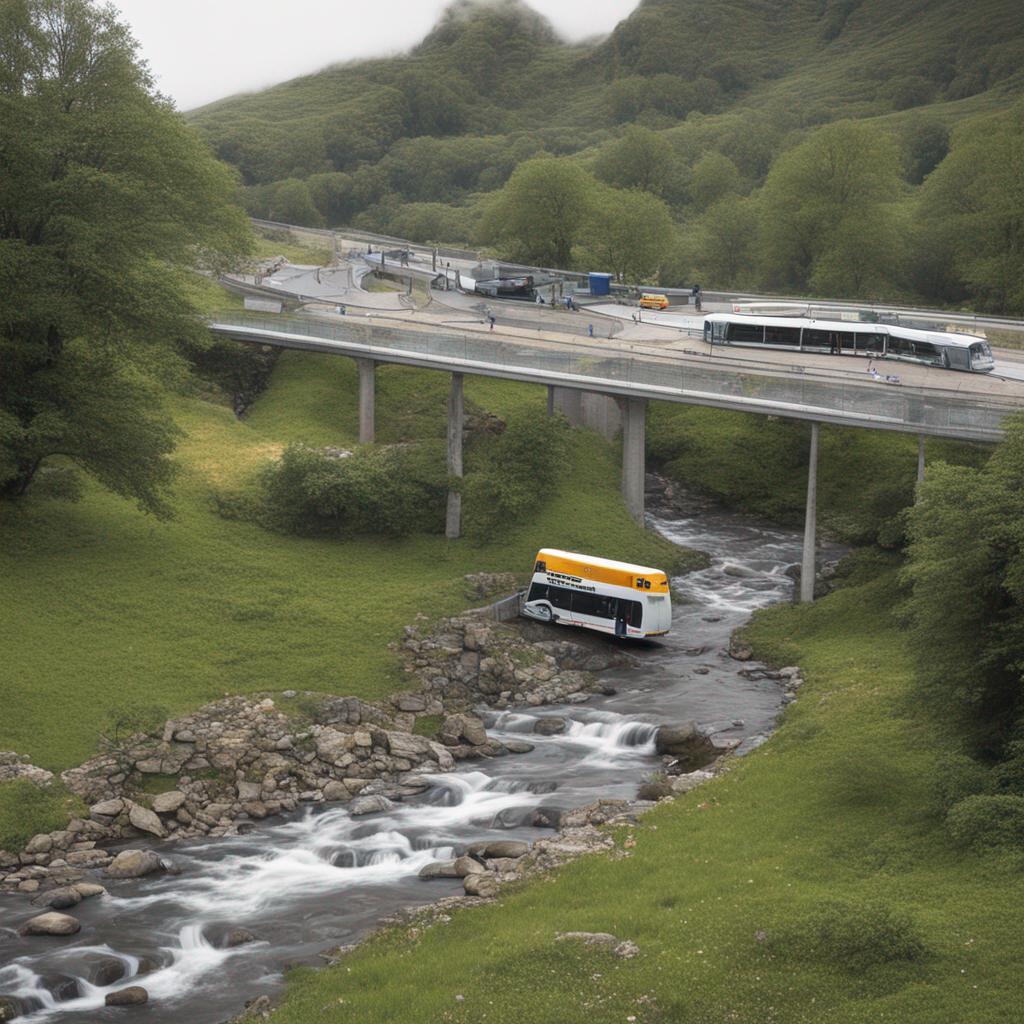} &
\includegraphics[width=1.8cm]{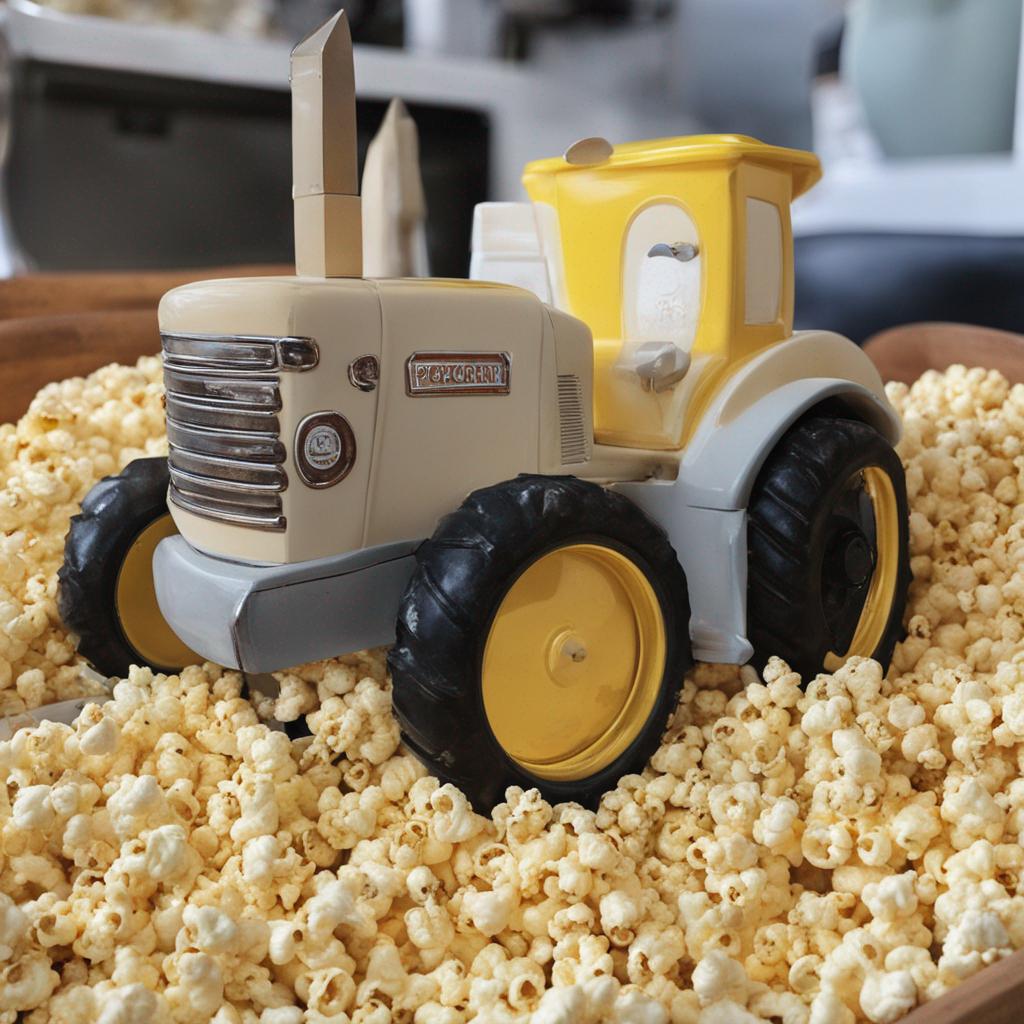} &
\includegraphics[width=1.8cm]{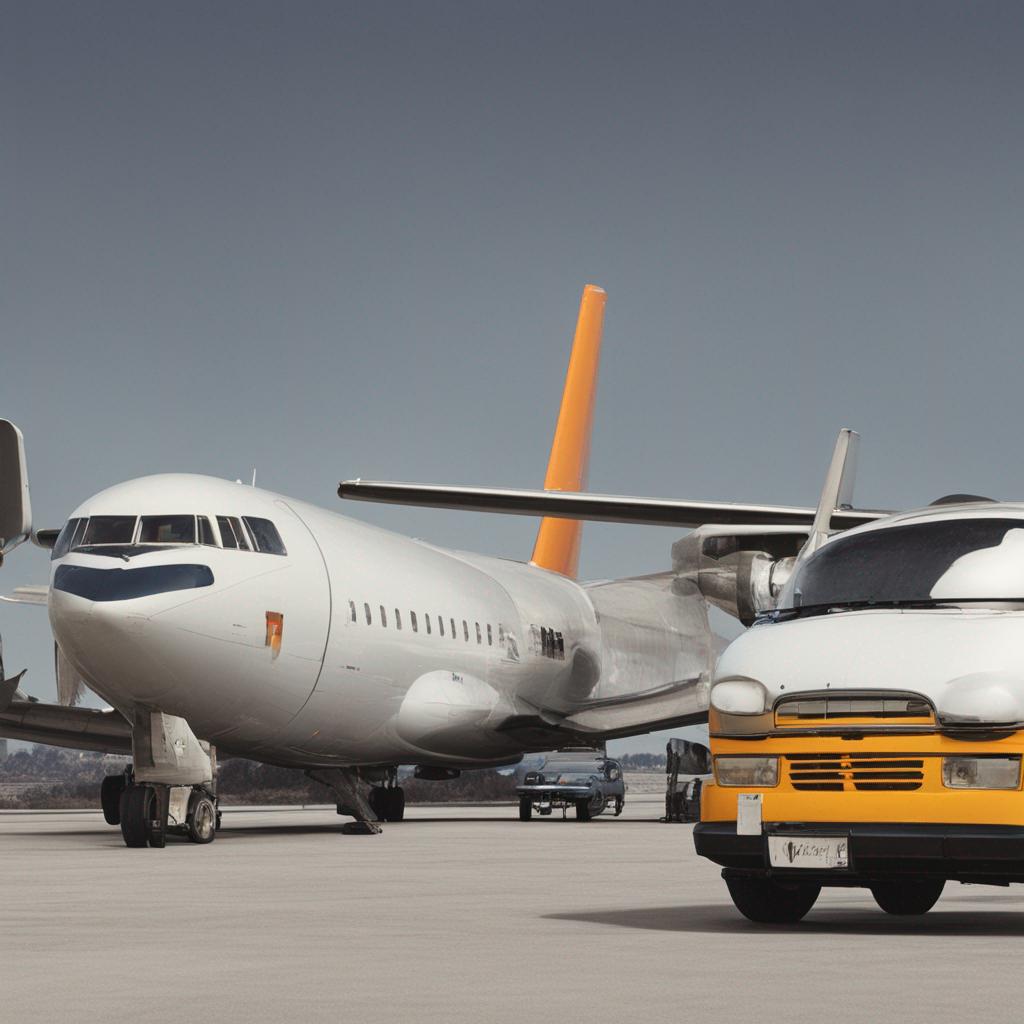} \\
& \scriptsize \hspace{0.35cm} \raisebox{-0.17cm}{\includegraphics[height=0.5cm]{Figures/equalizer.png}} Bus 
& \scriptsize \hspace{0.35cm} \raisebox{-0.17cm}{\includegraphics[height=0.5cm]{Figures/equalizer.png}} Bus 
& \scriptsize \hspace{0.35cm} \raisebox{-0.17cm}{\includegraphics[height=0.5cm]{Figures/equalizer.png}} Tractor 
& \scriptsize \hspace{0.35cm} \raisebox{-0.17cm}{\includegraphics[height=0.5cm]{Figures/equalizer.png}} Bus \\

\includegraphics[width=1.8cm]{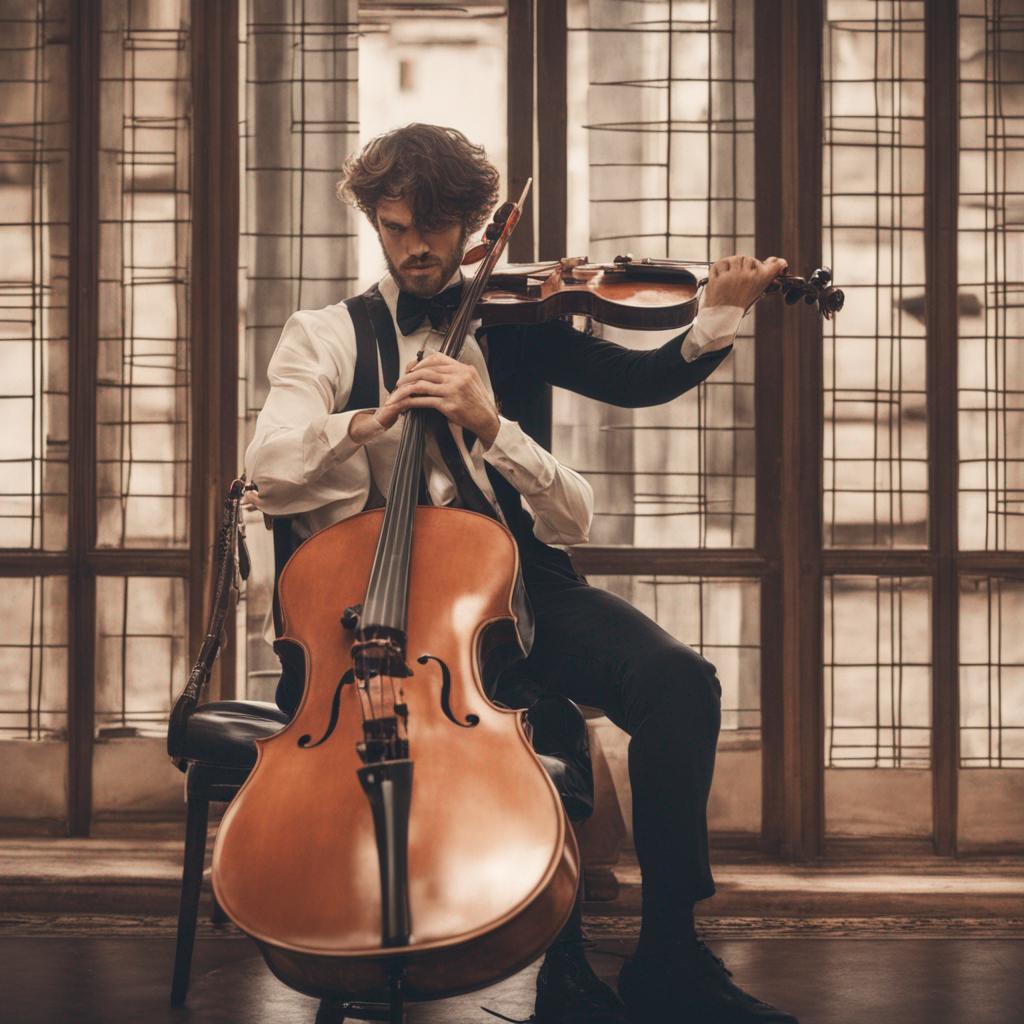} &
\includegraphics[width=1.8cm]{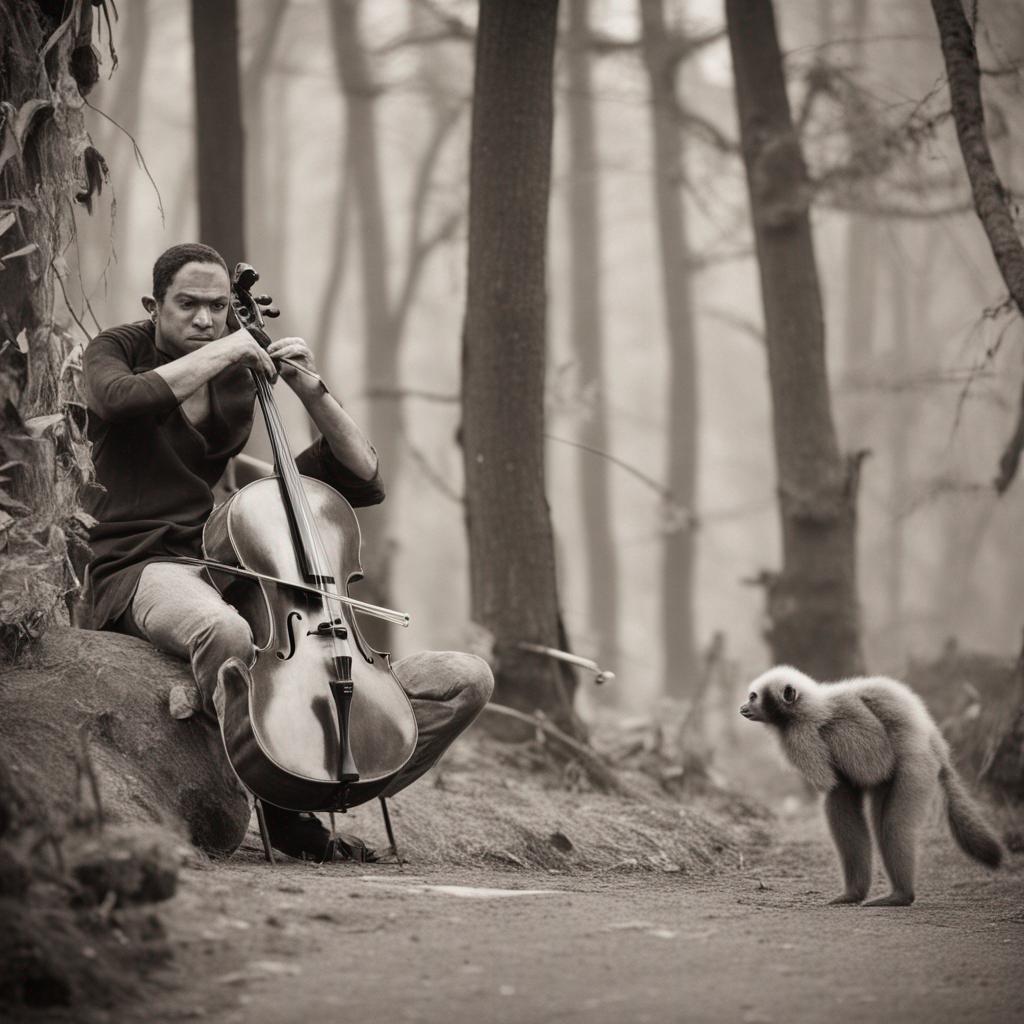} &
\includegraphics[width=1.8cm]{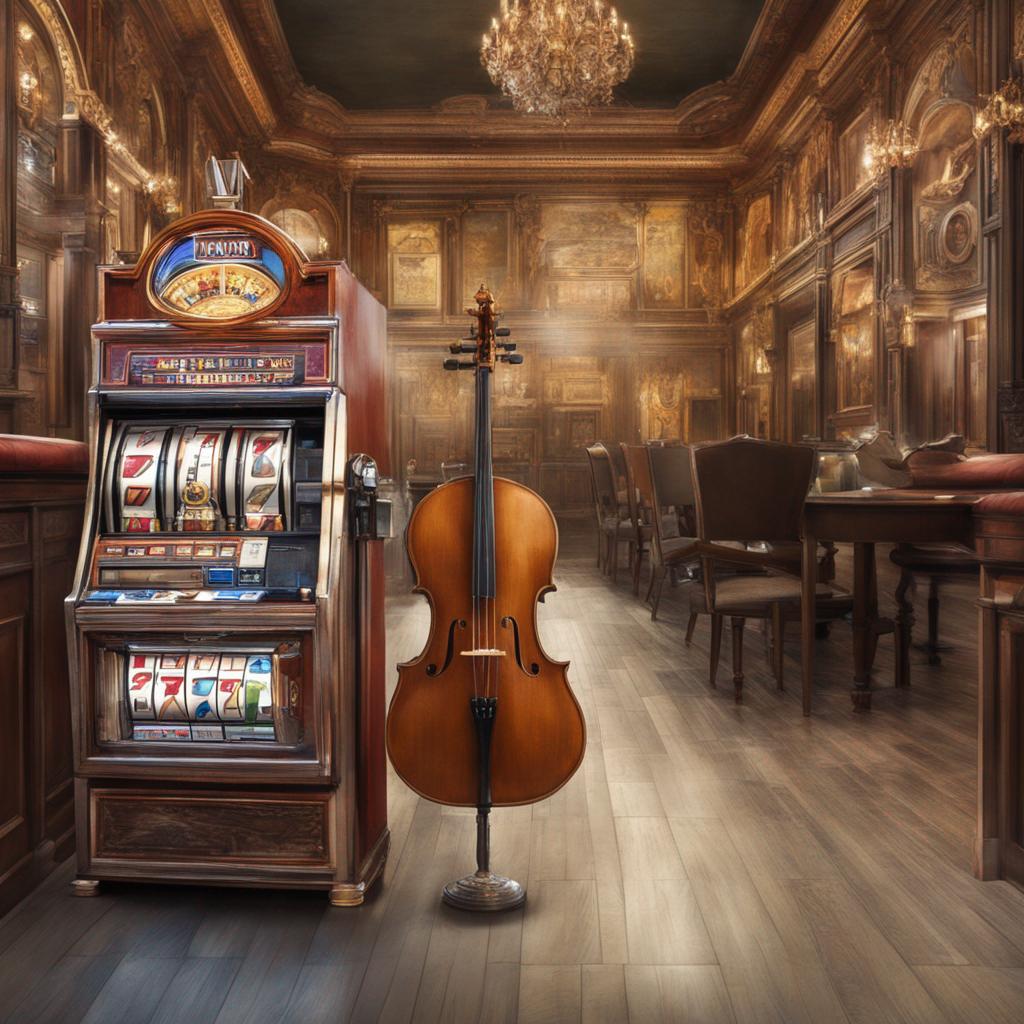} &
\includegraphics[width=1.8cm]{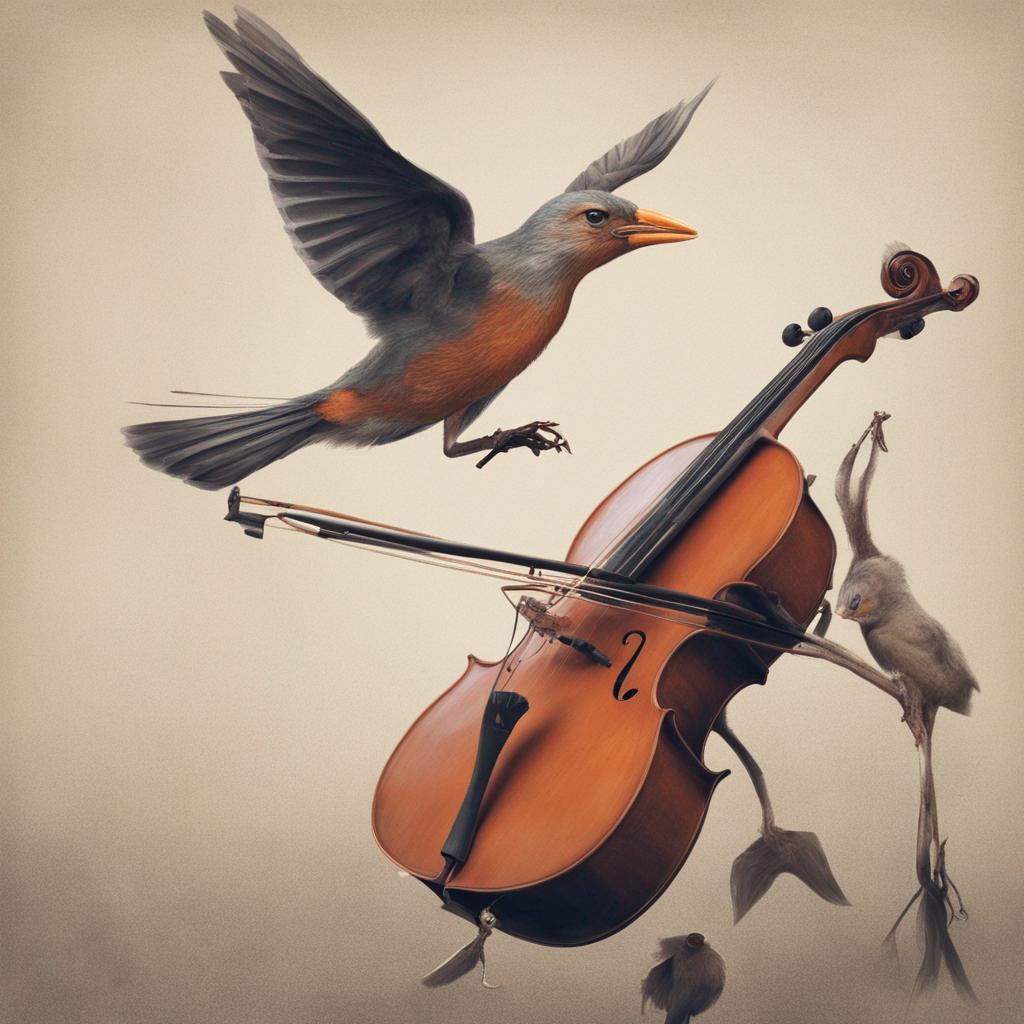} &
\includegraphics[width=1.8cm]{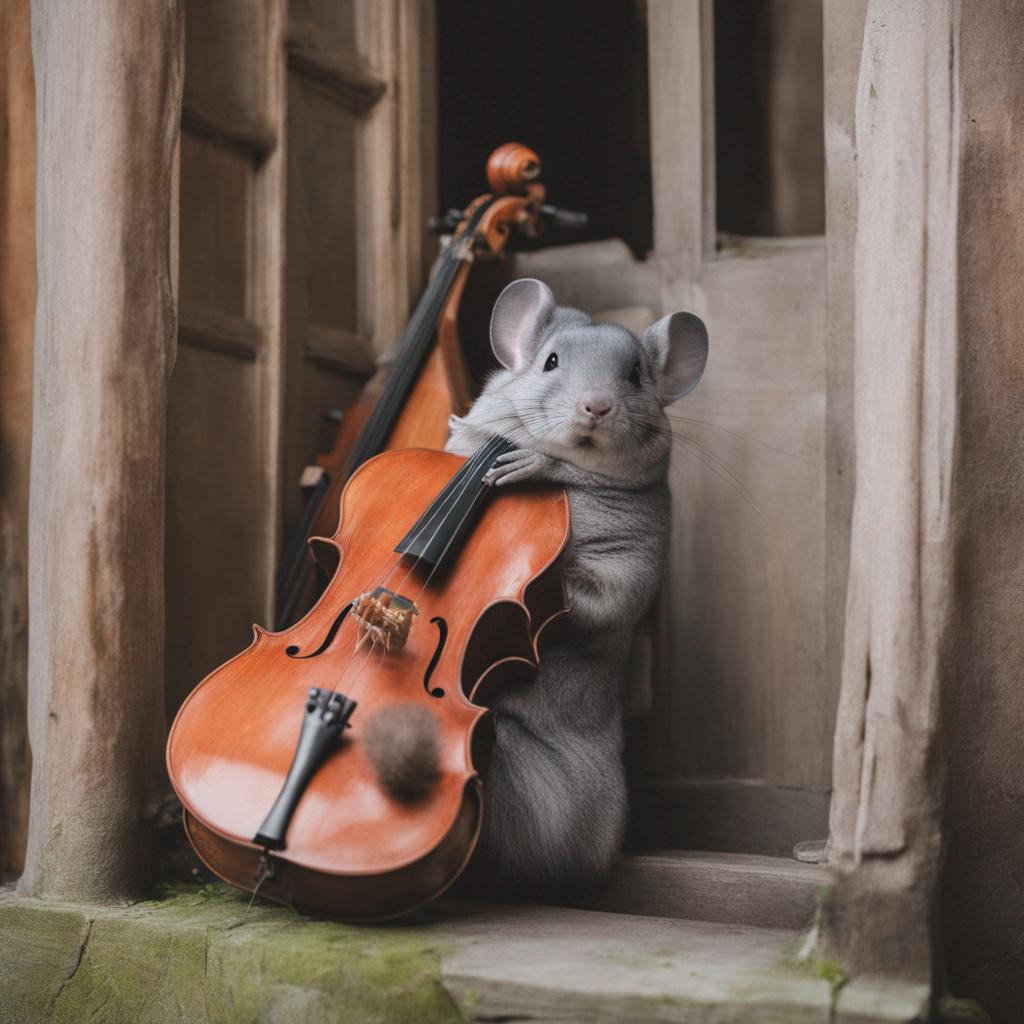} \\
& \scriptsize \hspace{0.35cm} \raisebox{-0.17cm}{\includegraphics[height=0.5cm]{Figures/equalizer.png}} Cello 
& \scriptsize \hspace{0.35cm} \raisebox{-0.17cm}{\includegraphics[height=0.5cm]{Figures/equalizer.png}} Cello
& \scriptsize \hspace{0.35cm} \raisebox{-0.17cm}{\includegraphics[height=0.5cm]{Figures/equalizer.png}} Cello 
& \scriptsize \hspace{0.35cm} \raisebox{-0.17cm}{\includegraphics[height=0.5cm]{Figures/equalizer.png}} Cello \\

\end{tabular}
\vspace{0.3cm}
\caption{Image to Audio Retrieval in IS3+.}
\label{fig:audio_retrieval_is3+}
\end{figure}

\begin{figure}[ht]
\centering
\begin{tabular}{m{1.8cm} | m{1.8cm} m{1.8cm} m{1.8cm} m{1.8cm}}
\multicolumn{1}{c|}{\textbf{Query Audio}} &
\multicolumn{4}{c}{\textbf{Retrieved Images}} \\
& \centering \textbf{Top 1} & \centering \textbf{Top 2} & \centering \textbf{Top 3} & \centering \textbf{Top 4} \\
\end{tabular}

\vspace{0.2em}

\begin{tabular}{m{1.8cm} | m{1.8cm} m{1.8cm} m{1.8cm} m{1.8cm}}
\includegraphics[width=1.8cm]{Figures/retrieval/IS3+/case1/accordion_female_10438_accordion.jpg} &
\includegraphics[width=1.8cm]{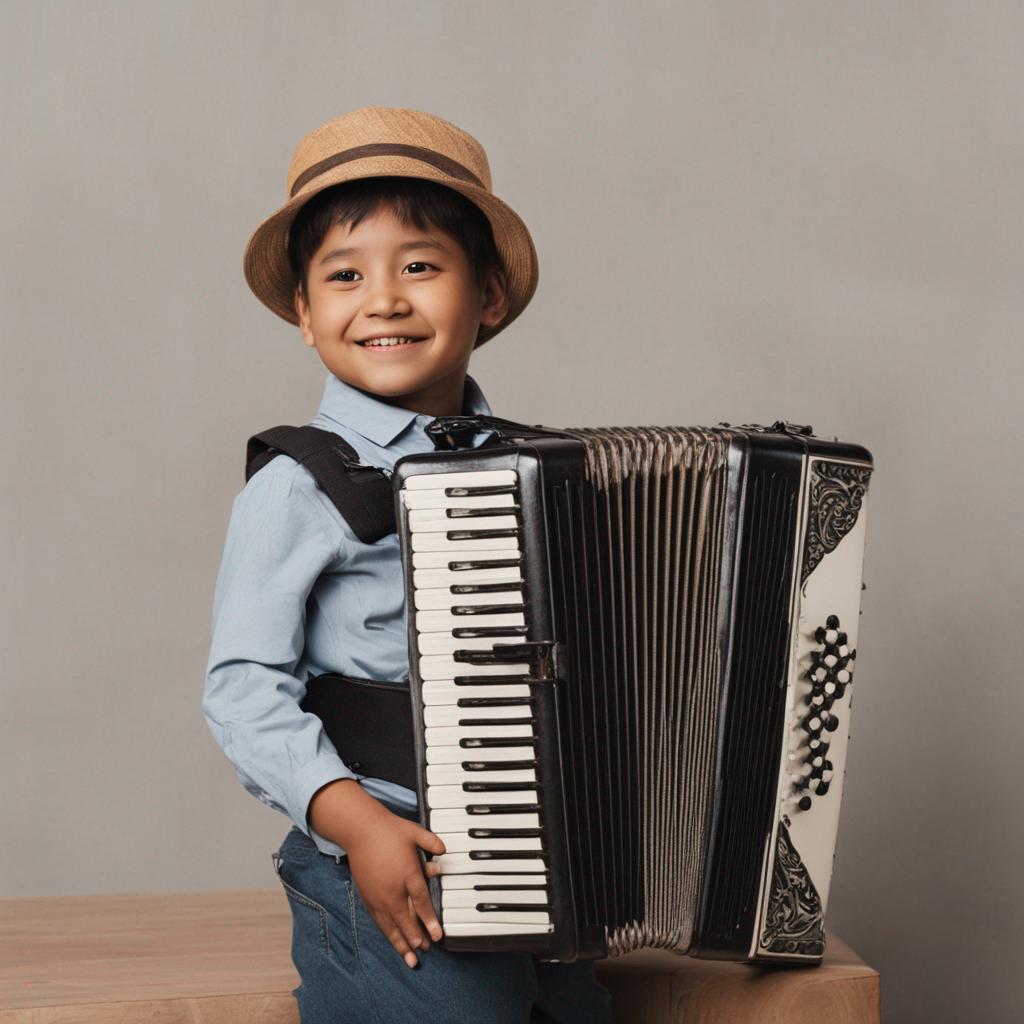} &
\includegraphics[width=1.8cm]{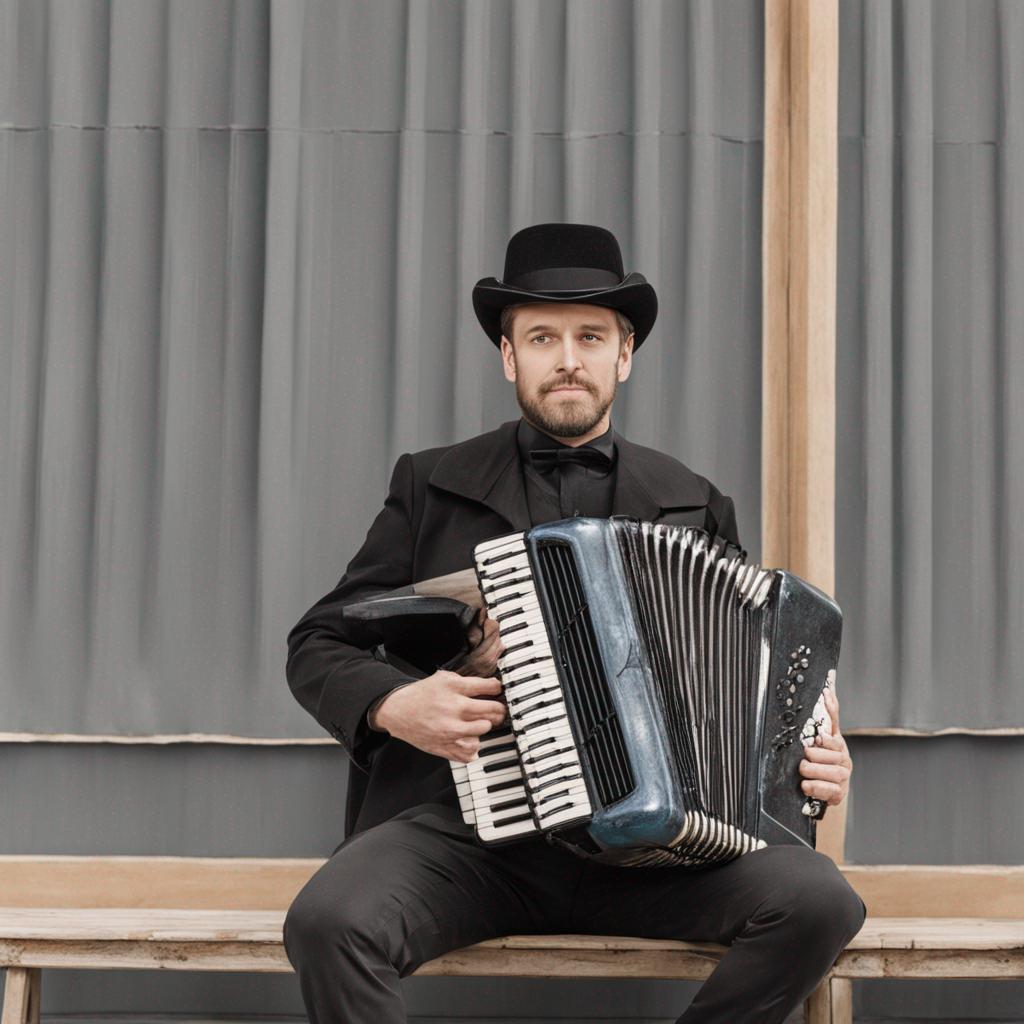} &
\includegraphics[width=1.8cm]{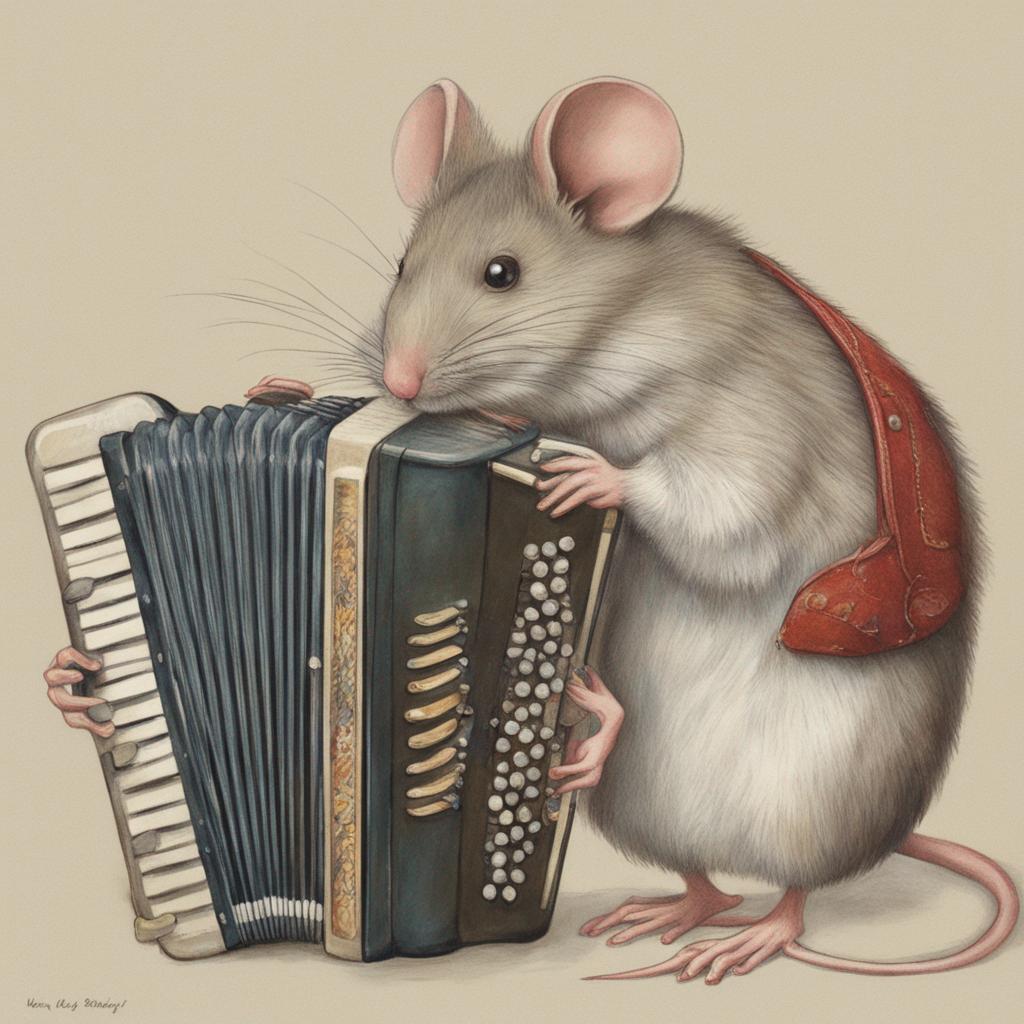} &
\includegraphics[width=1.8cm]{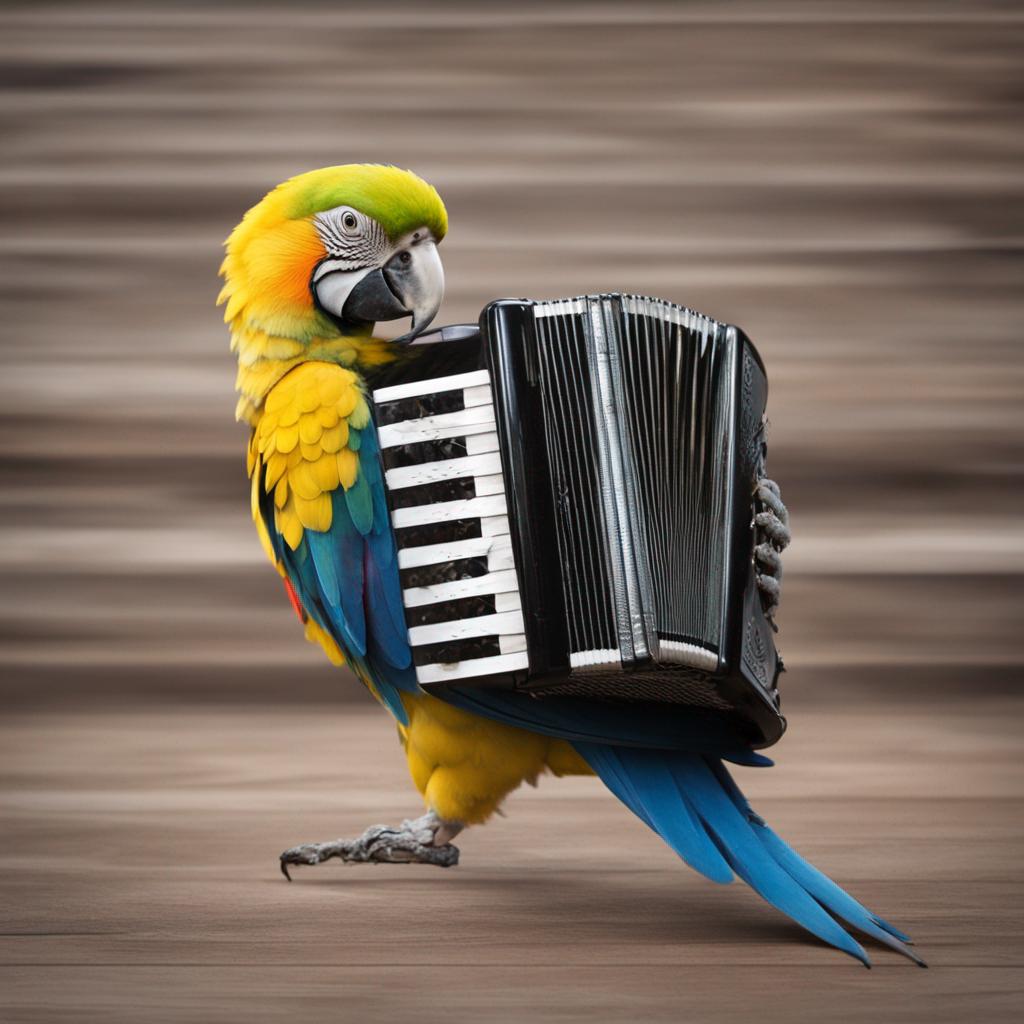} \\
\scriptsize \hspace{0.1cm}\raisebox{-0.17cm}{\includegraphics[height=0.5cm]{Figures/equalizer.png}} Accordion \\

\includegraphics[width=1.8cm]{Figures/retrieval/IS3+/case2/lion_chainsaw_756_lion.jpg} &
\includegraphics[width=1.8cm]{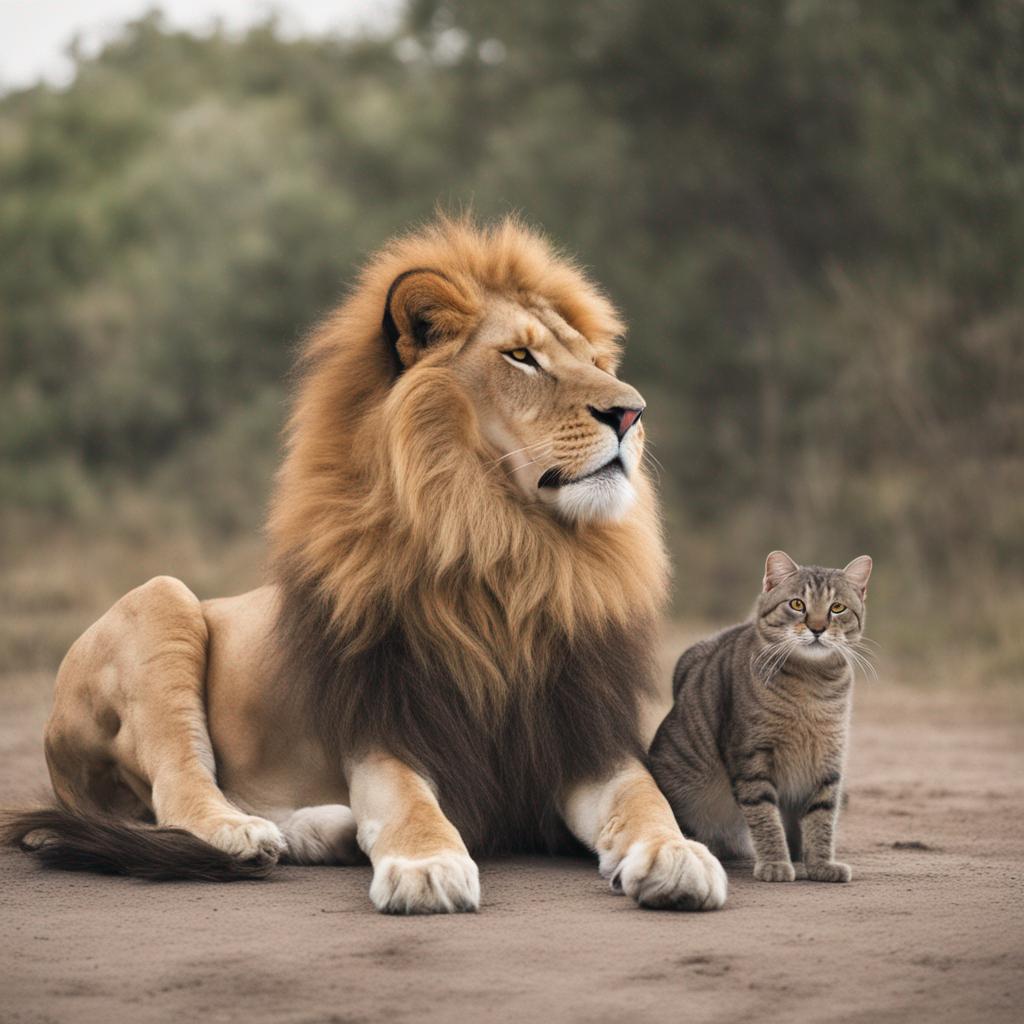} &
\includegraphics[width=1.8cm]{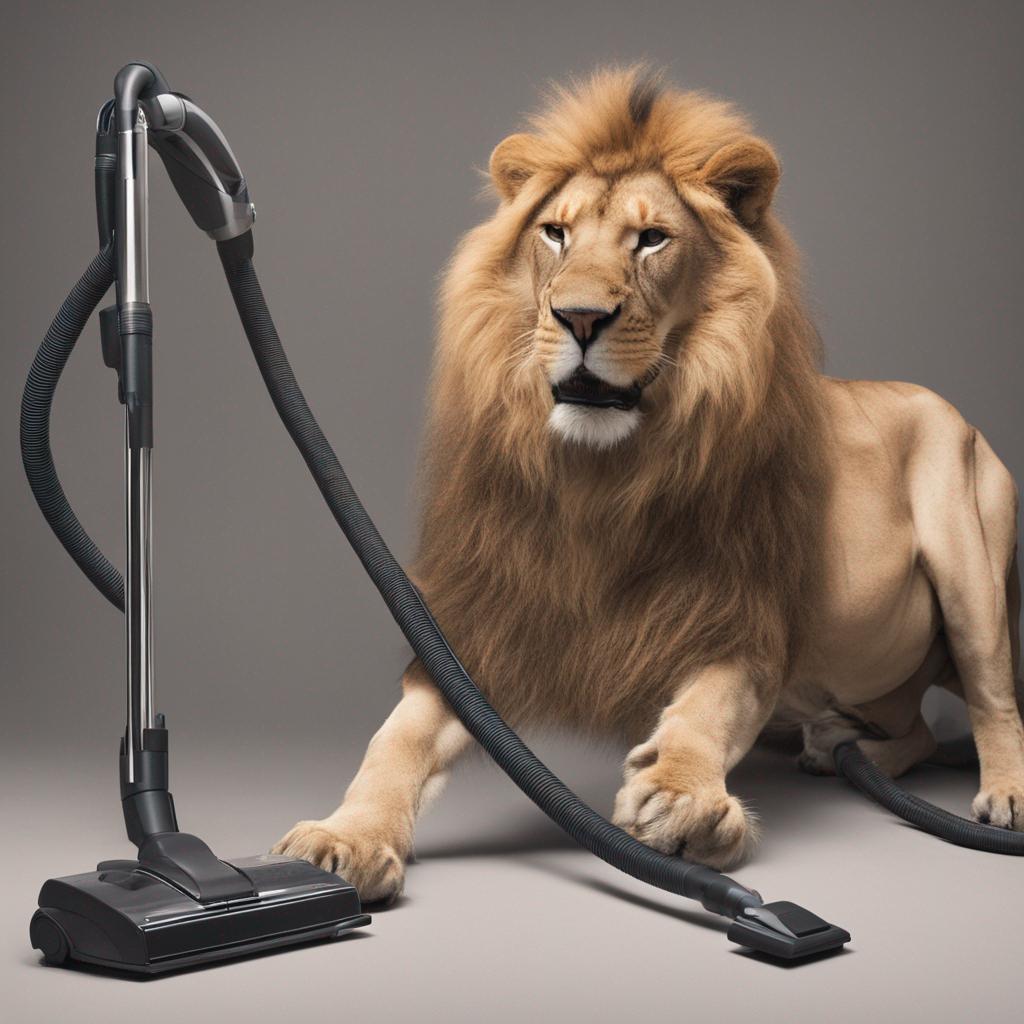} &
\includegraphics[width=1.8cm]{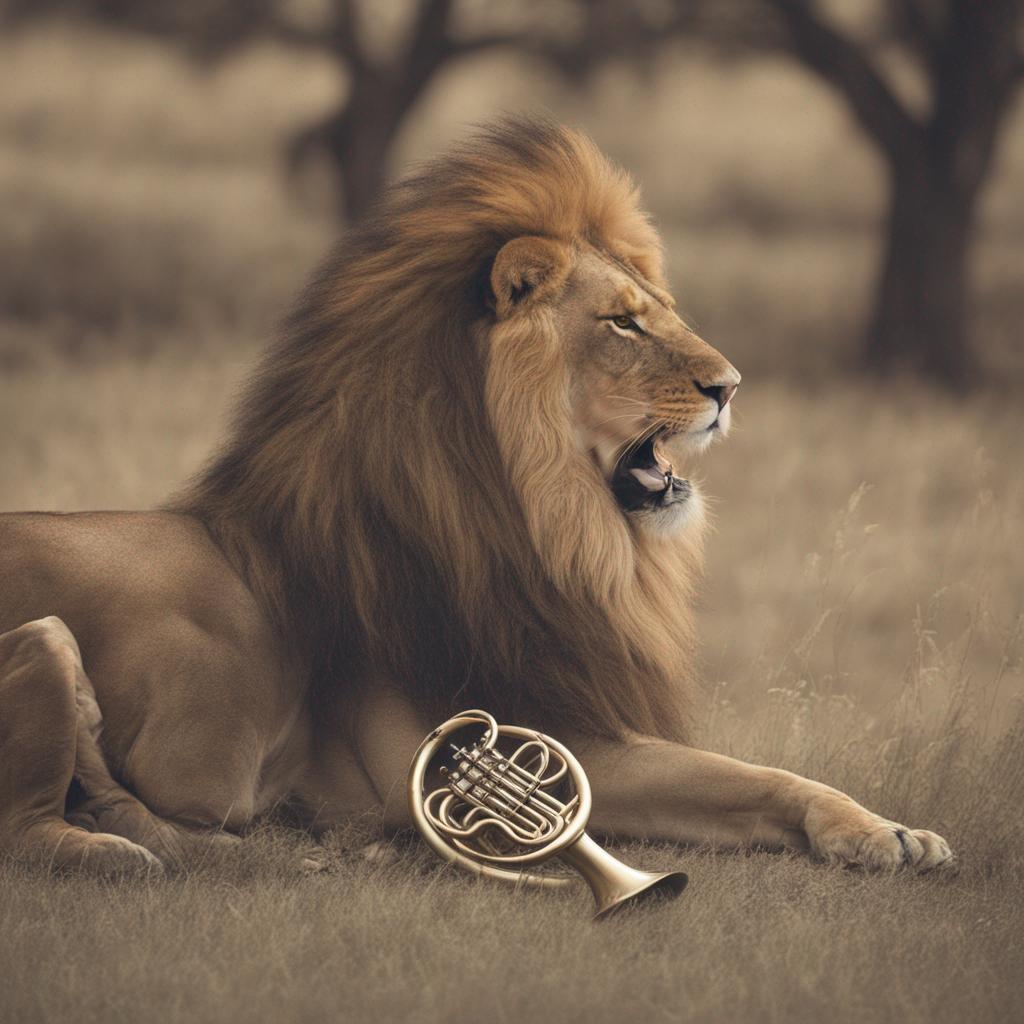} &
\includegraphics[width=1.8cm]{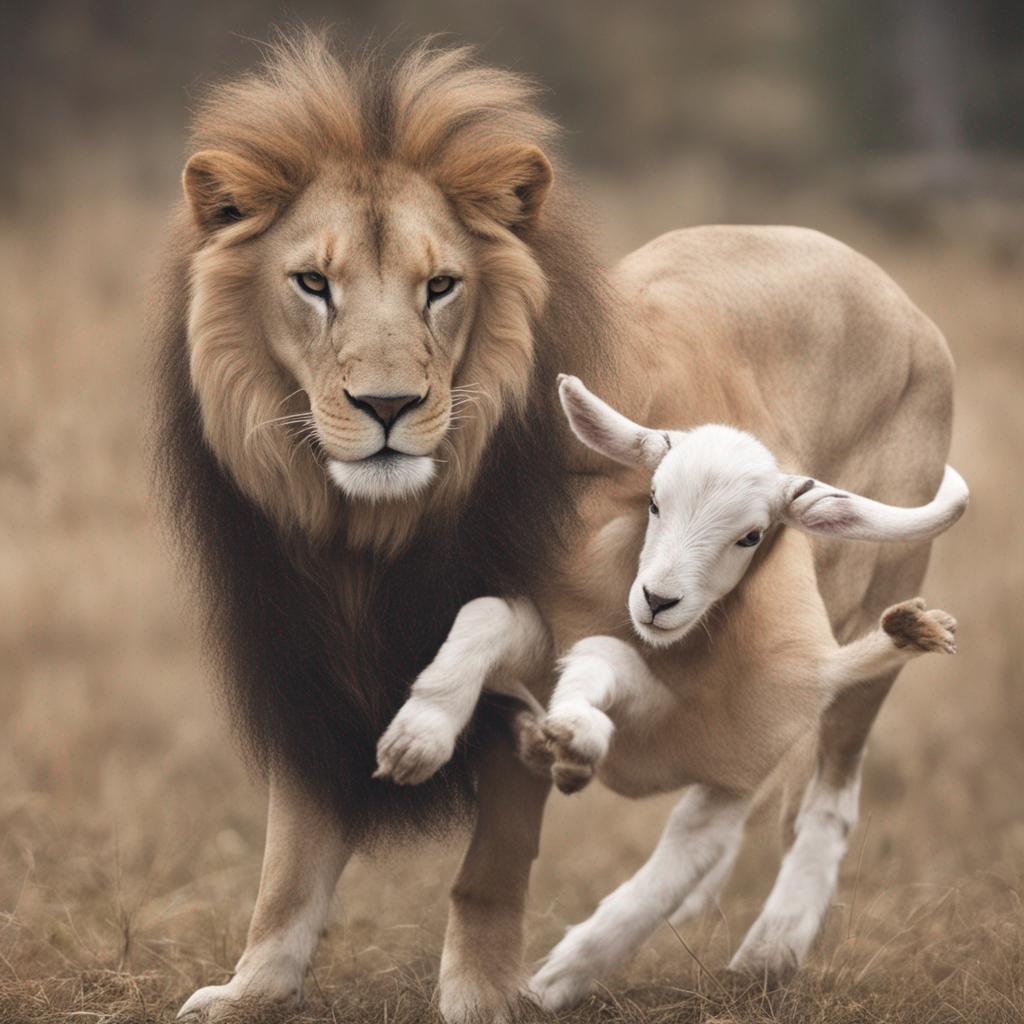} \\
\scriptsize \hspace{0.35cm} \raisebox{-0.17cm}{\includegraphics[height=0.5cm]{Figures/equalizer.png}} Lion \\

\includegraphics[width=1.8cm]{Figures/retrieval/IS3+/case3/chicken_bus_8050_chicken.jpg} &
\includegraphics[width=1.8cm]{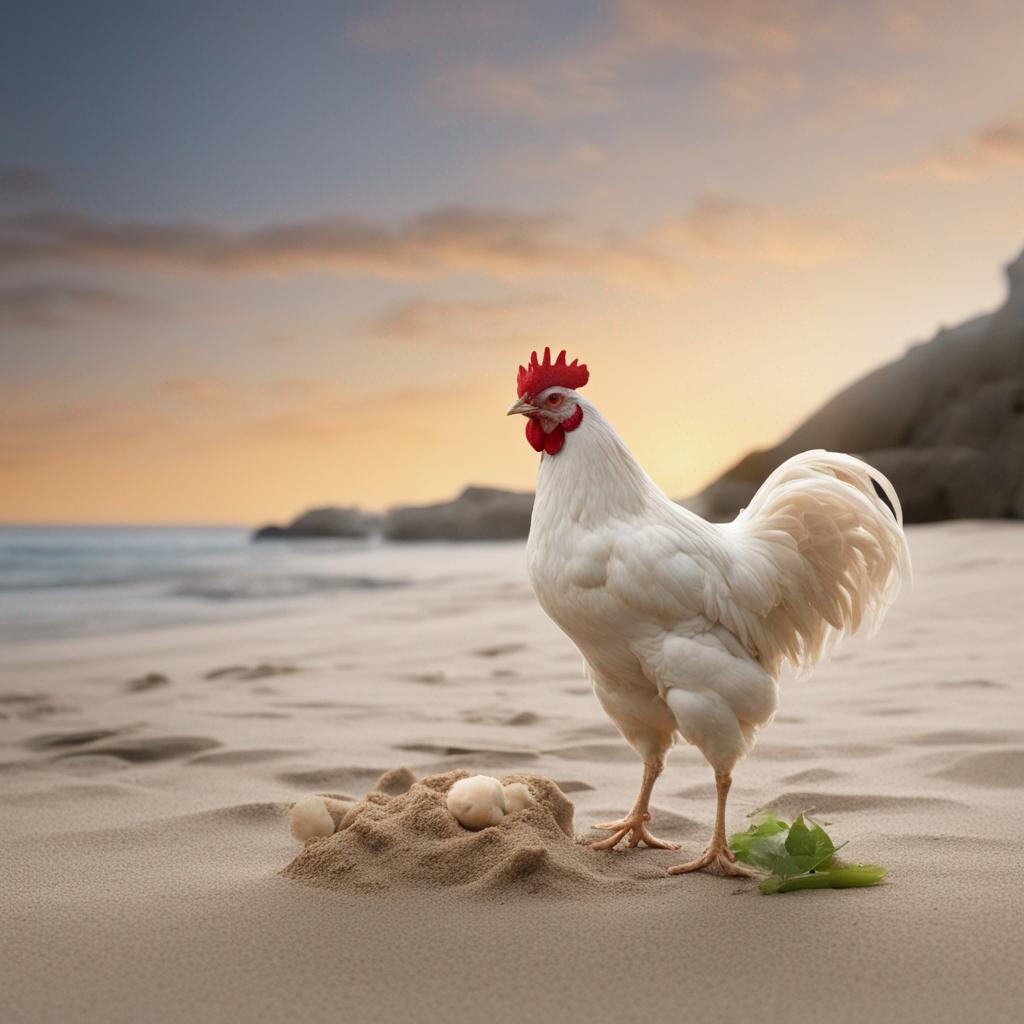} &
\includegraphics[width=1.8cm]{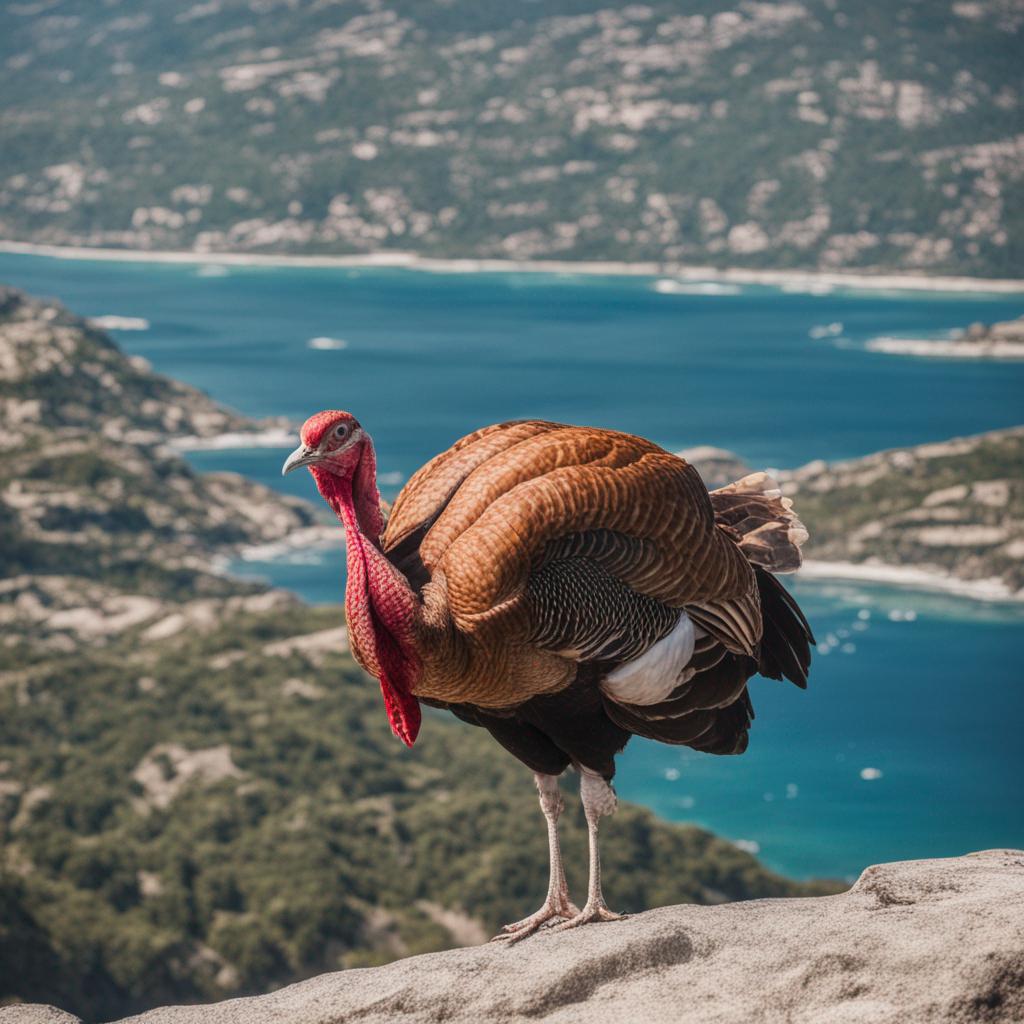} &
\includegraphics[width=1.8cm]{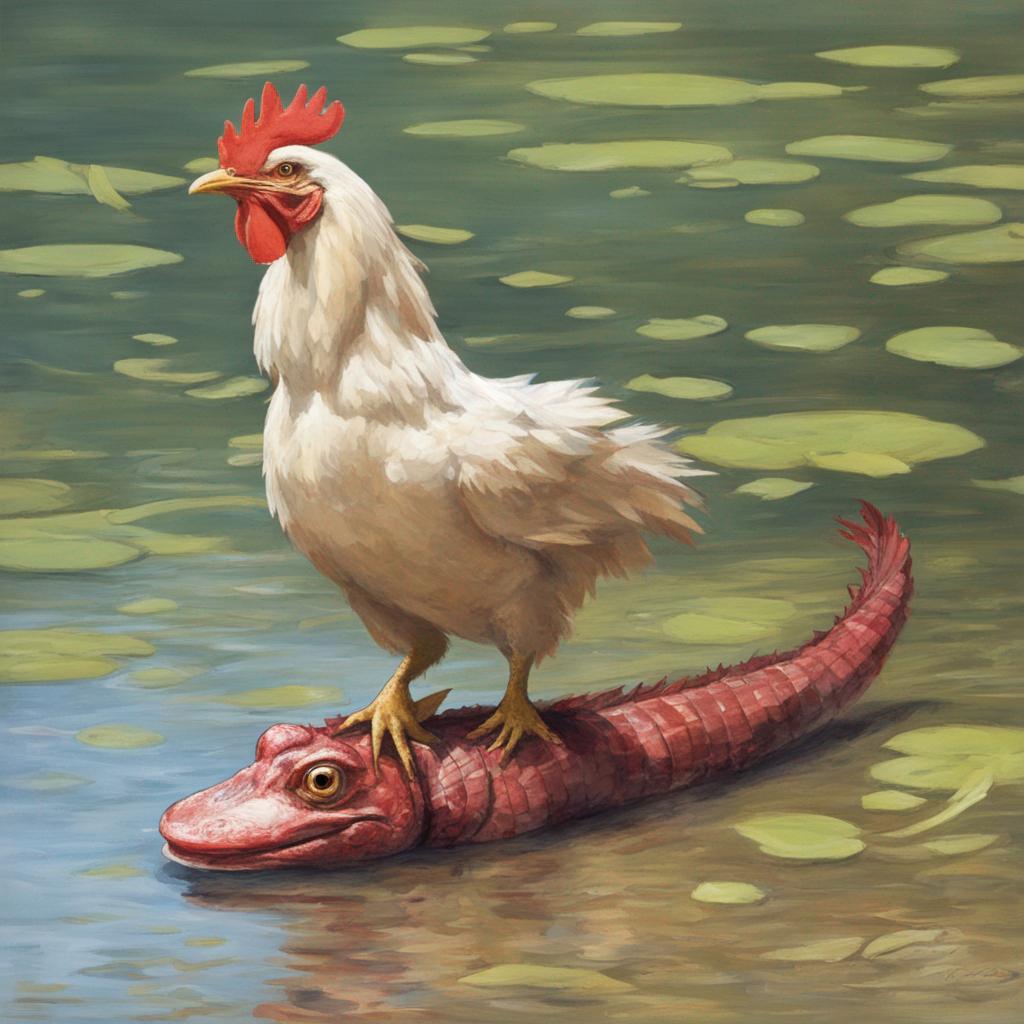} &
\includegraphics[width=1.8cm]{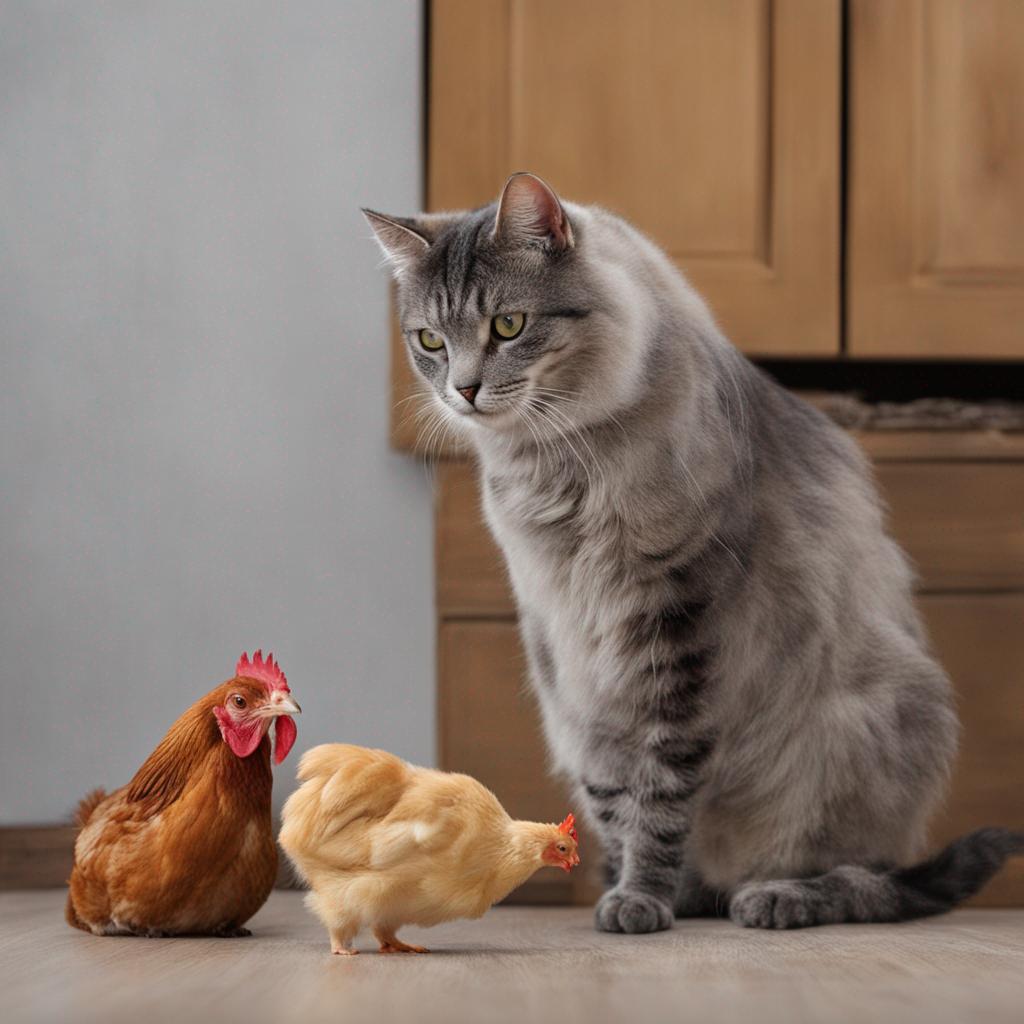} \\
\scriptsize \hspace{0.2cm} \raisebox{-0.17cm}{\includegraphics[height=0.5cm]{Figures/equalizer.png}} Chicken \\

\includegraphics[width=1.8cm]{Figures/retrieval/IS3+/case4/cello_male_3879_cello.jpg} &
\includegraphics[width=1.8cm]{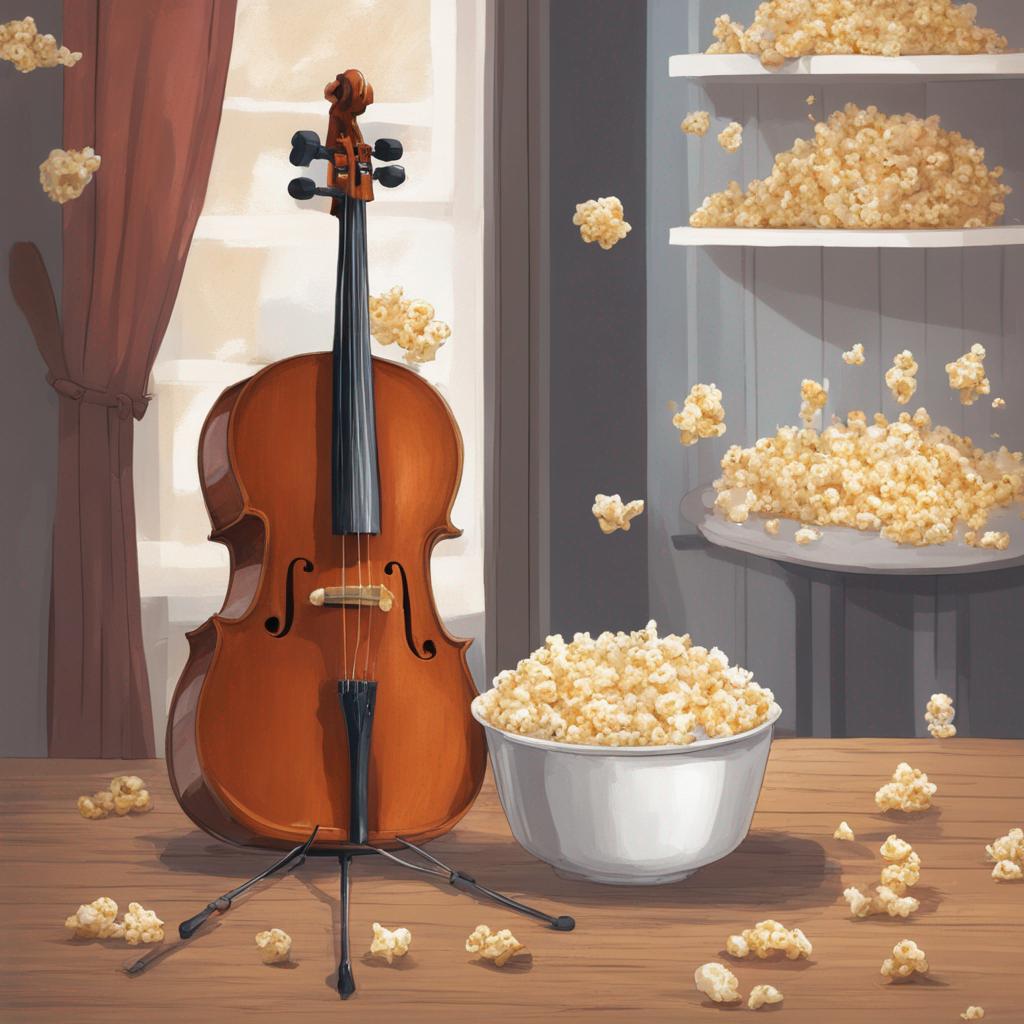} &
\includegraphics[width=1.8cm]{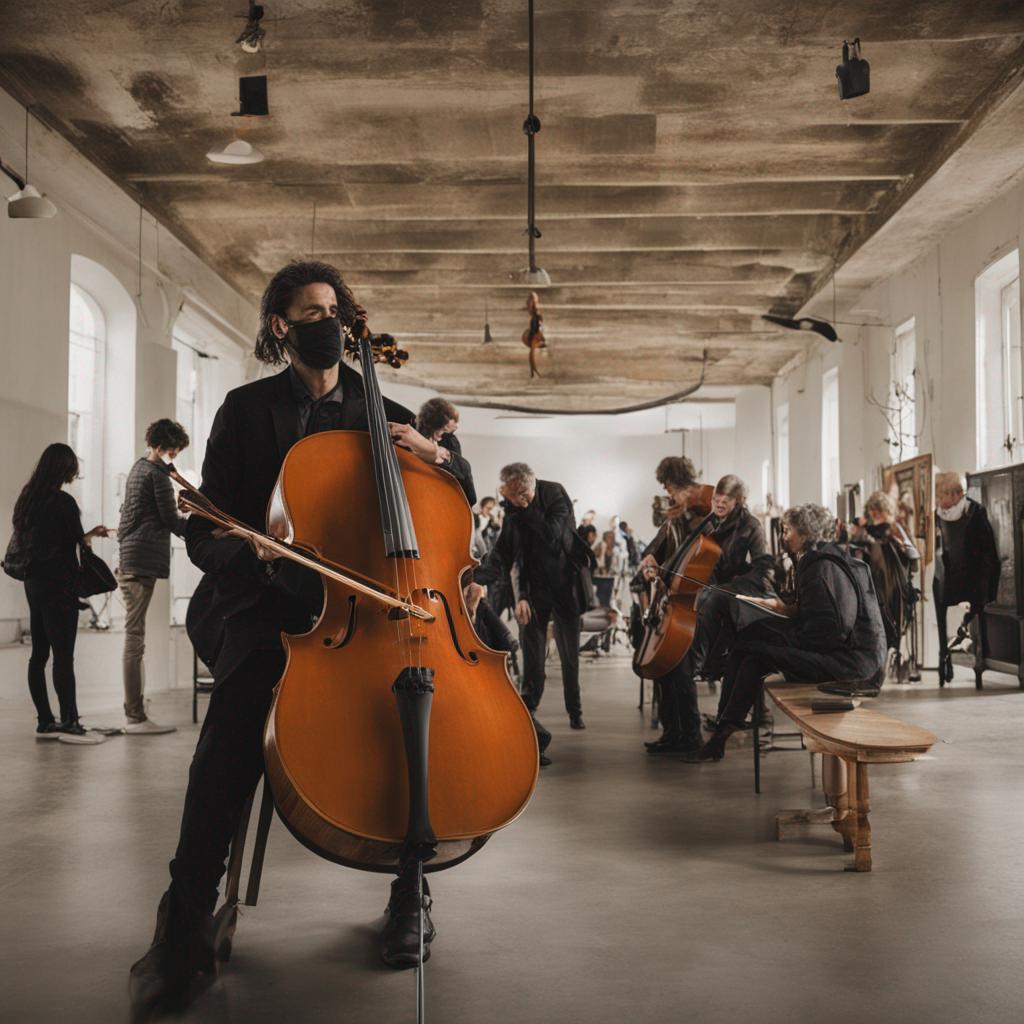} &
\includegraphics[width=1.8cm]{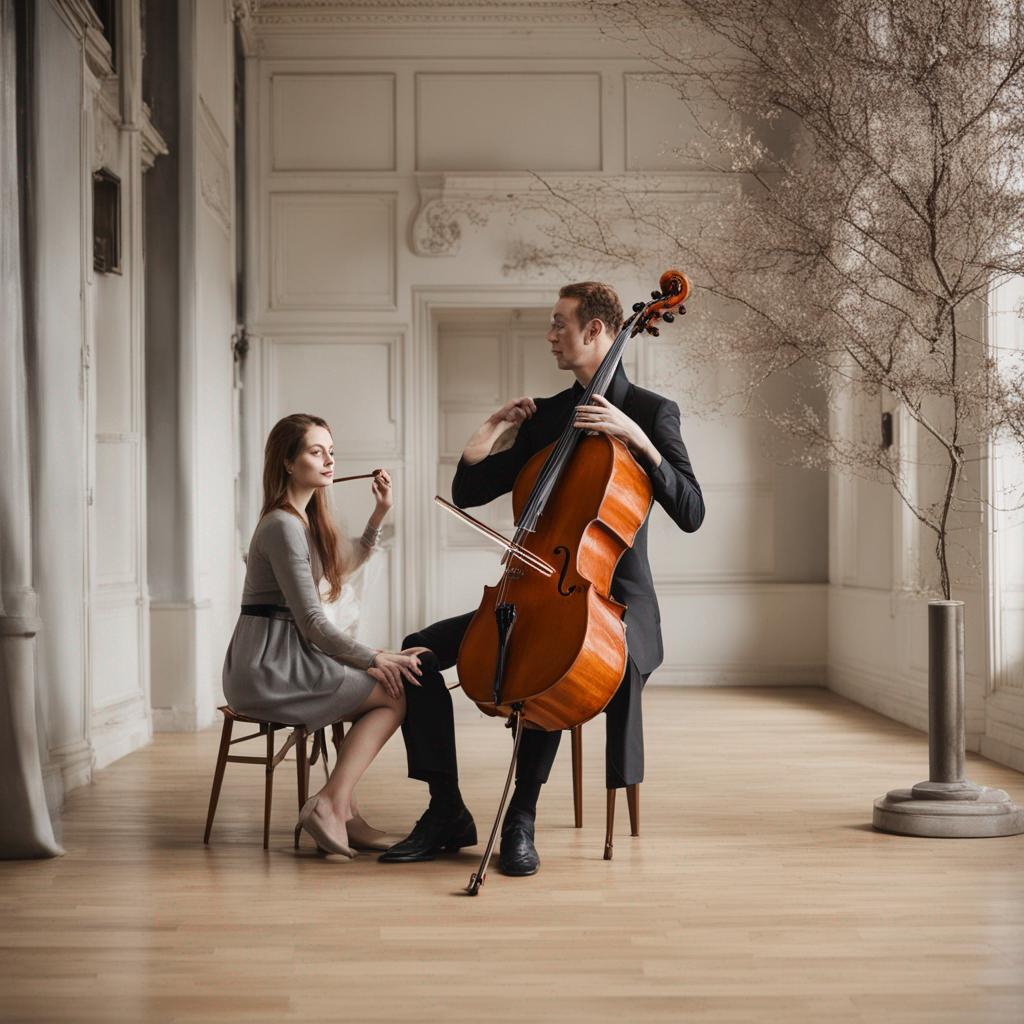} &
\includegraphics[width=1.8cm]{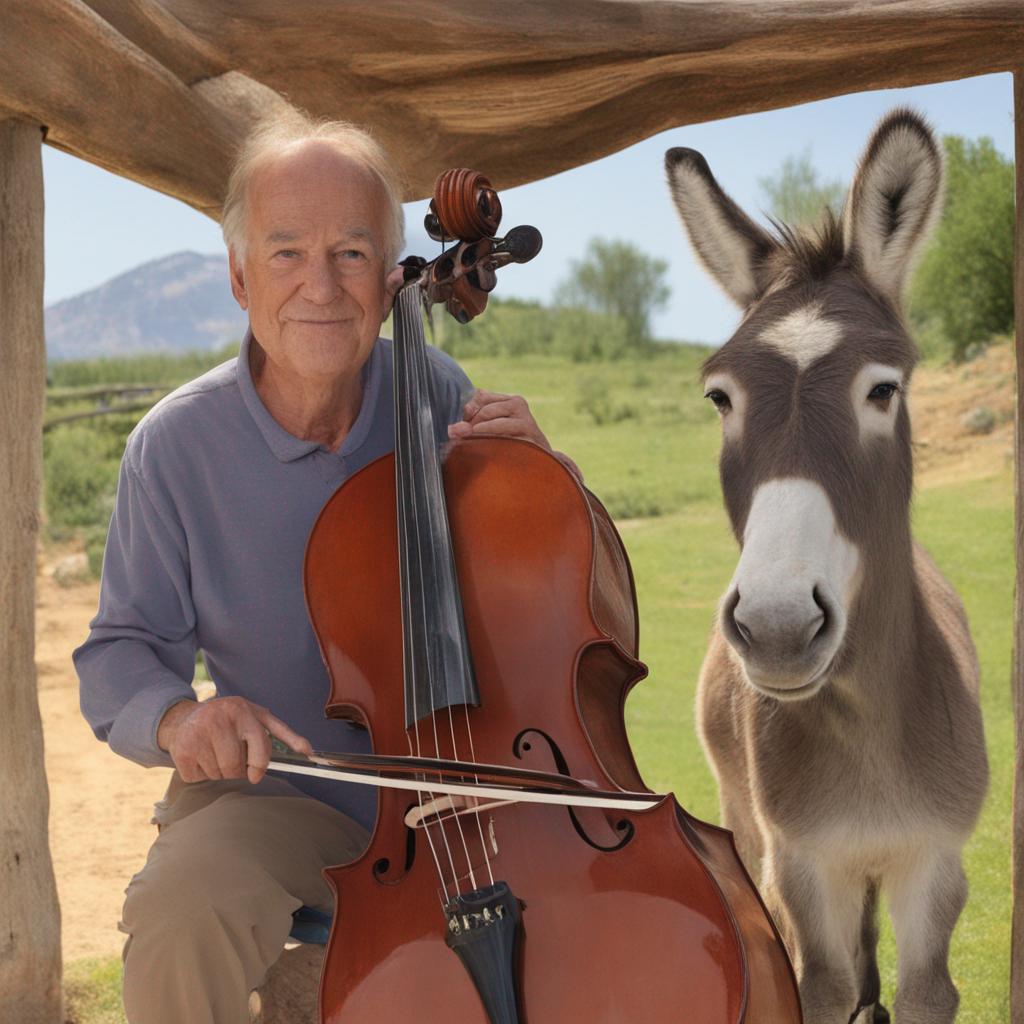} \\
\scriptsize \hspace{0.3cm} \raisebox{-0.17cm}{\includegraphics[height=0.5cm]{Figures/equalizer.png}} Cello \\
\end{tabular}

\caption{Audio to Image Retrieval in IS3+}
\label{fig:image_retrieval_is3+}
\end{figure}

After presenting the quantitative results, we now include some qualitative examples illustrating the cross-modal retrieval task. Specifically, we show, given an audio input, the top 4 images retrieved  by the model, and vice versa when doing image to audio.  
These results are shown in Figures \ref{fig:audio_retrieval_is3+} and \ref{fig:image_retrieval_is3+}.
In the case of audio, we specify its corresponding class in words and show its corresponding image. The audio samples (and images) are available in the project page: \href{https://xavijuanola.github.io/SSL-SaN/}{https://xavijuanola.github.io/SSL-SaN/} 

Figure \ref{fig:audio_retrieval_is3+} illustrates the good  performance of our model doing cross-modal retrieval from image to audio in IS3+. 
In all cases, except for one, the top four retrieved audio samples  are perfect matches. 
It is  interesting to note that in the failure case, where the target image contains a \textit{bus} and a \textit{chicken}, the top four retrieved audio samples correspond to \textit{bus}, except for one, that  contains a \textit{tractor}, another vehicle that can produce a very similar sound. 

A similar example can be seen in Figure \ref{fig:image_retrieval_is3+}, where all retrieved images are correct except for one (third row),  where the query audio is from a \textit{chicken}, and the second retrieved image is of a \textit{turkey}. These two animals produce similar sounds, which demonstrates how our model understands the sources of different sounds.

\begin{figure}[ht]
\centering
\begin{tabular}{m{1.8cm} | m{1.8cm} m{1.8cm} m{1.8cm} m{1.8cm}}
\multicolumn{1}{c|}{\textbf{Query Image}} &
\multicolumn{4}{c}{\textbf{Retrieved Audios}} \\
& \centering \textbf{Top 1} & \centering \textbf{Top 2} & \centering \textbf{Top 3} & \centering \textbf{Top 4} \\
\end{tabular}

\vspace{0.2em}

\begin{tabular}{m{1.8cm} | m{1.8cm} m{1.8cm} m{1.8cm} m{1.8cm}}
\includegraphics[width=1.8cm]{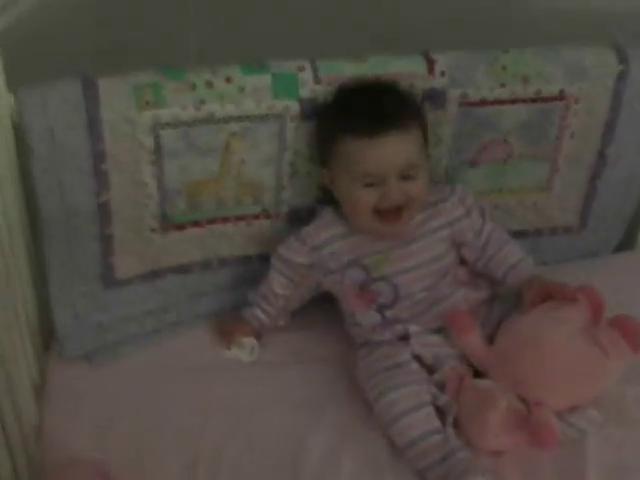} &
\includegraphics[width=1.8cm]{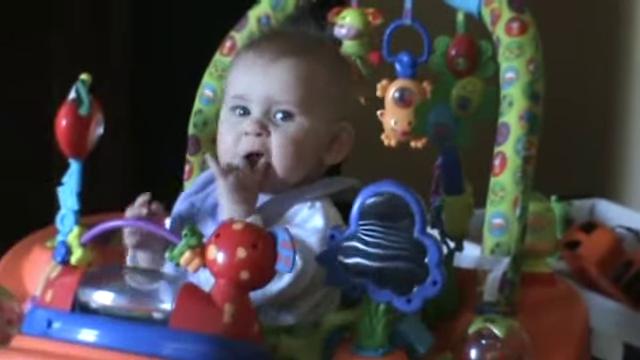} &
\includegraphics[width=1.8cm]{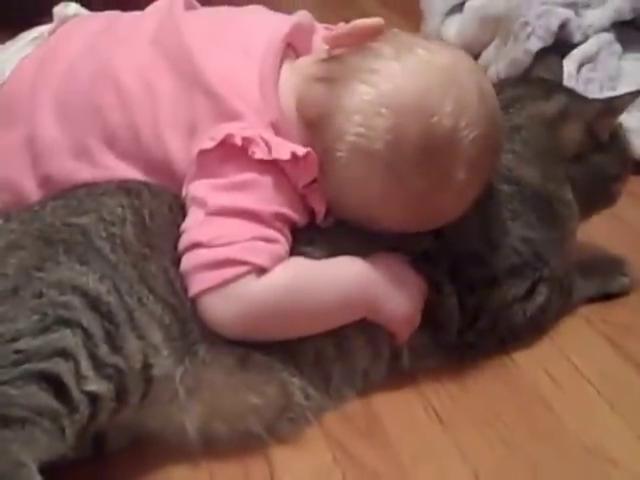} &
\includegraphics[width=1.8cm]{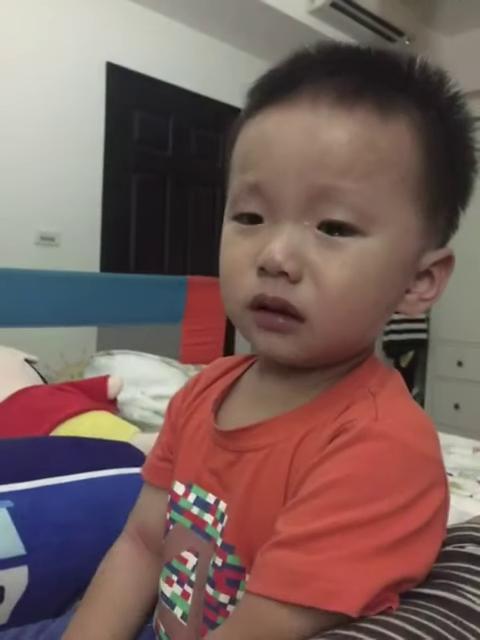} &
\includegraphics[width=1.5cm]{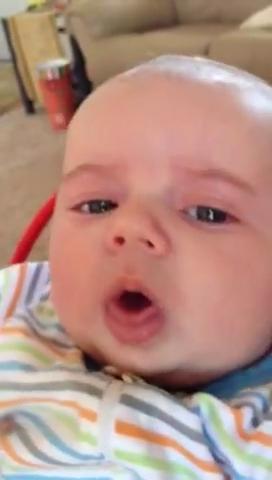} \\
& \scriptsize people babbling 
& \scriptsize people babbling 
& \scriptsize people coughing 
& \scriptsize people coughing \\

\includegraphics[width=1.8cm]{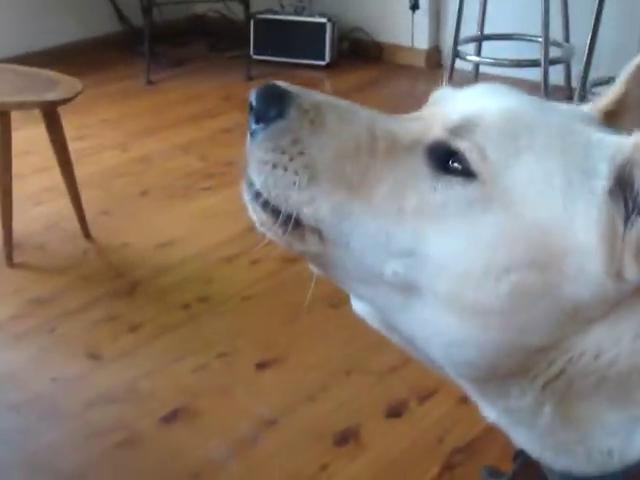} &
\includegraphics[width=1.8cm]{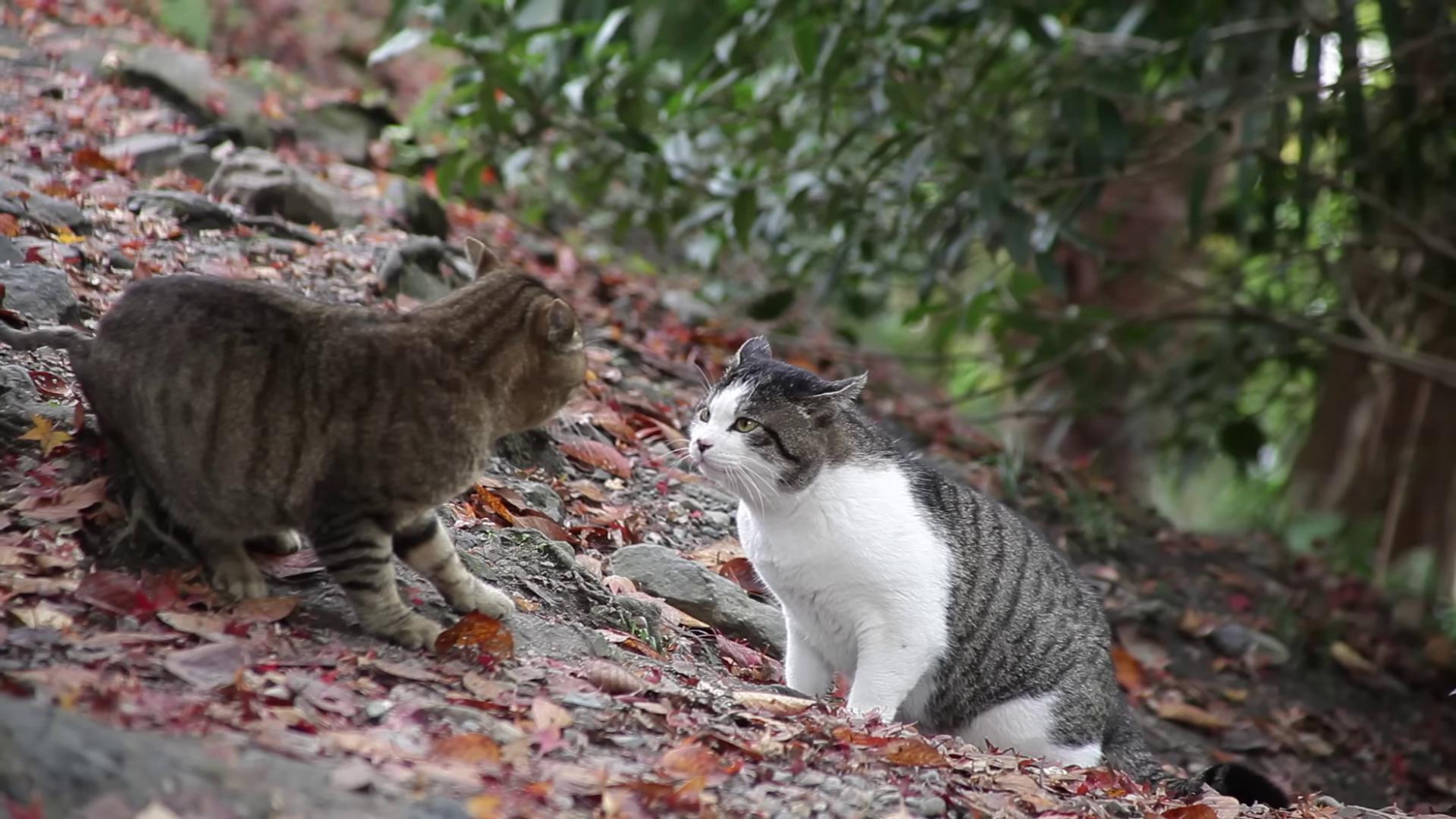} &
\includegraphics[width=1.8cm]{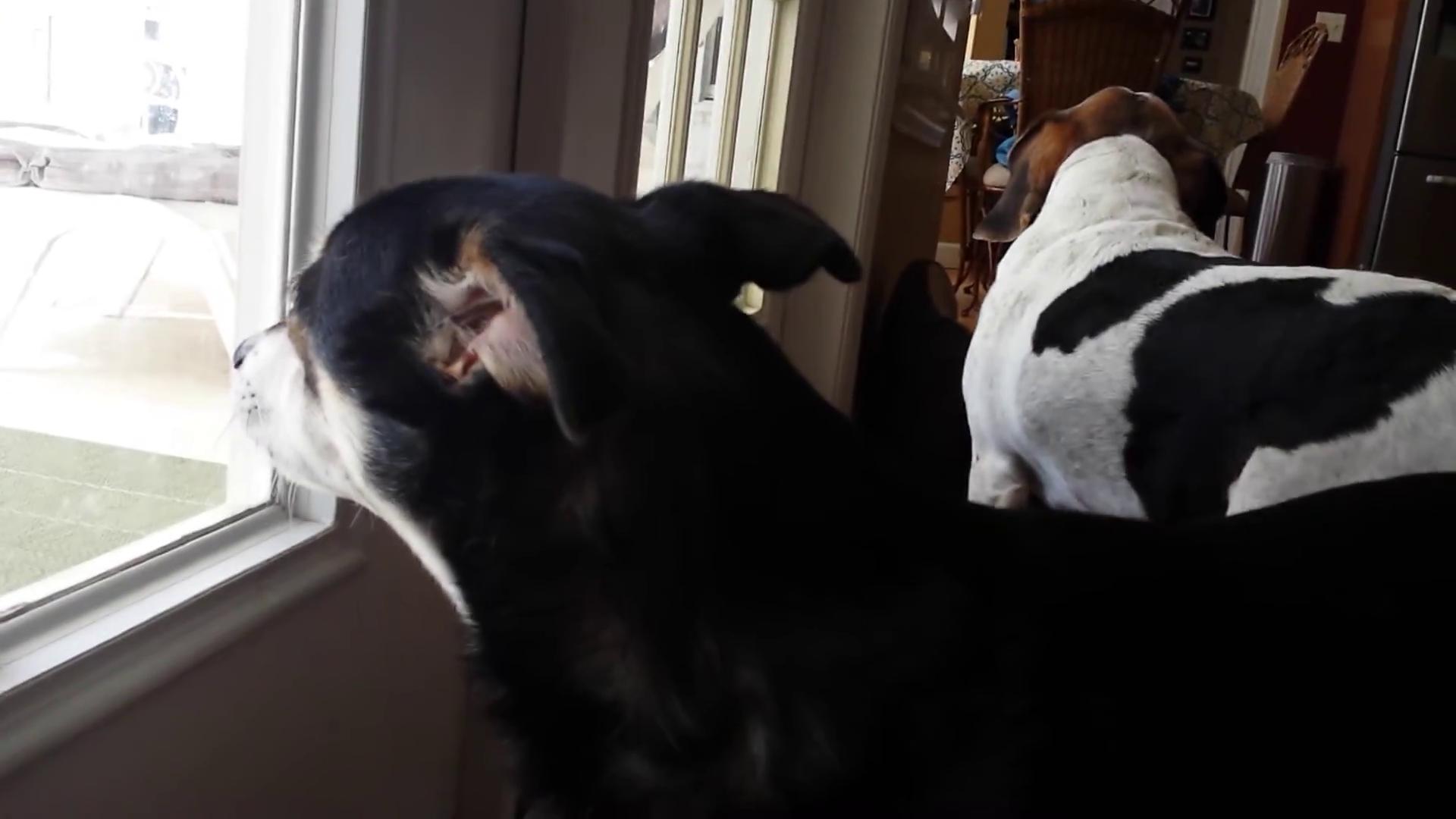} &
\includegraphics[width=1.8cm]{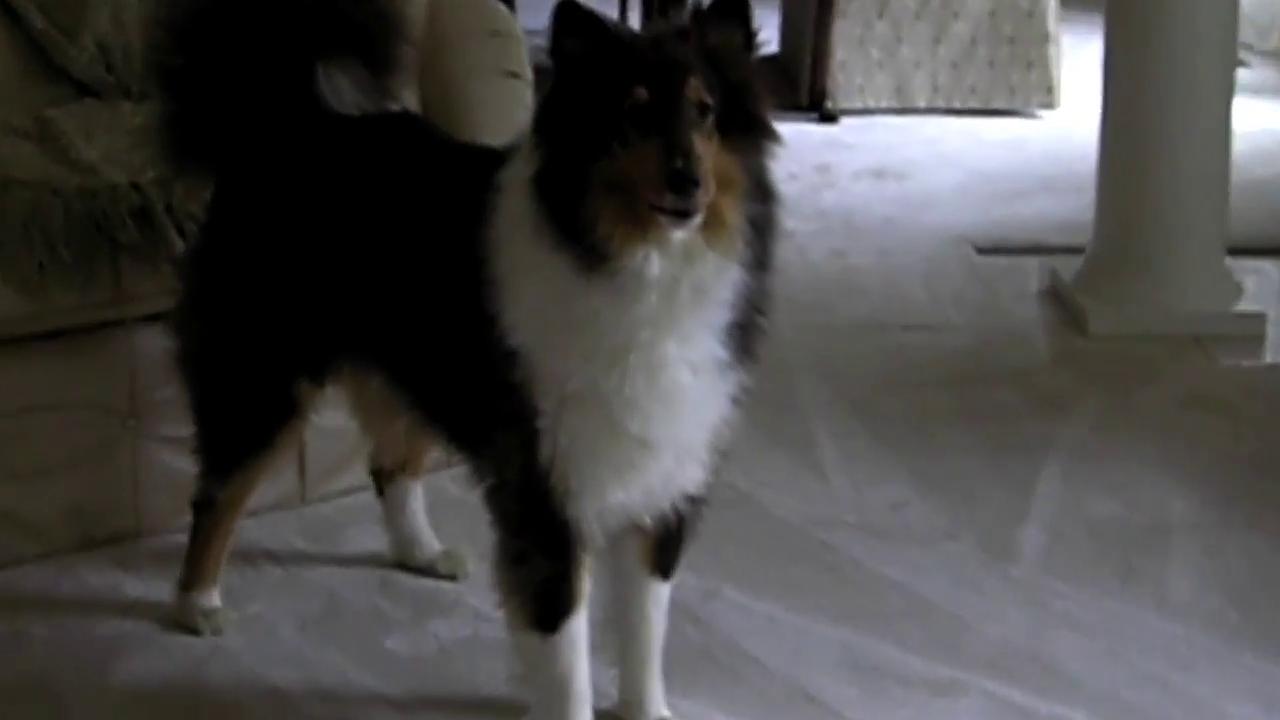} &
\includegraphics[width=1.8cm]{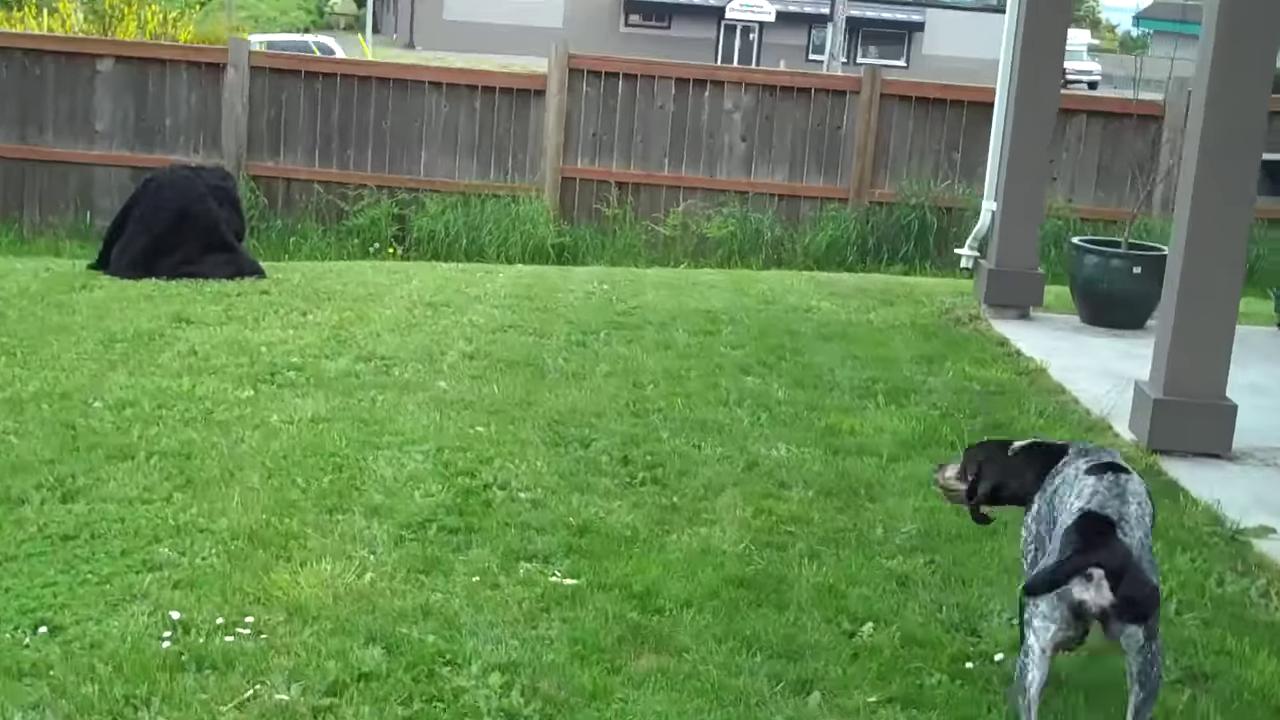} \\
& \scriptsize cat caterwauling 
& \scriptsize dog howling
& \scriptsize coyote howling 
& \scriptsize dog baying \\

\includegraphics[width=1.8cm]{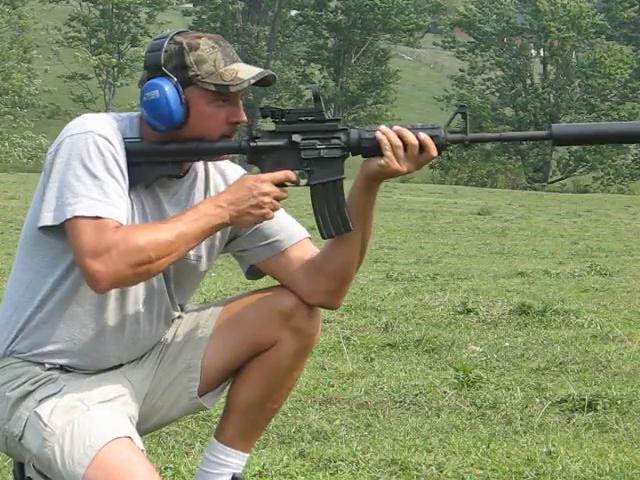} &
\includegraphics[width=1.8cm]{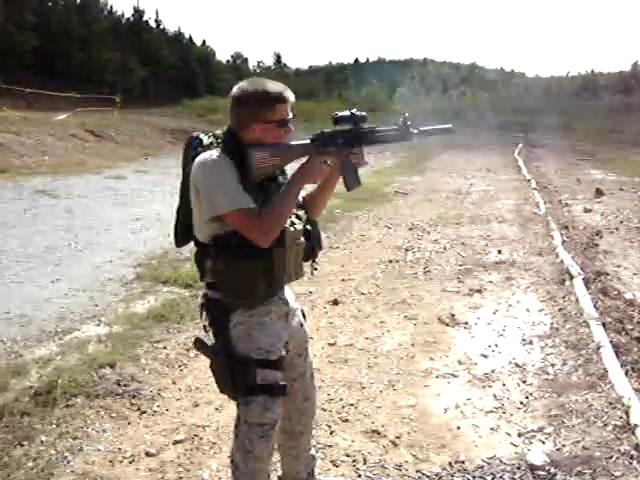} &
\includegraphics[width=1.8cm]{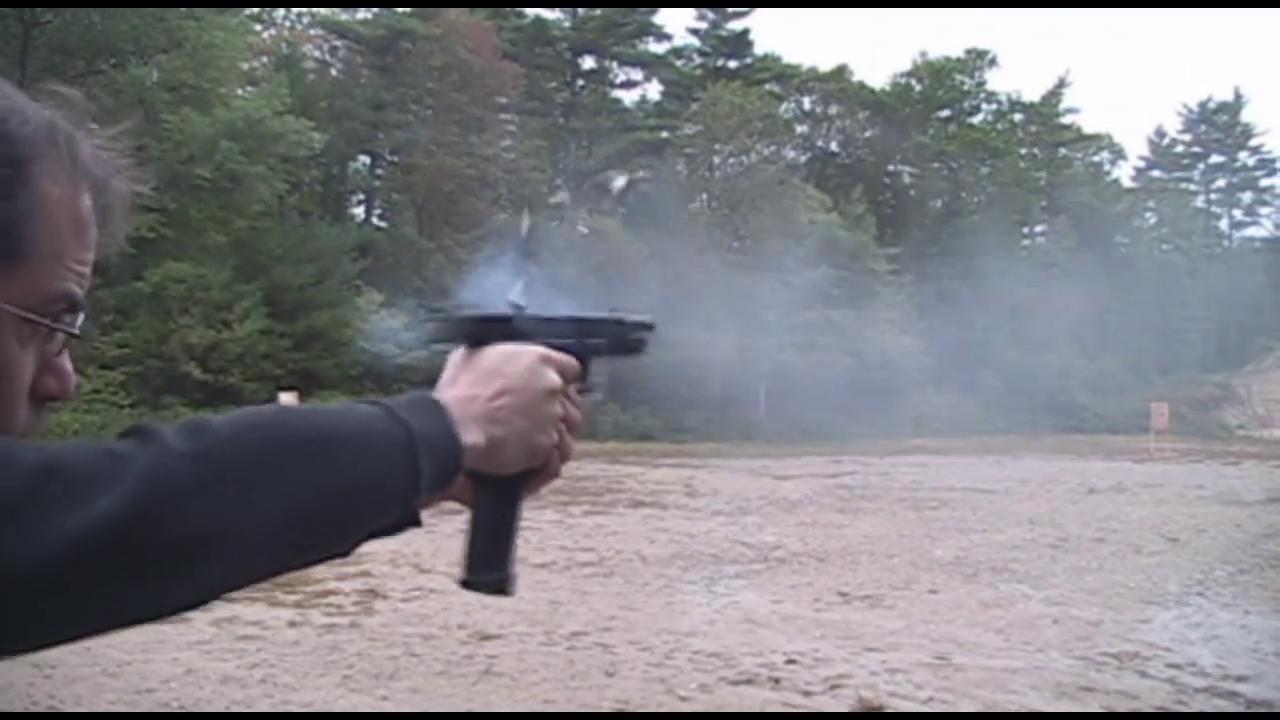} &
\includegraphics[width=1.8cm]{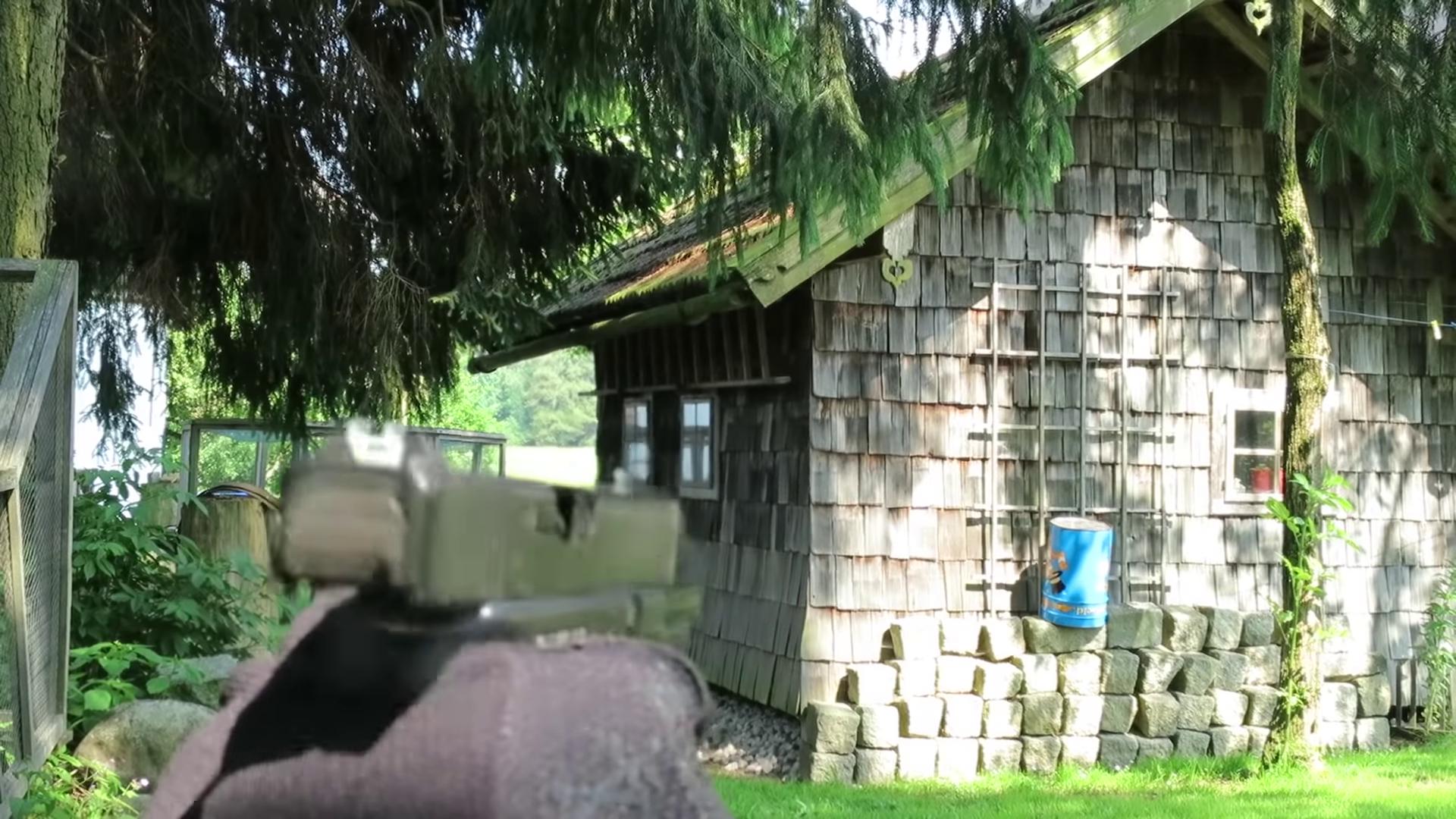} &
\includegraphics[width=1.8cm]{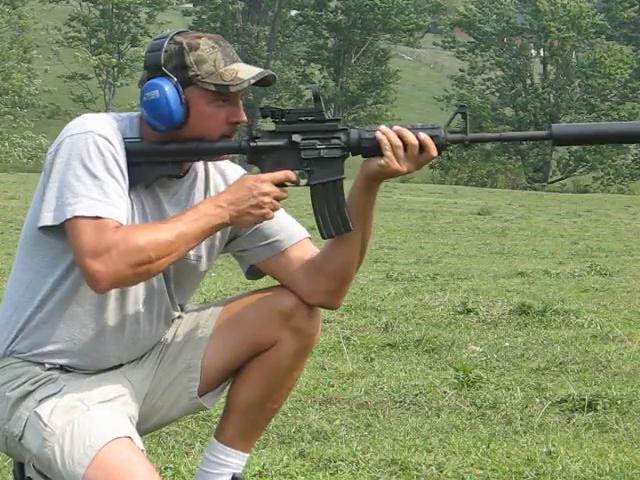} \\
& \scriptsize machine gun shooting 
& \scriptsize machine gun shooting
& \scriptsize machine gun shooting 
& \scriptsize machine gun shooting \\

\includegraphics[width=1.8cm]{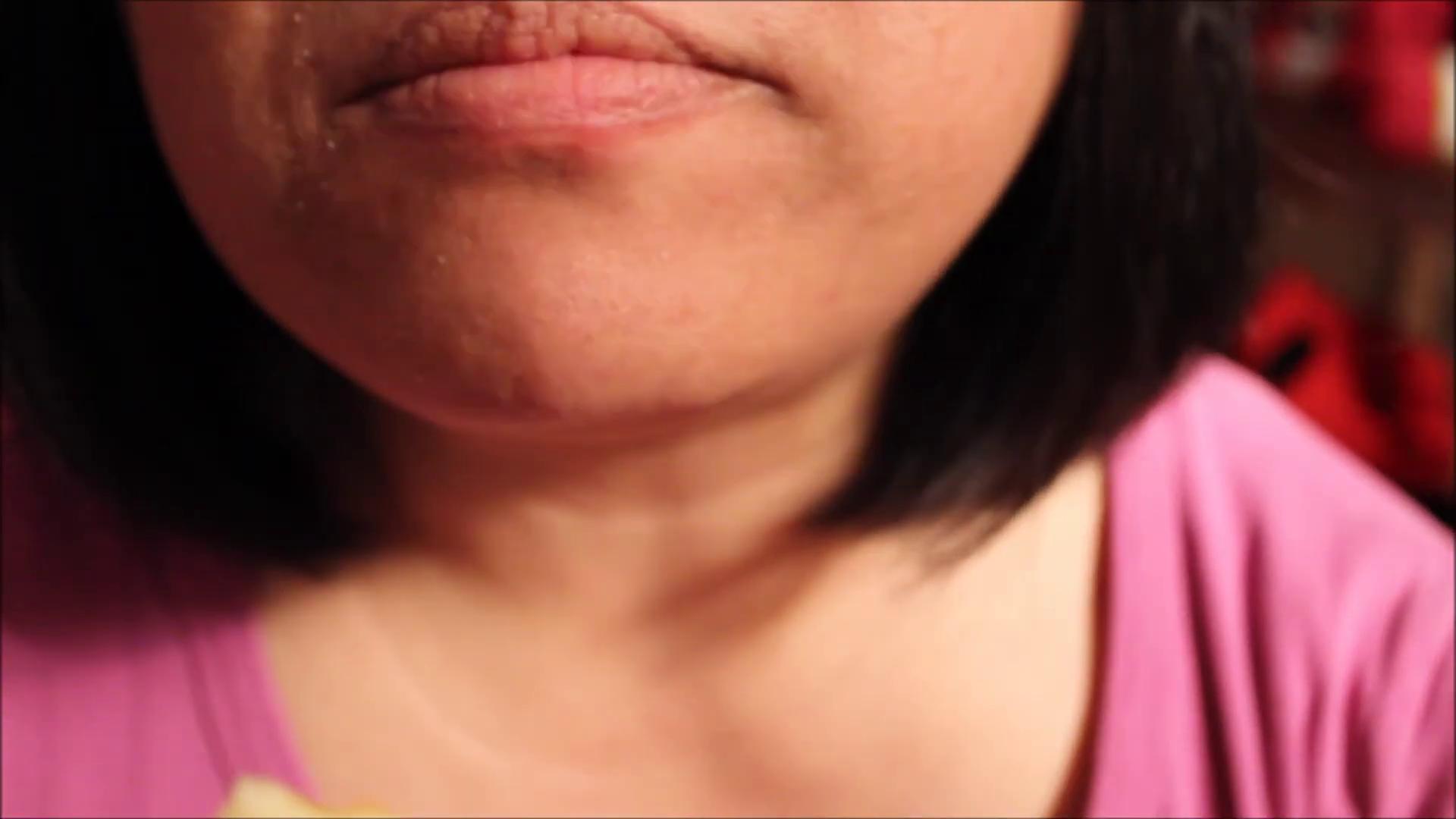} &
\includegraphics[width=1.8cm]{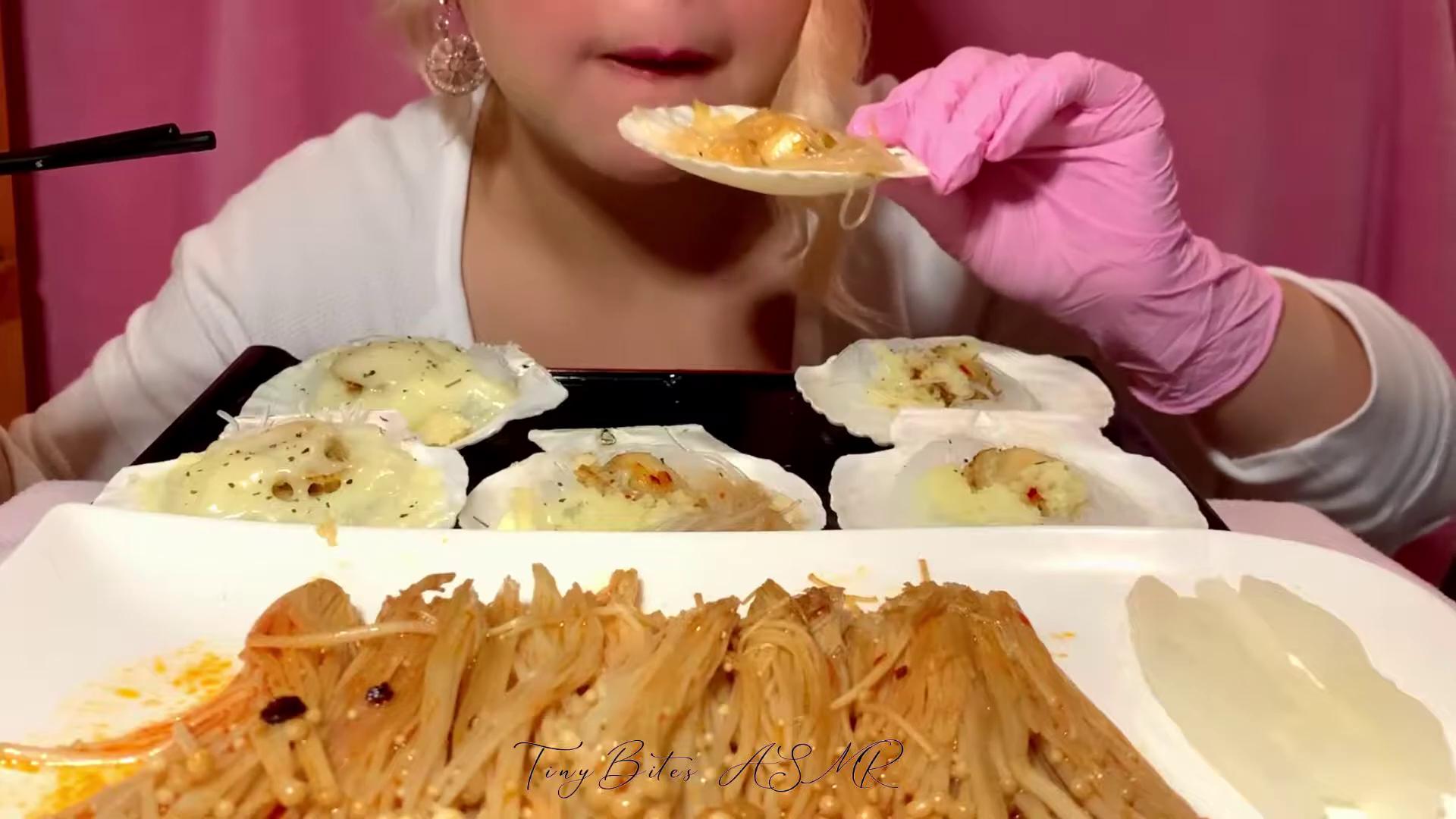} &
\includegraphics[width=1.8cm]{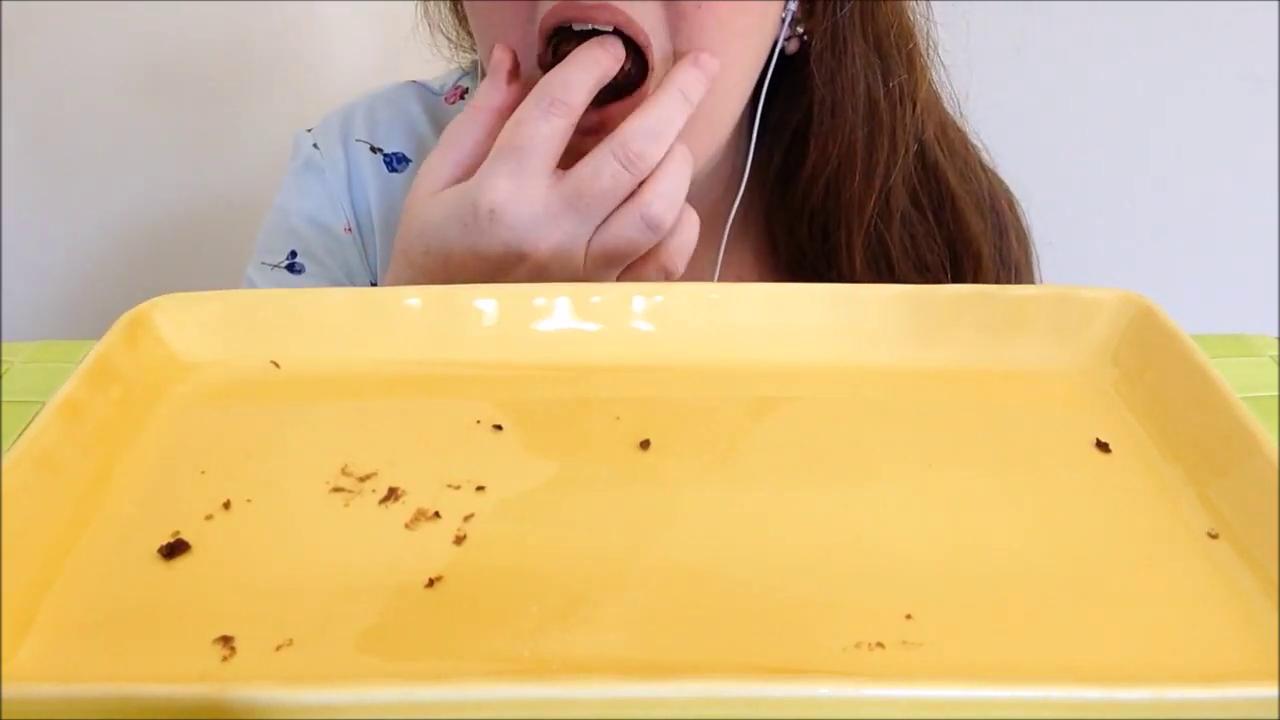} &
\includegraphics[width=1.8cm]{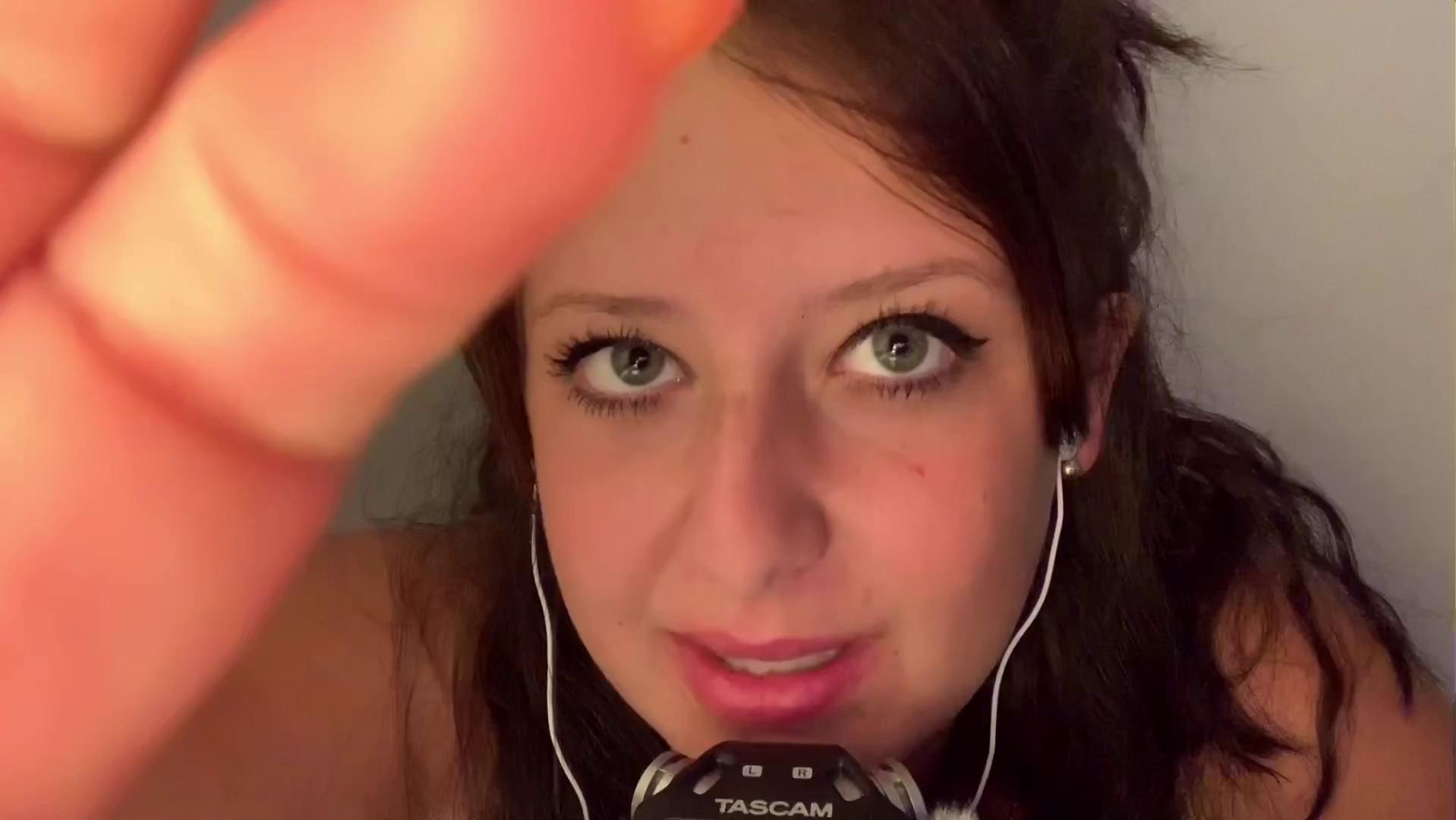} &
\includegraphics[width=1.8cm]{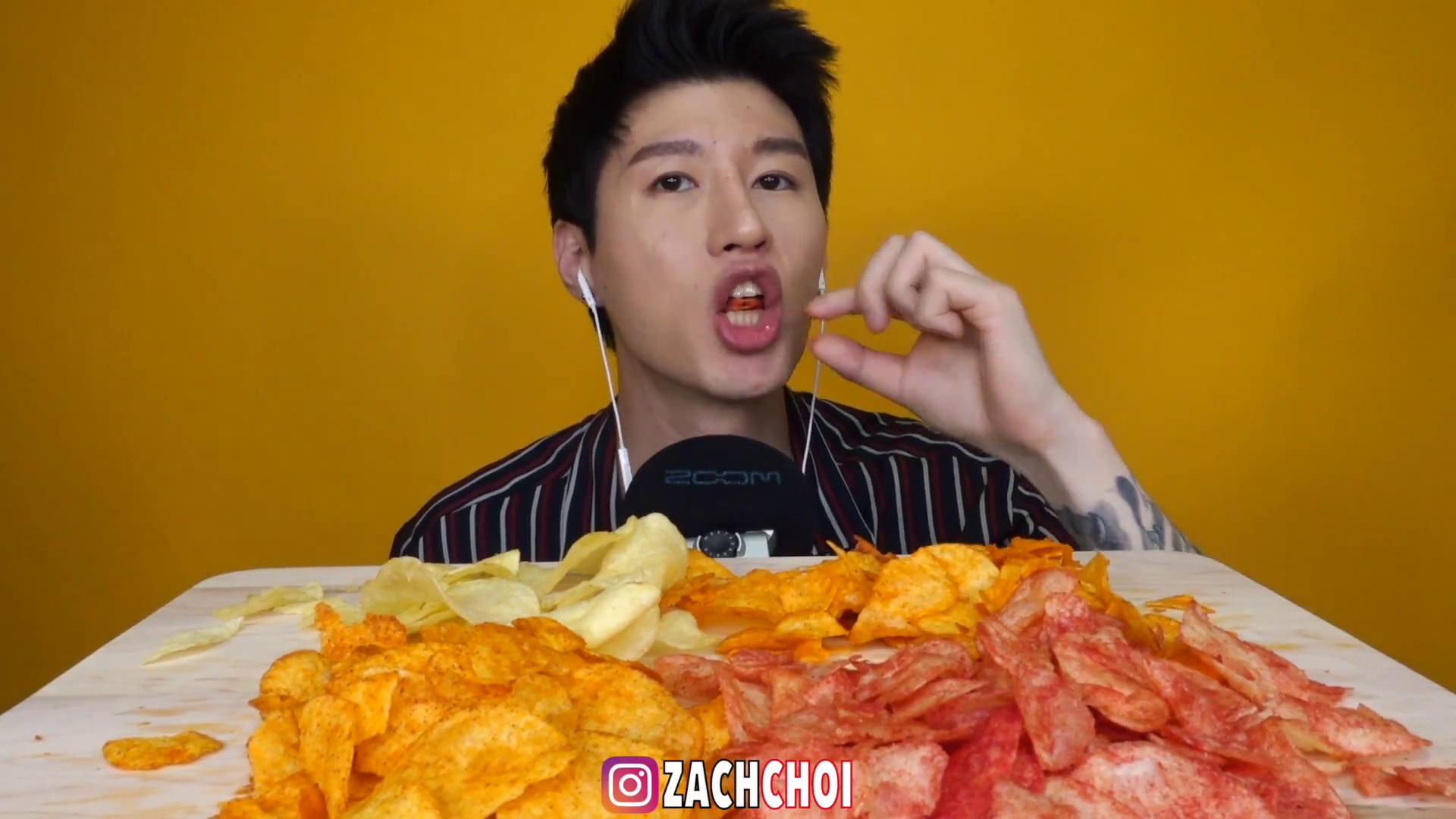} \\
& \scriptsize people slurping 
& \scriptsize people slurping 
& \scriptsize lip smacking 
& \scriptsize people eating crisps \\

\end{tabular}
\vspace{0.3cm}
\caption{Image to Audio Retrieval in VGG-SS.}
\label{fig:audio_retrieval_vggss}
\end{figure}

\begin{figure}[ht]
\centering
\begin{tabular}{m{1.8cm} | m{1.8cm} m{1.8cm} m{1.8cm} m{1.8cm}}
\multicolumn{1}{c|}{\textbf{Query Audio}} &
\multicolumn{4}{c}{\textbf{Retrieved Images}} \\
& \centering \textbf{Top 1} & \centering \textbf{Top 2} & \centering \textbf{Top 3} & \centering \textbf{Top 4} \\
\end{tabular}

\begin{tabular}{m{1.8cm} | m{1.8cm} m{1.8cm} m{1.8cm} m{1.8cm}}
\includegraphics[width=1.8cm]{Figures/retrieval/VGG-SS/case1/oSUCIMc3Vp8_000055.jpg} &
\includegraphics[width=1.8cm]{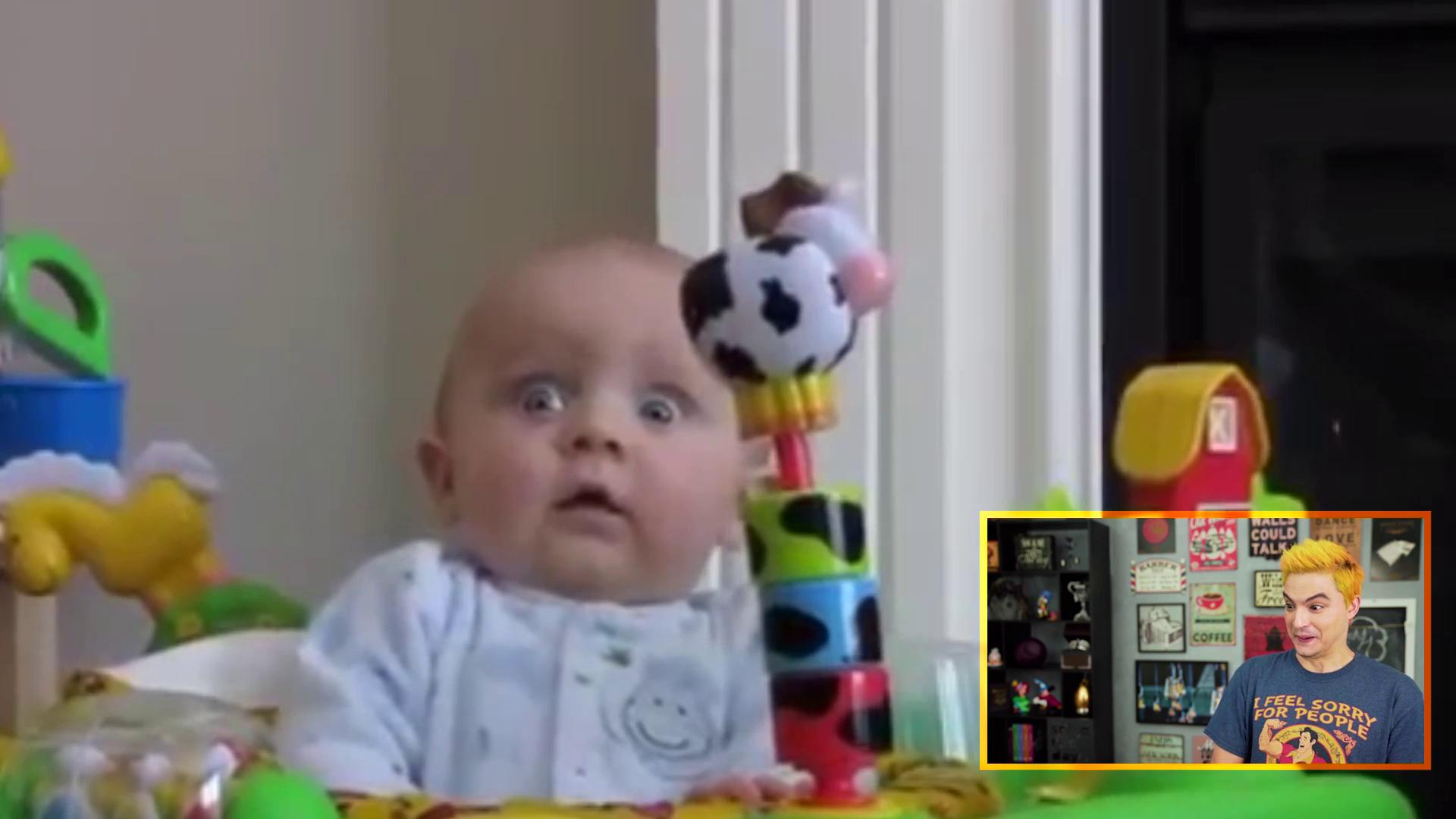} &
\includegraphics[width=1.8cm]{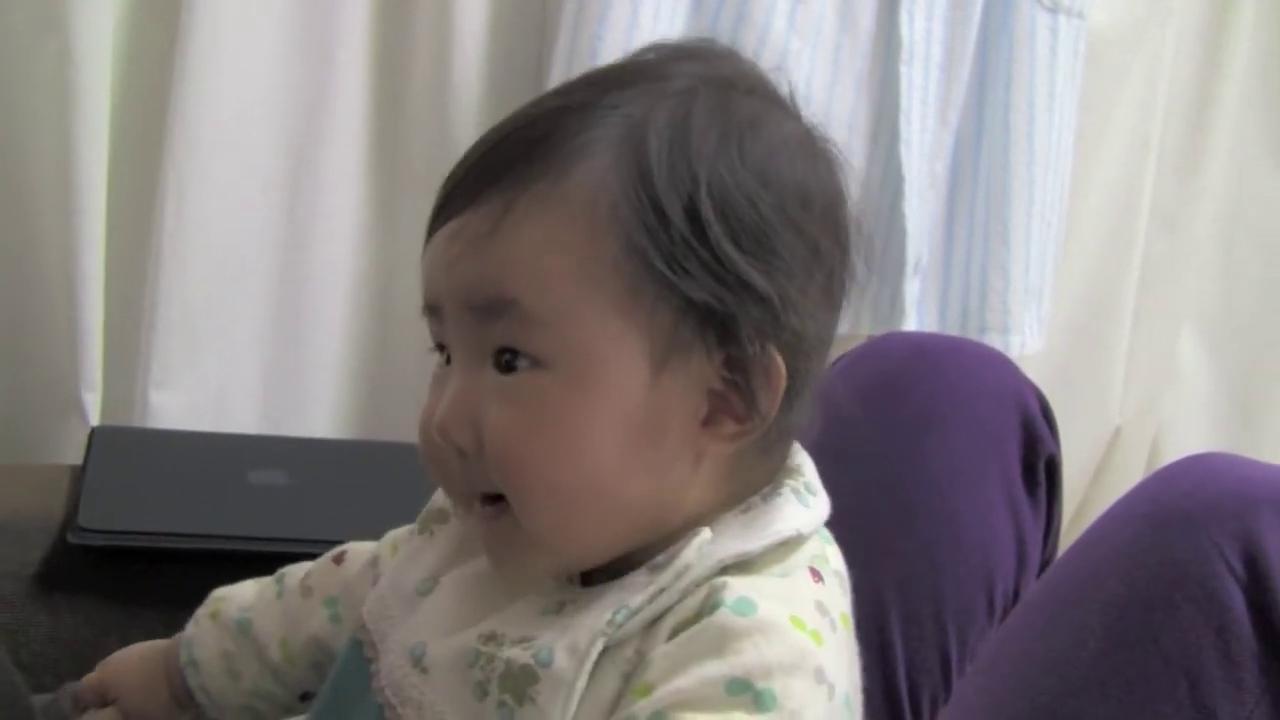} &
\includegraphics[width=1.8cm]{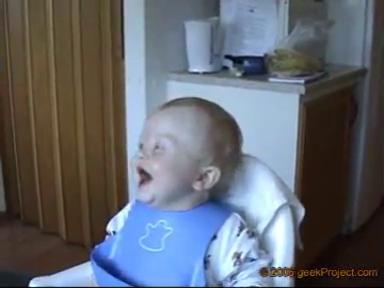} &
\includegraphics[width=1.8cm]{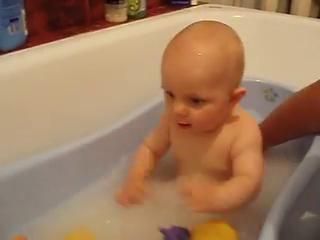} \\
\scriptsize People giggling \\

\includegraphics[width=1.8cm]{Figures/retrieval/VGG-SS/case2/0F04c_rY4aw_000000.jpg} &
\includegraphics[width=1.8cm]{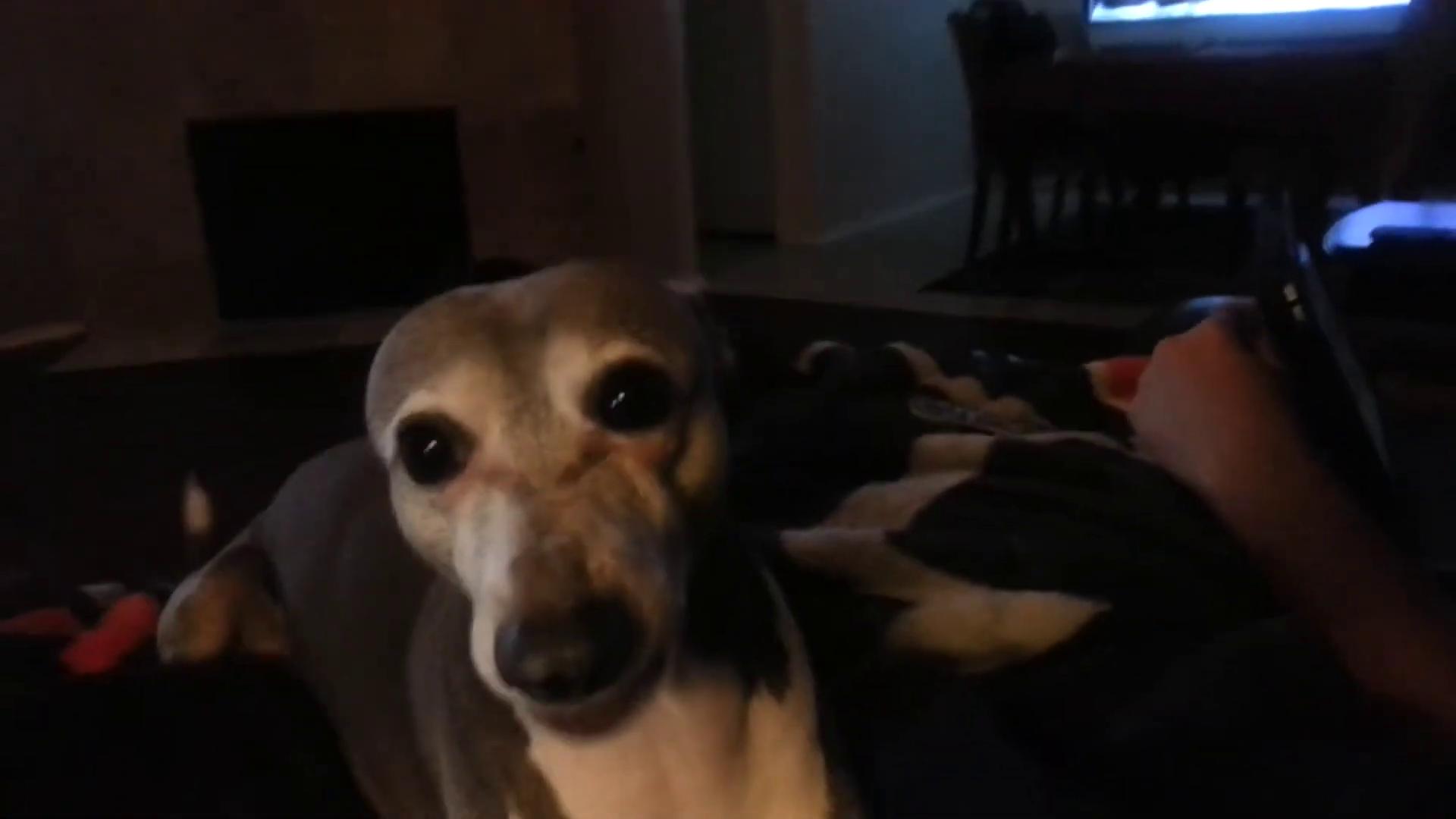} &
\includegraphics[width=1.8cm]{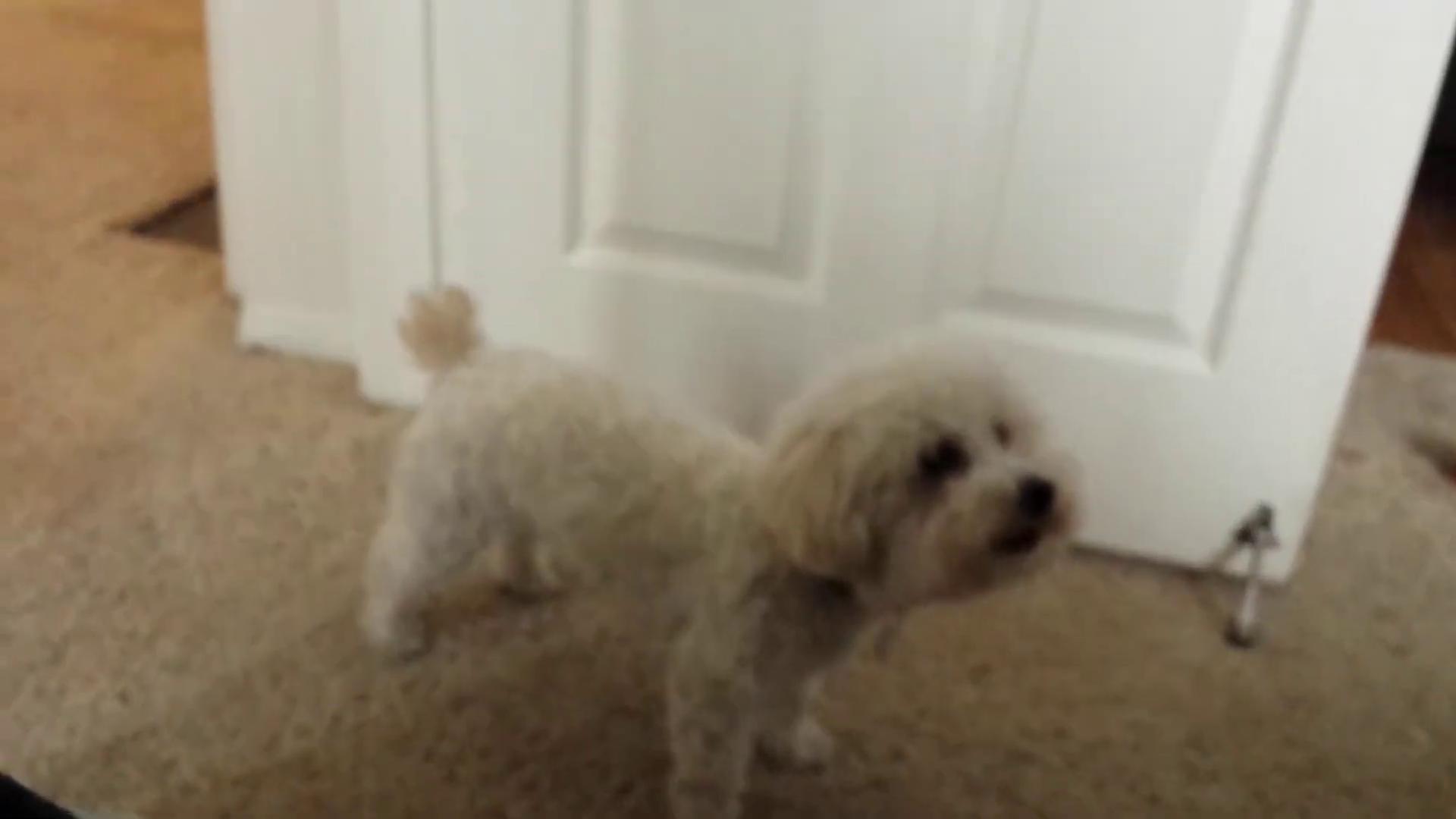} &
\includegraphics[width=1.8cm]{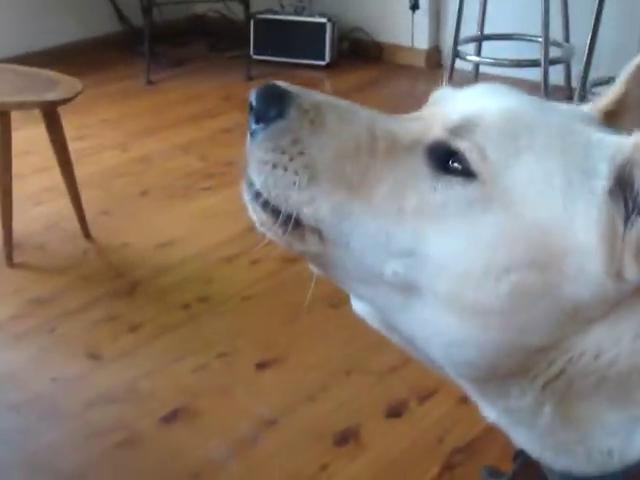} &
\includegraphics[width=1.8cm]{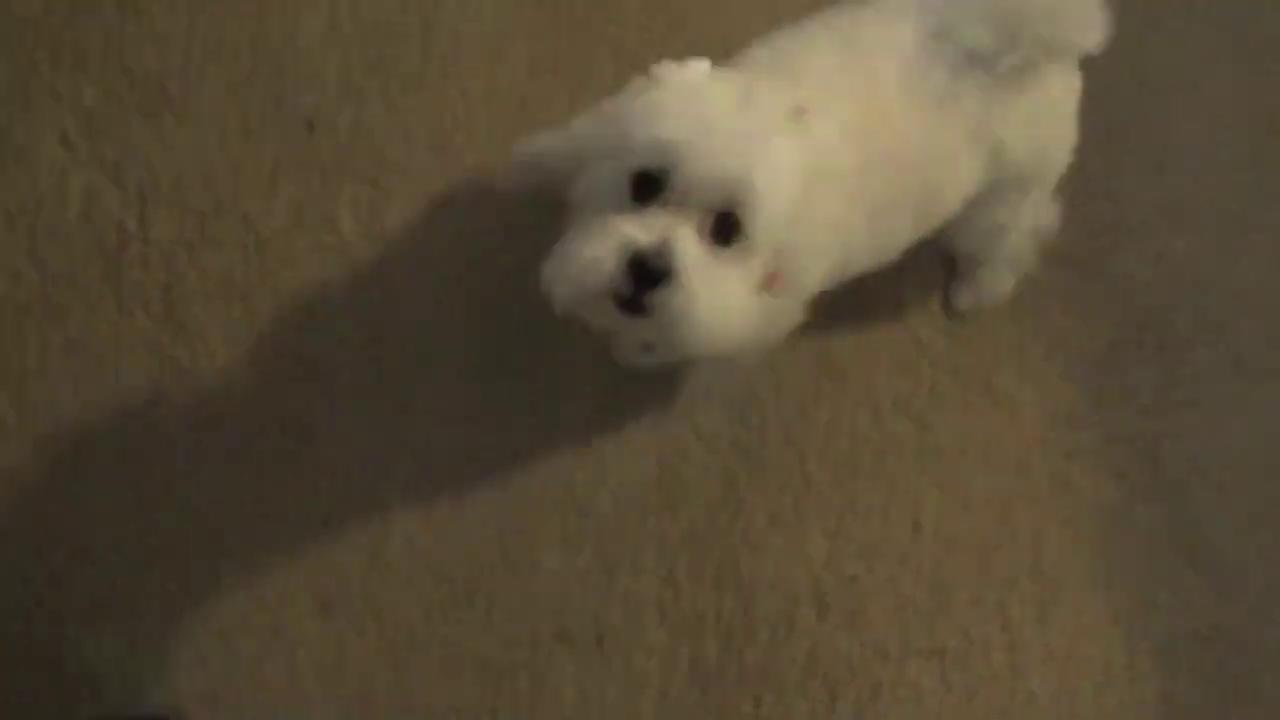} \\
\scriptsize dog howling \\

\includegraphics[width=1.8cm]{Figures/retrieval/VGG-SS/case3/0YTieIiZNN4_000010.jpg} &
\includegraphics[width=1.8cm]{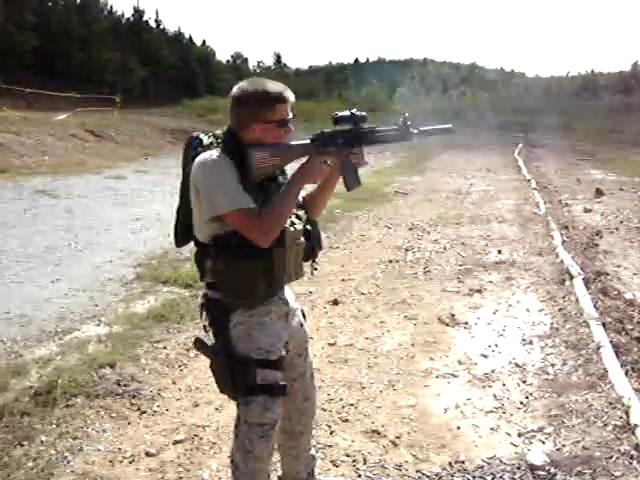} &
\includegraphics[width=1.8cm]{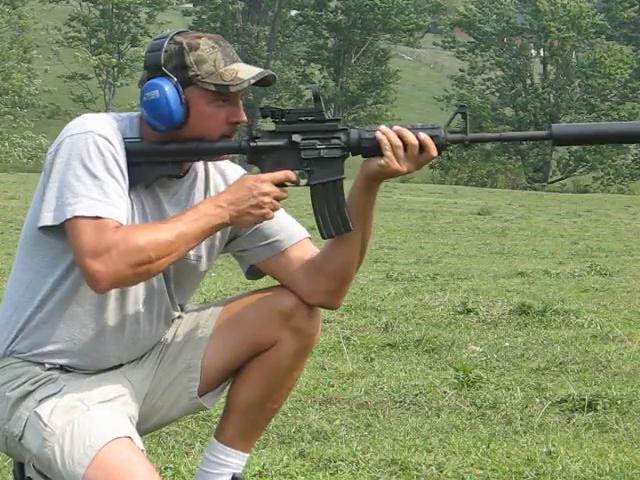} &
\includegraphics[width=1.8cm]{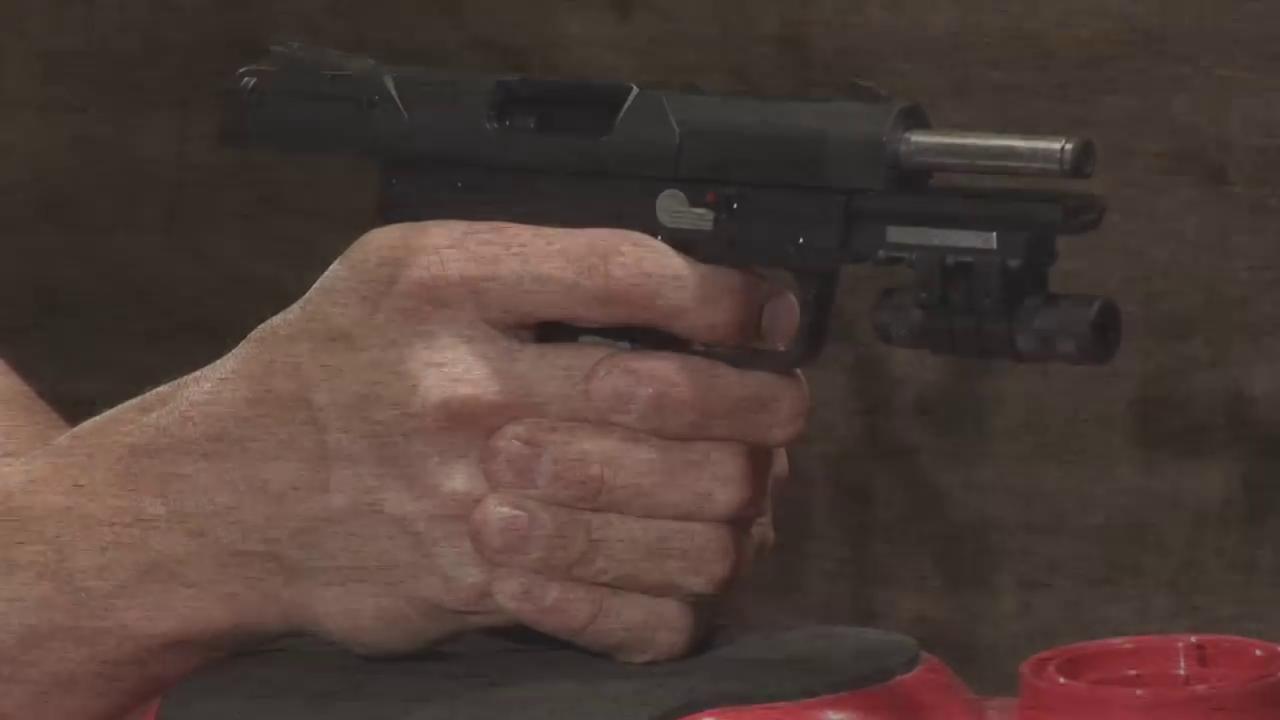} &
\includegraphics[width=1.8cm]{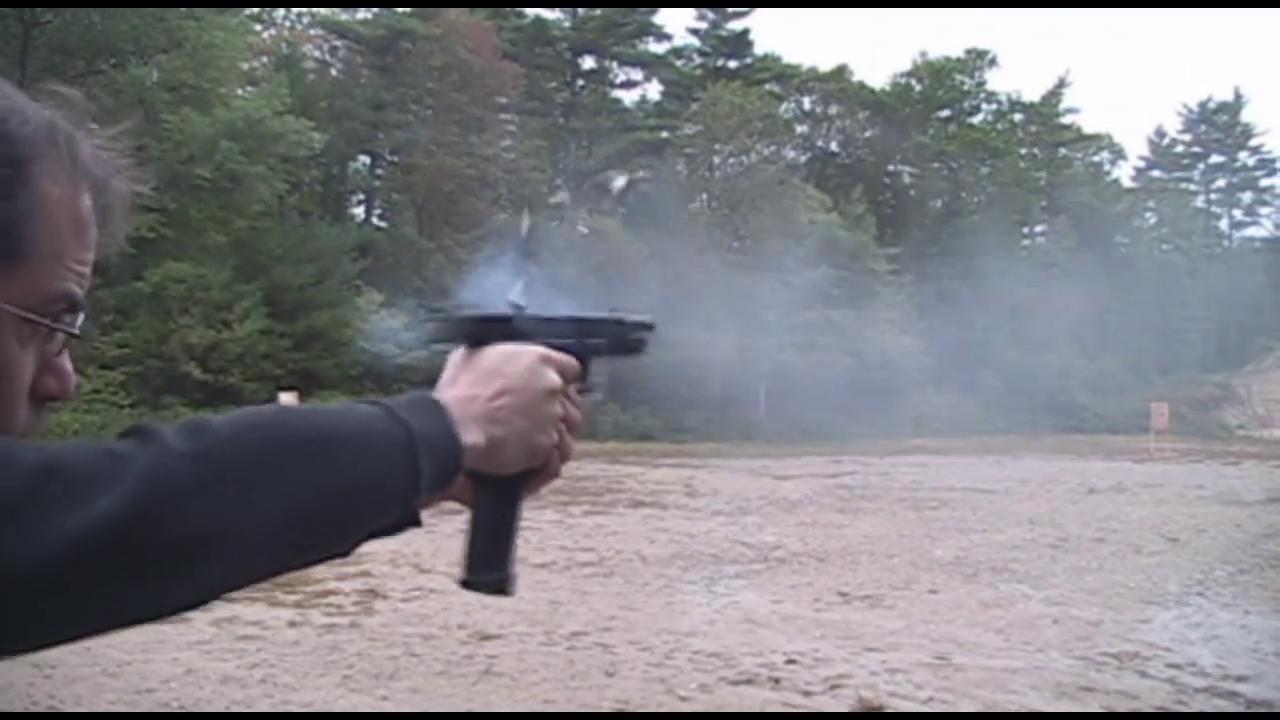} \\
\scriptsize machine gun shooting \\

\includegraphics[width=1.8cm]{Figures/retrieval/VGG-SS/case4/CWxZNhucY8o_000573.jpg} &
\includegraphics[width=1.8cm]{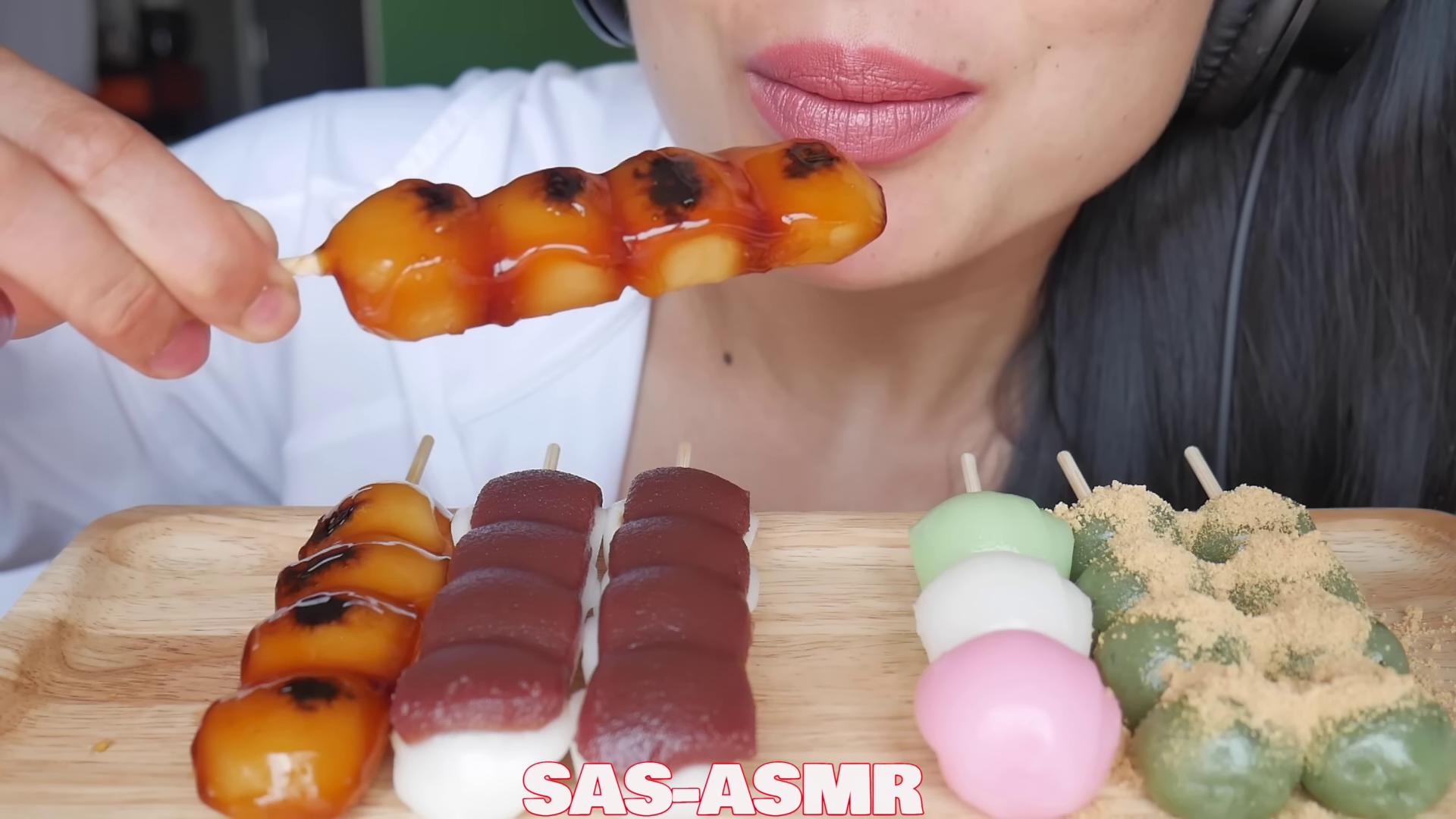} &
\includegraphics[width=1.8cm]{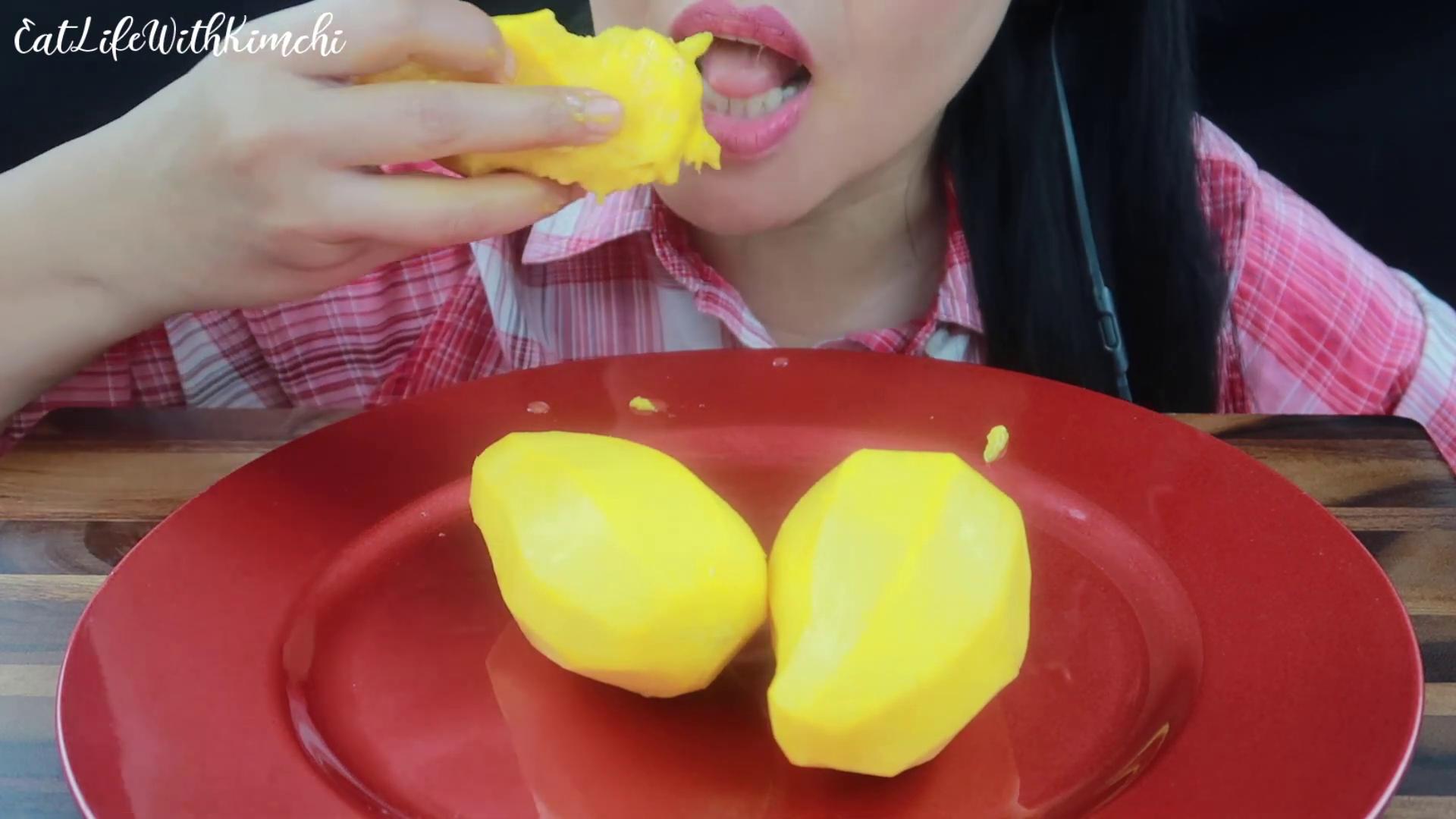} &
\includegraphics[width=1.8cm]{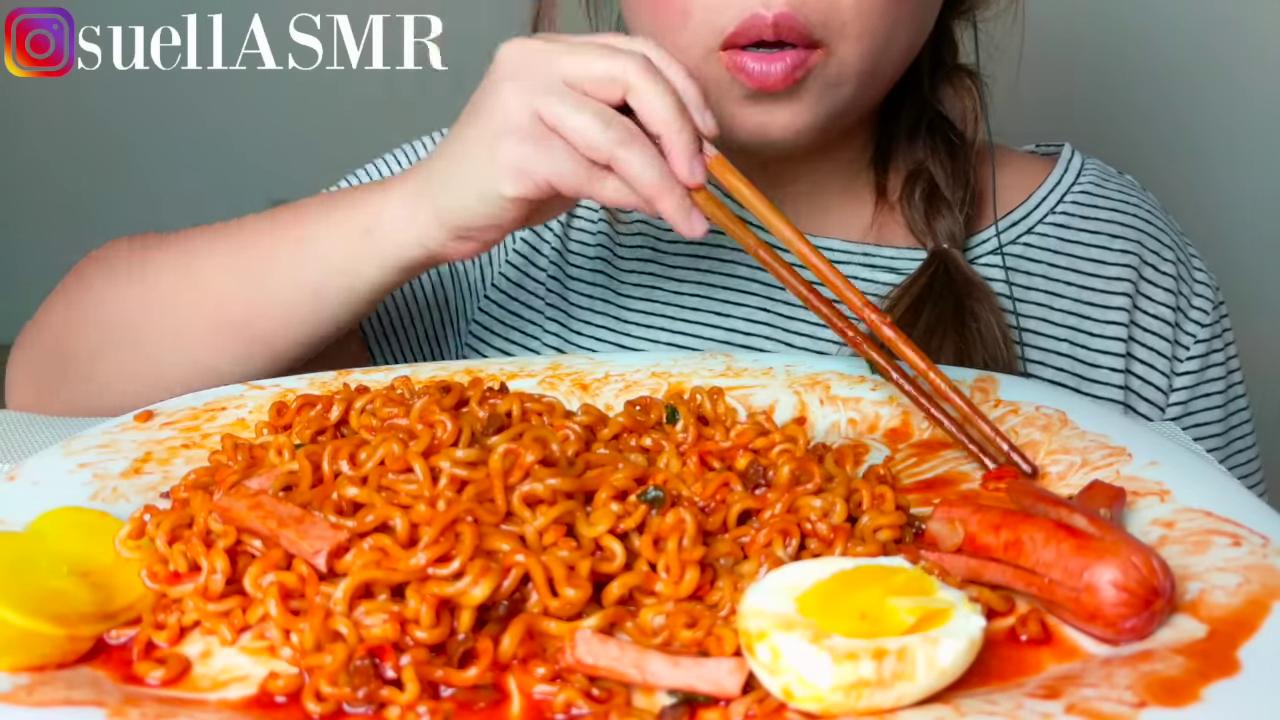} &
\includegraphics[width=1.8cm]{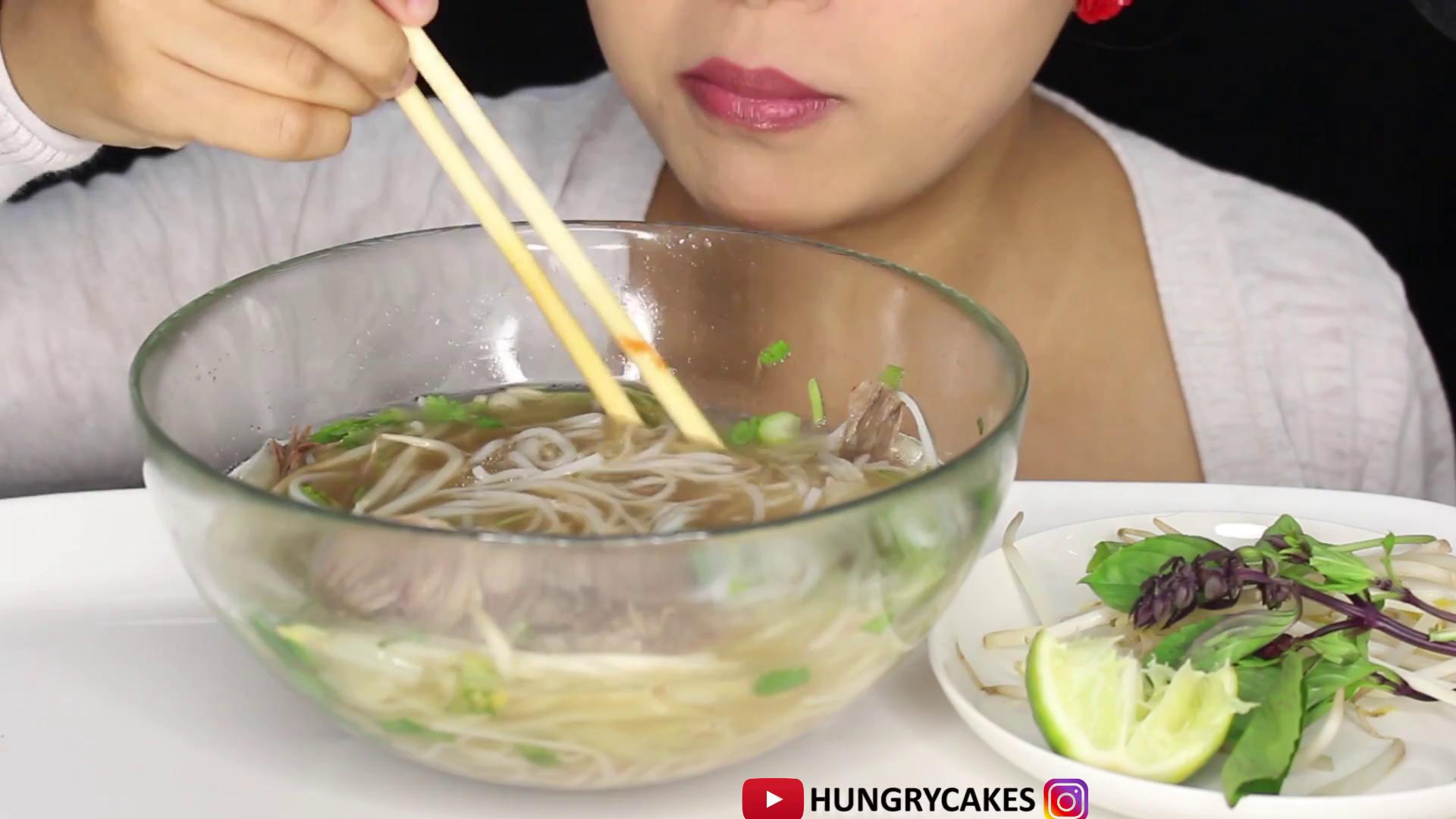} \\
\scriptsize people eating apple \\

\end{tabular}
\vspace{0.3cm}
\caption{Audio to Image Retrieval in VGG-SS.}
\label{fig:image_retrieval_vggss}
\end{figure}

Figures \ref{fig:audio_retrieval_vggss} and \ref{fig:image_retrieval_vggss} show qualitative results of cross-modal retrieval in the VGG-SS dataset. The labels in VGG-SS are highly specific, making it extremely challenging for models to differentiate accurately between certain classes using only single-query images instead of videos. For instance, in the last row of Figure \ref{fig:audio_retrieval_vggss}, the frame is labeled \textit{people eating apple}, yet the depicted image shows a woman with her mouth closed, offering no visual indication of an apple being eaten. Consequently, the model struggles to retrieve appropriate audio clips corresponding to this precise class and instead returns audio associated with subtle mouth movements, typically with a closed mouth, such as \textit{people slurping} or \textit{lip smacking}. This issue is similarly evident in the Audio to Image retrieval scenario depicted in Figure \ref{fig:image_retrieval_vggss}.

\clearpage
\subsection{More ablation studies}
\label{sup:ablations}

As mentioned in the paper, we conducted an ablation study by training multiple models with different values of the $\lambda_{SN}$ parameter, which controls the relative importance of the $\mathcal{L}_S$ and $\mathcal{L}_N$ loss terms with respect to the contrastive loss. The results of this analysis is presented in Table \ref{tab:ablation_lambda}. It indicates that the optimal value is $\lambda_{SN} = 1$, which yields the highest value for the positive cIoU metric, fully filters out silence and noise negatives, and maintains a relatively low offscreen pIA value.

\begin{table}[ht]
\begin{center}
\resizebox{0.6\textwidth}{!}{
\begin{tabular}{c|c c c c c c c c}
\hline
& \multicolumn{1}{c}{{\bf Positive audio input}} & & \multicolumn{3}{c}{{\bf Negative audio input}} & & \multicolumn{1}{c}{{\bf Global metric}} \\
\cline{2-2} \cline{4-6} \cline{8-8}
\textbf{$\lambda_{SN}$} & \textbf{cIoU {\text{Uth}}} & & \textbf{pIA$_S$} & \textbf{pIA$_N$} & \textbf{pIA$_O$} & & \textbf{F$_{\text{LOC}}$}\\
\hline

\textbf{0} & 
20.18 & 
 & 
0.03 & 
\textbf{0.00} & 
1.95 & 
 & 
33.55 \\ 

\textbf{0.01} & 
20.37 & 
 & 
0.61 & 
0.40 & 
2.18 & 
 & 
33.79 \\ 

\textbf{0.1} &  
17.90 & 
 & 
\textbf{0.00} &  
\textbf{0.00} & 
\textbf{1.54} & 
 & 
30.34 \\ 

\textbf{0.25} &  
19.43 & 
 & 
\textbf{0.00} &  
\textbf{0.00} & 
2.03 & 
 & 
32.50 \\ 

\textbf{0.5} &  
\underline{20.38} & 
 & 
\textbf{0.00} &  
\textbf{0.00} & 
2.18 & 
 & 
\underline{33.81} \\ 

\textbf{0.75} &  
18.67 & 
 & 
\textbf{0.00} &  
\textbf{0.00} & 
1.84 & 
 & 
31.43 \\ 

\textbf{1} &  
\textbf{20.47} & 
 & 
\textbf{0.00} &  
\textbf{0.00} & 
2.10 & 
 & 
\textbf{33.94} \\ 

\textbf{1.25} &  
18.94 & 
 & 
\textbf{0.00} &  
\textbf{0.00} & 
\underline{1.77} & 
 & 
31.81 \\ 

\textbf{1.5} &  
\underline{20.38} & 
 & 
\textbf{0.00} &  
\textbf{0.00} & 
2.18 & 
 & 
\underline{33.81} \\ 

\end{tabular}}
\end{center}
\caption{Results of the ablation of $\lambda_{SN}$ that multiplies the term $(\mathcal{L}_S + \mathcal{L}_N$).}
\label{tab:ablation_lambda}
\end{table}

    \begin{table}[ht]
    \begin{center}
    \resizebox{0.7\columnwidth}{!}{
    \begin{tabular}{@{}c|lcccccclcccc@{}}\hline
    & &  & & & & & \multicolumn{3}{c}{{\bf Negative audio input}} &\\
    \cline{8-10}
    {\bf Test set} & {\bf S} & {\bf N} & {\bf $\mathcal{L}_S$} & {\bf $\mathcal{L}_N$} & & {\bf cIoU$_{Uth}$ $\uparrow$} & {\bf pIA$_{\text{S}}$ $\downarrow$} & {\bf pIA$_{\text{N}}$ $\downarrow$}& {\bf pIA$_{\text{O}}$ $\downarrow$}& {\bf F$_{\text{LOC}}$ $\uparrow$}& {\bf Sep $\uparrow$} \\
    \hline

    \multirow{7}{*}{\STAB{\rotatebox[origin=c]{90}{\textbf{VGG-SS}}}}
        &  
        &  
        &  
        &  
        &
        & \small 28.38
        & \small 0.71
        & \small 0.71
        & \small 1.54
        & \small 44.11 
        & \small 0.0896 \\

        & \ding{51} 
        &  
        &  
        &  
        &
        & \small 28.21
        & \small 0.69
        & \small 0.61
        & \small \underline{1.47}
        & \small 43.91 
        & \small 0.1019 \\

        & \ding{51} 
        &  
        & \ding{51} 
        &  
        &
        & \small 27.90
        & \small 0.72
        & \small 0.76
        & \small \textbf{1.32}
        & \small 43.54 
        & \small 0.0982 \\

        &  
        & \ding{51} 
        &  
        &  
        &
        & \small 29.14
        & \small 0.89
        & \small 1.01
        & \small 1.56
        & \small 45.01 
        & \small \underline{0.1047} \\

        &  
        & \ding{51} 
        &  
        & \ding{51} 
        &
        & \small 29.30
        & \small 0.71
        & \small 0.72
        & \small 1.50
        & \small 45.22 
        & \small 0.0975 \\

        & \ding{51} 
        & \ding{51} 
        &  
        &  
        &
        & \small \textbf{29.99}
        & \small \underline{0.68}
        & \small \underline{0.35}
        & \small 1.63
        & \small \textbf{46.05} 
        & \small \textbf{0.1059} \\

        & \ding{51} 
        & \ding{51} 
        & \ding{51} 
        & \ding{51} 
        &
        & \small \underline{29.61}
        & \small \textbf{0.01}
        & \small \textbf{0.00}
        & \small 1.72
        & \small \underline{45.63} 
        & \small 0.0971 \\

    \hline 

    \multirow{7}{*}{\STAB{\rotatebox[origin=c]{90}{\textbf{IS3}}}}

        &  
        &  
        &  
        &  
        &
        & \small 18.83
        & \small 0.46
        & \small 0.47
        & \small 2.23
        & \small 31.63 
        & \small -0.0324 \\

        & \ding{51} 
        &  
        &  
        &  
        &
        & \small 17.80
        & \small 0.51
        & \small 0.50
        & \small 2.10
        & \small 30.17 
        & \small \underline{-0.0279} \\

        & \ding{51} 
        &  
        & \ding{51} 
        &  
        &
        & \small 17.96
        & \small 0.41
        & \small 0.45
        & \small \textbf{1.74}
        & \small 30.40 
        & \small -0.0303 \\

        &  
        & \ding{51} 
        &  
        &  
        &
        & \small 18.07
        & \small 0.91
        & \small 0.99
        & \small 1.98
        & \small 30.54 
        & \small -0.0300 \\

        &  
        & \ding{51} 
        &  
        & \ding{51} 
        &
        & \small \underline{19.06}
        & \small \underline{0.34}
        & \small 0.35
        & \small 2.30
        & \small \underline{31.96} 
        & \small -0.0301 \\

        & \ding{51} 
        & \ding{51} 
        &  
        &  
        &
        & \small \textbf{19.83}
        & \small 0.36
        & \small \underline{0.17}
        & \small 2.42
        & \small \textbf{33.05} 
        & \small \textbf{-0.0261} \\

        & \ding{51} 
        & \ding{51} 
        & \ding{51} 
        & \ding{51} 
        &
        & \small 18.87
        & \small \textbf{0.00}
        & \small \textbf{0.00}
        & \small \underline{1.97}
        & \small 31.72 
        & \small -0.0327 \\

    \hline 

    \multirow{7}{*}{\STAB{\rotatebox[origin=c]{90}{\textbf{IS3+}}}}
        &  
        &  
        &  
        &  
        &
        & \small 18.30
        & \small 0.45
        & \small 0.46
        & \small 2.04
        & \small 30.89 
        & \small -0.0303 \\

        & \ding{51} 
        & 
        & 
        & 
        &
        & \small 17.52
        & \small 0.50
        & \small 0.49
        & \small 1.99
        & \small 29.77 
        & \small \underline{-0.0301} \\

        & \ding{51} 
        &  
        & \ding{51} 
        &  
        &
        & \small 17.72
        & \small 0.43
        & \small 0.48
        & \small \textbf{1.84}
        & \small 30.06 
        & \small -0.0347 \\

        &  
        & \ding{51} 
        &  
        &  
        &
        & \small 17.34
        & \small 0.91
        & \small 0.99
        & \small \underline{1.92}
        & \small 29.50 
        & \small -0.0360 \\

        &  
        & \ding{51} 
        & 
        & \ding{51} 
        &
        & \small 18.61
        & \small \underline{0.33}
        & \small 0.34
        & \small 2.36
        & \small 31.33 
        & \small -0.0369 \\

        & \ding{51} 
        & \ding{51} 
        &  
        &  
        &
        & \small \textbf{19.67}
        & \small 0.36
        & \small \underline{0.18}
        & \small 2.23
        & \small \textbf{32.82} 
        & \small \textbf{-0.0209} \\

        & \ding{51} 
        & \ding{51} 
        & \ding{51} 
        & \ding{51} 
        &
        & \small \underline{19.13}
        & \small \textbf{0.00}
        & \small \textbf{0.00}
        & \small 2.09
        & \small \underline{32.08} 
        & \small -0.0327 \\
    
    \hline 

    \multirow{7}{*}{\STAB{\rotatebox[origin=c]{90}{\textbf{AVS-Bench S4}}}}
        &  
        &  
        &  
        &  
        &
        & \small 31.57
        & \small 2.53
        & \small 2.49
        & \small 1.00
        & \small 47.76 
        & \small 0.2384 \\

        & \ding{51} 
        & 
        & 
        & 
        &
        & \small 31.76
        & \small 2.39
        & \small 1.85
        & \small \underline{0.82}
        & \small 48.01 
        & \small 0.2350 \\

        & \ding{51} 
        &  
        & \ding{51} 
        &  
        &
        & \small \underline{32.49}
        & \small 2.54
        & \small 2.58
        & \small \textbf{0.81}
        & \small \underline{48.80} 
        & \small 0.2458 \\

        &  
        & \ding{51} 
        &  
        &  
        &
        & \small 31.97
        & \small 2.26
        & \small 2.59
        & \small 0.88
        & \small 48.22 
        & \small \textbf{0.2558} \\

        &  
        & \ding{51} 
        & 
        & \ding{51} 
        &
        & \small 32.07
        & \small 2.21
        & \small 2.26
        & \small 1.07
        & \small 48.34 
        & \small 0.2375 \\

        & \ding{51} 
        & \ding{51} 
        &  
        &  
        &
        & \small 32.05
        & \small \underline{1.75}
        & \small \underline{1.01}
        & \small 0.84
        & \small 48.40 
        & \small 0.2393 \\

        & \ding{51} 
        & \ding{51} 
        & \ding{51} 
        & \ding{51} 
        &
        & \small \textbf{32.76}
        & \small \textbf{0.05}
        & \small \textbf{0.00}
        & \small 0.95
        & \small \textbf{49.31} 
        & \small \underline{0.2479} \\
    
    \hline
    \end{tabular}%
    }
    \end{center}
    \caption{Ablation table of the silence, noise, $\mathcal{L}_S$ and $\mathcal{L}_N$. Best results in \textbf{bold}, second best \underline{underlined}.}
    \label{tab:ab_snm_cf_complete}
    \end{table}

\new{
On the other hand, the Table \ref{tab:ab_snm_cf_complete} shows the impact of training with samples of noise and silence, as well as with the new loss terms $\mathcal{L}_S$ and $\mathcal{L}_N$. 
Overall, the model trained with silence, noise, and $\mathcal{L}_S$ and $\mathcal{L}_N$ delivers the best balance between negative audio filtering and localization. It is the only setup that drives negative leaks essentially to zero across datasets: in VGG-SS it achieves $pIA_S{=}0.01$ and $pIA_N{=}0.00$, and in IS3/IS3+ both are exactly $0.00$. At the same time, it is top on S4  with the best cIoU$_{Uth}$ and F$_\text{LOC}$, and remains very close to the best on VGG-SS and IS3/IS3+ (e.g., second-best cIoU and F$_\text{LOC}$ in VGG-SS). Compared to the version with silence and noise but without the loss terms, the difference in negative suppression is massive (e.g., VGG-SS pIA$_S$: $0.68 \rightarrow 0.01$, pIA$_N$: $0.35 \rightarrow 0.00$) for small differences in cIoU/F$_\text{LOC}$.

Conceptually, combining both types of negative samples  (silence and noise) with explicit supervision, $\mathcal{L}_S$ and $\mathcal{L}_N$,  teaches the model to filter negative audio samples, preventing hallucinations, tightening activation thresholds and separating the values of the similarity maps of the positives from the negatives (see Figure \ref{sup_fig:boxplots}). This regularization yields better generalization across datasets and tasks.
}
\subsection{Qualitative results}
\label{sup:top_cases_sslsan}

\subsubsection{VGG-SS}
\begin{figure*}[!ht]
\label{fig:inferences}
\centering
\resizebox{\textwidth}{!}{%
\begin{tabular}{ccccccccccc}
& \scriptsize \hspace{-0.35cm} \raisebox{-0.17cm}{\includegraphics[height=0.5cm]{Figures/equalizer.png}} Piano & \scriptsize \hspace{-0.5cm} \raisebox{-0.17cm}{\includegraphics[height=0.5cm]{Figures/equalizer.png}} Silence & \scriptsize \hspace{-0.6cm} \raisebox{-0.17cm}{\includegraphics[height=0.5cm]{Figures/equalizer.png}} Noise & \hspace{-0.52cm} \raisebox{-0.17cm}{\includegraphics[height=0.5cm]{Figures/equalizer.png}} \scriptsize Offscreen & & \scriptsize \hspace{-0.35cm} \raisebox{-0.17cm}{\includegraphics[height=0.5cm]{Figures/equalizer.png}} Chicken & \scriptsize \hspace{-0.5cm} \raisebox{-0.17cm}{\includegraphics[height=0.5cm]{Figures/equalizer.png}} Silence & \scriptsize \hspace{-0.6cm} \raisebox{-0.17cm}{\includegraphics[height=0.5cm]{Figures/equalizer.png}} Noise & \hspace{-0.52cm} \raisebox{-0.17cm}{\includegraphics[height=0.5cm]{Figures/equalizer.png}} \scriptsize Offscreen \\
\vspace{0.05cm} 
\rotatebox[origin=c]{90}{\scriptsize LVS } & 
\hspace{-0.45cm} \includegraphics[align=c, width=1.45cm]{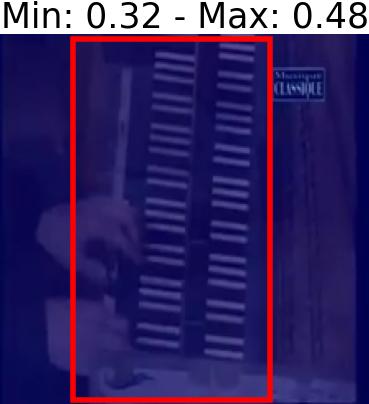}&
\hspace{-0.45cm} \includegraphics[align=c, width=1.45cm]{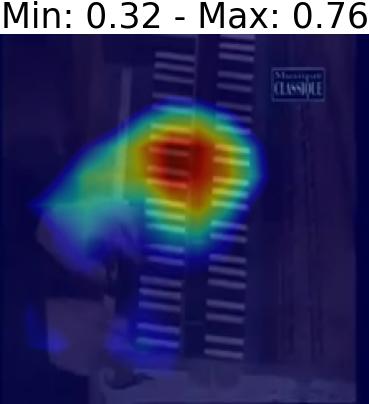}&
\hspace{-0.45cm} \includegraphics[align=c, width=1.45cm]{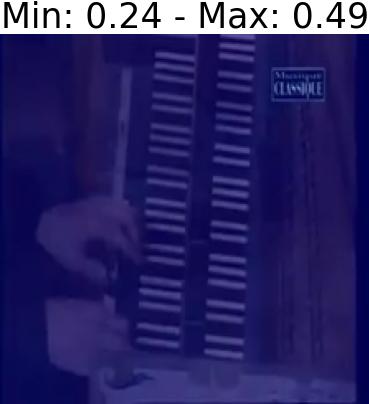}&
\hspace{-0.45cm} \includegraphics[align=c, width=1.45cm]{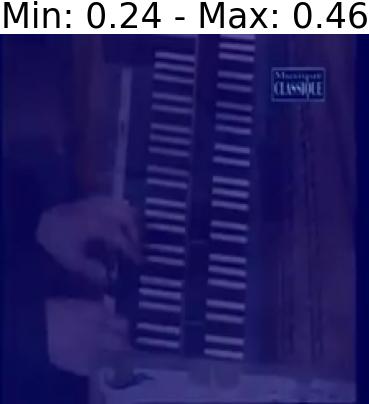}&
&
\hspace{-0.45cm} \includegraphics[align=c, width=1.45cm]{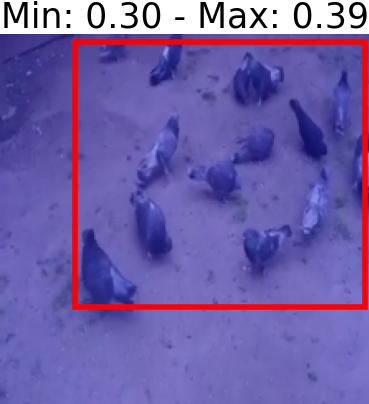}&
\hspace{-0.45cm} \includegraphics[align=c, width=1.45cm]{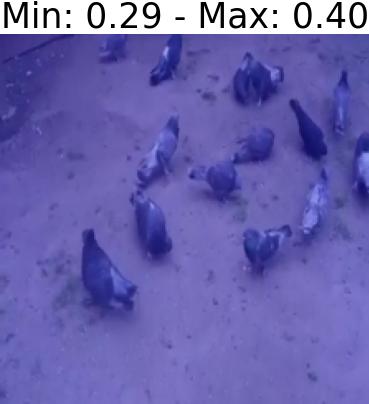}&
\hspace{-0.45cm} \includegraphics[align=c, width=1.45cm]{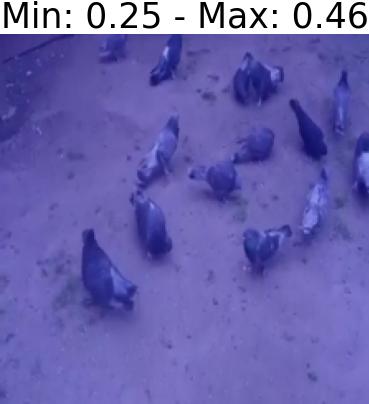}&
\hspace{-0.45cm} \includegraphics[align=c, width=1.45cm]{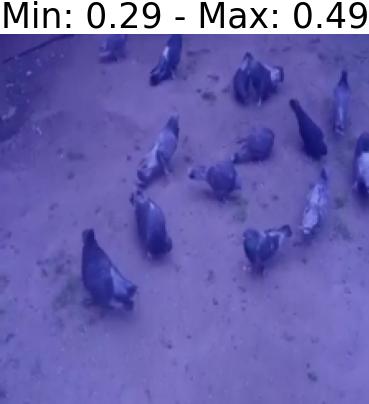}\\ 
\vspace{0.05cm}
\rotatebox[origin=c]{90}{\scriptsize EZ-VSL } & 
\hspace{-0.45cm} \includegraphics[align=c, width=1.45cm]{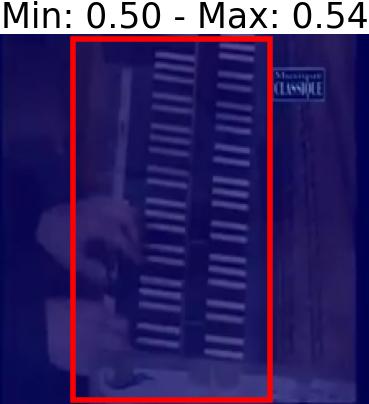}&
\hspace{-0.45cm} \includegraphics[align=c, width=1.45cm]{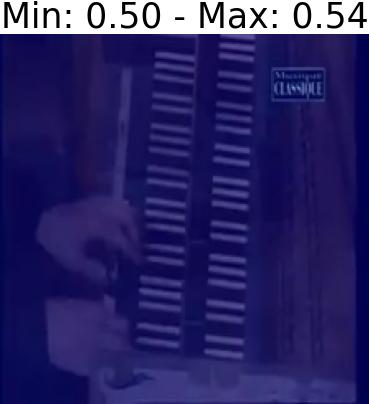}&
\hspace{-0.45cm} \includegraphics[align=c, width=1.45cm]{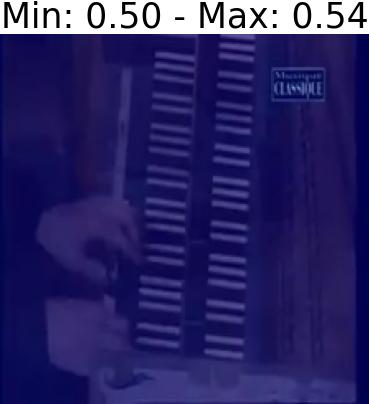}&
\hspace{-0.45cm} \includegraphics[align=c, width=1.45cm]{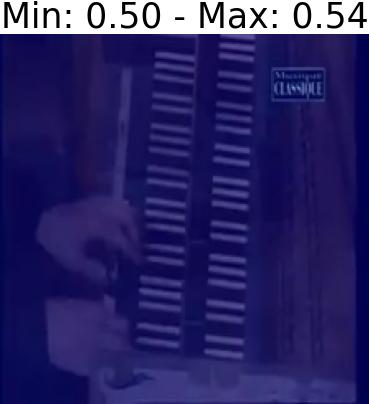}&
&
\hspace{-0.45cm} \includegraphics[align=c, width=1.45cm]{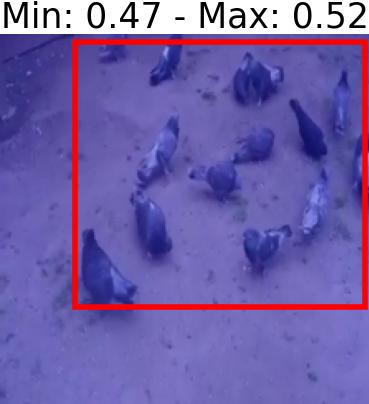}&
\hspace{-0.45cm} \includegraphics[align=c, width=1.45cm]{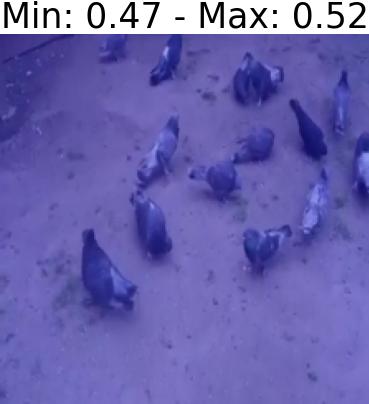}&
\hspace{-0.45cm} \includegraphics[align=c, width=1.45cm]{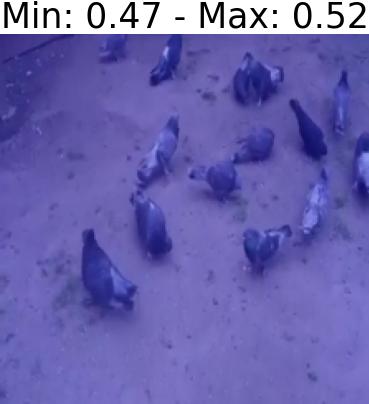}&
\hspace{-0.45cm} \includegraphics[align=c, width=1.45cm]{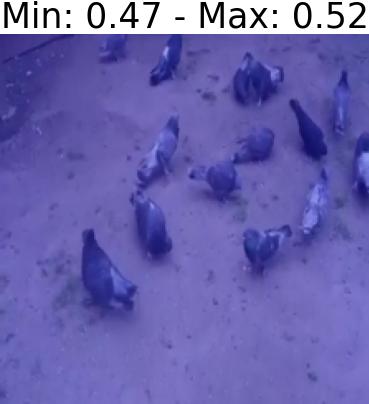}\\
\vspace{0.05cm}
\rotatebox[origin=c]{90}{\scriptsize FNAC } & 
\hspace{-0.45cm} \includegraphics[align=c, width=1.45cm]{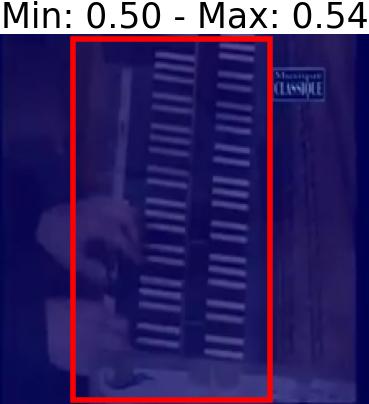}&
\hspace{-0.45cm} \includegraphics[align=c, width=1.45cm]{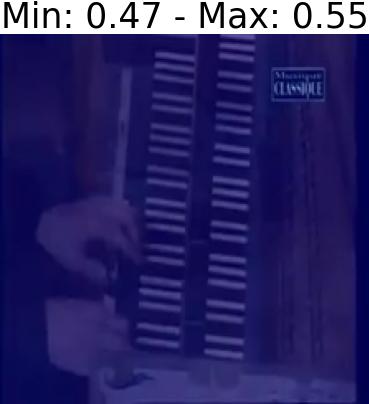}&
\hspace{-0.45cm} \includegraphics[align=c, width=1.45cm]{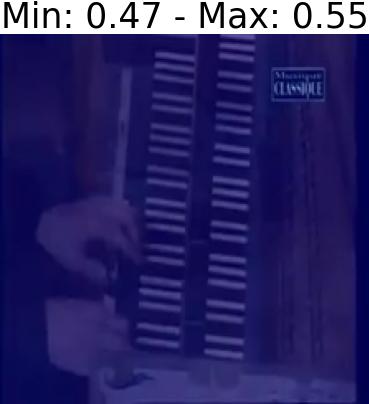}&
\hspace{-0.45cm} \includegraphics[align=c, width=1.45cm]{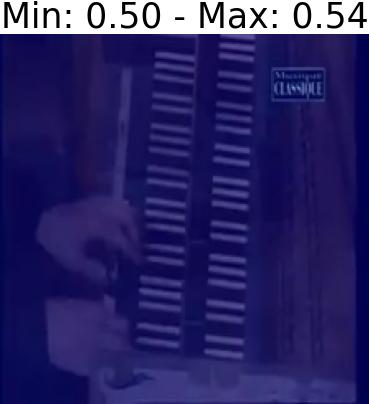}&
&
\hspace{-0.45cm} \includegraphics[align=c, width=1.45cm]{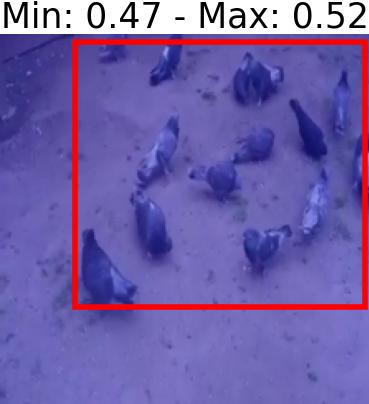}&
\hspace{-0.45cm} \includegraphics[align=c, width=1.45cm]{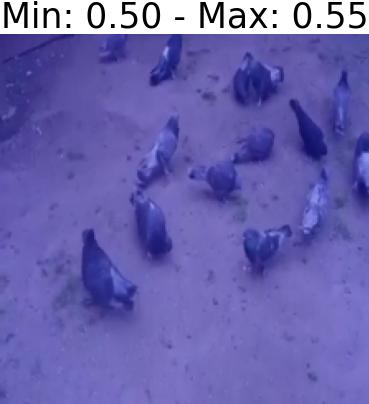}&
\hspace{-0.45cm} \includegraphics[align=c, width=1.45cm]{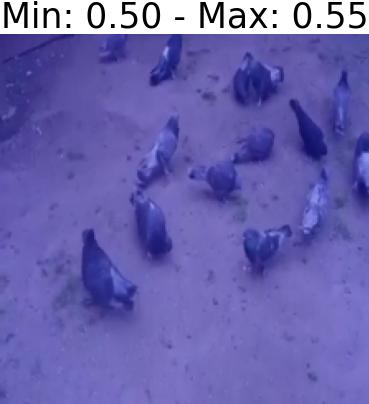}&
\hspace{-0.45cm} \includegraphics[align=c, width=1.45cm]{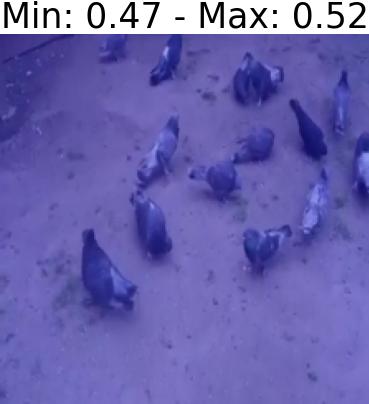}\\
\vspace{0.05cm}
\rotatebox[origin=c]{90}{\scriptsize SLAVC } & 
\hspace{-0.45cm} \includegraphics[align=c, width=1.45cm]{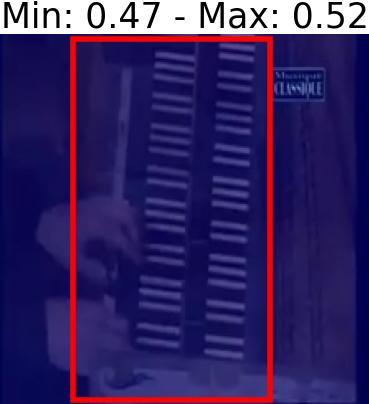}&
\hspace{-0.45cm} \includegraphics[align=c, width=1.45cm]{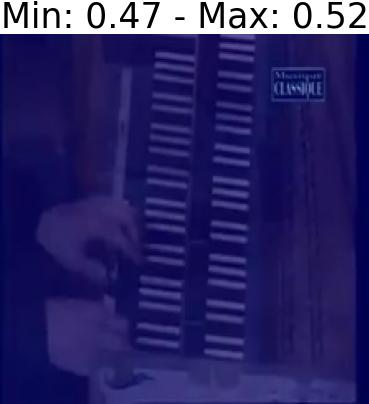}&
\hspace{-0.45cm} \includegraphics[align=c, width=1.45cm]{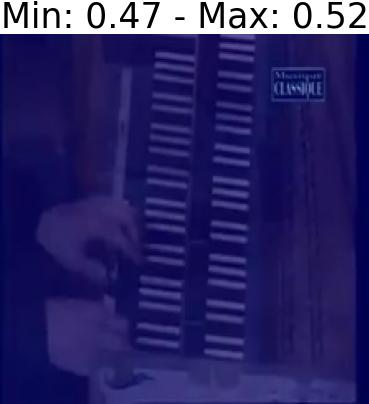}&
\hspace{-0.45cm} \includegraphics[align=c, width=1.45cm]{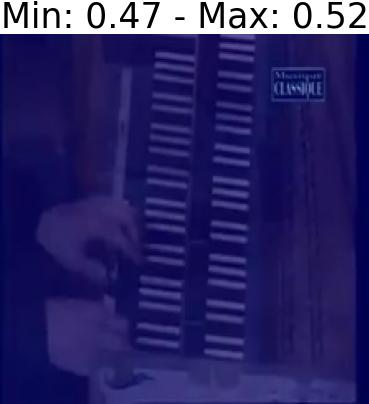}&
&
\hspace{-0.45cm} \includegraphics[align=c, width=1.45cm]{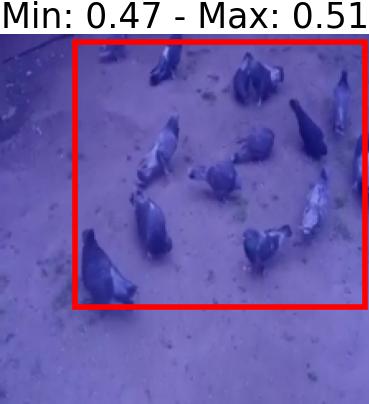}&
\hspace{-0.45cm} \includegraphics[align=c, width=1.45cm]{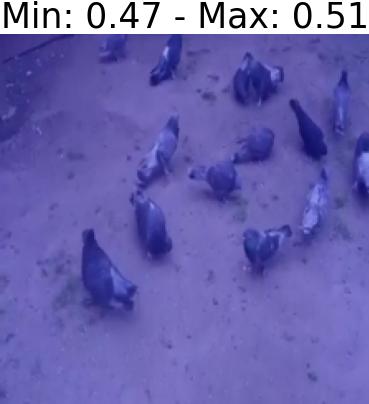}&
\hspace{-0.45cm} \includegraphics[align=c, width=1.45cm]{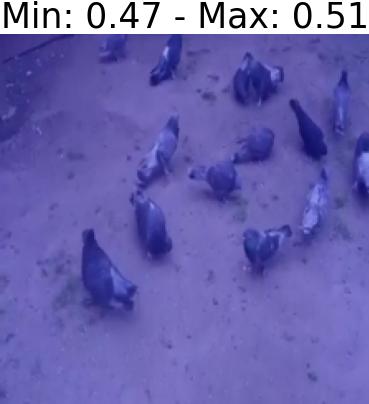}&
\hspace{-0.45cm} \includegraphics[align=c, width=1.45cm]{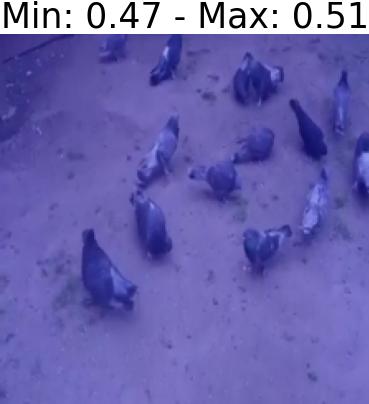}\\
\vspace{0.05cm}
\rotatebox[origin=c]{90}{\scriptsize SSL-TIE } & 
\hspace{-0.45cm} \includegraphics[align=c, width=1.45cm]{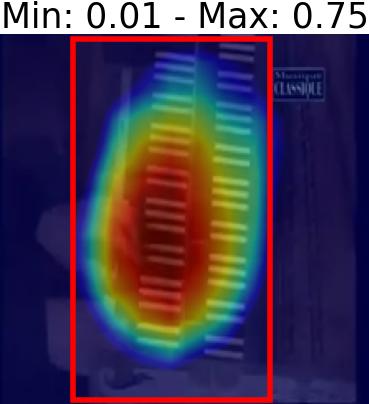}&
\hspace{-0.45cm} \includegraphics[align=c, width=1.45cm]{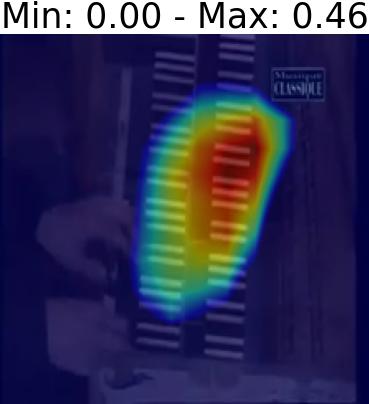}&
\hspace{-0.45cm} \includegraphics[align=c, width=1.45cm]{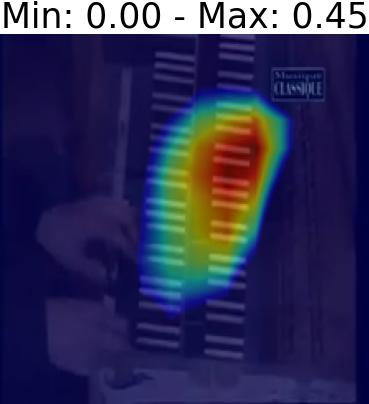}&
\hspace{-0.45cm} \includegraphics[align=c, width=1.45cm]{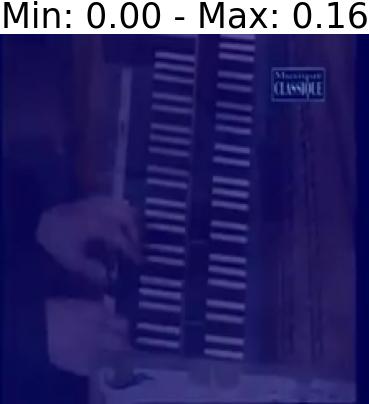}&
&
\hspace{-0.45cm} \includegraphics[align=c, width=1.45cm]{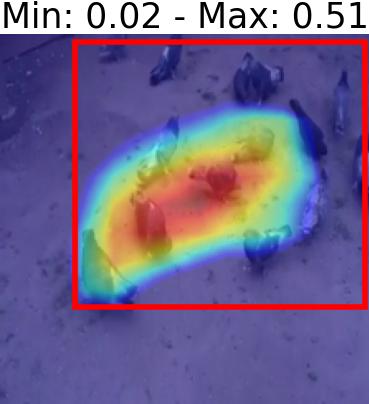}&
\hspace{-0.45cm} \includegraphics[align=c, width=1.45cm]{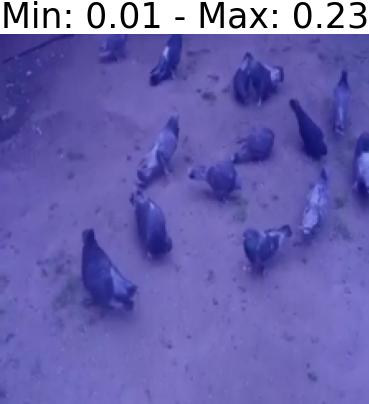}&
\hspace{-0.45cm} \includegraphics[align=c, width=1.45cm]{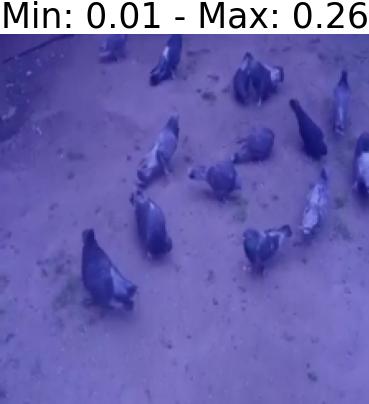}&
\hspace{-0.45cm} \includegraphics[align=c, width=1.45cm]{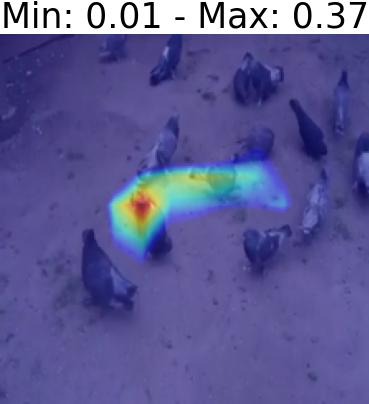}\\
\vspace{0.05cm}
\rotatebox[origin=c]{90}{\scriptsize SSL-Align (S.S.) } & 
\hspace{-0.45cm} \includegraphics[align=c, width=1.45cm]{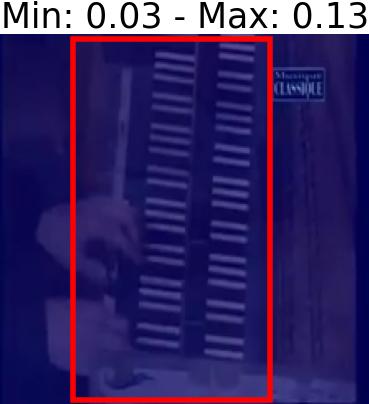}&
\hspace{-0.45cm} \includegraphics[align=c, width=1.45cm]{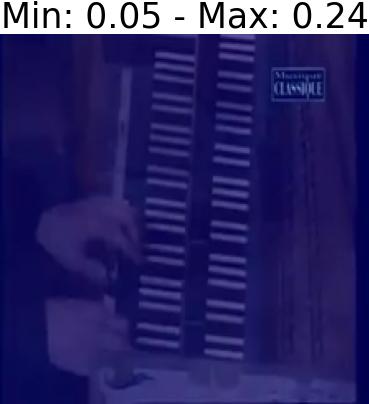}&
\hspace{-0.45cm} \includegraphics[align=c, width=1.45cm]{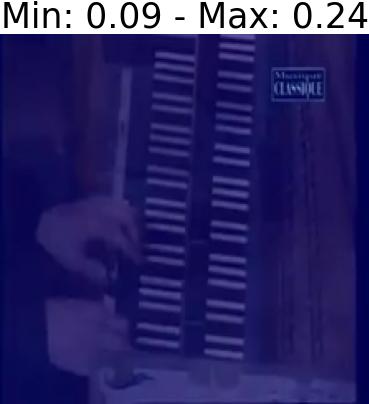}&
\hspace{-0.45cm} \includegraphics[align=c, width=1.45cm]{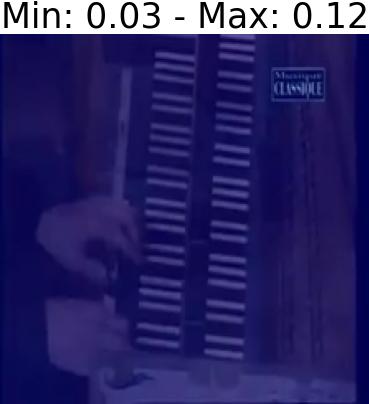}&
&
\hspace{-0.45cm} \includegraphics[align=c, width=1.45cm]{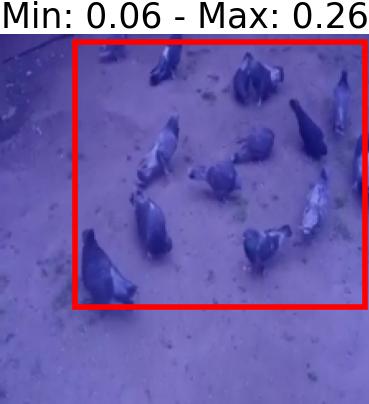}&
\hspace{-0.45cm} \includegraphics[align=c, width=1.45cm]{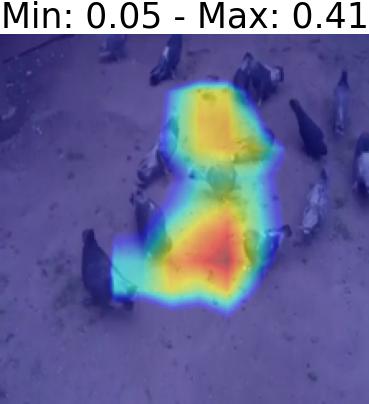}&
\hspace{-0.45cm} \includegraphics[align=c, width=1.45cm]{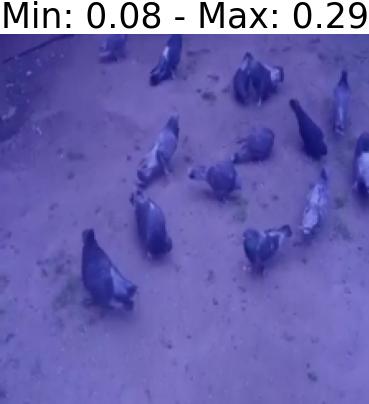}&
\hspace{-0.45cm} \includegraphics[align=c, width=1.45cm]{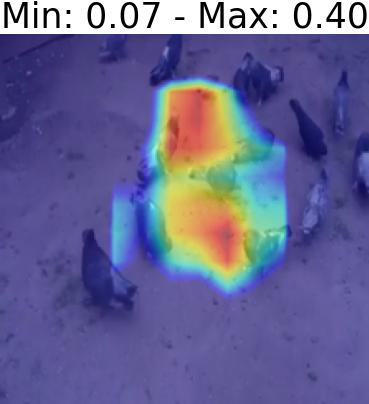}\\ \vspace{0.05cm}
\rotatebox[origin=c]{90}{\scriptsize ACL } & 
\hspace{-0.45cm} \includegraphics[align=c, width=1.45cm]{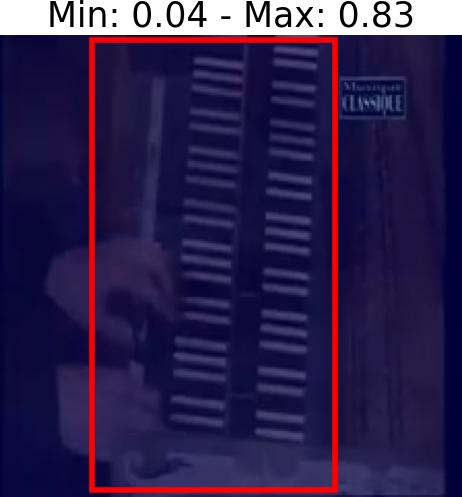}&
\hspace{-0.45cm} \includegraphics[align=c, width=1.45cm]{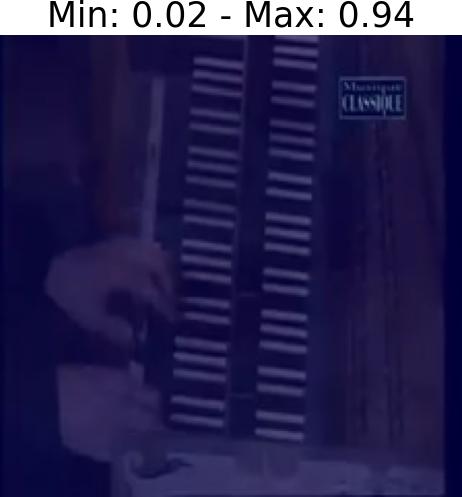}&
\hspace{-0.45cm} \includegraphics[align=c, width=1.45cm]{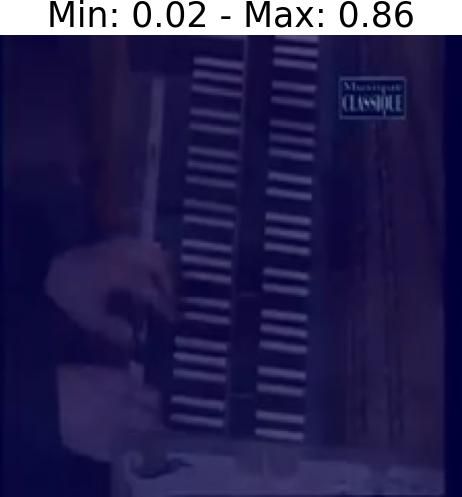}&
\hspace{-0.45cm} \includegraphics[align=c, width=1.45cm]{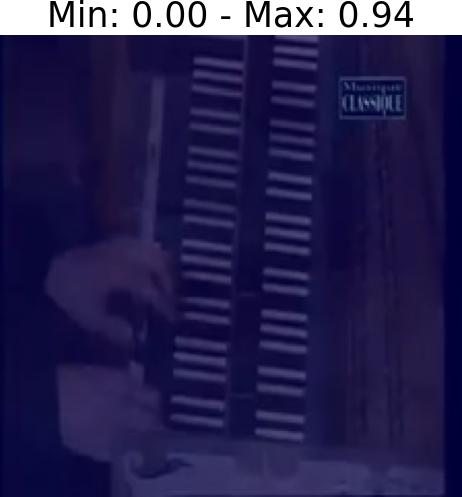}&
&
\hspace{-0.45cm} \includegraphics[align=c, width=1.45cm]{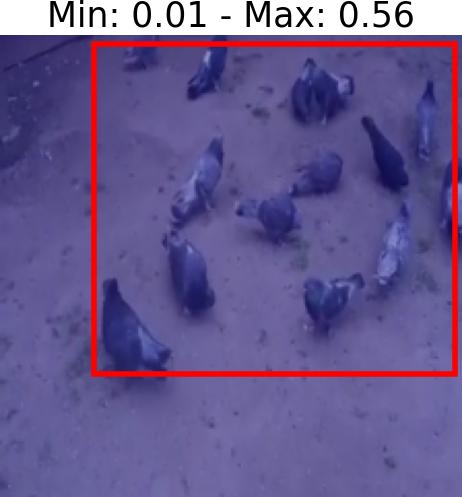}&
\hspace{-0.45cm} \includegraphics[align=c, width=1.45cm]{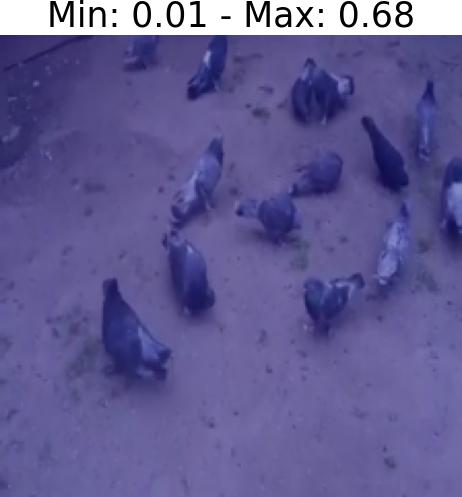}&
\hspace{-0.45cm} \includegraphics[align=c, width=1.45cm]{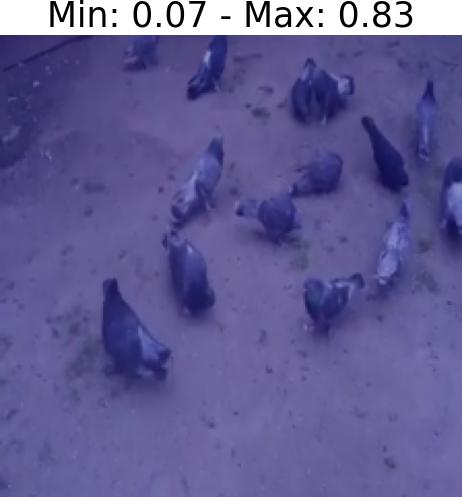}&
\hspace{-0.45cm} \includegraphics[align=c, width=1.45cm]{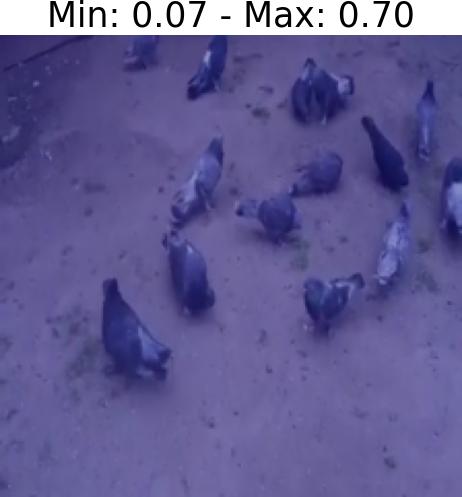}\\ \vspace{0.05cm}
\rotatebox[origin=c]{90}{\scriptsize \textbf{SSL-SaN} } & 
\hspace{-0.45cm} \includegraphics[align=c, width=1.45cm]{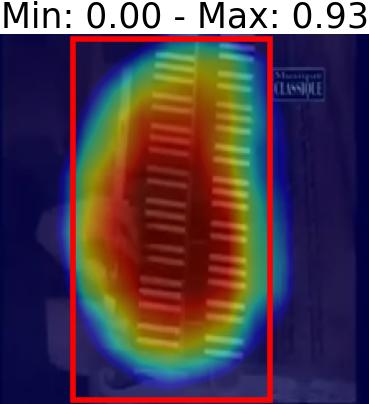}&
\hspace{-0.45cm} \includegraphics[align=c, width=1.45cm]{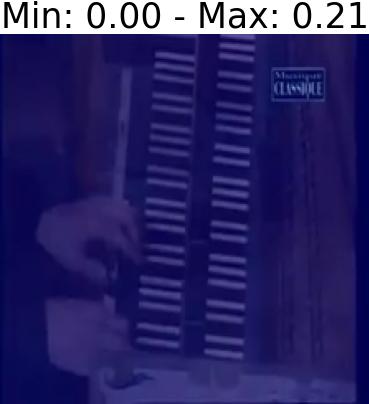}&
\hspace{-0.45cm} \includegraphics[align=c, width=1.45cm]{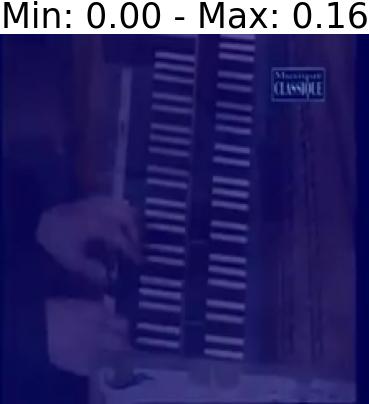}&
\hspace{-0.45cm} \includegraphics[align=c, width=1.45cm]{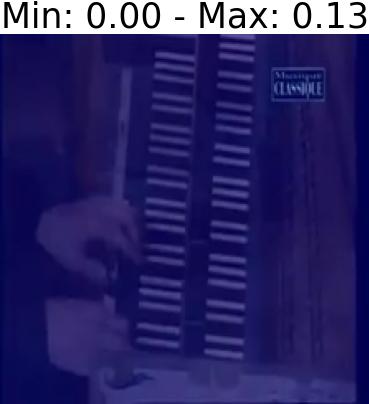}&
&
\hspace{-0.45cm} \includegraphics[align=c, width=1.45cm]{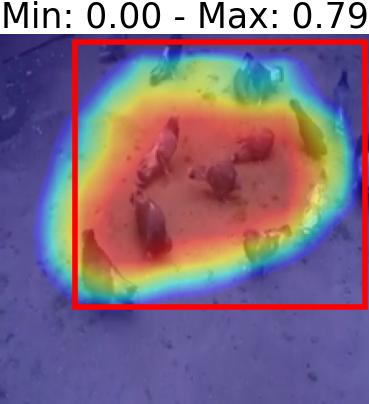}&
\hspace{-0.45cm} \includegraphics[align=c, width=1.45cm]{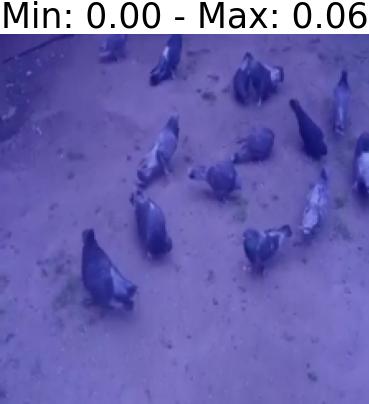}&
\hspace{-0.45cm} \includegraphics[align=c, width=1.45cm]{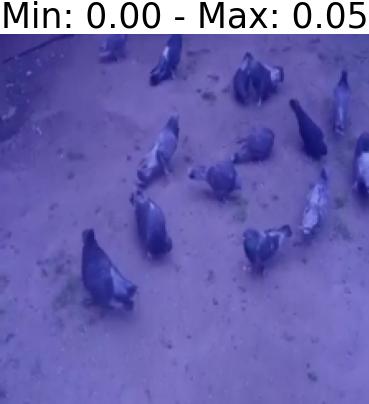}&
\hspace{-0.45cm} \includegraphics[align=c, width=1.45cm]{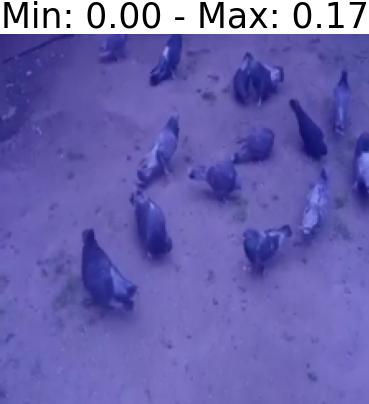}
\end{tabular}
}
\vspace{0.01cm}
\caption{Example of localization results of the different models in both \textit{positive} and \textit{negative} audio samples in VGG-SS test set. The audio-visual similarities below the universal threshold of each model are clipped, normalized to the interval $[0, 1]$, and overlaid with the original image We show the bounding box of the positive cases for better understanding. The min. and max. values of the audio-visual similarities are reported on top of each image.}
\label{fig:sup_qualitative_vggss}
\end{figure*}

\new{We present in Figure \ref{fig:sup_qualitative_vggss} two examples illustrating how our model outperforms others in the VGG-SS test set. SSL-TIE and our model are the only ones correctly localizing the sound coming from the piano, but SSL-TIE incorrectly filters the silence and noise. Moreover, our model highlights a bigger part of the piano, giving a better localization.

The second example depicts multiple chickens. Similar to the previous case, SSL-TIE and SSL-SaN are the only models able to localize the correct region for the positive sound, but SSL-TIE highlights part of the chickens when an offscreen sound is played.}
\clearpage

\subsubsection{IS3+}
\begin{figure*}[!ht]
\label{fig:inferences}
\centering
\resizebox{\textwidth}{!}{%
\begin{tabular}{ccccccccccccc}
& \scriptsize \hspace{-0.35cm} \raisebox{-0.17cm}{\includegraphics[height=0.5cm]{Figures/equalizer.png}} Cattle & \scriptsize \hspace{-0.35cm} \raisebox{-0.17cm}{\includegraphics[height=0.5cm]{Figures/equalizer.png}} Stream & \scriptsize \hspace{-0.5cm} \raisebox{-0.17cm}{\includegraphics[height=0.5cm]{Figures/equalizer.png}} Silence & \scriptsize \hspace{-0.6cm} \raisebox{-0.17cm}{\includegraphics[height=0.5cm]{Figures/equalizer.png}} Noise & \hspace{-0.52cm} \raisebox{-0.17cm}{\includegraphics[height=0.5cm]{Figures/equalizer.png}} \scriptsize Offscreen & & \scriptsize \hspace{-0.35cm} \raisebox{-0.17cm}{\includegraphics[height=0.5cm]{Figures/equalizer.png}} Fireworks & \scriptsize \hspace{-0.35cm} \raisebox{-0.17cm}{\includegraphics[height=0.5cm]{Figures/equalizer.png}} Firetruck & \scriptsize \hspace{-0.5cm} \raisebox{-0.17cm}{\includegraphics[height=0.5cm]{Figures/equalizer.png}} Silence & \scriptsize \hspace{-0.6cm} \raisebox{-0.17cm}{\includegraphics[height=0.5cm]{Figures/equalizer.png}} Noise & \hspace{-0.52cm} \raisebox{-0.17cm}{\includegraphics[height=0.5cm]{Figures/equalizer.png}} \scriptsize Offscreen \\
\vspace{0.05cm} 
\rotatebox[origin=c]{90}{\scriptsize LVS } & 
\hspace{-0.45cm} \includegraphics[align=c, width=1.45cm]{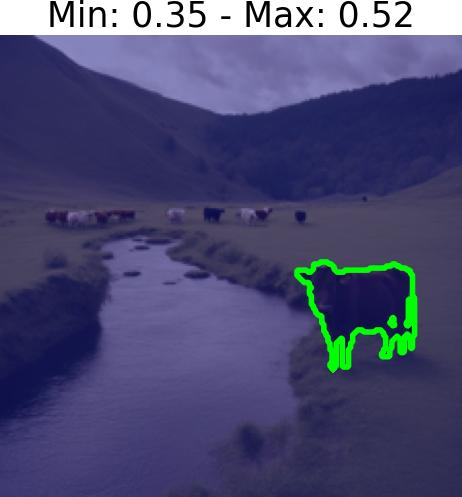}&
\hspace{-0.45cm} \includegraphics[align=c, width=1.45cm]{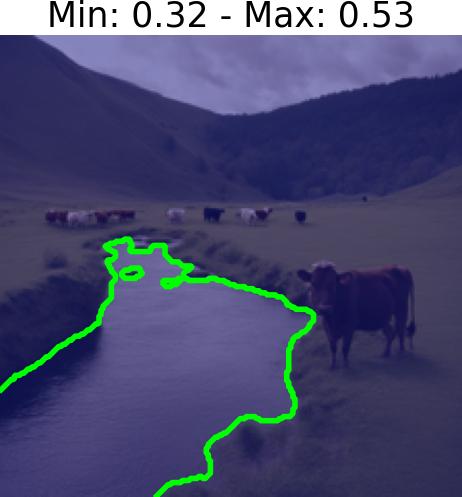}&
\hspace{-0.45cm} \includegraphics[align=c, width=1.45cm]{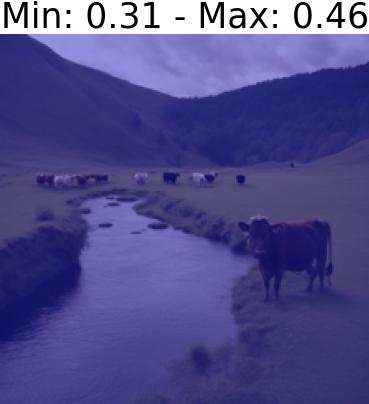}&
\hspace{-0.45cm} \includegraphics[align=c, width=1.45cm]{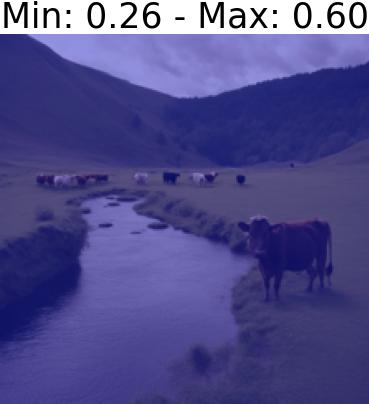}&
\hspace{-0.45cm} \includegraphics[align=c, width=1.45cm]{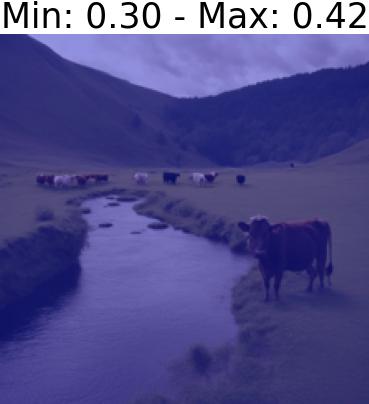}&
&
\hspace{-0.45cm} \includegraphics[align=c, width=1.45cm]{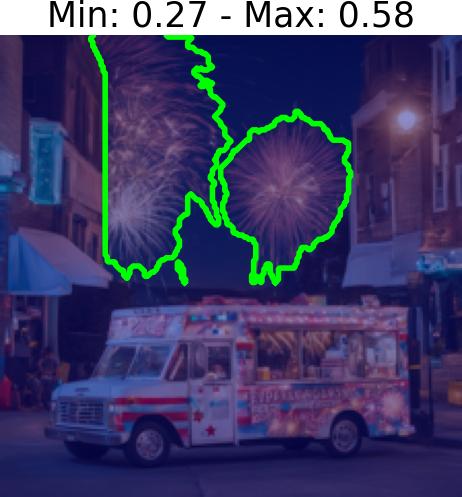}&
\hspace{-0.45cm} \includegraphics[align=c, width=1.45cm]{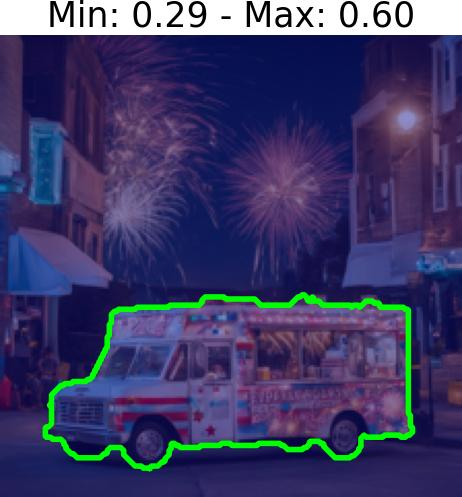}&
\hspace{-0.45cm} \includegraphics[align=c, width=1.45cm]{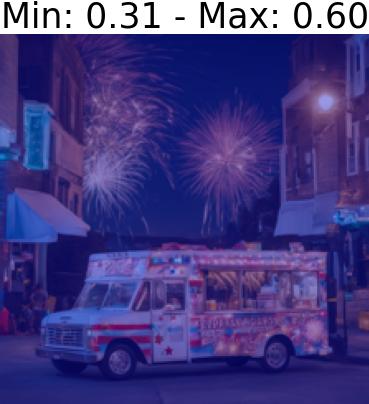}&
\hspace{-0.45cm} \includegraphics[align=c, width=1.45cm]{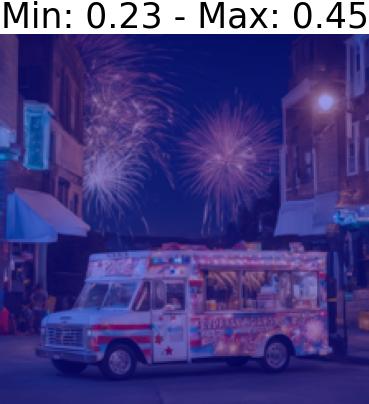}&
\hspace{-0.45cm} \includegraphics[align=c, width=1.45cm]{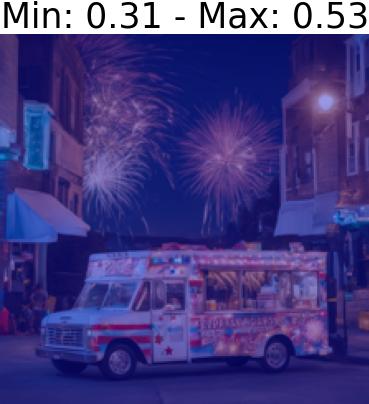}\\ 
\vspace{0.05cm}
\rotatebox[origin=c]{90}{\scriptsize EZ-VSL } & 
\hspace{-0.45cm} \includegraphics[align=c, width=1.45cm]{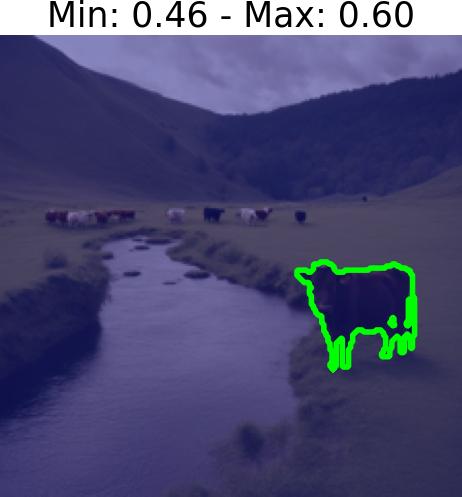}&
\hspace{-0.45cm} \includegraphics[align=c, width=1.45cm]{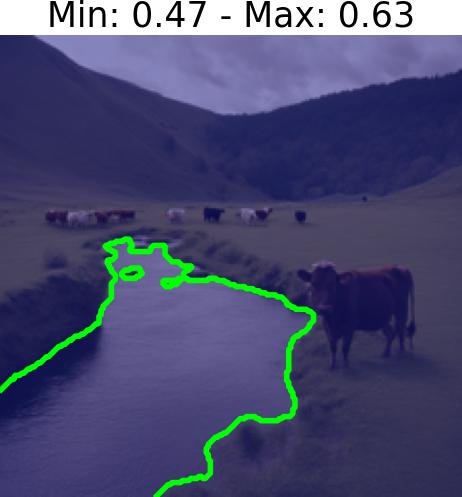}&
\hspace{-0.45cm} \includegraphics[align=c, width=1.45cm]{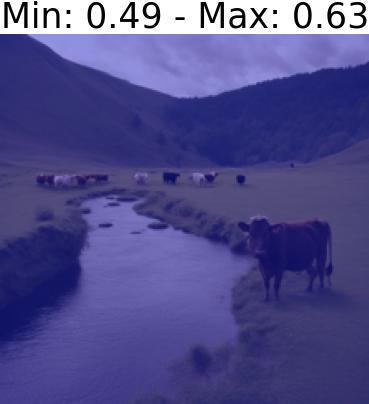}&
\hspace{-0.45cm} \includegraphics[align=c, width=1.45cm]{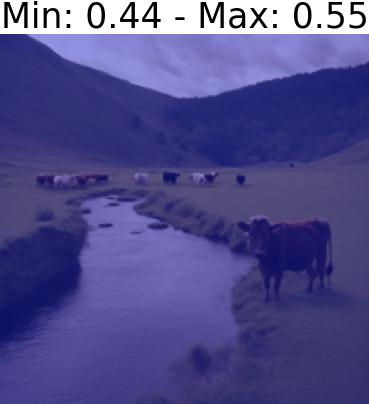}&
\hspace{-0.45cm} \includegraphics[align=c, width=1.45cm]{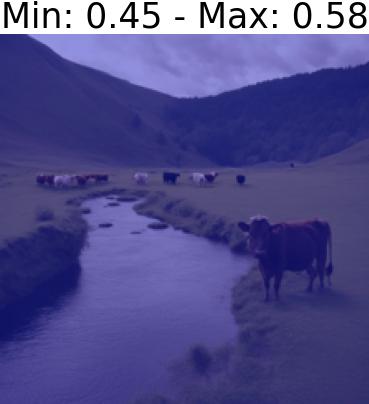}&
&
\hspace{-0.45cm} \includegraphics[align=c, width=1.45cm]{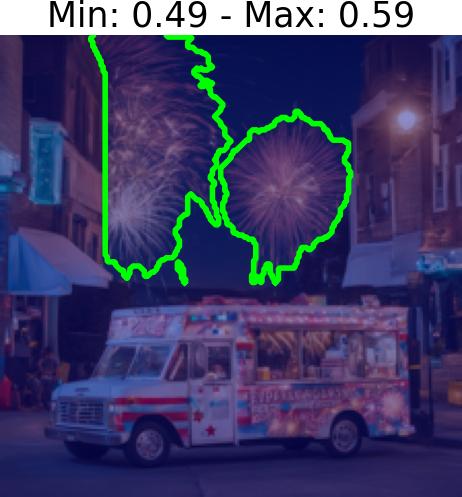}&
\hspace{-0.45cm} \includegraphics[align=c, width=1.45cm]{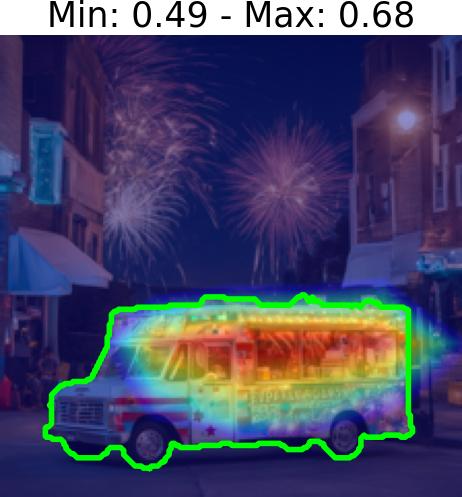}&
\hspace{-0.45cm} \includegraphics[align=c, width=1.45cm]{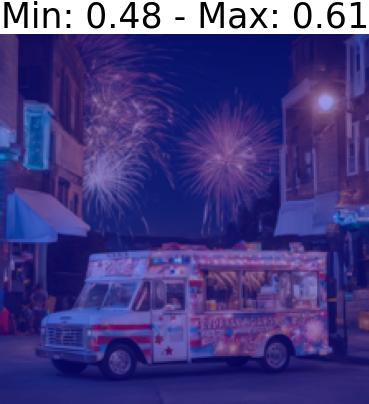}&
\hspace{-0.45cm} \includegraphics[align=c, width=1.45cm]{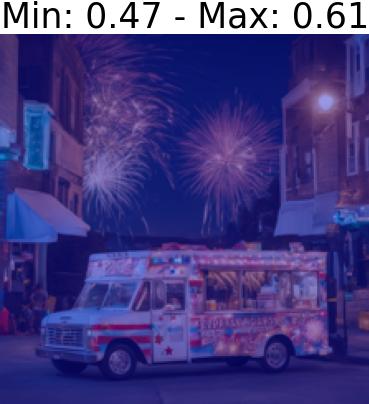}&
\hspace{-0.45cm} \includegraphics[align=c, width=1.45cm]{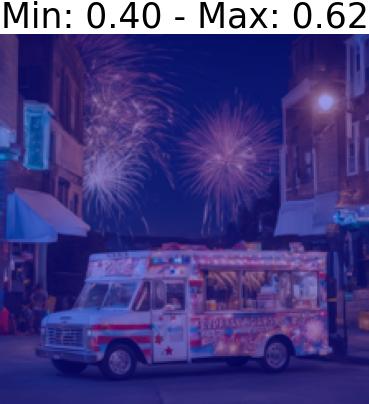}\\
\vspace{0.05cm}
\rotatebox[origin=c]{90}{\scriptsize FNAC } & 
\hspace{-0.45cm} \includegraphics[align=c, width=1.45cm]{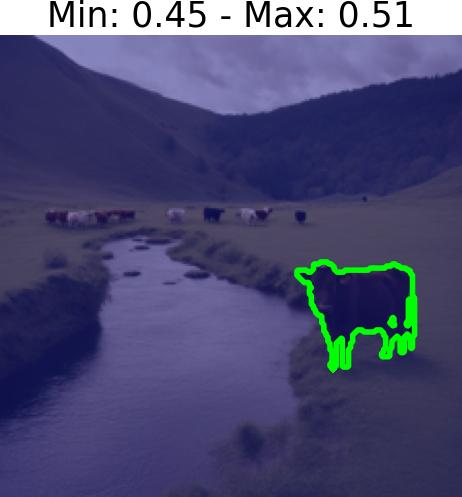}&
\hspace{-0.45cm} \includegraphics[align=c, width=1.45cm]{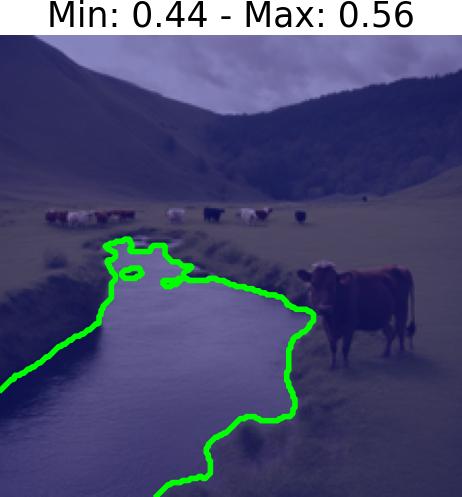}&
\hspace{-0.45cm} \includegraphics[align=c, width=1.45cm]{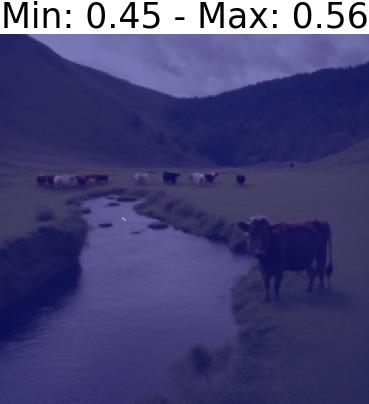}&
\hspace{-0.45cm} \includegraphics[align=c, width=1.45cm]{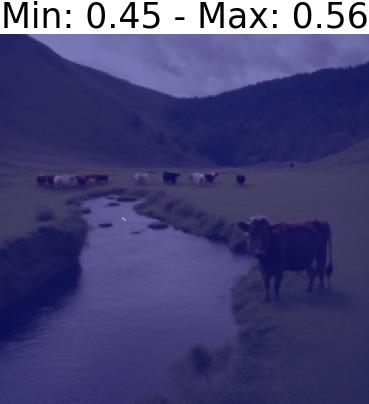}&
\hspace{-0.45cm} \includegraphics[align=c, width=1.45cm]{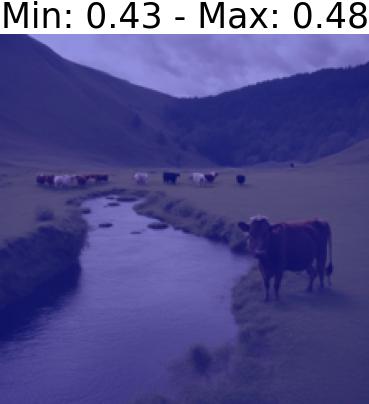}&
&
\hspace{-0.45cm} \includegraphics[align=c, width=1.45cm]{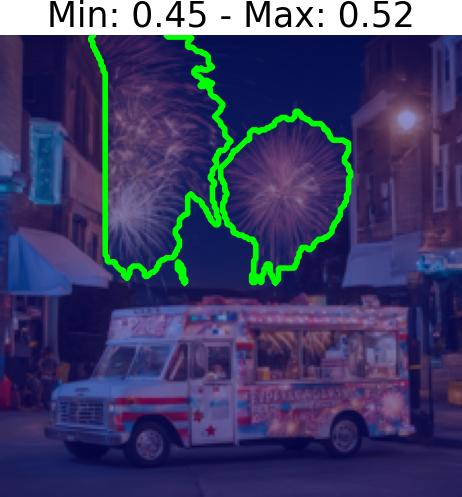}&
\hspace{-0.45cm} \includegraphics[align=c, width=1.45cm]{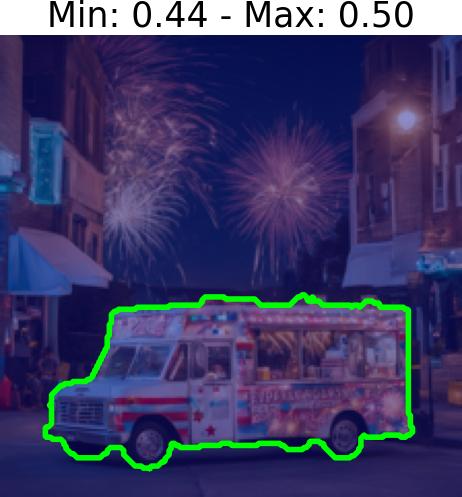}&
\hspace{-0.45cm} \includegraphics[align=c, width=1.45cm]{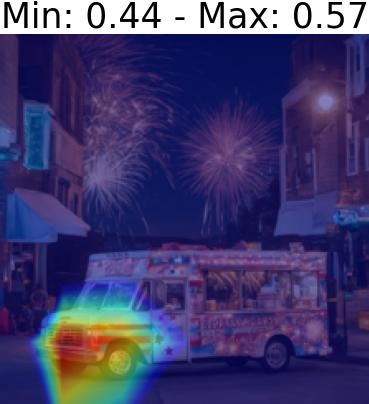}&
\hspace{-0.45cm} \includegraphics[align=c, width=1.45cm]{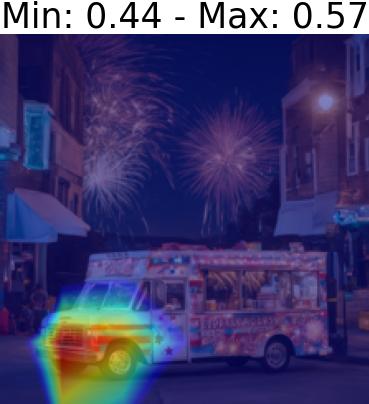}&
\hspace{-0.45cm} \includegraphics[align=c, width=1.45cm]{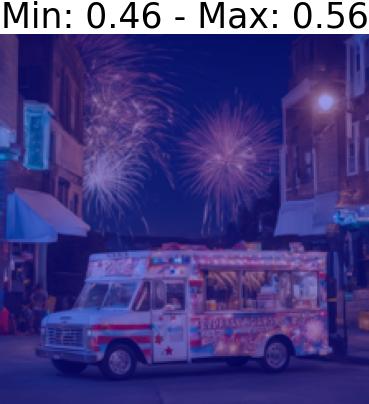}\\
\vspace{0.05cm}
\rotatebox[origin=c]{90}{\scriptsize SLAVC } & 
\hspace{-0.45cm} \includegraphics[align=c, width=1.45cm]{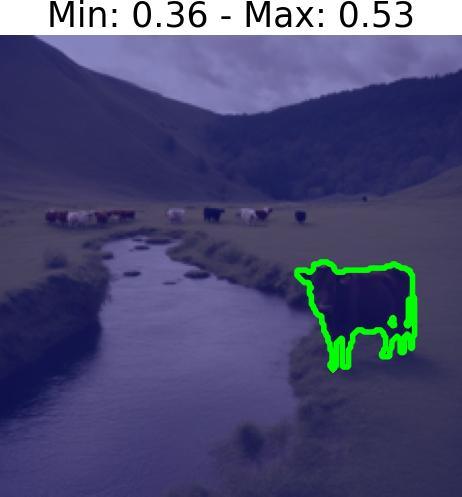}&
\hspace{-0.45cm} \includegraphics[align=c, width=1.45cm]{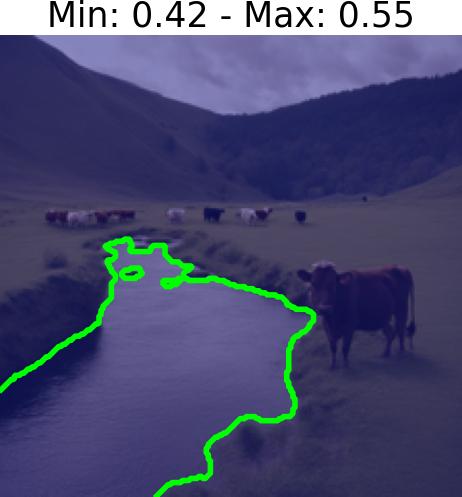}&
\hspace{-0.45cm} \includegraphics[align=c, width=1.45cm]{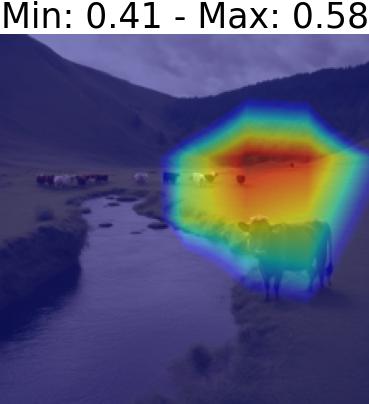}&
\hspace{-0.45cm} \includegraphics[align=c, width=1.45cm]{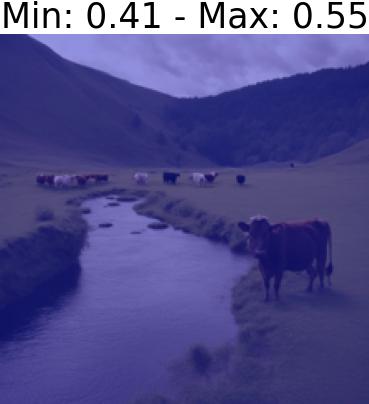}&
\hspace{-0.45cm} \includegraphics[align=c, width=1.45cm]{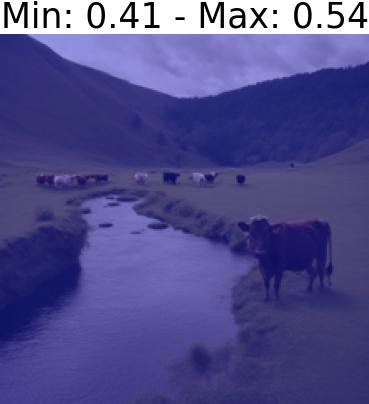}&
&
\hspace{-0.45cm} \includegraphics[align=c, width=1.45cm]{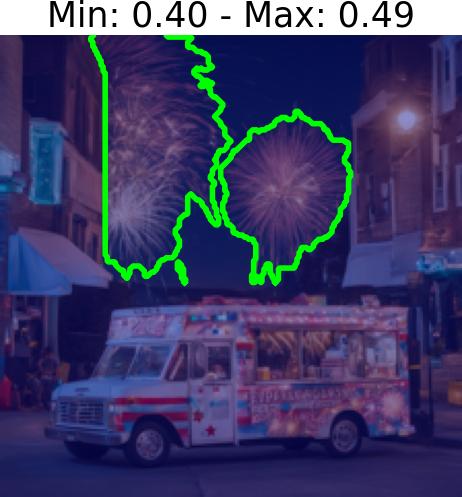}&
\hspace{-0.45cm} \includegraphics[align=c, width=1.45cm]{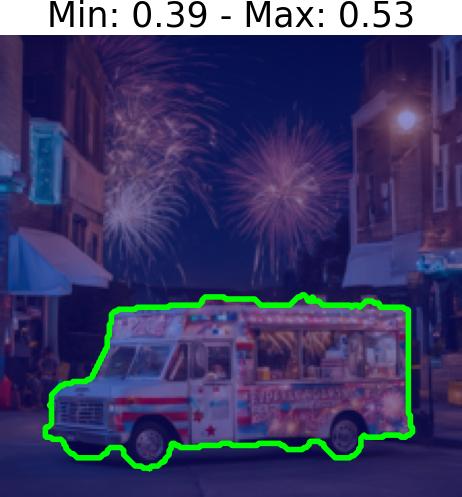}&
\hspace{-0.45cm} \includegraphics[align=c, width=1.45cm]{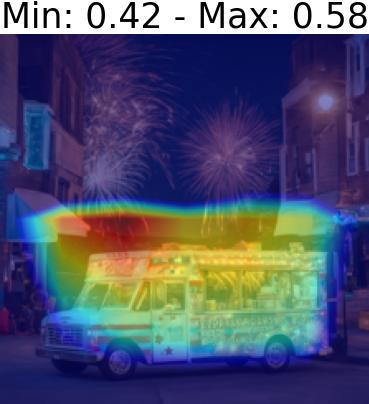}&
\hspace{-0.45cm} \includegraphics[align=c, width=1.45cm]{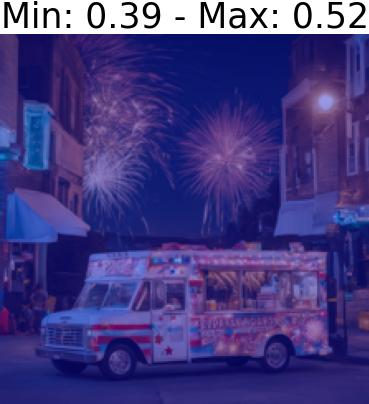}&
\hspace{-0.45cm} \includegraphics[align=c, width=1.45cm]{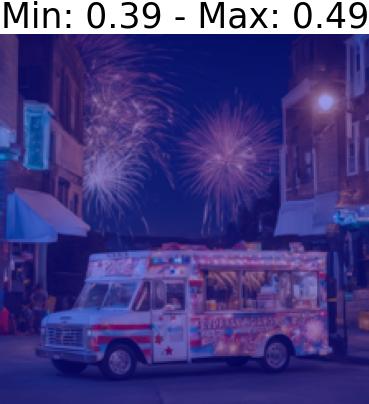}\\
\vspace{0.05cm}
\rotatebox[origin=c]{90}{\scriptsize SSL-TIE } & 
\hspace{-0.45cm} \includegraphics[align=c, width=1.45cm]{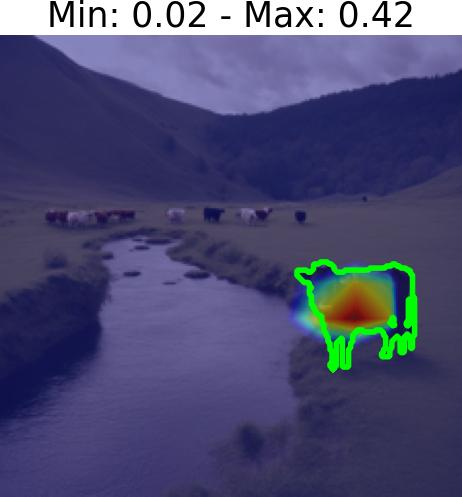}&
\hspace{-0.45cm} \includegraphics[align=c, width=1.45cm]{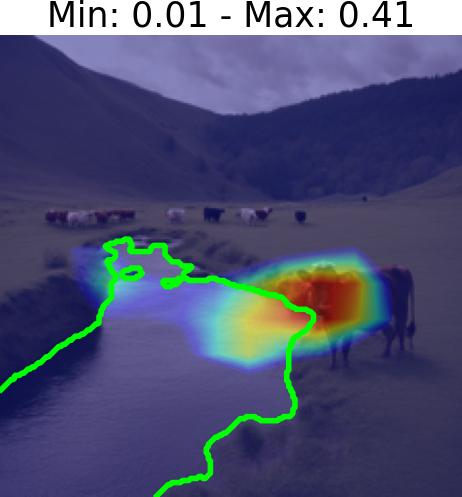}&
\hspace{-0.45cm} \includegraphics[align=c, width=1.45cm]{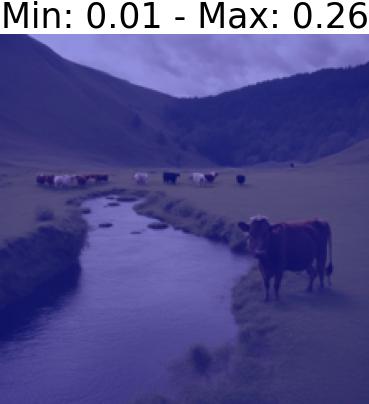}&
\hspace{-0.45cm} \includegraphics[align=c, width=1.45cm]{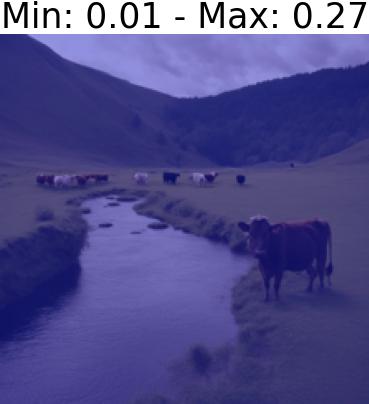}&
\hspace{-0.45cm} \includegraphics[align=c, width=1.45cm]{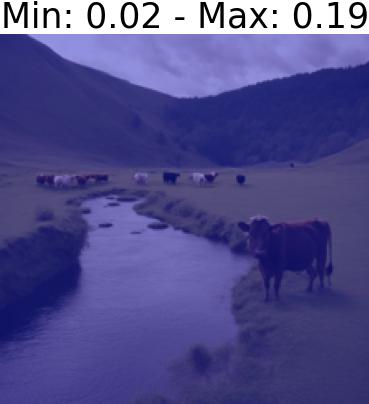}&
&
\hspace{-0.45cm} \includegraphics[align=c, width=1.45cm]{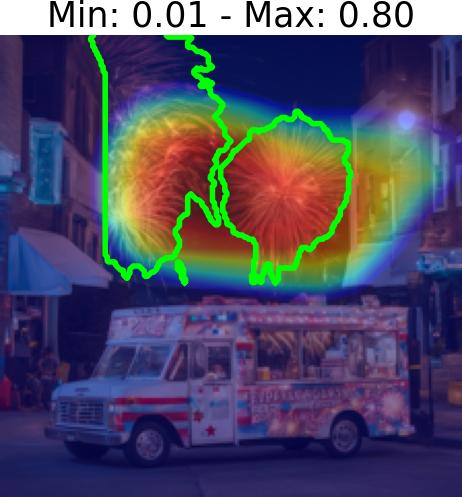}&
\hspace{-0.45cm} \includegraphics[align=c, width=1.45cm]{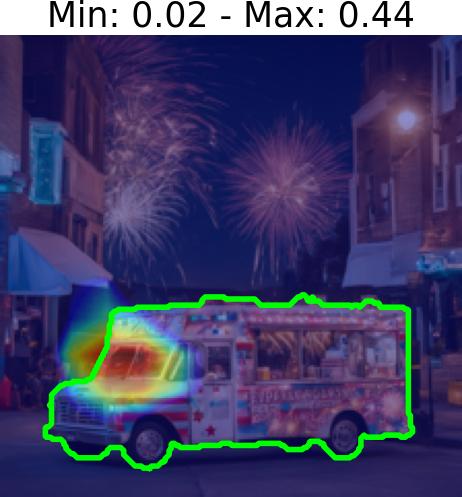}&
\hspace{-0.45cm} \includegraphics[align=c, width=1.45cm]{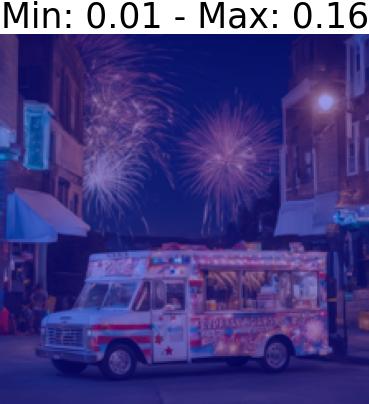}&
\hspace{-0.45cm} \includegraphics[align=c, width=1.45cm]{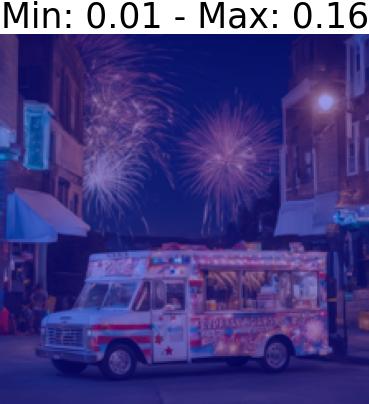}&
\hspace{-0.45cm} \includegraphics[align=c, width=1.45cm]{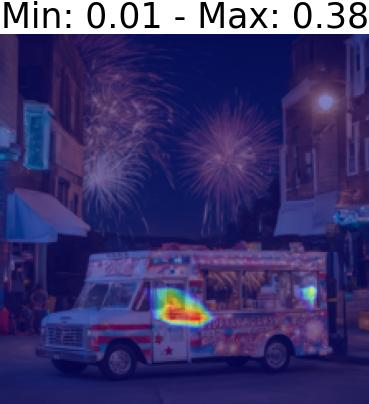}\\
\vspace{0.05cm}
\rotatebox[origin=c]{90}{\scriptsize SSL-Align (S.S.) } & 
\hspace{-0.45cm} \includegraphics[align=c, width=1.45cm]{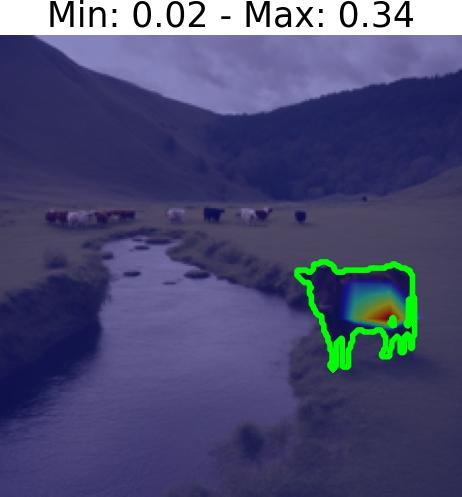}&
\hspace{-0.45cm} \includegraphics[align=c, width=1.45cm]{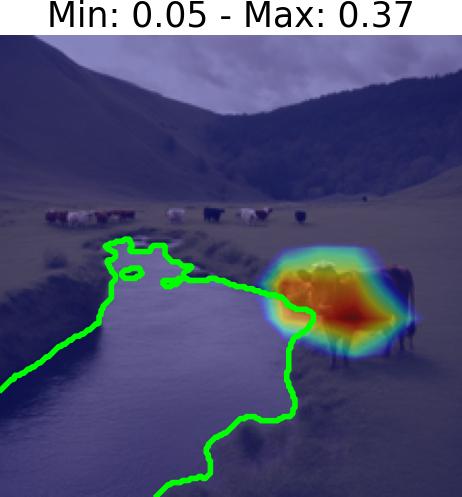}&
\hspace{-0.45cm} \includegraphics[align=c, width=1.45cm]{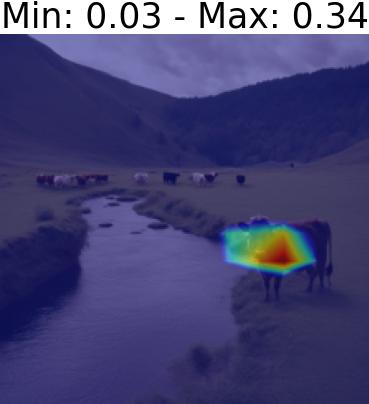}&
\hspace{-0.45cm} \includegraphics[align=c, width=1.45cm]{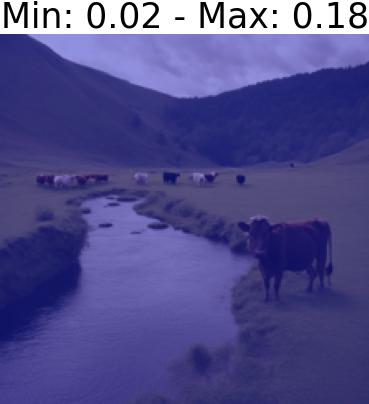}&
\hspace{-0.45cm} \includegraphics[align=c, width=1.45cm]{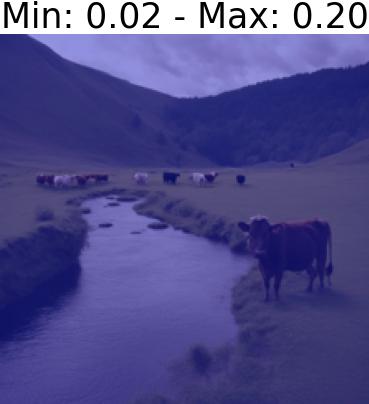}&
&
\hspace{-0.45cm} \includegraphics[align=c, width=1.45cm]{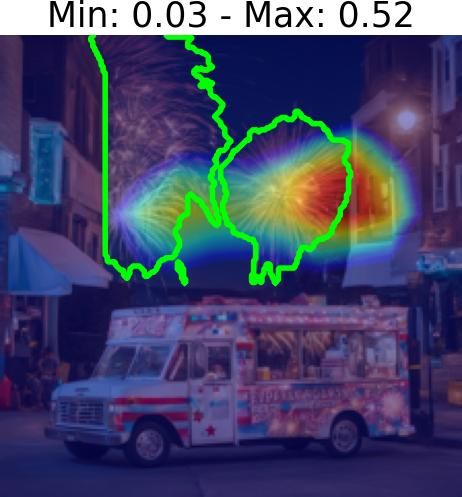}&
\hspace{-0.45cm} \includegraphics[align=c, width=1.45cm]{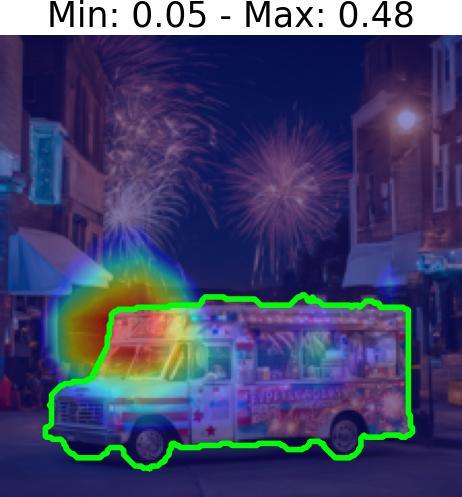}&
\hspace{-0.45cm} \includegraphics[align=c, width=1.45cm]{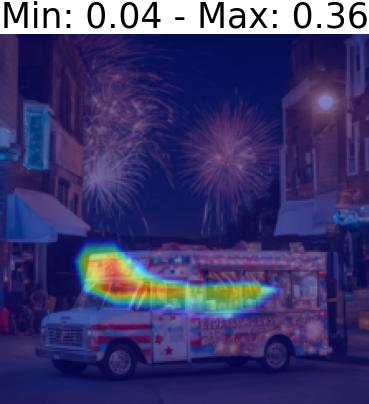}&
\hspace{-0.45cm} \includegraphics[align=c, width=1.45cm]{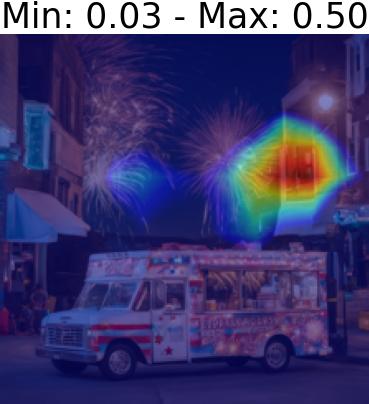}&
\hspace{-0.45cm} \includegraphics[align=c, width=1.45cm]{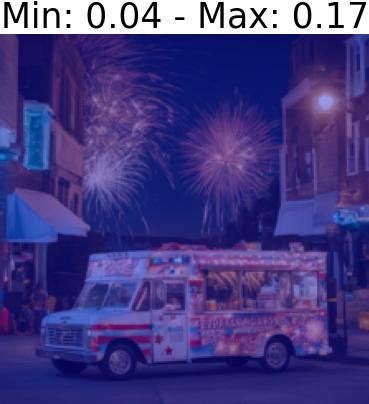}\\ \vspace{0.05cm}
\rotatebox[origin=c]{90}{\scriptsize ACL } & 
\hspace{-0.45cm} \includegraphics[align=c, width=1.45cm]{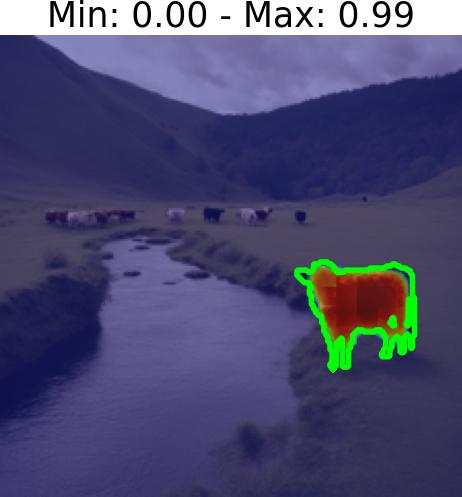}&
\hspace{-0.45cm} \includegraphics[align=c, width=1.45cm]{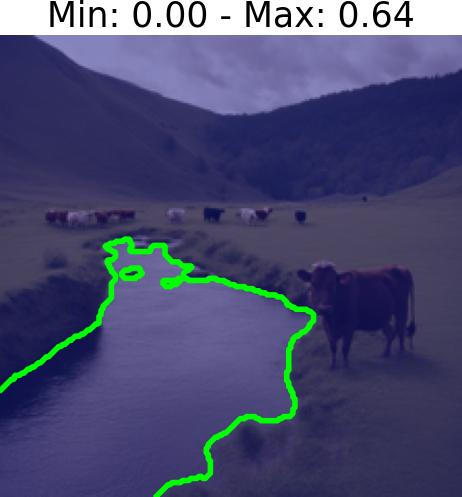}&
\hspace{-0.45cm} \includegraphics[align=c, width=1.45cm]{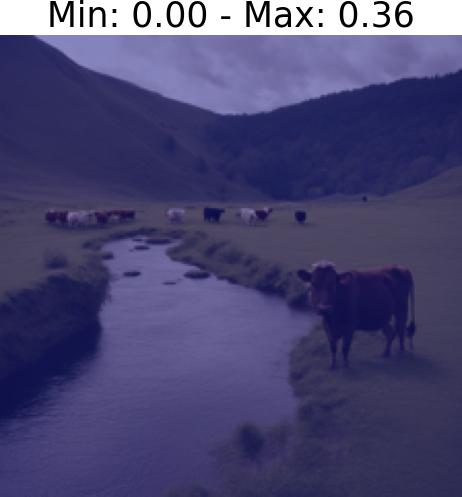}&
\hspace{-0.45cm} \includegraphics[align=c, width=1.45cm]{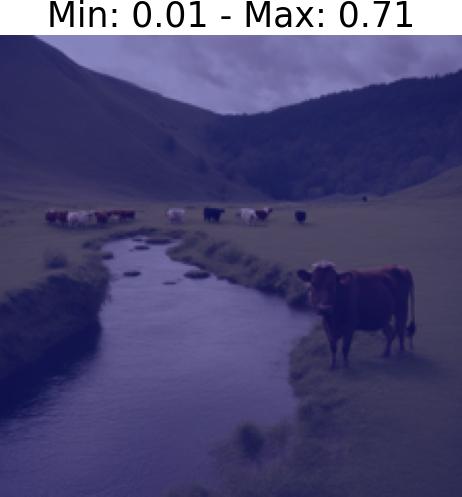}&
\hspace{-0.45cm} \includegraphics[align=c, width=1.45cm]{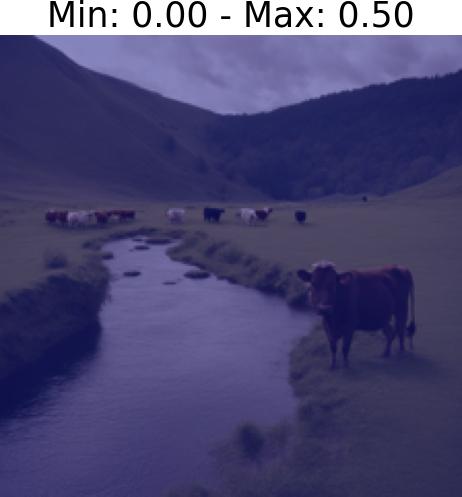}&
&
\hspace{-0.45cm} \includegraphics[align=c, width=1.45cm]{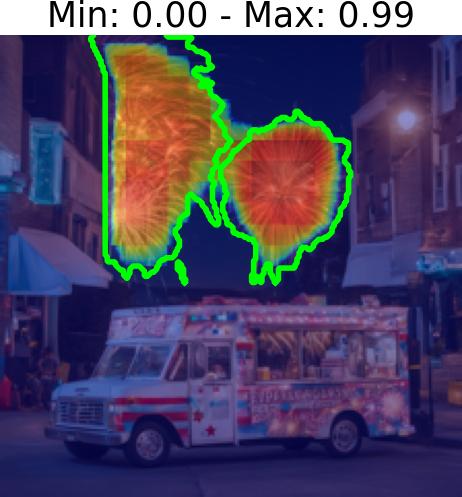}&
\hspace{-0.45cm} \includegraphics[align=c, width=1.45cm]{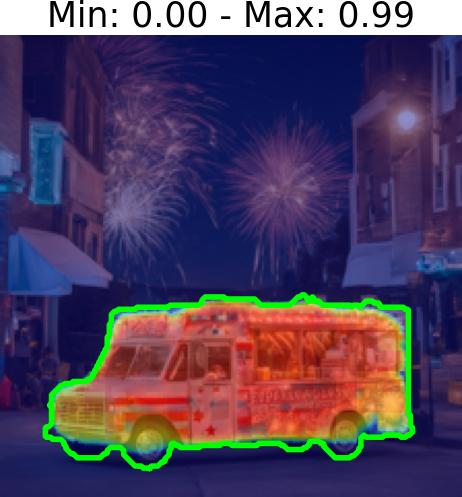}&
\hspace{-0.45cm} \includegraphics[align=c, width=1.45cm]{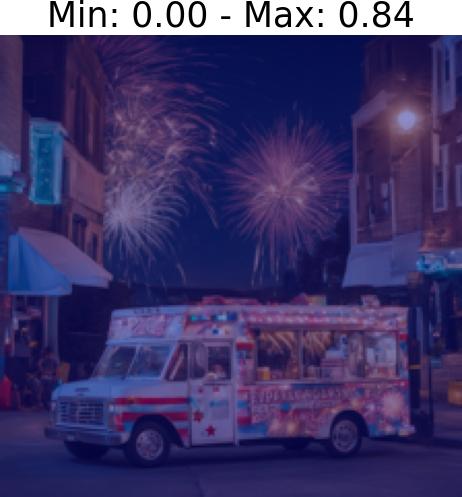}&
\hspace{-0.45cm} \includegraphics[align=c, width=1.45cm]{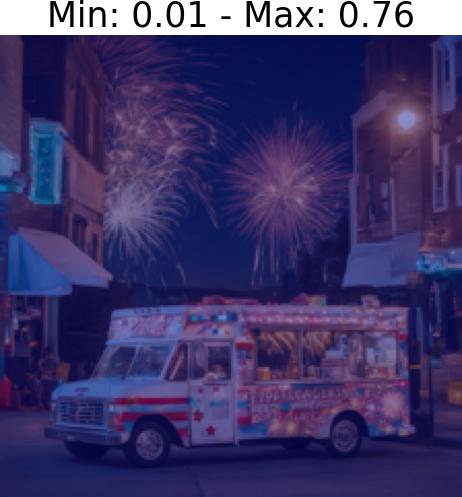}&
\hspace{-0.45cm} \includegraphics[align=c, width=1.45cm]{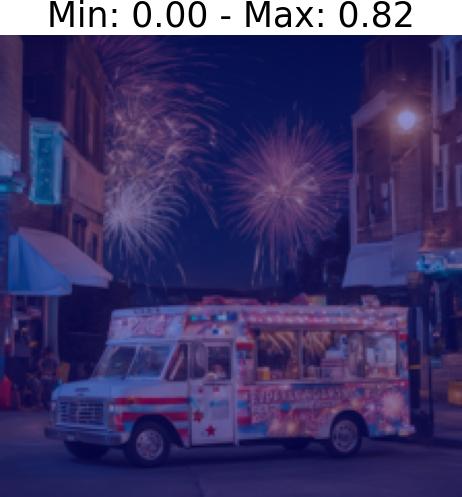}\\ \vspace{0.05cm}
\rotatebox[origin=c]{90}{\scriptsize \textbf{SSL-SaN} } & 
\hspace{-0.45cm} \includegraphics[align=c, width=1.45cm]{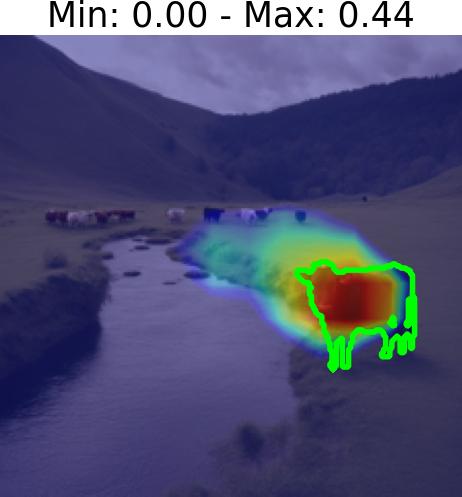}&
\hspace{-0.45cm} \includegraphics[align=c, width=1.45cm]{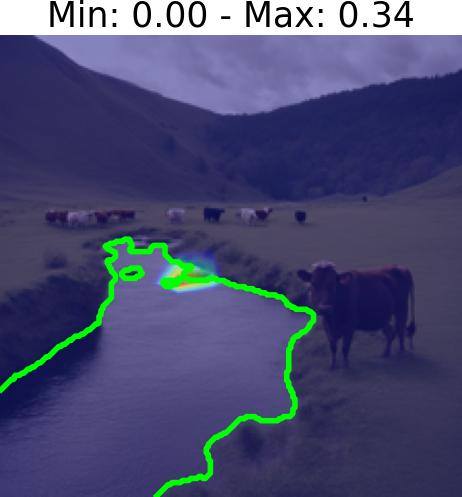}&
\hspace{-0.45cm} \includegraphics[align=c, width=1.45cm]{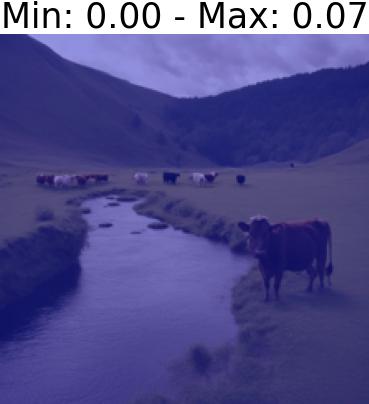}&
\hspace{-0.45cm} \includegraphics[align=c, width=1.45cm]{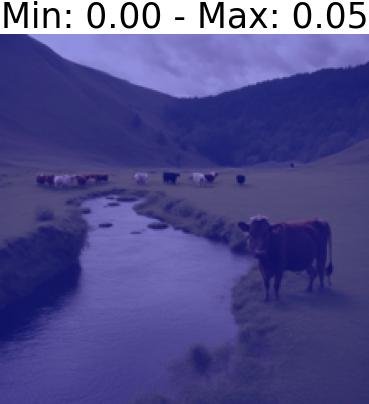}&
\hspace{-0.45cm} \includegraphics[align=c, width=1.45cm]{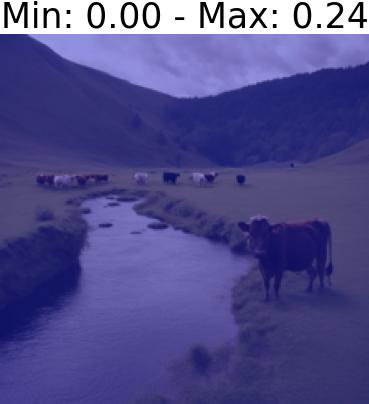}&
&
\hspace{-0.45cm} \includegraphics[align=c, width=1.45cm]{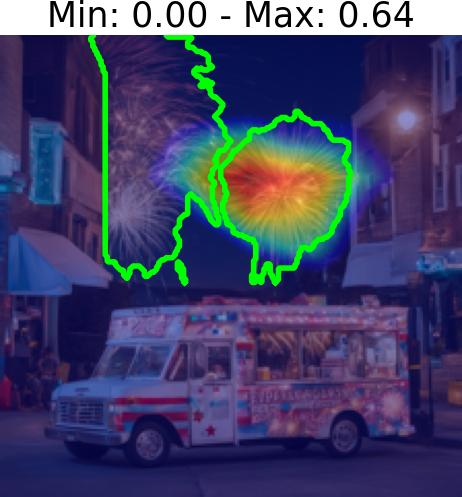}&
\hspace{-0.45cm} \includegraphics[align=c, width=1.45cm]{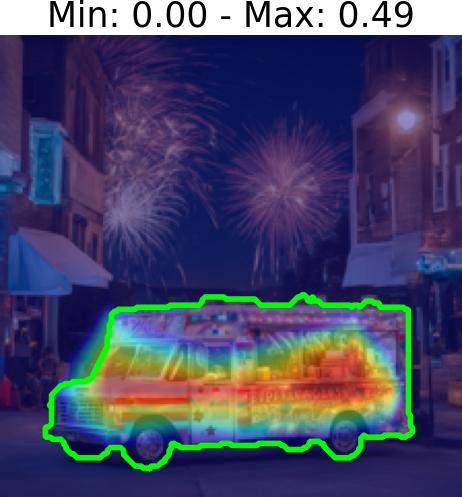}&
\hspace{-0.45cm} \includegraphics[align=c, width=1.45cm]{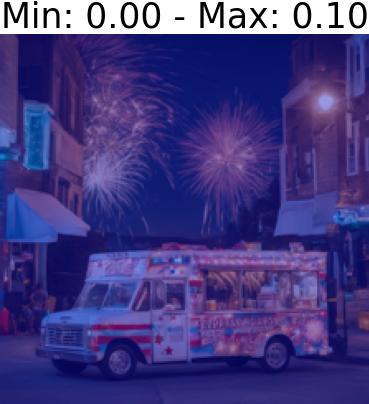}&
\hspace{-0.45cm} \includegraphics[align=c, width=1.45cm]{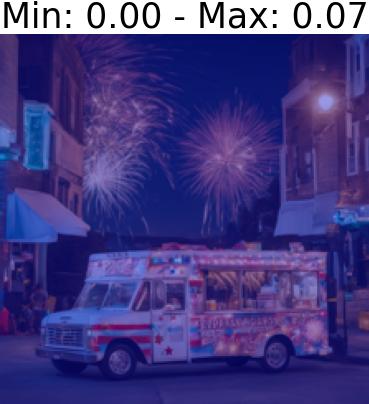}&
\hspace{-0.45cm} \includegraphics[align=c, width=1.45cm]{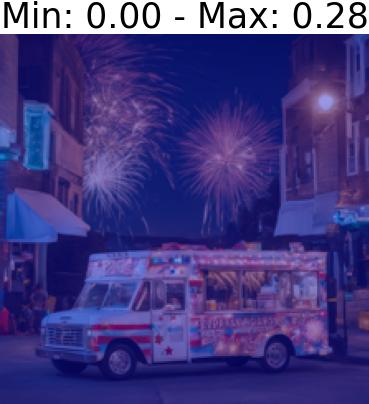}
\end{tabular}
}
\vspace{0.01cm}
\caption{Example of localization results of the different models in both \textit{positive} and \textit{negative} audio samples in the extended IS3+ test set. The audio-visual similarities below the universal threshold of each model are clipped, normalized to the interval [0, 1], and overlaid with the original image. We show the outline of the segmentation mask (in green) of the positive cases for better understanding. The min. and max. values of the audio-visual similarities are reported on top of each image.}
\label{fig:sup_qualitative_is3+}
\end{figure*}

\new{
In the first example of Figure \ref{fig:sup_qualitative_is3+}, the scene shows a cow near a river with additional cows in the background. When the sound corresponds to the cow, SSL-SaN, ACL, and SSL-TIE all correctly localize it. However, when the sound corresponds to the river, only SSL-TIE and SSL-SaN produce activations: SSL-TIE localizes closer to the cow, while SSL-SaN focuses on the rocks in the river. None of the models successfully highlight the background cows.

In the second example, the image contains a firetruck in the foreground and fireworks in the sky. ACL delivers nearly perfect localization, accurately identifying the firetruck while filtering out negative sounds. SSL-SaN also localizes both positive sounds and filters negatives, though with slightly less precision. Notice that ACL leverages a network trained with a supervised segmentation loss, while our model has been trained completely from scratch in a contrastive way. On the other hand, SSL-TIE and SSL-Align, while correctly detecting the positive sources, fail to fully suppress the negative sounds.}

\subsubsection{AVSBench S4}
\begin{figure*}[!ht]
\label{fig:inferences}
\centering
\resizebox{\textwidth}{!}{%
\begin{tabular}{ccccccccccc}
& \scriptsize \hspace{-0.35cm} \raisebox{-0.17cm}{\includegraphics[height=0.5cm]{Figures/equalizer.png}} Race car & \scriptsize \hspace{-0.5cm} \raisebox{-0.17cm}{\includegraphics[height=0.5cm]{Figures/equalizer.png}} Silence & \scriptsize \hspace{-0.6cm} \raisebox{-0.17cm}{\includegraphics[height=0.5cm]{Figures/equalizer.png}} Noise & \hspace{-0.52cm} \raisebox{-0.17cm}{\includegraphics[height=0.5cm]{Figures/equalizer.png}} \scriptsize Offscreen & & \scriptsize \hspace{-0.35cm} \raisebox{-0.17cm}{\includegraphics[height=0.5cm]{Figures/equalizer.png}} Keyboard & \scriptsize \hspace{-0.5cm} \raisebox{-0.17cm}{\includegraphics[height=0.5cm]{Figures/equalizer.png}} Silence & \scriptsize \hspace{-0.6cm} \raisebox{-0.17cm}{\includegraphics[height=0.5cm]{Figures/equalizer.png}} Noise & \hspace{-0.52cm} \raisebox{-0.17cm}{\includegraphics[height=0.5cm]{Figures/equalizer.png}} \scriptsize Offscreen \\
\vspace{0.05cm}
\rotatebox[origin=c]{90}{\scriptsize LVS } & 
\hspace{-0.45cm} \includegraphics[align=c, width=1.45cm]{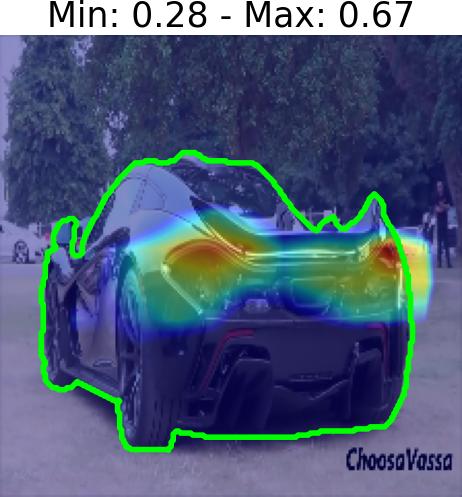}&
\hspace{-0.45cm} \includegraphics[align=c, width=1.45cm]{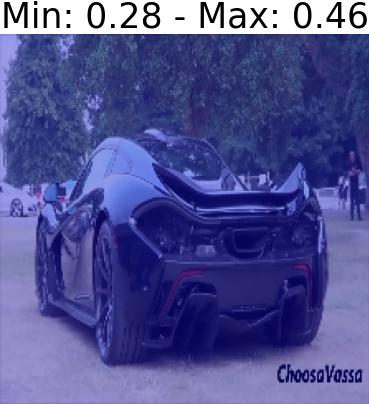}&
\hspace{-0.45cm} \includegraphics[align=c, width=1.45cm]{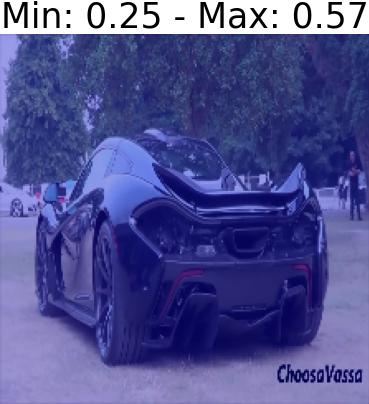}&
\hspace{-0.45cm} \includegraphics[align=c, width=1.45cm]{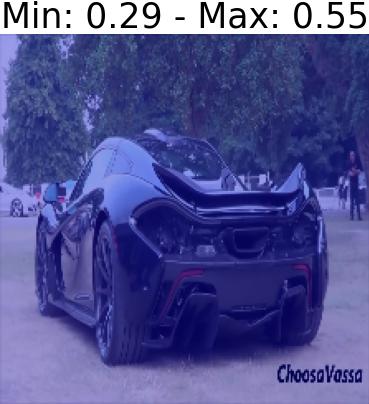}&
&
\hspace{-0.45cm} \includegraphics[align=c, width=1.45cm]{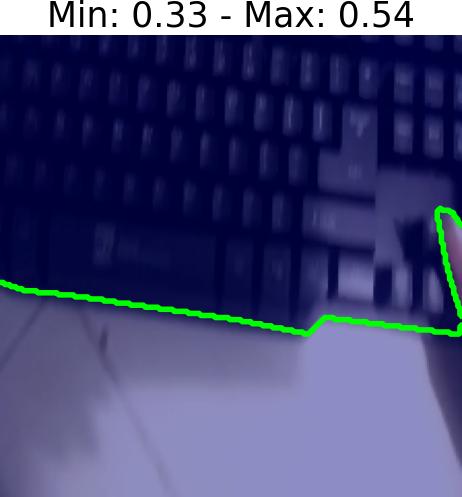}&
\hspace{-0.45cm} \includegraphics[align=c, width=1.45cm]{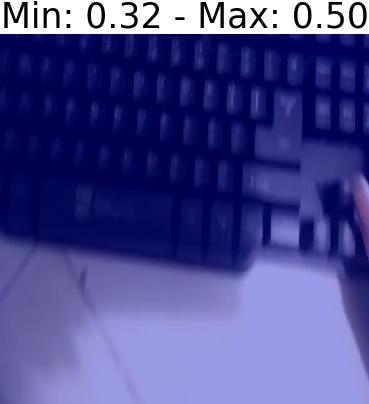}&
\hspace{-0.45cm} \includegraphics[align=c, width=1.45cm]{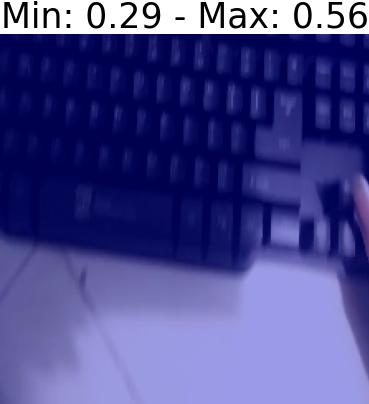}&
\hspace{-0.45cm} \includegraphics[align=c, width=1.45cm]{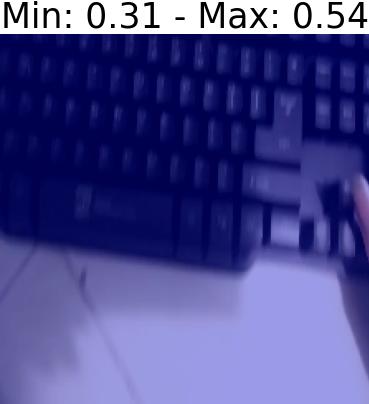}\\ \vspace{0.05cm}
\rotatebox[origin=c]{90}{\scriptsize EZ-VSL } & 
\hspace{-0.45cm} \includegraphics[align=c, width=1.45cm]{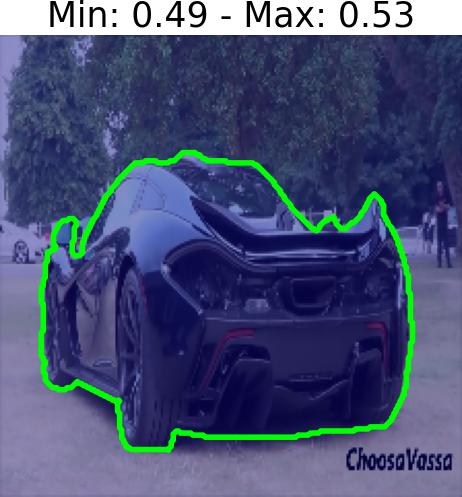}&
\hspace{-0.45cm} \includegraphics[align=c, width=1.45cm]{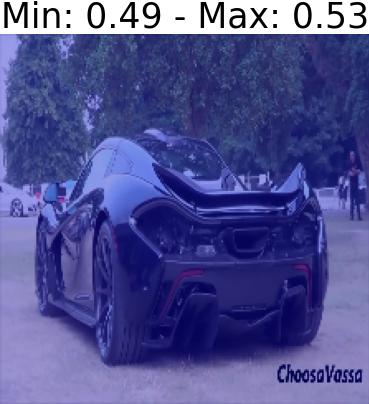}&
\hspace{-0.45cm} \includegraphics[align=c, width=1.45cm]{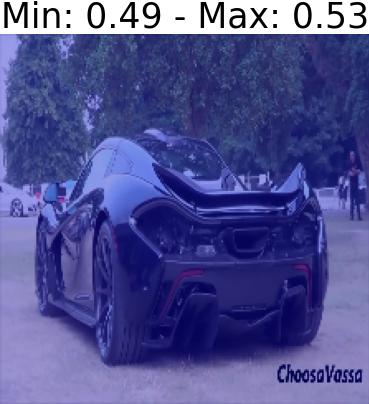}&
\hspace{-0.45cm} \includegraphics[align=c, width=1.45cm]{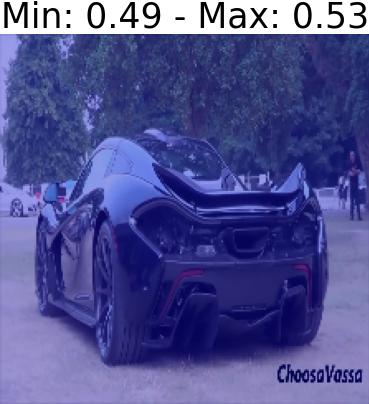}&
&
\hspace{-0.45cm} \includegraphics[align=c, width=1.45cm]{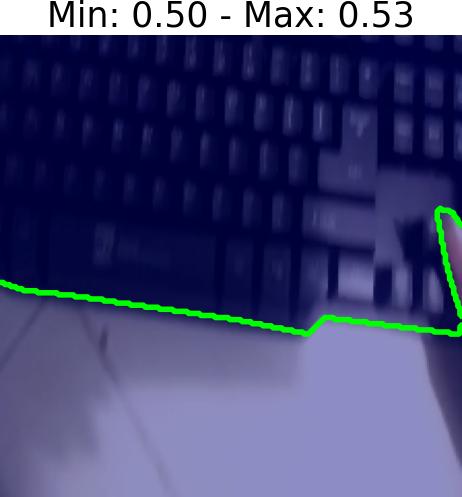}&
\hspace{-0.45cm} \includegraphics[align=c, width=1.45cm]{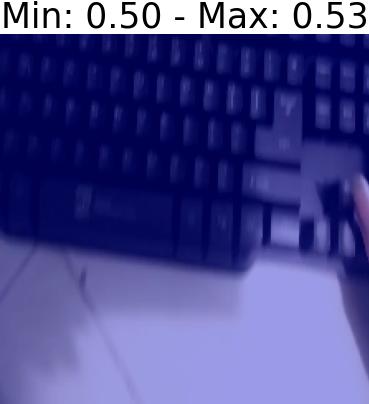}&
\hspace{-0.45cm} \includegraphics[align=c, width=1.45cm]{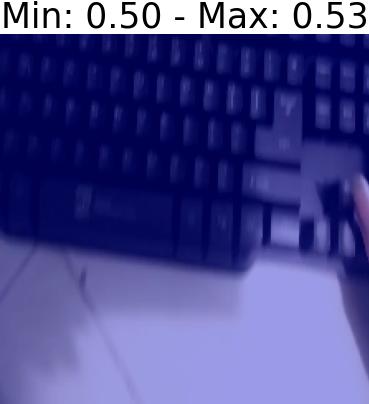}&
\hspace{-0.45cm} \includegraphics[align=c, width=1.45cm]{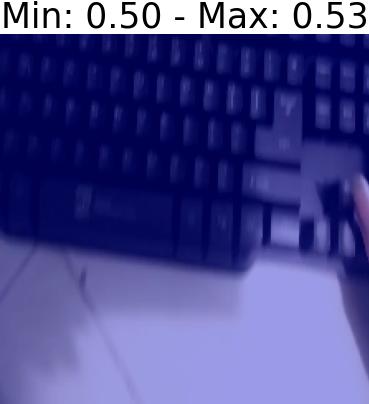}\\
\vspace{0.05cm}
\rotatebox[origin=c]{90}{\scriptsize FNAC } & 
\hspace{-0.45cm} \includegraphics[align=c, width=1.45cm]{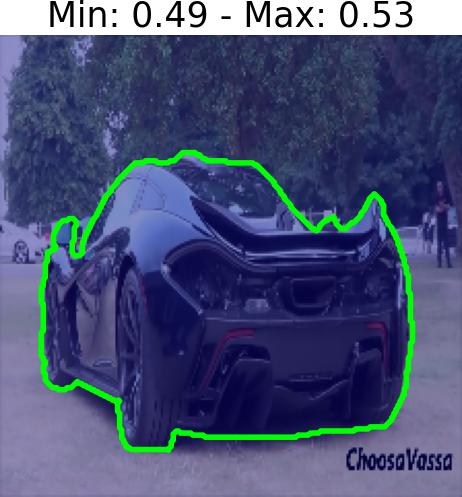}&
\hspace{-0.45cm} \includegraphics[align=c, width=1.45cm]{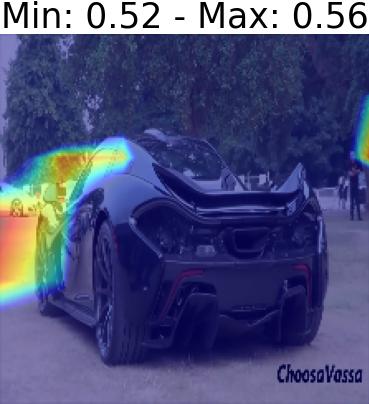}&
\hspace{-0.45cm} \includegraphics[align=c, width=1.45cm]{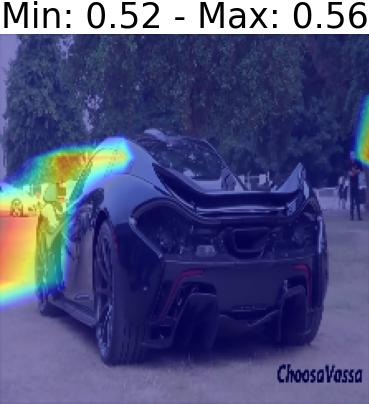}&
\hspace{-0.45cm} \includegraphics[align=c, width=1.45cm]{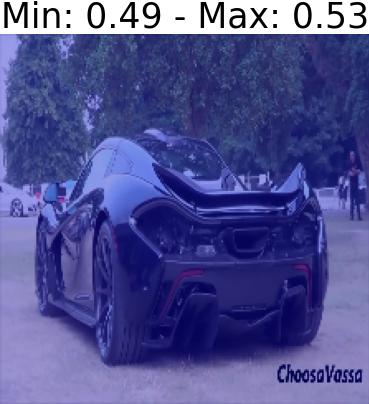}&
&
\hspace{-0.45cm} \includegraphics[align=c, width=1.45cm]{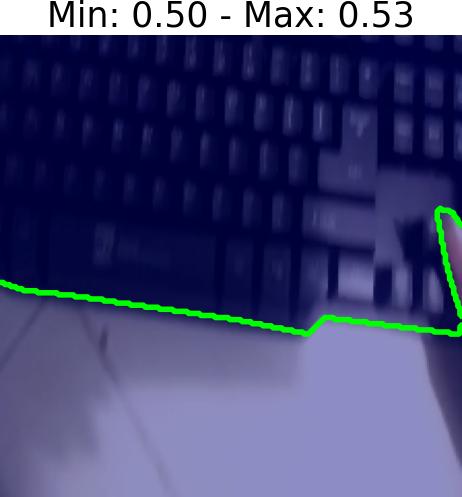}&
\hspace{-0.45cm} \includegraphics[align=c, width=1.45cm]{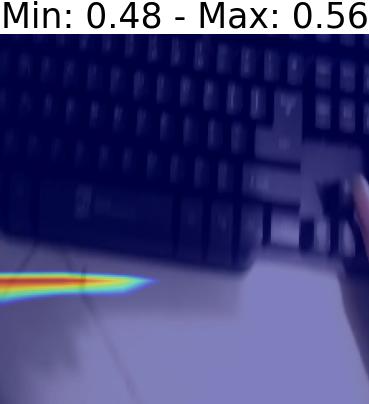}&
\hspace{-0.45cm} \includegraphics[align=c, width=1.45cm]{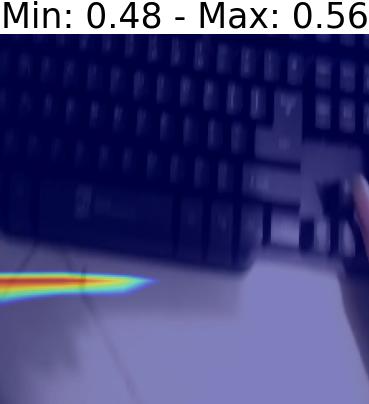}&
\hspace{-0.45cm} \includegraphics[align=c, width=1.45cm]{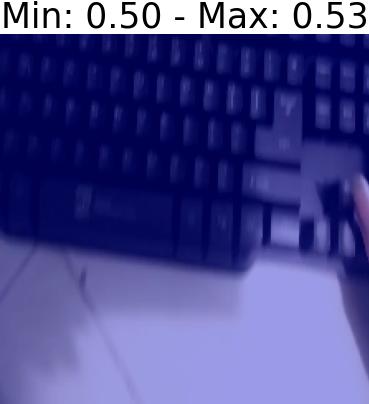}\\
\vspace{0.05cm}
\rotatebox[origin=c]{90}{\scriptsize SLAVC } & 
\hspace{-0.45cm} \includegraphics[align=c, width=1.45cm]{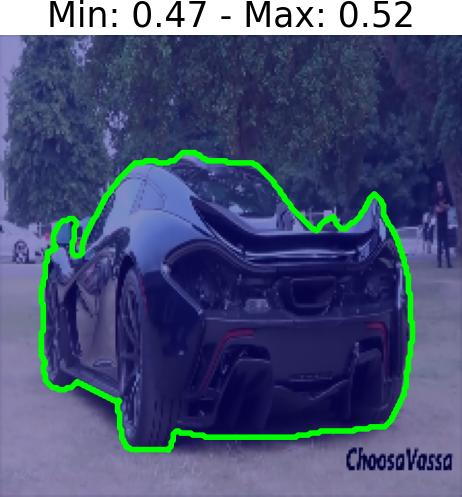}&
\hspace{-0.45cm} \includegraphics[align=c, width=1.45cm]{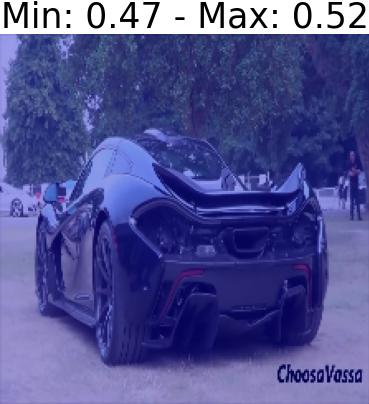}&
\hspace{-0.45cm} \includegraphics[align=c, width=1.45cm]{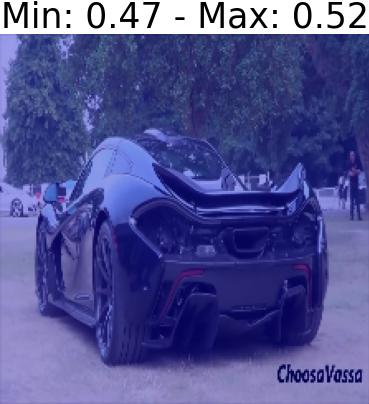}&
\hspace{-0.45cm} \includegraphics[align=c, width=1.45cm]{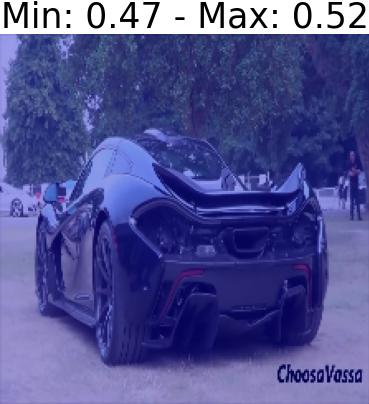}&
&
\hspace{-0.45cm} \includegraphics[align=c, width=1.45cm]{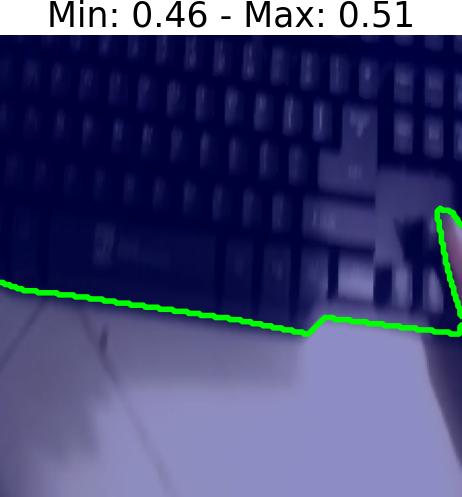}&
\hspace{-0.45cm} \includegraphics[align=c, width=1.45cm]{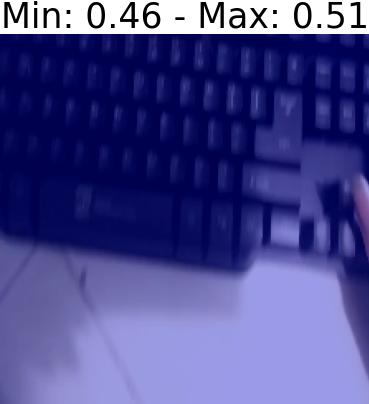}&
\hspace{-0.45cm} \includegraphics[align=c, width=1.45cm]{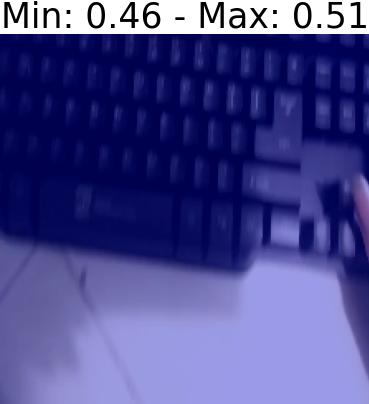}&
\hspace{-0.45cm} \includegraphics[align=c, width=1.45cm]{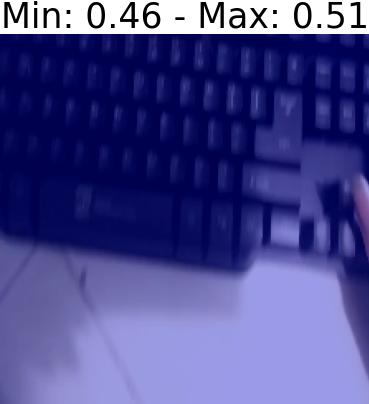}\\
\vspace{0.05cm}
\rotatebox[origin=c]{90}{\scriptsize SSL-TIE } & 
\hspace{-0.45cm} \includegraphics[align=c, width=1.45cm]{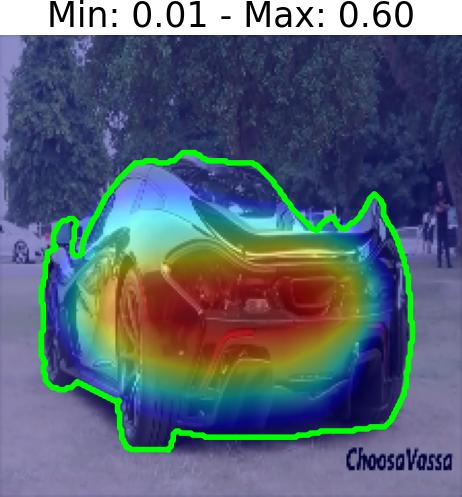}&
\hspace{-0.45cm} \includegraphics[align=c, width=1.45cm]{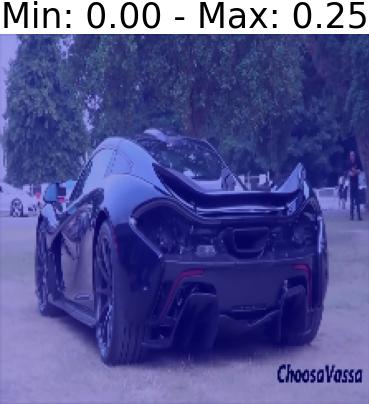}&
\hspace{-0.45cm} \includegraphics[align=c, width=1.45cm]{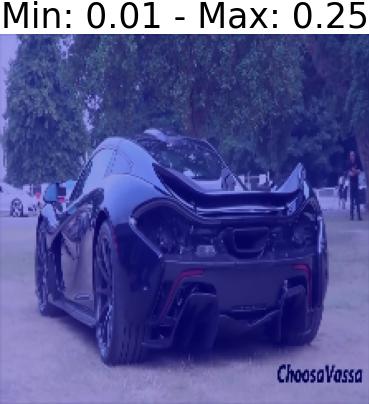}&
\hspace{-0.45cm} \includegraphics[align=c, width=1.45cm]{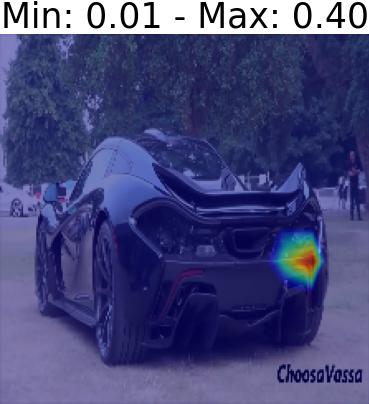}&
&
\hspace{-0.45cm} \includegraphics[align=c, width=1.45cm]{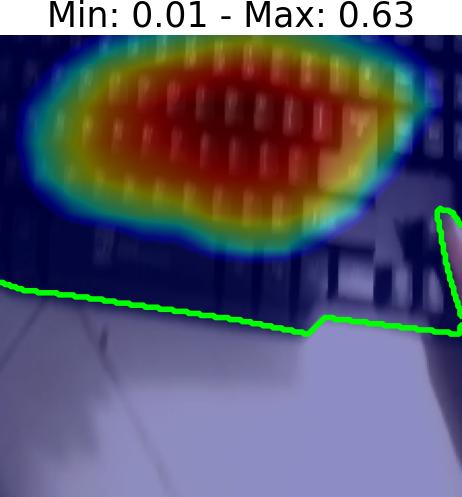}&
\hspace{-0.45cm} \includegraphics[align=c, width=1.45cm]{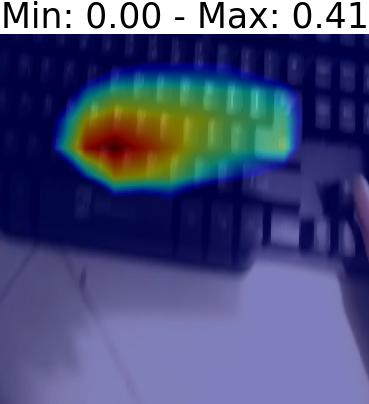}&
\hspace{-0.45cm} \includegraphics[align=c, width=1.45cm]{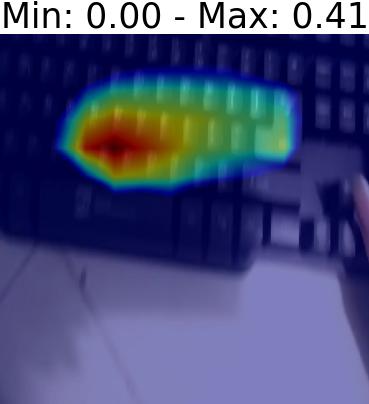}&
\hspace{-0.45cm} \includegraphics[align=c, width=1.45cm]{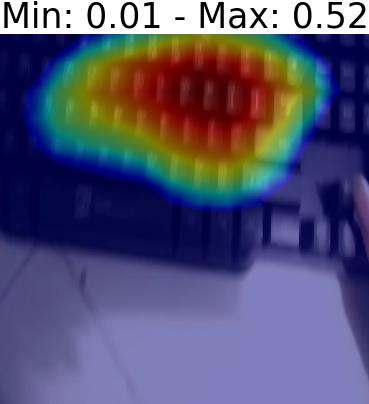}\\
\vspace{0.05cm}
\rotatebox[origin=c]{90}{\scriptsize SSL-Align (S.S.) } & 
\hspace{-0.45cm} \includegraphics[align=c, width=1.45cm]{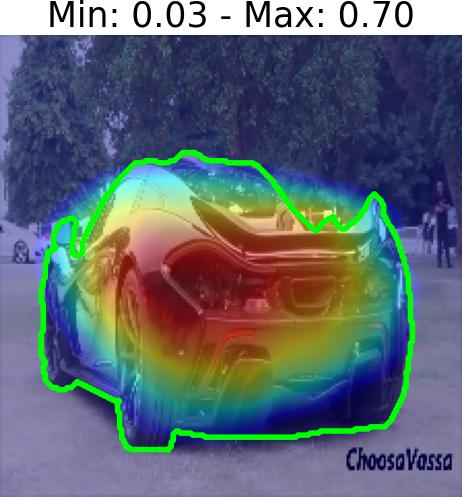}&
\hspace{-0.45cm} \includegraphics[align=c, width=1.45cm]{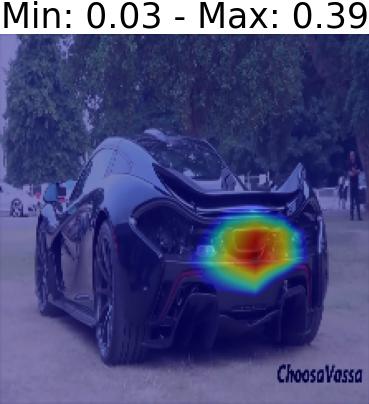}&
\hspace{-0.45cm} \includegraphics[align=c, width=1.45cm]{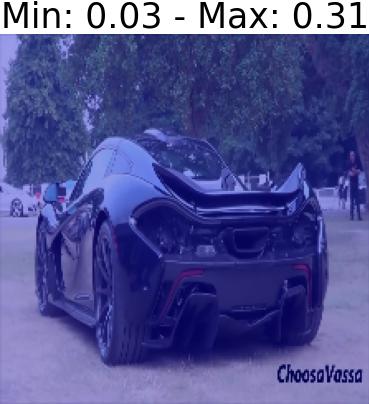}&
\hspace{-0.45cm} \includegraphics[align=c, width=1.45cm]{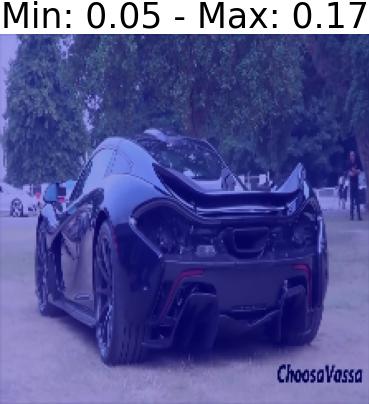}&
&
\hspace{-0.45cm} \includegraphics[align=c, width=1.45cm]{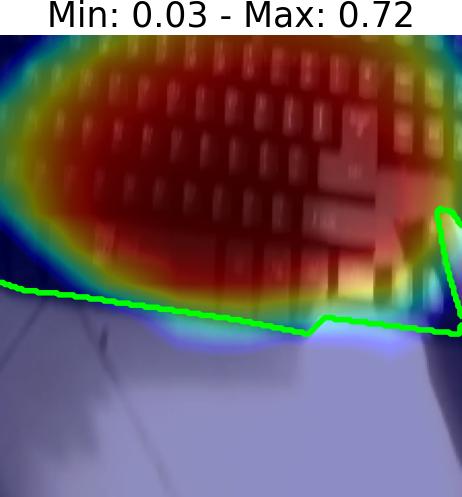}&
\hspace{-0.45cm} \includegraphics[align=c, width=1.45cm]{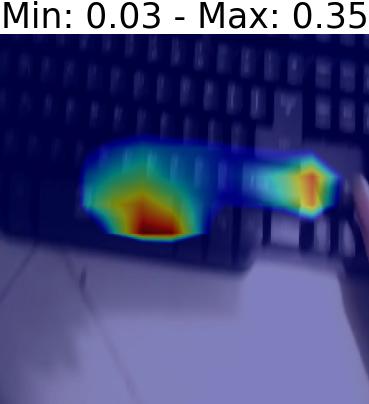}&
\hspace{-0.45cm} \includegraphics[align=c, width=1.45cm]{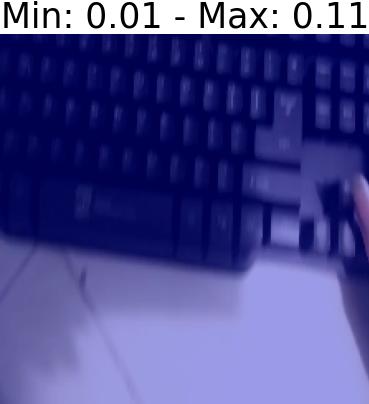}&
\hspace{-0.45cm} \includegraphics[align=c, width=1.45cm]{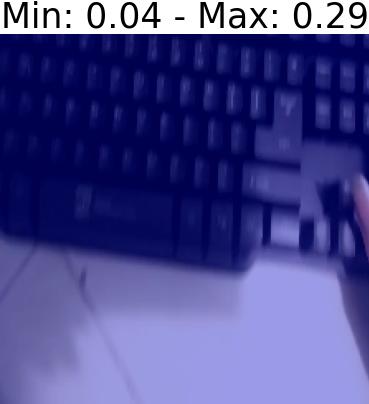}\\ \vspace{0.05cm}
\rotatebox[origin=c]{90}{\scriptsize ACL } & 
\hspace{-0.45cm} \includegraphics[align=c, width=1.45cm]{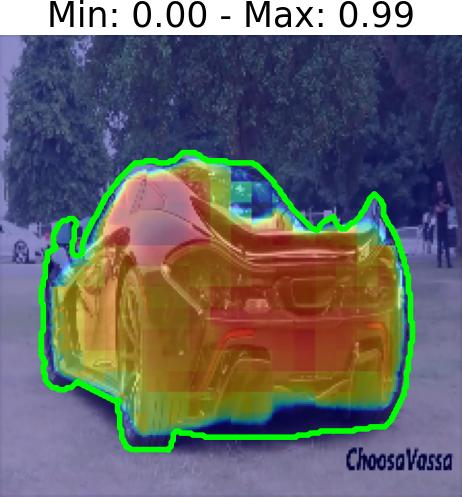}&
\hspace{-0.45cm} \includegraphics[align=c, width=1.45cm]{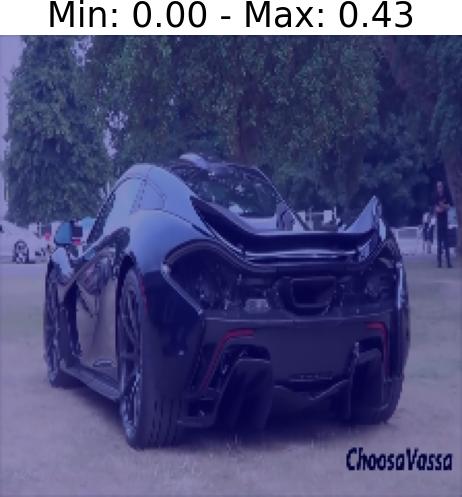}&
\hspace{-0.45cm} \includegraphics[align=c, width=1.45cm]{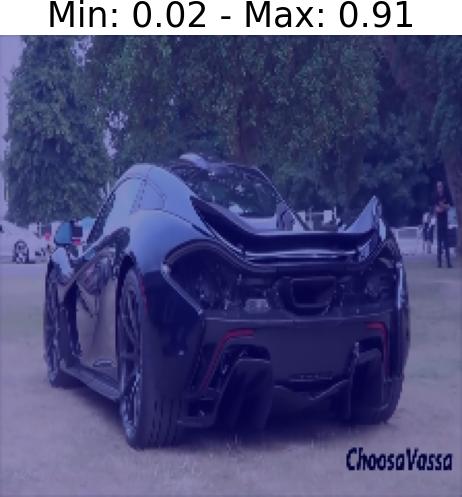}&
\hspace{-0.45cm} \includegraphics[align=c, width=1.45cm]{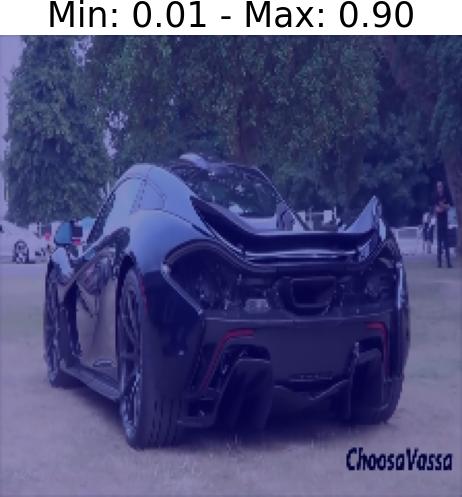}&
&
\hspace{-0.45cm} \includegraphics[align=c, width=1.45cm]{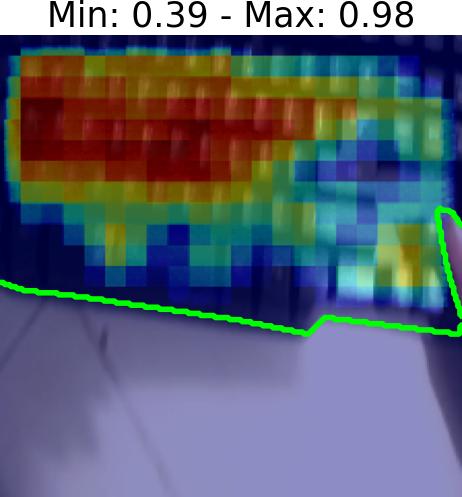}&
\hspace{-0.45cm} \includegraphics[align=c, width=1.45cm]{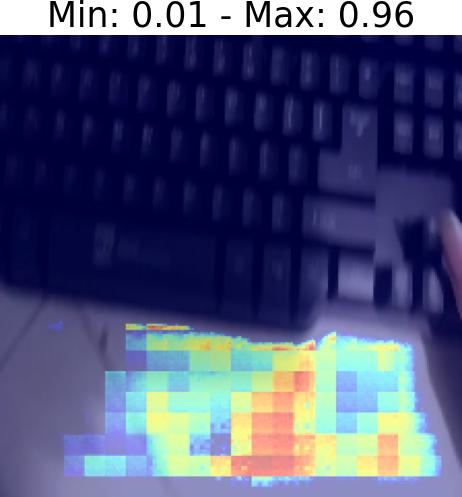}&
\hspace{-0.45cm} \includegraphics[align=c, width=1.45cm]{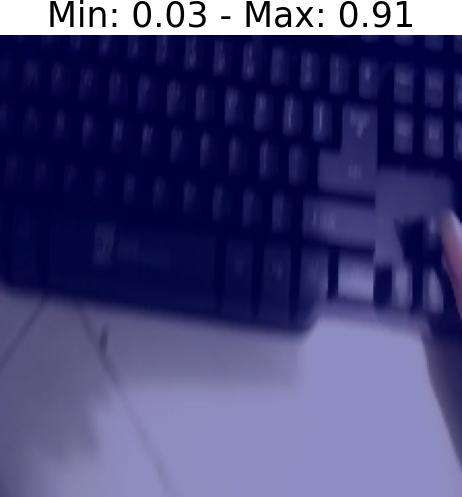}&
\hspace{-0.45cm} \includegraphics[align=c, width=1.45cm]{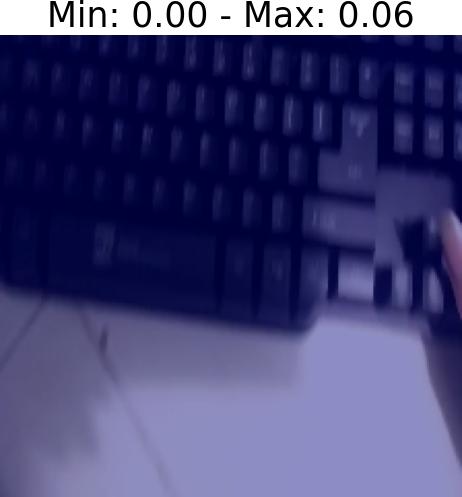}\\ \vspace{0.05cm}
\rotatebox[origin=c]{90}{\scriptsize \textbf{SSL-SaN} } & 
\hspace{-0.45cm} \includegraphics[align=c, width=1.45cm]{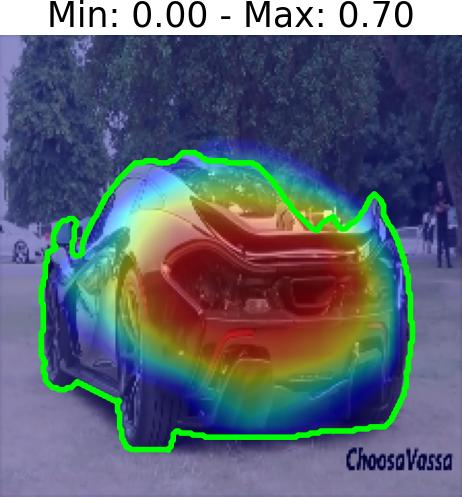}&
\hspace{-0.45cm} \includegraphics[align=c, width=1.45cm]{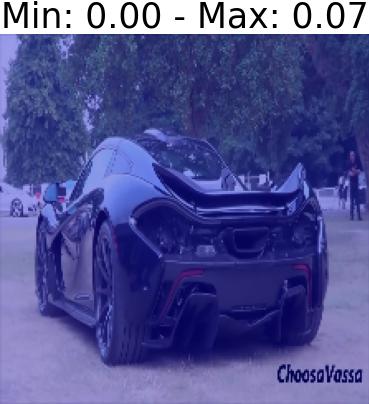}&
\hspace{-0.45cm} \includegraphics[align=c, width=1.45cm]{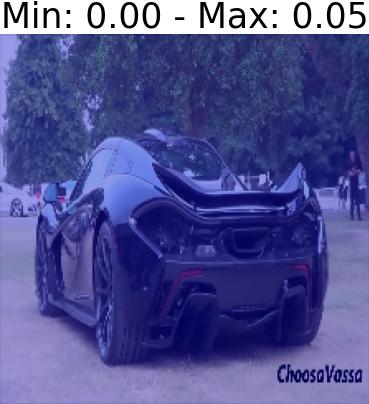}&
\hspace{-0.45cm} \includegraphics[align=c, width=1.45cm]{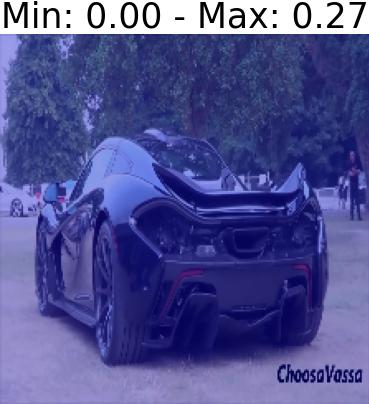} &
&
\hspace{-0.45cm} \includegraphics[align=c, width=1.45cm]{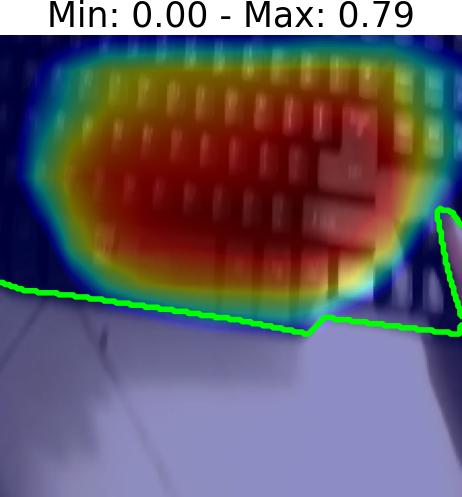}&
\hspace{-0.45cm} \includegraphics[align=c, width=1.45cm]{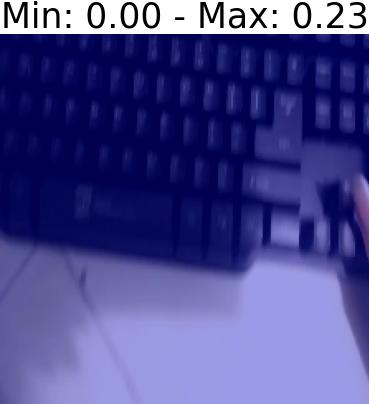}&
\hspace{-0.45cm} \includegraphics[align=c, width=1.45cm]{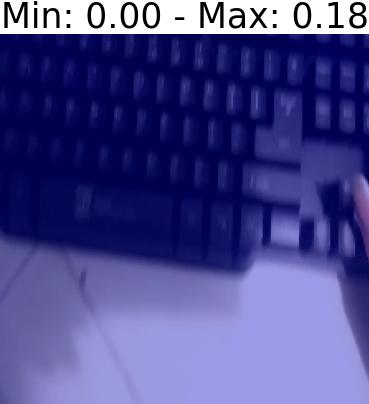}&
\hspace{-0.45cm} \includegraphics[align=c, width=1.45cm]{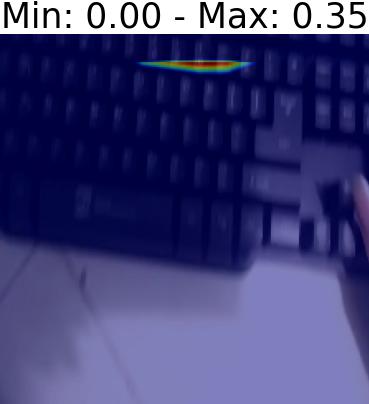} 
\end{tabular}
}
\vspace{0.01cm}
\caption{Example of localization results of the different models in both \textit{positive} and \textit{negative} audio samples in AVS-Bench S4 test set. The audio-visual similarities below the universal threshold of each model are clipped, normalized to the interval [0, 1], and overlaid with the original image. We show the outline of the segmentation mask (in green) of the positive cases for better understanding. The min. and max. values of the audio-visual similarities are reported on top of each image.}
\label{fig:sup_qualitative_s4}
\end{figure*}

\new{
In Figure \ref{fig:sup_qualitative_s4}, we present two examples from the AVS-Bench S4 test set that highlight how our model outperforms competing approaches.

In the first example, the scene features a race car. Similar to the IS3+ case, ACL achieves nearly perfect localization while filtering out all negative sounds. SSL-TIE, SSL-Align, and SSL-SaN also localize the positive sound correctly; however, only our model successfully suppresses all negatives. Specifically, SSL-TIE fails to filter the offscreen sound, while SSL-Align fails to filter the silence. 

In the second example, showing a computer keyboard, SSL-SaN, ACL, SSL-Align, and SSL-TIE all produce correct localizations of the keyboard. Yet, SSL-SaN is the only model that does not produce false activations for the negative sounds. SSL-Align and ACL fail to filter the silence, while SSL-TIE fails across all three negative cases.

}
\subsection{Failure cases}
\label{sup:failure_cases}

\subsubsection{VGG-SS}

\begin{figure*}[!ht]
\centering
\resizebox{\textwidth}{!}{%
\begin{tabular}{ccccccc}

\scriptsize \raisebox{-0.17cm}{\includegraphics[height=0.5cm]{Figures/equalizer.png}} Positive case &
\includegraphics[width=1.45cm]{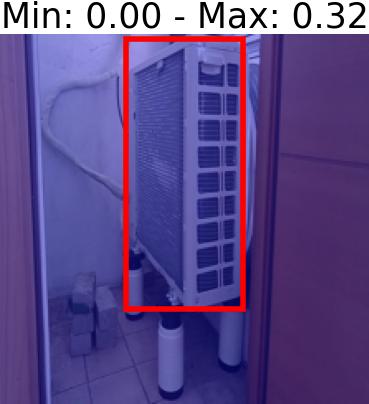} &
\includegraphics[width=1.45cm]{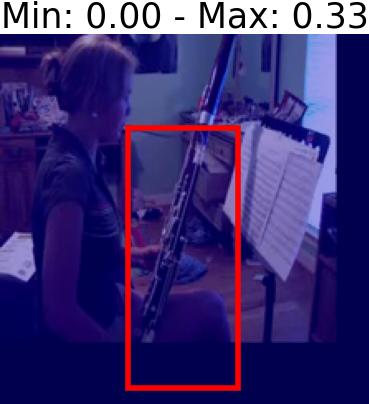} &
\includegraphics[width=1.45cm]{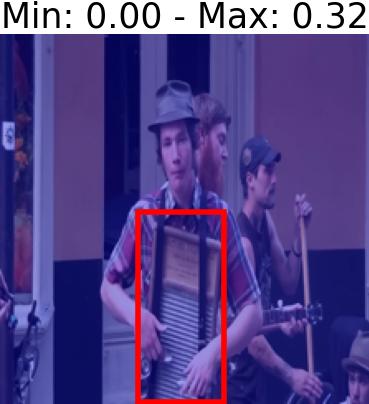} &
\includegraphics[width=1.45cm]{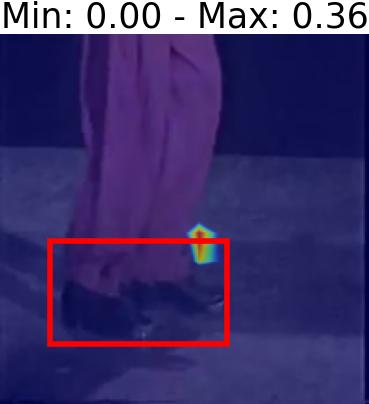} &
\includegraphics[width=1.45cm]{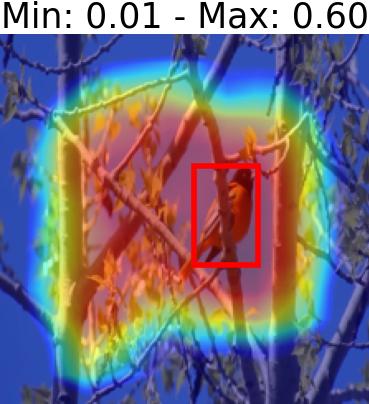} &
\includegraphics[width=1.45cm]{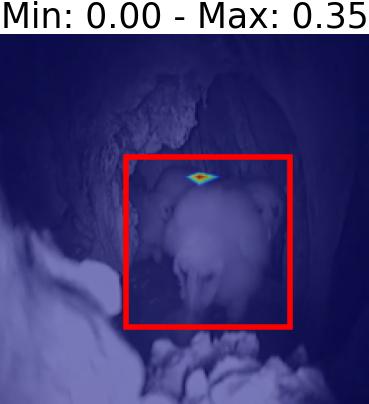} \\
\scriptsize \raisebox{-0.17cm}{\includegraphics[height=0.5cm]{Figures/equalizer.png}} Silence &
\includegraphics[width=1.45cm]{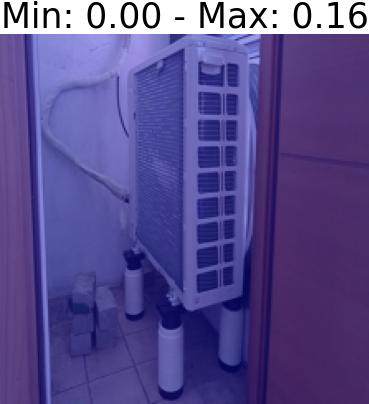} &
\includegraphics[width=1.45cm]{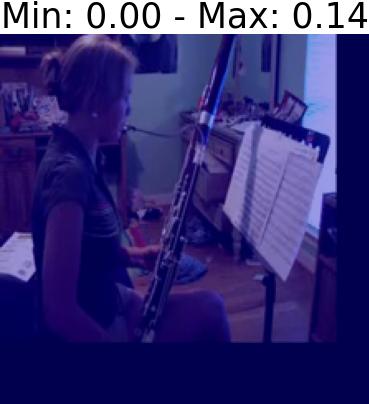} &
\includegraphics[width=1.45cm]{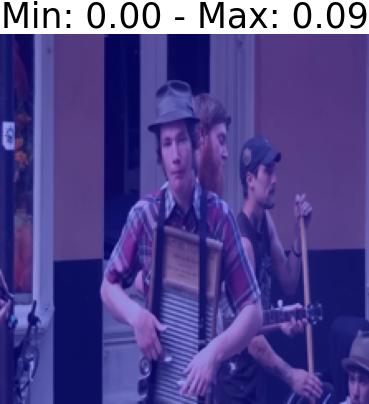} &
\includegraphics[width=1.45cm]{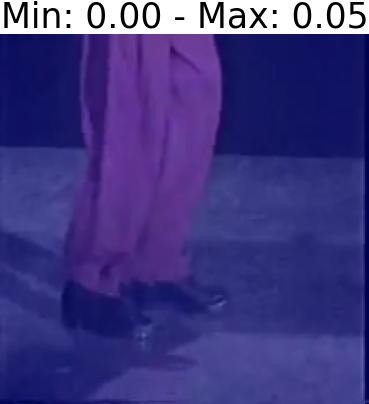} &
\includegraphics[width=1.45cm]{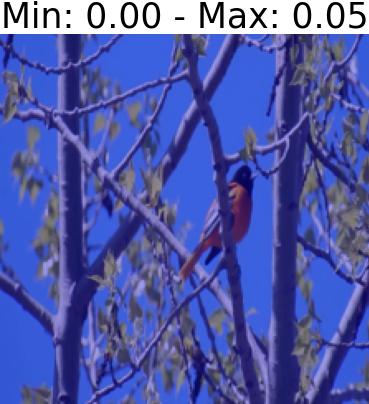} &
\includegraphics[width=1.45cm]{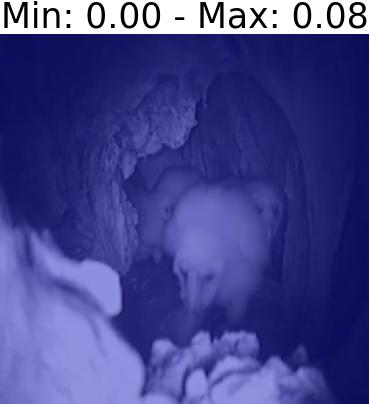} \\
\scriptsize \raisebox{-0.17cm}{\includegraphics[height=0.5cm]{Figures/equalizer.png}} Noise &
\includegraphics[width=1.45cm]{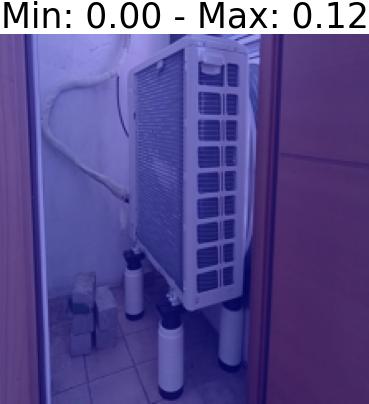} &
\includegraphics[width=1.45cm]{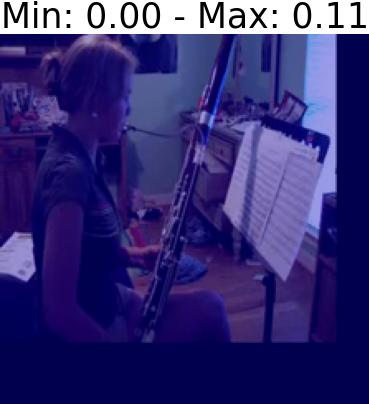} &
\includegraphics[width=1.45cm]{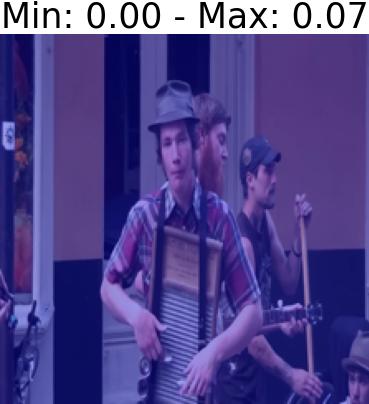} &
\includegraphics[width=1.45cm]{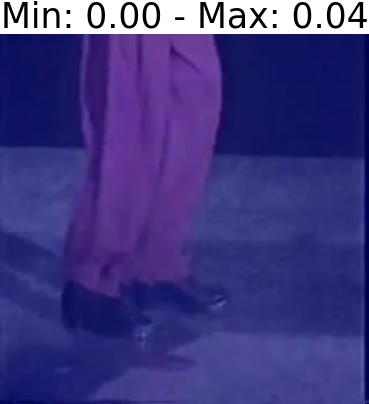} &
\includegraphics[width=1.45cm]{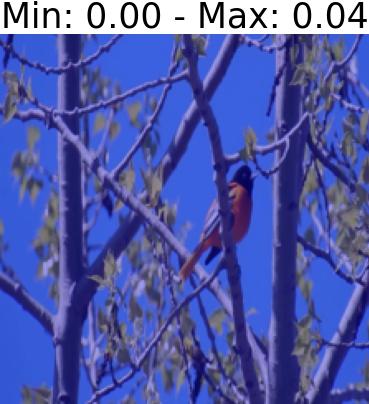} &
\includegraphics[width=1.45cm]{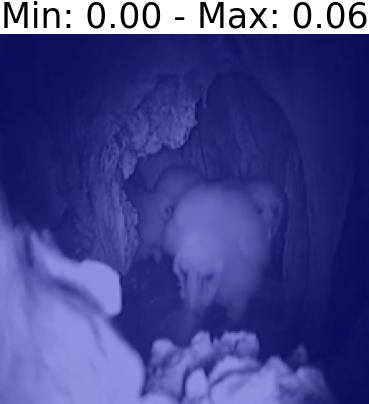} \\
\scriptsize \raisebox{-0.17cm}{\includegraphics[height=0.5cm]{Figures/equalizer.png}} Offscreen &
\includegraphics[width=1.45cm]{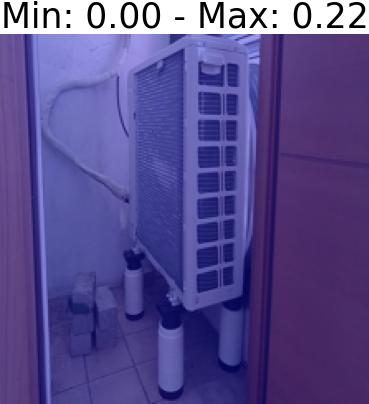} &
\includegraphics[width=1.45cm]{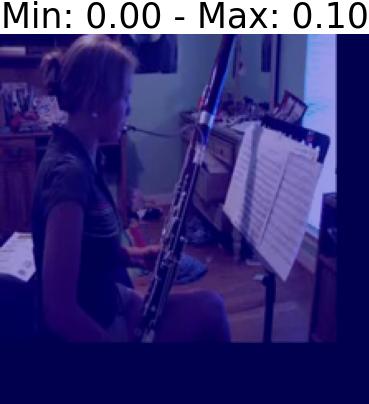} &
\includegraphics[width=1.45cm]{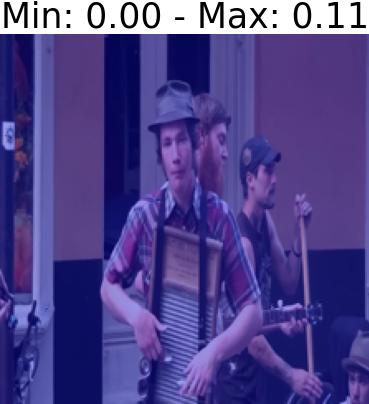} &
\includegraphics[width=1.45cm]{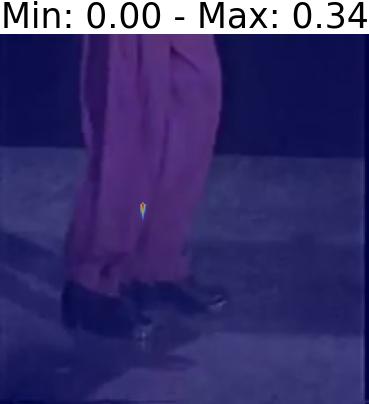} &
\includegraphics[width=1.45cm]{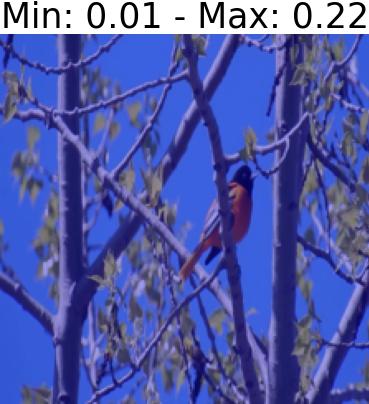} &
\includegraphics[width=1.45cm]{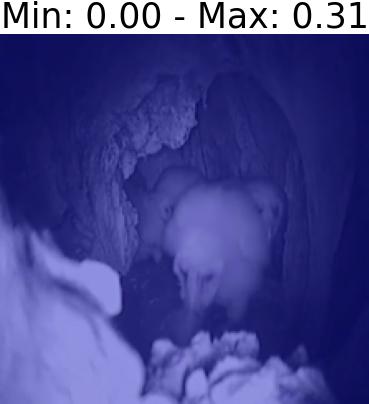} \\
\end{tabular}
}
\vspace{0.01cm}
\caption{Example of localization results of the different models in both positive and negative audio samples of some failure cases in VGG-SS test set. The audio-visual similarities below the universal threshold of each model are clipped, normalized to the interval [0, 1], and overlaid with the original image. We show the bounding box (in red) of the positive cases for better understanding. The min. and max. values of the audio-visual similarities are reported on top of each image.}
\label{fig:sup_failure_vggss}
\end{figure*}

\new{In Figure \ref{fig:sup_failure_vggss}, we present six failure cases from the VGG-Sound Sources test set. In the first example, the video shows an air conditioner: although the sound is identifiable as the machine, it is weak, and the model fails to localize the source in the image. In the second and third examples, the model struggles to localize the different instruments. In the fourth example, the audio is the sound of the shoes against the floor (slap dancing) and our model is not able to identify the sound source in the image. In the last two examples, the sound is also clear, but the model appears to interpret the tree’s leaves as birds and fails to detect the owl in the final image.}
\clearpage

\subsubsection{IS3+}
\begin{figure*}[!ht]
\centering
\resizebox{\textwidth}{!}{%
\begin{tabular}{ccccccc}

\scriptsize \raisebox{-0.17cm}{\includegraphics[height=0.5cm]{Figures/equalizer.png}} Positive case &
\includegraphics[width=1.45cm]{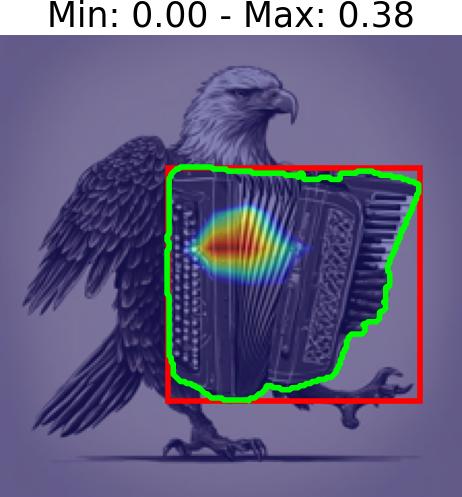} &
\includegraphics[width=1.45cm]{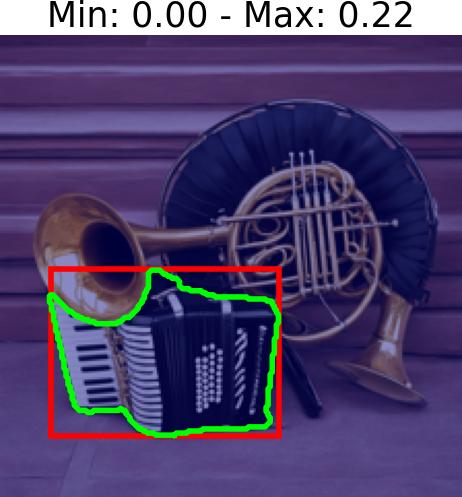} &
\includegraphics[width=1.45cm]{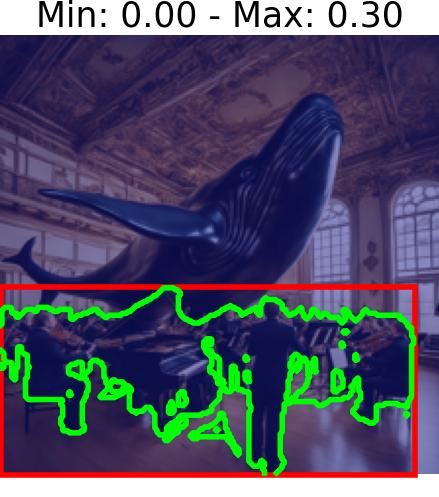} &
\includegraphics[width=1.45cm]{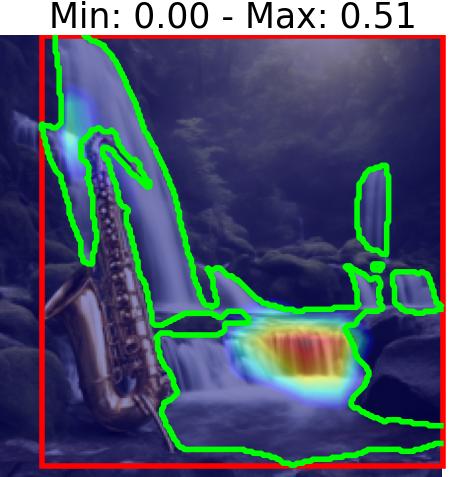} &
\includegraphics[width=1.45cm]{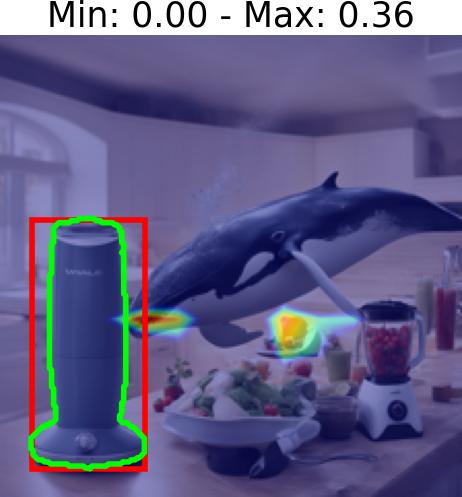} &
\includegraphics[width=1.45cm]{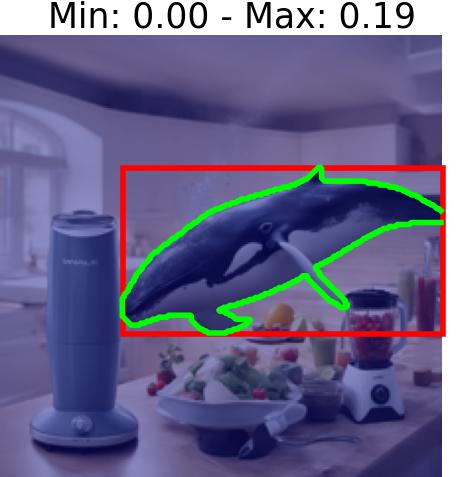} \\
\scriptsize \raisebox{-0.17cm}{\includegraphics[height=0.5cm]{Figures/equalizer.png}} Silence &
\includegraphics[width=1.45cm]{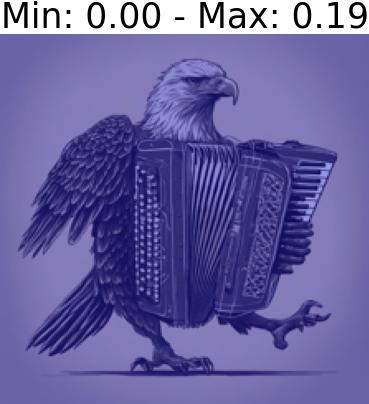} &
\includegraphics[width=1.45cm]{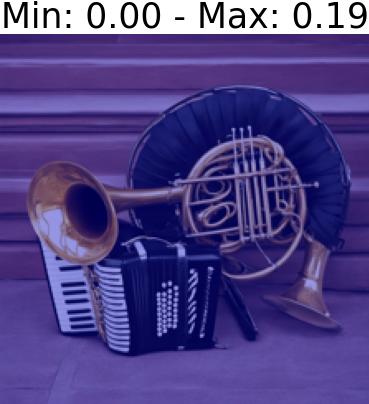} &
\includegraphics[width=1.45cm]{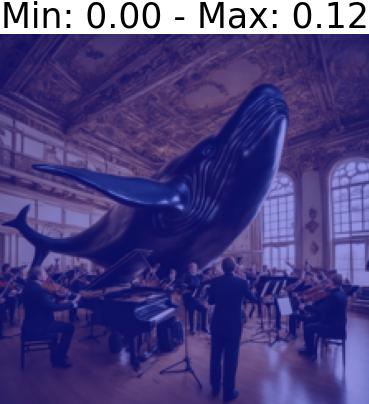} &
\includegraphics[width=1.45cm]{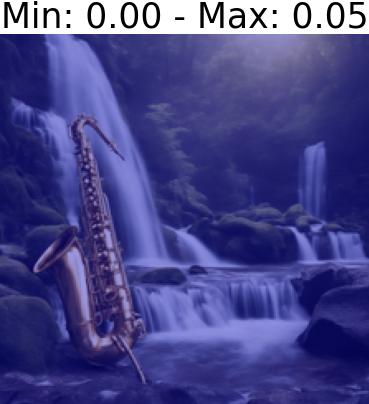} &
\includegraphics[width=1.45cm]{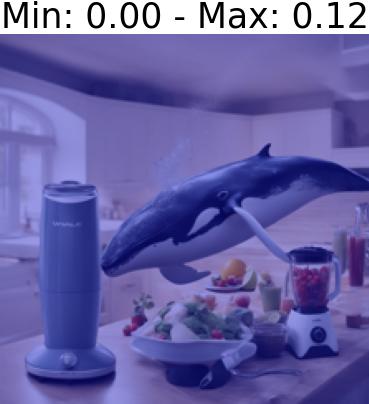} &
\includegraphics[width=1.45cm]{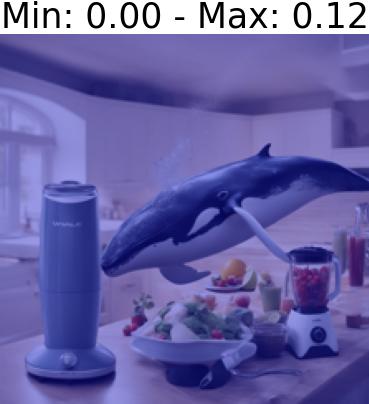} \\
\scriptsize \raisebox{-0.17cm}{\includegraphics[height=0.5cm]{Figures/equalizer.png}} Noise &
\includegraphics[width=1.45cm]{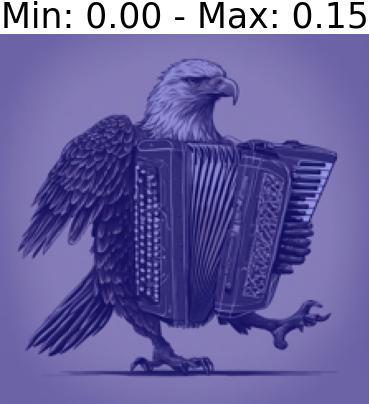} &
\includegraphics[width=1.45cm]{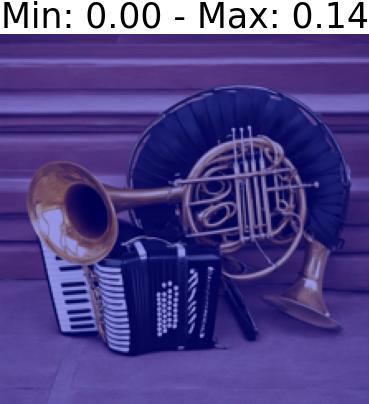} &
\includegraphics[width=1.45cm]{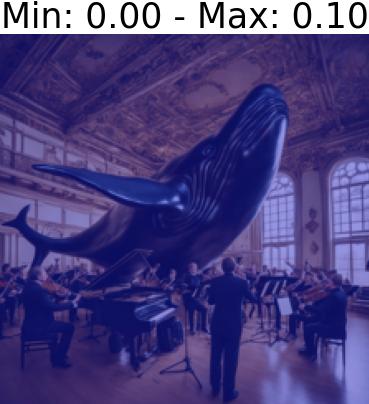} &
\includegraphics[width=1.45cm]{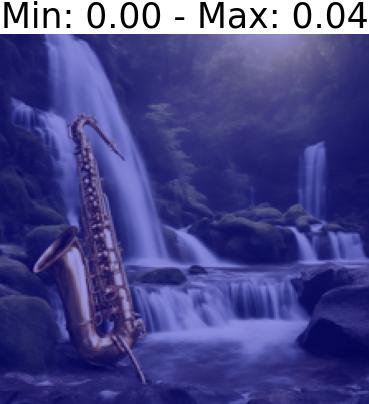} &
\includegraphics[width=1.45cm]{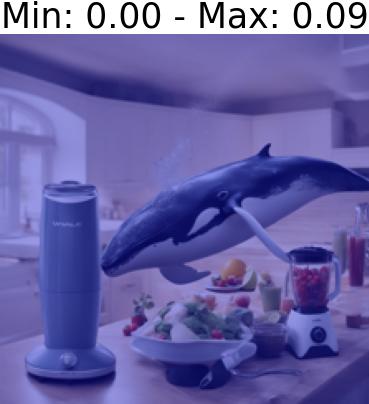} &
\includegraphics[width=1.45cm]{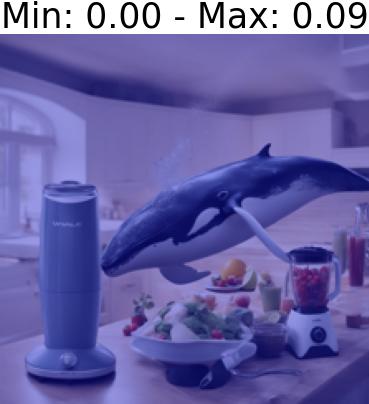} \\
\scriptsize \raisebox{-0.17cm}{\includegraphics[height=0.5cm]{Figures/equalizer.png}} Offscreen &
\includegraphics[width=1.45cm]{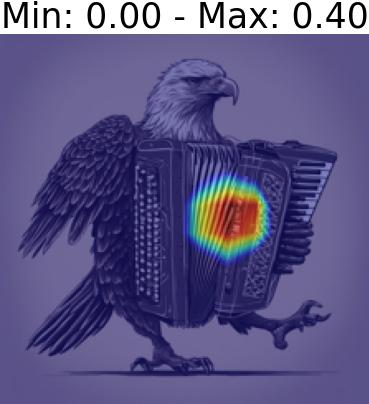} &
\includegraphics[width=1.45cm]{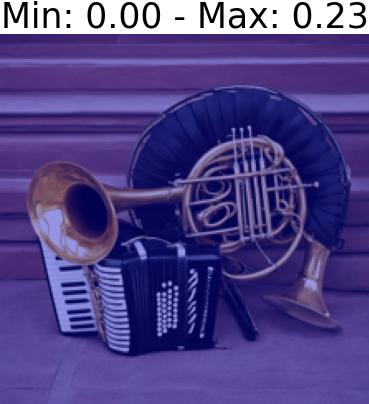} &
\includegraphics[width=1.45cm]{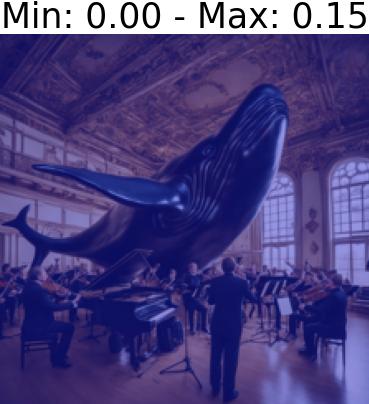} &
\includegraphics[width=1.45cm]{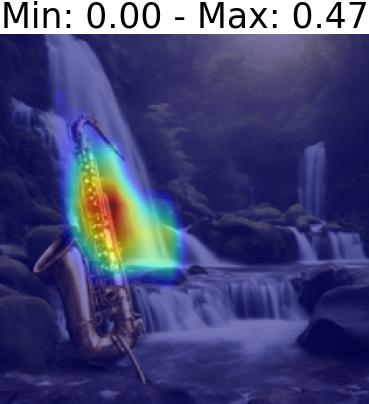} &
\includegraphics[width=1.45cm]{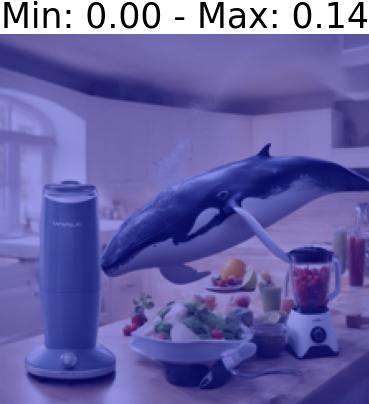} &
\includegraphics[width=1.45cm]{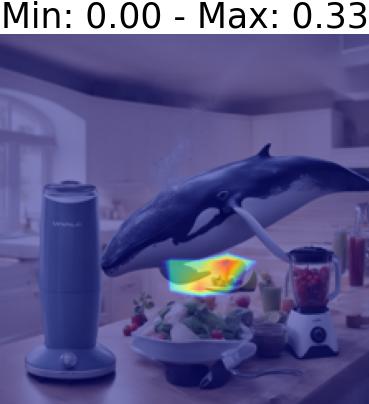} \\
\end{tabular}
}
\vspace{0.01cm}
\caption{Example of localization results of the different models in both positive and negative audio samples of some failure cases in IS3$^+$ test set. The audio-visual similarities below the universal threshold of each model are clipped, normalized to the interval [0, 1], and overlaid with the original image. We show the outline of the segmentation mask (in green) of the positive cases for better understanding. The min. and max. values of the audio-visual similarities are reported on top of each image.}
\label{fig:sup_failure_is3+}
\end{figure*}

\new{Figure \ref{fig:sup_failure_is3+}, shows six examples where our model fails to detect the sounding object in the IS3+  curated test set. In the first and fourth examples, the model correctly identifies the sound sources (accordion and waterfall), but produces very limited or sparse localization maps compared to the actual size of the objects. In the remaining cases, the model completely fails to detect the sounding source and does not localize any relevant region of the image. We also see that our model tends to fail more towards the offscreen negative sound. This is an expected result if we observe Figure \ref{fig:boxplots}, where we can see that in IS3+, the Universal threshold in SSL\_SaN doesn't completely filter the interquartile range of offscreen sounds (the separibility of positive and offscreen is much better in the other two datasets). }
\clearpage

\subsubsection{AVSBench S4}

\begin{figure*}[!ht]
\centering
\resizebox{\textwidth}{!}{%
\begin{tabular}{ccccccc}

\scriptsize \raisebox{-0.17cm}{\includegraphics[height=0.5cm]{Figures/equalizer.png}} Positive case &
\includegraphics[width=1.45cm]{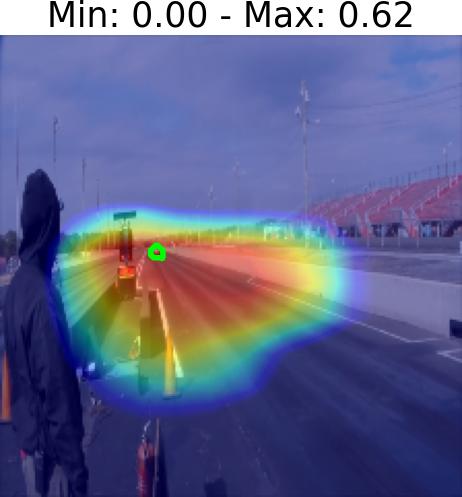} &
\includegraphics[width=1.45cm]{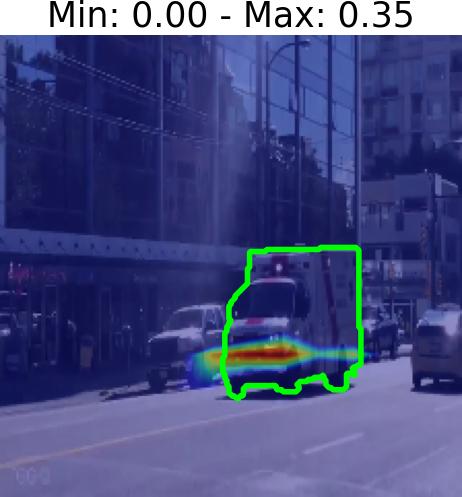} &
\includegraphics[width=1.45cm]{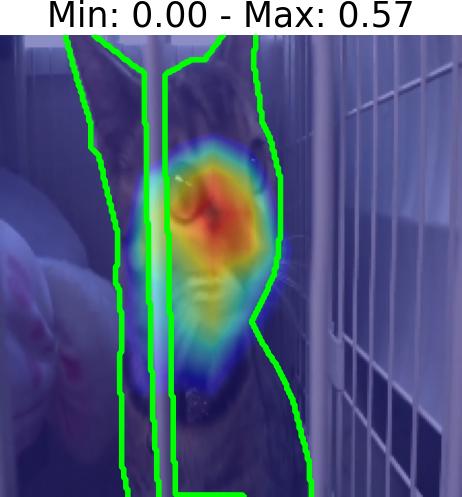} &
\includegraphics[width=1.45cm]{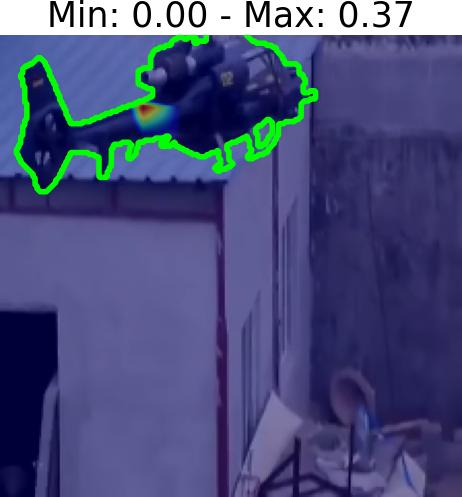} &
\includegraphics[width=1.45cm]{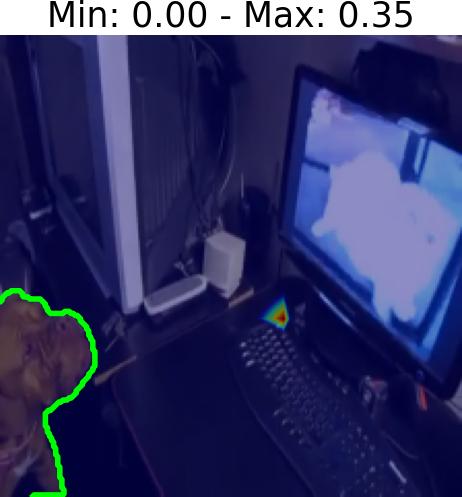} &
\includegraphics[width=1.45cm]{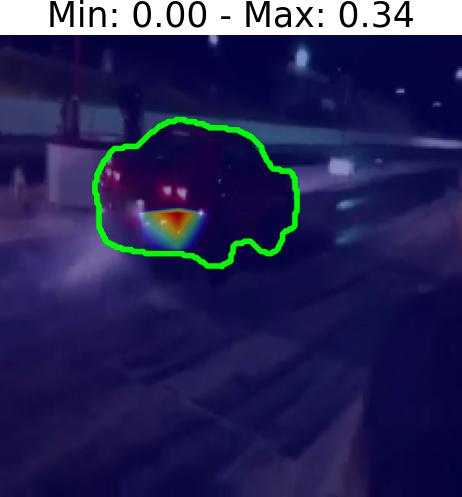} \\
\scriptsize \raisebox{-0.17cm}{\includegraphics[height=0.5cm]{Figures/equalizer.png}} Silence &
\includegraphics[width=1.45cm]{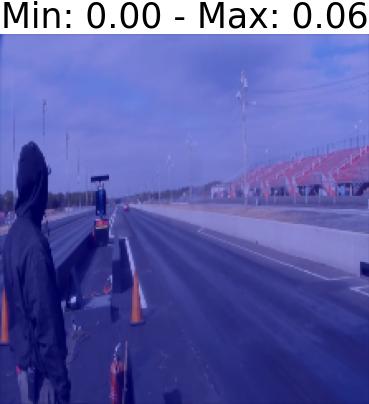} &
\includegraphics[width=1.45cm]{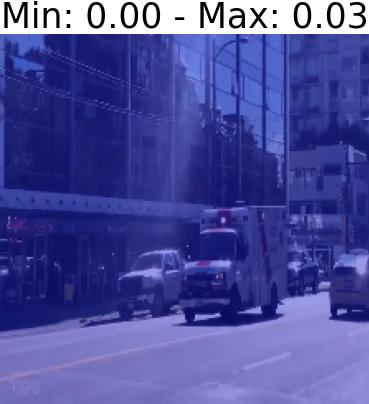} &
\includegraphics[width=1.45cm]{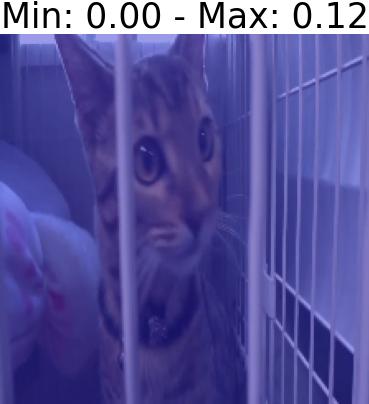} &
\includegraphics[width=1.45cm]{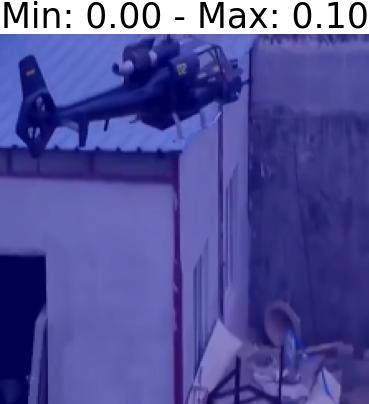} &
\includegraphics[width=1.45cm]{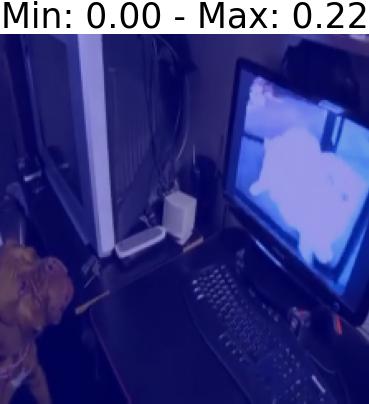} &
\includegraphics[width=1.45cm]{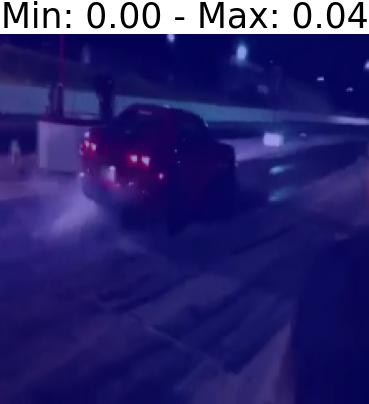} \\
\scriptsize \raisebox{-0.17cm}{\includegraphics[height=0.5cm]{Figures/equalizer.png}} Noise &
\includegraphics[width=1.45cm]{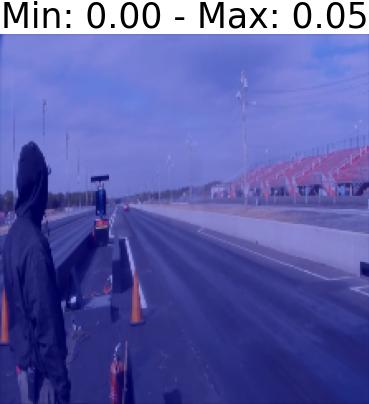} &
\includegraphics[width=1.45cm]{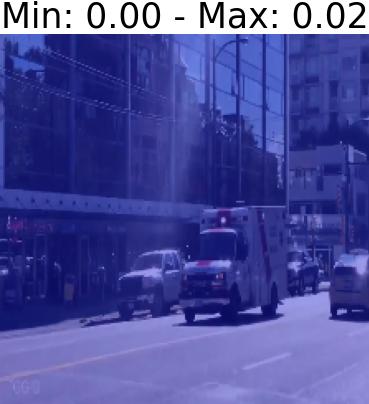} &
\includegraphics[width=1.45cm]{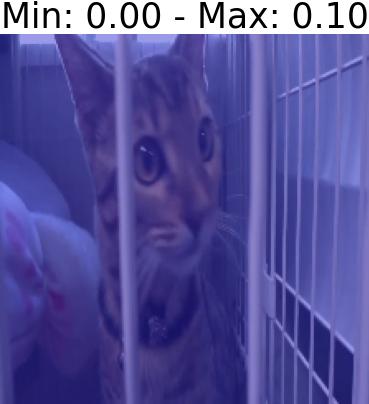} &
\includegraphics[width=1.45cm]{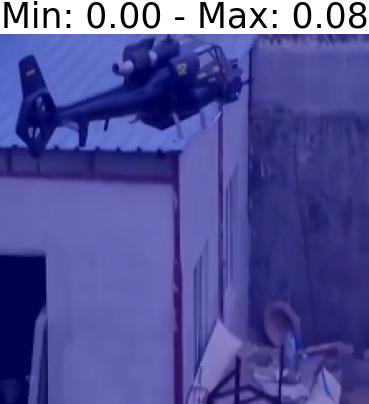} &
\includegraphics[width=1.45cm]{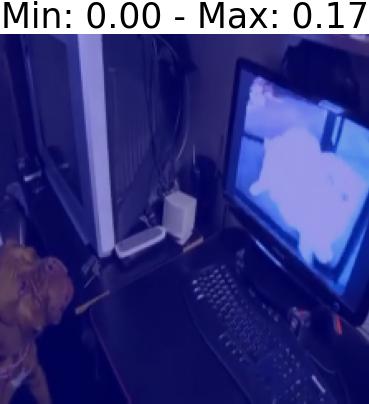} &
\includegraphics[width=1.45cm]{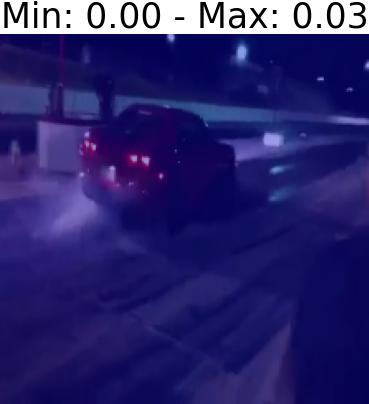} \\
\scriptsize \raisebox{-0.17cm}{\includegraphics[height=0.5cm]{Figures/equalizer.png}} Offscreen &
\includegraphics[width=1.45cm]{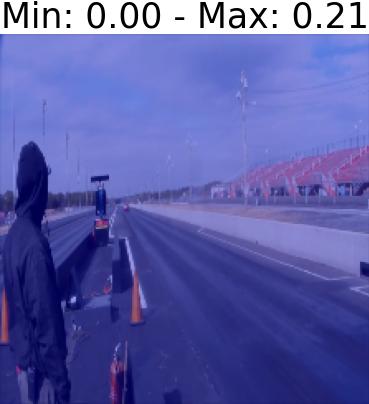} &
\includegraphics[width=1.45cm]{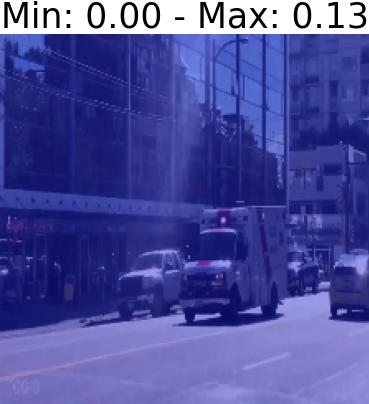} &
\includegraphics[width=1.45cm]{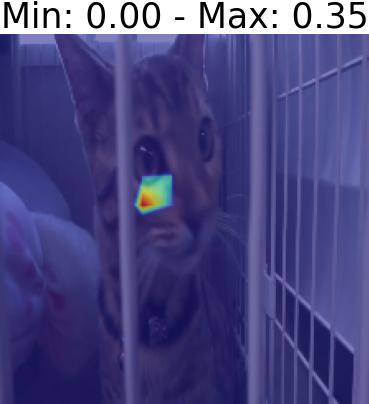} &
\includegraphics[width=1.45cm]{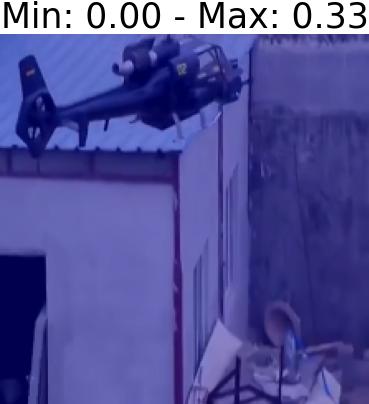} &
\includegraphics[width=1.45cm]{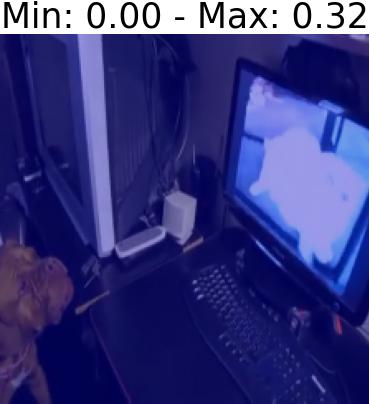} &
\includegraphics[width=1.45cm]{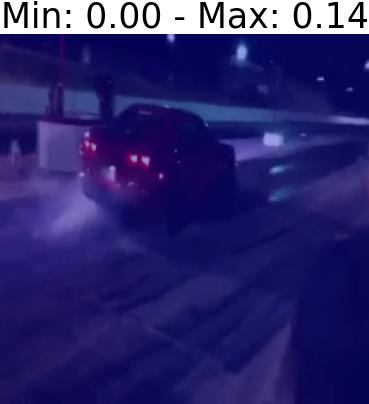} \\
\end{tabular}
}
\vspace{0.01cm}
\caption{Example of localization results of the different models in both positive and negative audio samples of some failure cases in AVS-Bench S4 test set. The audio-visual similarities below the universal threshold of each model are clipped, normalized to the interval [0, 1], and overlaid with the original image. We show the outline of the segmentation mask (in green) of the positive cases for better understanding. The min. and max. values of the audio-visual similarities are reported on top of each image.}
\label{fig:sup_failure_s4}
\end{figure*}

\new{
We present several failure cases from the AVSBench S4 test set in Figure \ref{fig:sup_failure_s4}. In the first two images, the target vehicles (a distant car and an ambulance) appear very small, and the model fails to extract sufficient detail for reliable localization. In the third and fifth columns, animals are clearly visible as foreground, yet the model does not localize them. In the third image, a pole occludes the scene and the cat’s face is only partially visible, which further hinders detection. The fourth and sixth images also contain vehicles that the model misses. Notably, in the first, second, third, and sixth images the model produces weak activations in approximately the correct region, but these responses are too diffuse or misplaced to demonstrate clear object-level localization.
}

\subsubsection{Discussion}

\new{

From the qualitative inspection of the failure cases across VGG-SS, IS3+, and AVS-Bench S4, several consistent trends can be identified. A common difficulty arises when the sounding object occupies only a small portion of the image, as with the distant vehicles in AVS-Bench S4, where the model fails to associate the weak visual evidence with the audio cue. Complex scenes also introduce confusion: for instance, musical instrument cases in VGG-SS often include people, sheet music, and microphones, all of which provide competing visual structures that the model mistakenly attends to. Similar effects are observed in nature scenes, such as the bird example where dense foliage is misinterpreted as additional animals. The model further struggles with monotonous sounds, where the acoustic signal provides little temporal variation or discriminative content, as seen with air conditioners or slap dance performances, leading to diffuse or misplaced activations. Partial occlusions represent another challenge, with examples such as cats or dogs hidden behind obstacles, or a helicopter where the blades are not visible, causing incomplete or absent localizations. These patterns suggest that limitations arise when the sound offers limited semantic cues, when the visual field is cluttered with distractors, or when the target object is visually ambiguous due to scale or occlusion.

}

\end{document}